\documentclass{article}

\usepackage{graphicx}
\usepackage[colorlinks=true, allcolors=blue]{hyperref}
\usepackage{booktabs}
\usepackage{caption}
\usepackage{subcaption}
\usepackage{tabularx}
\usepackage{amsmath,amssymb,mathtools,amsthm, bm}
\usepackage{float}
\usepackage{cleveref}
\usepackage{threeparttable}
\DeclareMathOperator*{\argmax}{arg\,max}
\usepackage{indentfirst}
\usepackage{makecell}
\usepackage{enumitem}

\newtheorem{definition}{Definition}

\newtheorem{hypothesis}{Hypothesis}

\usepackage[accepted, preprint]{icml2026}

\usepackage[textsize=tiny]{todonotes}

\icmltitlerunning{How to Achieve the Intended Aim of Deep Clustering Now, without Deep Learning}

\begin{document}

\twocolumn[
    \icmltitle{How to Achieve the Intended Aim of Deep Clustering Now, \texorpdfstring{\\}{ } without Deep Learning}



  \icmlsetsymbol{equal}{*}
  \icmlsetsymbol{corresp}{\textdagger}

  \begin{icmlauthorlist}
    \icmlauthor{Kai Ming Ting}{state,nju,corresp}
    \icmlauthor{Wei-Jie Xu}{state,nju}
    \icmlauthor{Hang Zhang}{state,nju}
  \end{icmlauthorlist}

  \icmlaffiliation{state}{State Key Laboratory for Novel Software Technology, Nanjing University, Nanjing, China}
  \icmlaffiliation{nju}{School of Artificial Intelligence, Nanjing University, Nanjing, China}

  \icmlcorrespondingauthor{Kai Ming Ting}{tingkm@nju.edu.cn}

  \icmlkeywords{Machine Learning, ICML}

  \vskip 0.3in
]



\printAffiliationsAndNotice{}  

\begin{abstract}
Deep clustering (DC) is often quoted to have a key advantage over $k$-means clustering. Yet, this advantage is often demonstrated using image datasets only, and it is unclear whether it addresses the fundamental limitations of $k$-means clustering. Deep Embedded Clustering (DEC) learns a latent representation via an autoencoder and performs clustering based on a $k$-means-like procedure, while the optimization is conducted in an end-to-end manner. This paper investigates  whether the deep-learned representation has enabled DEC to overcome the known fundamental limitations of $k$-means clustering, i.e., its inability to discover clusters of arbitrary shapes, varied sizes and densities. Our investigations on DEC have a wider implication on deep clustering methods in general. Notably, none of these methods exploit the underlying data distribution. We uncover that a non-deep learning approach achieves the intended aim of deep clustering by making use of distributional information of clusters in a dataset to effectively address these fundamental limitations. 
\end{abstract}

\section{Introduction}
Many methods of Deep Clustering (DC) have been proposed in the last few years, building upon the successes of deep learning in supervised learning tasks in Computer Vision \cite{krizhevsky2012imagenet, he2016deep}, Natural Language Processing \cite{vaswani2017attention, devlin2019bert} and Large Language Models \cite{bai2023qwen, achiam2023gpt}. Though some successes have been reported in the unsupervised learning clustering task, they have been largely constrained to image datasets only \cite{xie2016unsupervised, ijcai2017p243, li2021contrastive}; and the research has focused on different methods to learn a good latent representation for clustering, for example, DEC and IDEC employ autoencoders to learn low-dimensional data embeddings \cite{xie2016unsupervised, ijcai2017p243}, while Contrastive Clustering (CC) leverages a contrastive learning paradigm to learn both feature representations and cluster representations \cite{li2021contrastive}. 

Beginning with the age-old definition of clustering, we uncover that DC has implicitly employed the same definition which has one critical deficiency: The definition does not specify the characteristics of the clusters that
enable one to assess the ‘goodness’ of the clusters produced from a clustering method. Instead, the definition demands a point-to-point similarity measure to be used to find similar points in a cluster. This has influenced how a clustering method is designed, including DC.

The inadequacy of the age-old definition has prompted us to explore new definitions of clustering that explicitly state the kind of target clusters. These allow us to examine the fundamental limitations of DC. The investigation starts with Deep Embedded Clustering (DEC) \cite{xie2016unsupervised} and its improved version IDEC \cite{ijcai2017p243}, and we stipulate the intended aim of DEC, and also DC in general, that has yet to be achieved. The finding has a wide implication on DC.

We further show that the intended aim of DC can be achieved without deep learning using an approach we called Cluster-as-Distribution (CaD) Clustering. While there are recent clustering methods belonging to CaD Clustering, we are the first to show that this approach achieves the intended aim of DC now---this is achieved requiring neither deep learning nor a learned representation! This finding questions the focus of DC, i.e., the must-use deep learning in learning a latent representation. 

We summarize our contributions as follows:

\begin{enumerate}
    \item Uncover the harm the commonly used definition of clustering has inflicted on clustering methods created based on it---relying on a point-to-point similarity function to determine similar points in a cluster.
    \item Provide new definitions of clustering that can avoid such harm, i.e., by assuming that a cluster consists of  independent and identically distributed (i.i.d.) points sampled from a distribution. A distributional measure (instead of a point-to-point similarity function) shall be used to determine the members of a cluster according to this definition. 
    \item Reveal that deep clustering method DEC or IDEC has the same fundamental limitations of $k$-means clustering, i.e., its inability to discover clusters of  arbitrary shapes, varied data sizes and densities, despite having deep learning.
    \item Identify the main cause that leads to the fundamental limitations of DEC/IDEC, i.e., its latent representation learning does not produce a representation such that the clusters of arbitrary shapes, varied data sizes and densities in input space can be represented as centroids in the latent space. 
    \item Examine the hypothesis that the difficulty of many learned representation based clustering methods in achieving the intended aim of deep clustering is a result of ignoring the distributional information of clusters in a given dataset. Based on this hypothesis, we articulate a category of clustering approach, i.e., Cluster-as-Distribution (CaD) clustering, which can achieve the intended aim of deep clustering, without the difficulty faced by many current learned representation based clustering methods.
\end{enumerate}

\section{Preliminaries}
We consider a dataset $\mathcal{X} = \{x_1, \dots, x_n\}$, where each data point $x_i \in \mathbb{R}^d$ is drawn independently and identically distributed (i.i.d.) from an unknown distribution $\mathcal{D}$ over $\mathbb{R}^d$. Many classical clustering methods can be viewed as optimizing objectives that encourage high within-cluster similarity and low between-cluster similarity.

For example, \textbf{$k$-means} clustering objective seeks to partition the dataset into $K$ disjoint clusters by minimizing the within-cluster sum of squared distances to each cluster centroid. Formally, the objective is given as: 

\begin{equation}
\min_{\{C_k\}_{k=1}^K}
\sum_{k=1}^K \sum_{x_i \in C_k} \| x_i - \mu_k \|_2^2,
\end{equation}

where $\{C_k\}_{k=1}^K$ denotes a partition of $\mathcal{X}$, and $\mu_k = \frac{1}{|C_k|} \sum_{x \in C_k} x$ denotes the centroid of cluster $C_k$.

\textbf{Deep Embedded Clustering (DEC)} \cite{xie2016unsupervised} utilizes a stacked autoencoder (SAE) \cite{hinton2006reducing} to learn a non-linear mapping from the data space $\mathcal{X}$ to a lower-dimensional feature space $\mathcal{Z}$. The SAE consists of an encoder $f_\theta: \mathcal{X} \to \mathcal{Z}$ and a decoder $g_\phi: \mathcal{Z} \to \mathcal{X}$. In the pre-training phase, the network is trained by minimizing the reconstruction loss $\mathcal{L}_{r}$, which measures the discrepancy between the input $x_i$ and the reconstruction $\hat{x}_i = g_\phi(f_\theta(x_i))$:

\begin{equation}
\mathcal{L}_{r} = \frac{1}{n} \sum_{i=1}^n \| x_i - \hat{x}_i \|_2^2.
\end{equation}

After pre-training, DEC discards the decoder and fine-tunes the encoder by minimizing a clustering objective. This objective is defined as the Kullback-Leibler (KL) divergence \cite{kullback1997information} between a soft assignment distribution $Q$ and an auxiliary target distribution $P$:

\begin{equation}
\label{eq:clustering_loss}
\mathcal{L}_{c} = \mathrm{KL}(P \| Q) = \sum_{i=1}^n \sum_{j=1}^K p_{ij} \log \frac{p_{ij}}{q_{ij}}.
\end{equation}

Here, $q_{ij}$ represents the probability of assigning sample $i$ to cluster $j$, calculated using the Student's $t$-distribution \cite{student1908probable} kernel between the embedding $z_i$ and the cluster centroid $\mu_j$ and $p_{ij}$ serves as the target probability.






\textbf{Improved Deep Embedded Clustering (IDEC)} \cite{ijcai2017p243} argues that discarding the decoder in DEC may distort the embedded feature space and disregard the local structure of the data. To preserve the local structural information, IDEC retains the decoder and jointly optimizes the clustering loss and the reconstruction loss. The total objective function is formulated as:

\begin{equation}
\mathcal{L}_{total} = \mathcal{L}_{r} + \gamma \mathcal{L}_{c},
\end{equation}

where $\mathcal{L}_{c}$ is the clustering loss defined in Eq. (\ref{eq:clustering_loss}), $\mathcal{L}_{r}$ is the reconstruction loss, and $\gamma > 0$ is a coefficient that balances the trade-off between clustering optimization and feature preservation.

\textbf{Spectral Clustering} \cite{von2007tutorial} approaches the problem from a graph partitioning perspective. Given the dataset $\mathcal{X}$, SC constructs an undirected weighted graph $\mathcal{G} = (\mathcal{V}, \mathcal{E})$, where each vertex $v_i$ corresponds to a data point $x_i$. The edge weights are represented by an affinity matrix $\mathbf{W} \in \mathbb{R}^{n \times n}$, where $w_{ij} \ge 0$ measures the similarity between $x_i$ and $x_j$ (e.g., using a Gaussian kernel or $k$-nearest neighbors). The degree of a vertex $v_i$ is defined as $d_i = \sum_{j=1}^n w_{ij}$, and the degree matrix is denoted as $\mathbf{D} = \mathrm{diag}(d_1, \dots, d_n)$.

The goal is to partition the graph into $K$ disjoint sets $\{C_1, \dots, C_K\}$. A standard objective is to minimize the \textit{Cut}, defined as the sum of weights of edges connecting different partitions. However, simply minimizing the cut often leads to trivial solutions (e.g., separating a single outlier). 

To overcome this, \textbf{Normalized Cut (NCut)} \cite{ncut} normalizes the cut by the volume of each partition. 






\textbf{Kernel Bounded Clustering (KBC)} \cite{zhang2025kbc} re-examines the spectral objective through a simple yet fundamental mathematical transformation, bridging the gap between graph partitioning and clustering via a distribution kernel. 

Recall that the total weight of the graph, denoted as $W(\mathcal{X}, \mathcal{X}) = \sum_{i=1}^n \sum_{j=1}^n w_{ij}$, is a constant for a given dataset. Consequently, minimizing the cut between clusters (inter-cluster dissimilarity) is mathematically equivalent to maximizing the association within clusters (intra-cluster similarity). This relationship can be formally expressed as:

\begin{equation}
\setlength{\abovedisplayskip}{3pt}
\setlength{\belowdisplayskip}{3pt}
\min \sum\nolimits_{k=1}^K \mathrm{cut}(C_k, \bar{C}_k) \iff \max \sum\nolimits_{k=1}^K W(C_k, C_k),
\end{equation}

where $W(C_k, C_k) = \sum_{i \in C_k} \sum_{j \in C_k} w_{ij}$ is the self-similarity of cluster $C_k$. 

Therefore, the  discrete graph cut problem is transformed from minimizing the sum of all cuts between each pair of a cluster and the rest of the graph to the equivalent objective of maximizing the self-similarity of every cluster.

Based on this equivalence and $W$ can be expressed in  terms of a distributional kernel $\mathcal{K}$ \cite{scholkopf2002learning}, i.e., $W(X,Y) = |X| \times |Y|  \times \mathcal{K}(P_X, P_Y)$, KBC introduces a \textit{distributional approach} to clustering \cite{zhang2025kbc} that enables the objective of SC to be achieved without eigen-decomposition.

\section{Limitations of Current Definitions}

\begin{figure*}[t]
    \centering
    \includegraphics[width=\linewidth]{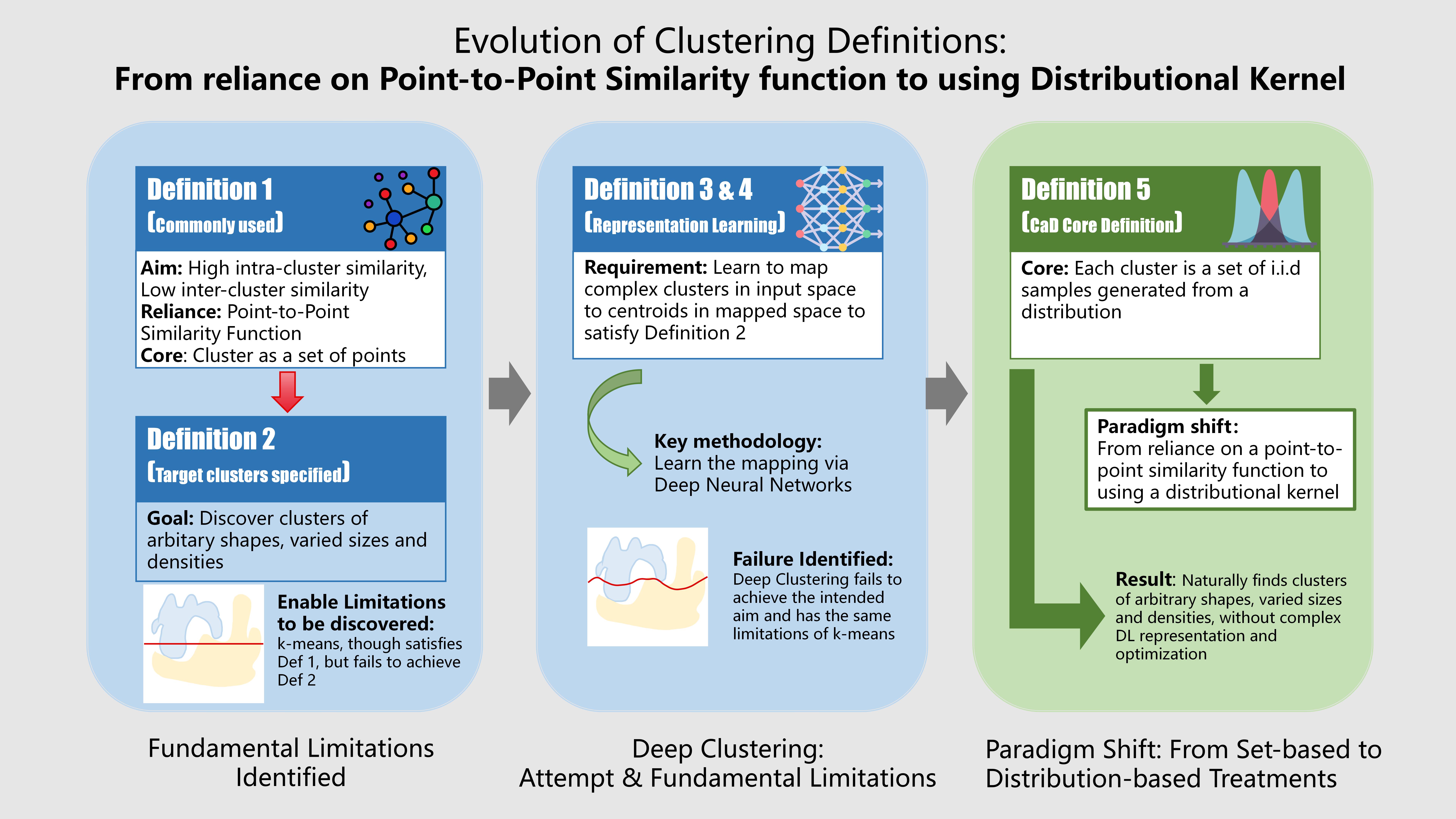}
    \caption{An illustration of the evolution from Definition \ref{def-typical} to Definition \ref{def-distribution-clustering}.}
    \label{fig:definition_shift}
\end{figure*}

\subsection{Current Definitions of Clustering}
\label{sec:definition_of_clustering}
Current definitions \cite{hartigan1975clustering, jain1999dataclustering, aggarwal2014dataclustering, aggarwal2015datamining, han2022data, lu2024survey} often do not specify the kind of clusters required for the clustering task. An example (paraphrased) definition\footnote{Examples of existing commonly used definitions are provided in the Appendix.} is given below:
\begin{definition}
    Given a dataset of data points in $\mathbb{R}^d$ without cluster labels, the aim of clustering is to discover clusters that have high similarity within each cluster, but low similarity between any two different clusters.
    \label{def-typical}
\end{definition}

It implies a reliance of a point-to-point similarity/distance function, and even Deep Clustering methods have used a similar definition i.e., `Clustering aims at grouping data instances into several clusters, where instances from the same cluster share similar semantics and instances from different clusters are dissimilar.' \cite{lu2024survey}.

The above definition is aimed to encompass all clustering methods, regardless of the kind of clusters a method can discover. 

A better definition is one which stipulates the kind of clusters a clustering method is expected to discover, that is: 
\begin{definition}
    Given a dataset of data points in $\mathbb{R}^d$ without cluster labels, the aim of clustering is to discover clusters of arbitrary shapes, varied sizes and densities in the dataset (knowing that the relation between data points and clusters is as stated in Definition \ref{def-typical}).
    \label{def-clustering-aim}
\end{definition}

Knowing the kind of clusters that $k$-means clustering can discover, it is clear that it is  unable to achieve the aim of clustering in Definition~\ref{def-clustering-aim}. Yet, $k$-means clustering, or any clustering method, achieves the aim stated in Definition~\ref{def-typical}! To facilitate an intuitive understanding of Definitions \ref{def-typical} and \ref{def-clustering-aim}, as well as the subsequent clustering definitions, we present a schematic illustration in Figure \ref{fig:definition_shift}. As depicted in the left panel, the $k$-means clustering successfully satisfies Definition \ref{def-typical}. Yet, Definition~\ref{def-clustering-aim} enables its fundamental limitations to be exposed.

A worse consequence of Definition \ref{def-typical} is that the clustering methods created based on it have been led to rely on a \emph{point-to-point similarity function}---due to the need to ensure that points in a cluster are similar to each other. We will see later that why this becomes the underlying reason why clustering methods created with this reliance have fundamental limitations.

This paper intends to answer the following questions:
\begin{enumerate}
    \item Has DEC or its improved version IDEC overcome the fundamental limitations of $k$-means clustering to such an extent that it  is able to achieve the aim of clustering stated in Definition \ref{def-clustering-aim}? If not, what is the cause that has led to this outcome?
    \item Do other deep clustering methods have  similar fundamental limitations as DEC?
    \item Is it possible to avoid the trap of many clustering methods, including deep clustering, that were constrained by the age-old Definition \ref{def-typical} of clustering?
    \item What are the characteristics of a clustering method which has avoided the trap? 
    \item Is there any evidence that deep clustering performs better in high dimensional datasets?
\end{enumerate}

Specifically, We define the fundamental limitation of a clustering method as the types of clusters it fails to discover. For example, $k$-means clustering can only discover clusters of spherical shapes that are well-separated; in addition, all clusters must have approximately the same  number of points and the same density \cite{ahmed2020k}. A smaller gap between clusters makes the clustering problem harder. Example clusters, which deviate from the $k$-means-discoverable clusters, and the clusters identified by $k$-means clustering, are shown in the first and second columns in Table~\ref{tab:clustering_results_examples}.

Using Deep Embedded Clustering (DEC) \cite{xie2016unsupervised} and IDEC \cite{ijcai2017p243} as the starting point, we show that they have fundamental limitations similar to those of $k$-means clustering empirically, and articulate the condition under which DEC must satisfy in order to claim that it does not have the fundamental limitations. Then, we argue that other deep clustering methods need to have the same condition.

\begin{table*}[htbp]
\centering
\caption{Clustering results comparison. Each reported Normalized Mutual Information (NMI) \cite{strehl2002nmi} value is averaged over 10 runs; the clustering visualization result is obtained from the run having the highest NMI. The fundamental limitations are arbitrary shapes, varied sizes and varied densities, shown in the 2Crescents, Diff-Sizes and AC datasets, respectively.}
\label{tab:clustering_results_examples}
\setlength{\tabcolsep}{1pt}
\renewcommand{\arraystretch}{0.5}
\resizebox{\textwidth}{!}{
\begin{tabular}{c c c c c c c}
    \toprule
    & \small \textbf{Original Data} & \small \textbf{$k$-means} & \small \textbf{IDEC} & \small \textbf{CC} & \small \textbf{SpectralNet} & \small \textbf{KBC} \\
    \midrule
    
    \rotatebox{90}{\footnotesize 2Crescents} & 
    \includegraphics[width=0.16\textwidth]{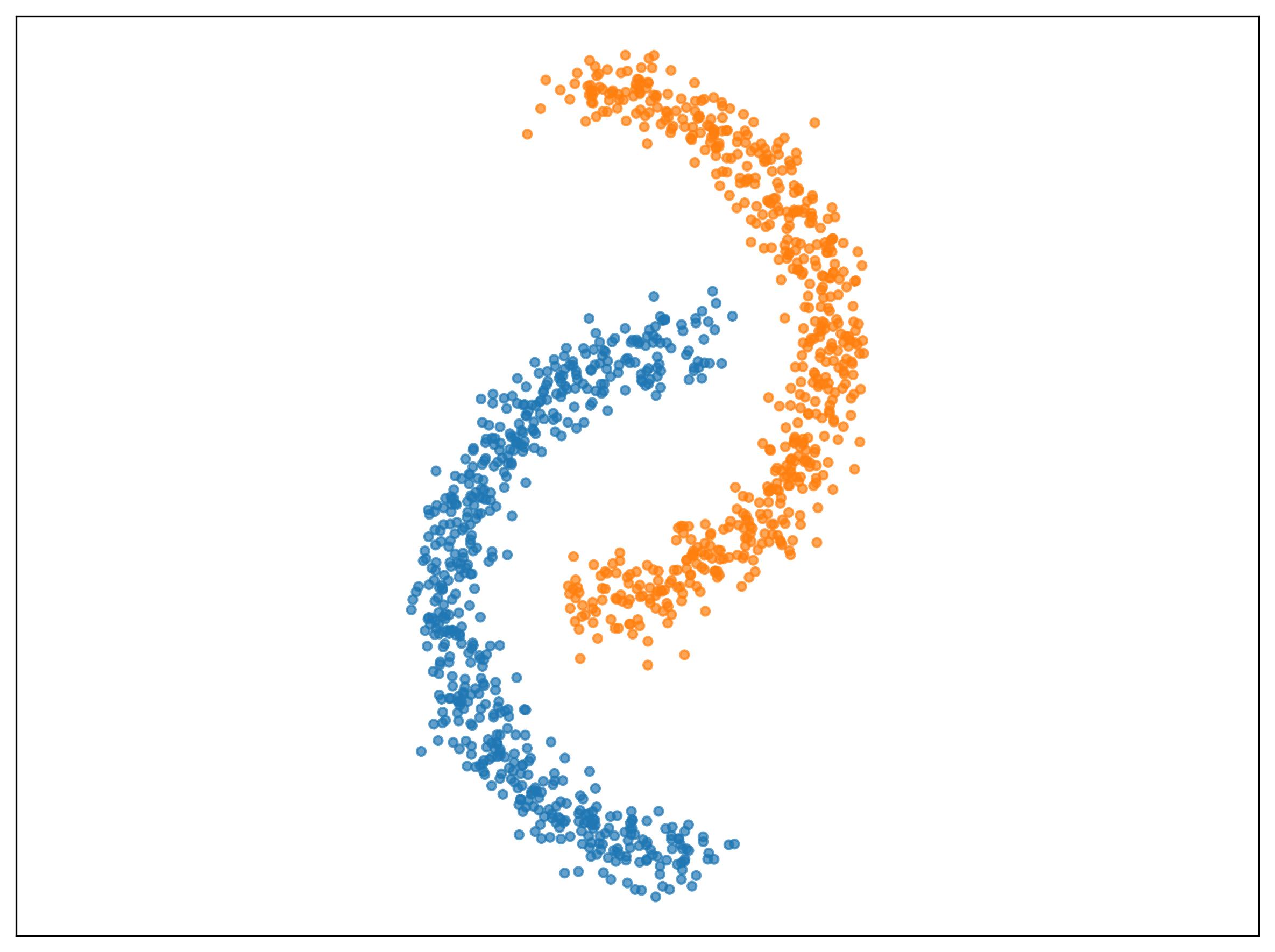} &
    \includegraphics[width=0.16\textwidth]{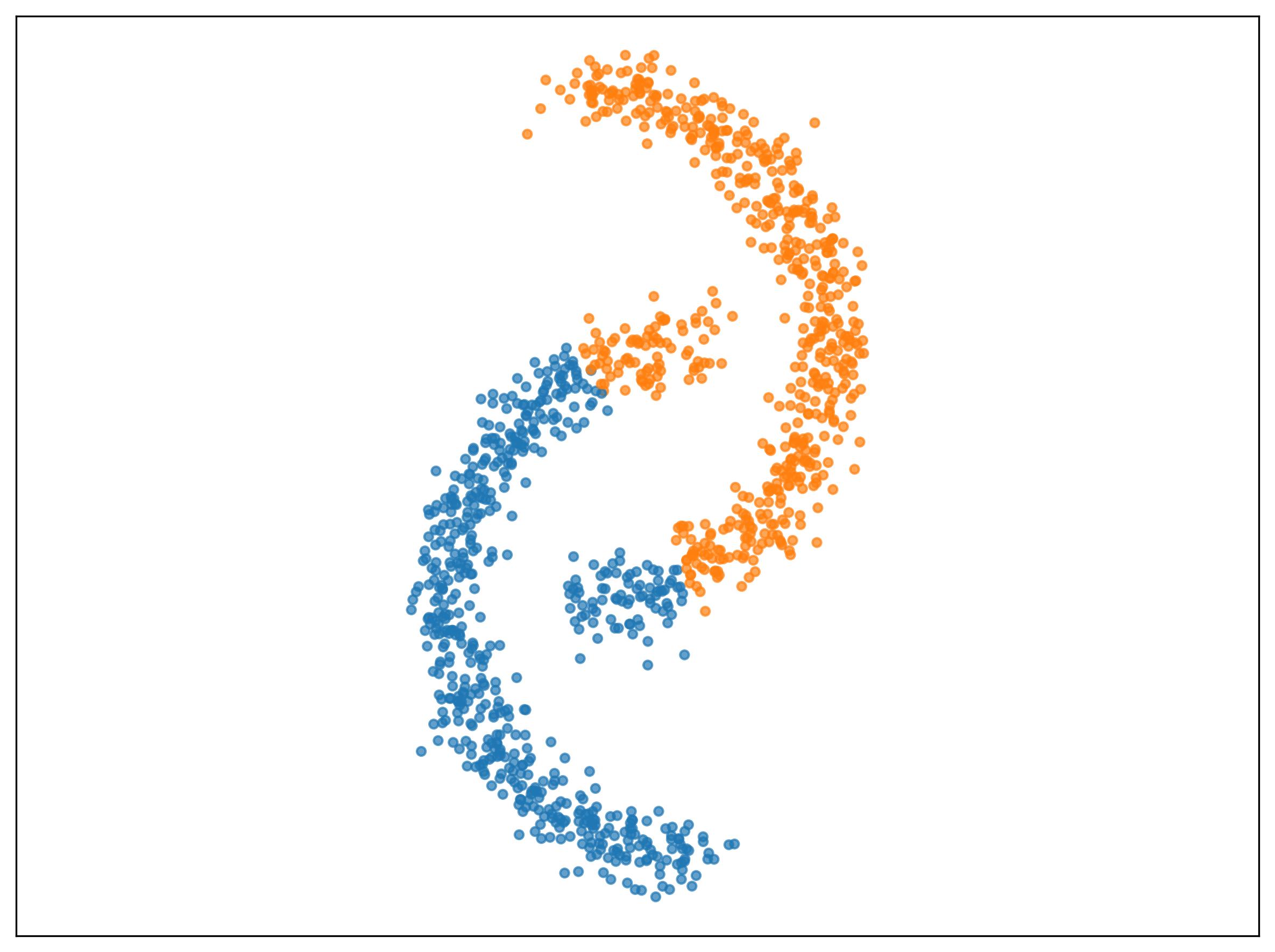} &
    \includegraphics[width=0.16\textwidth]{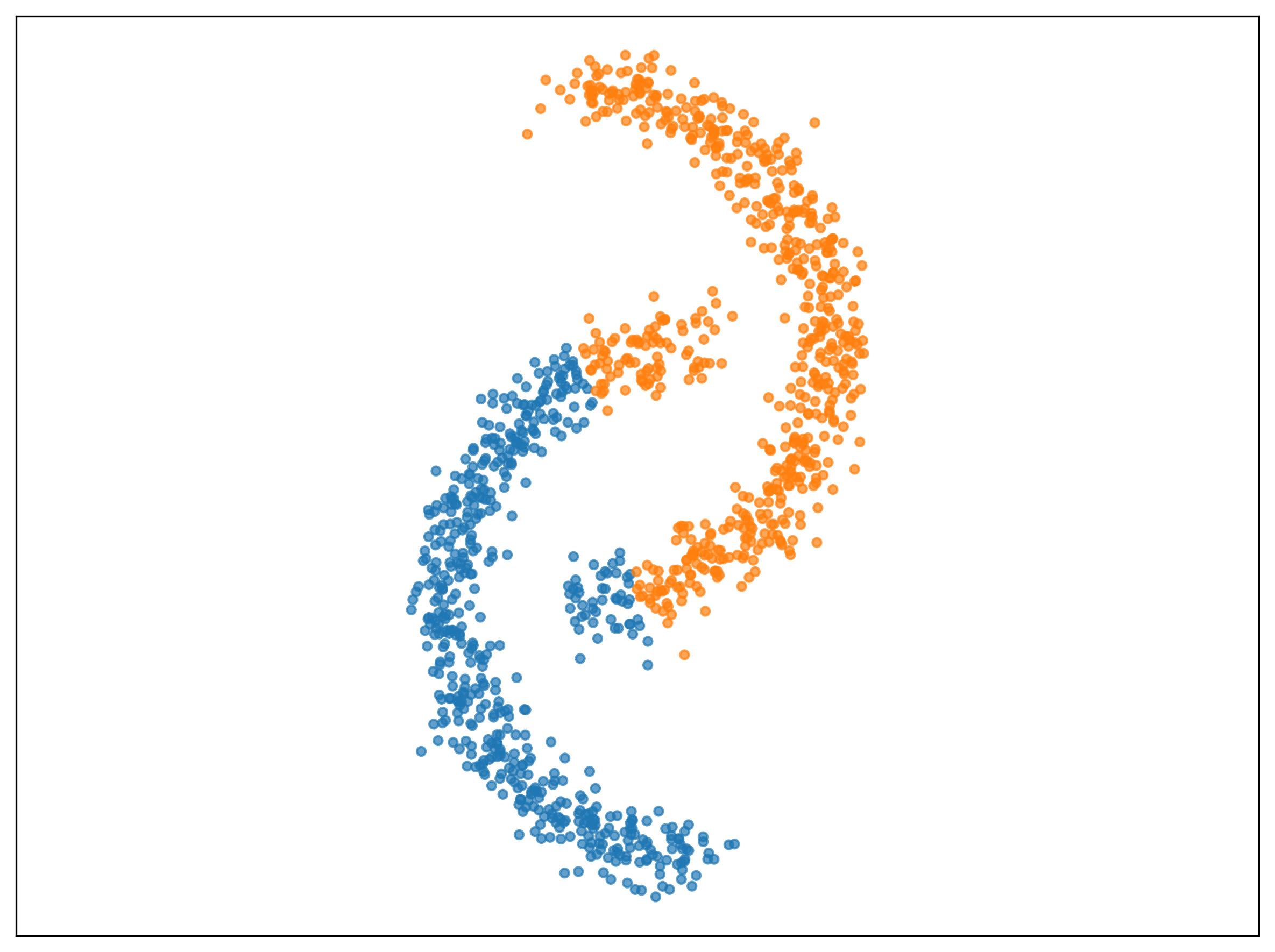} &
    \includegraphics[width=0.16\textwidth]{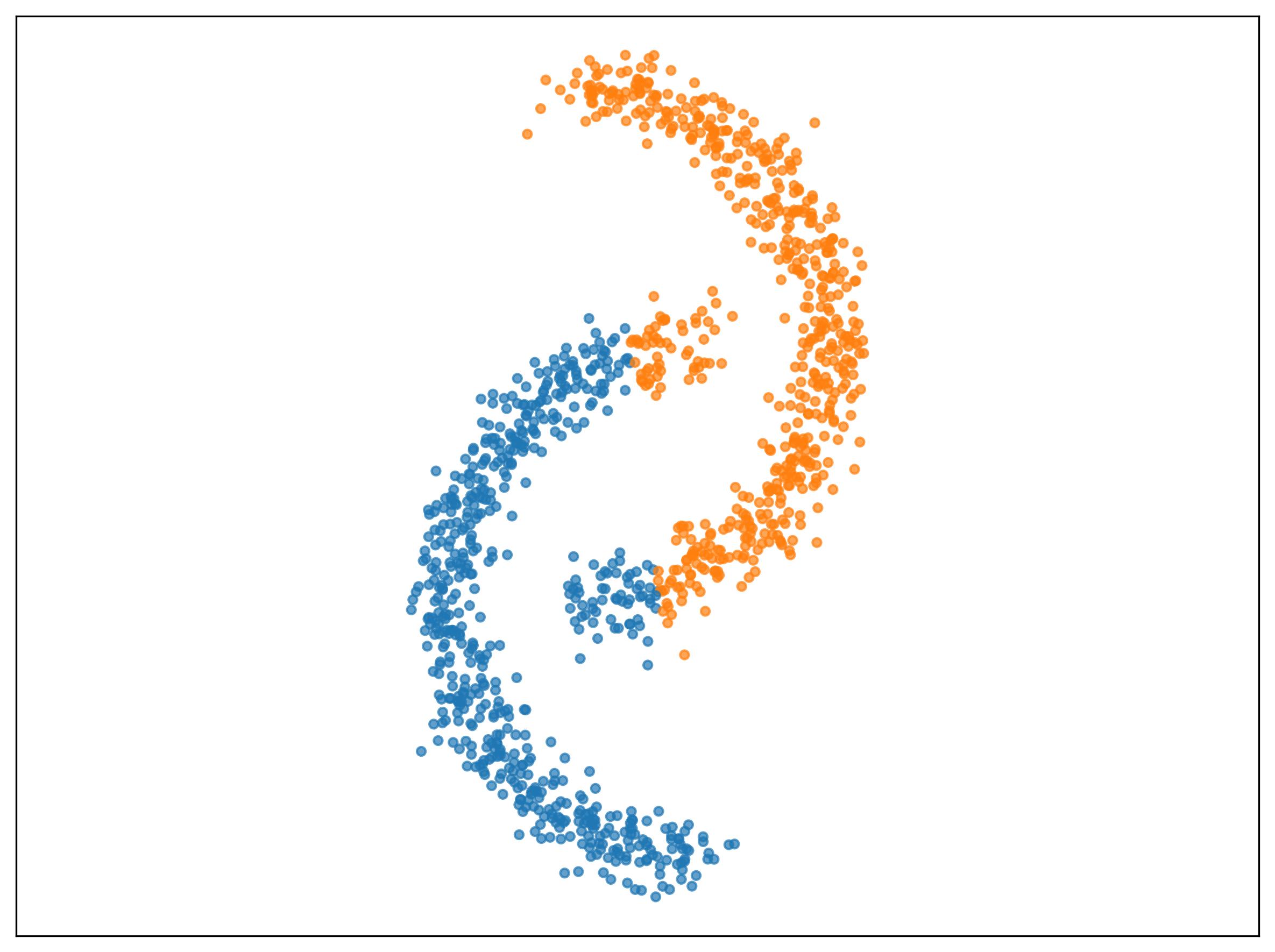} &
    \includegraphics[width=0.16\textwidth]{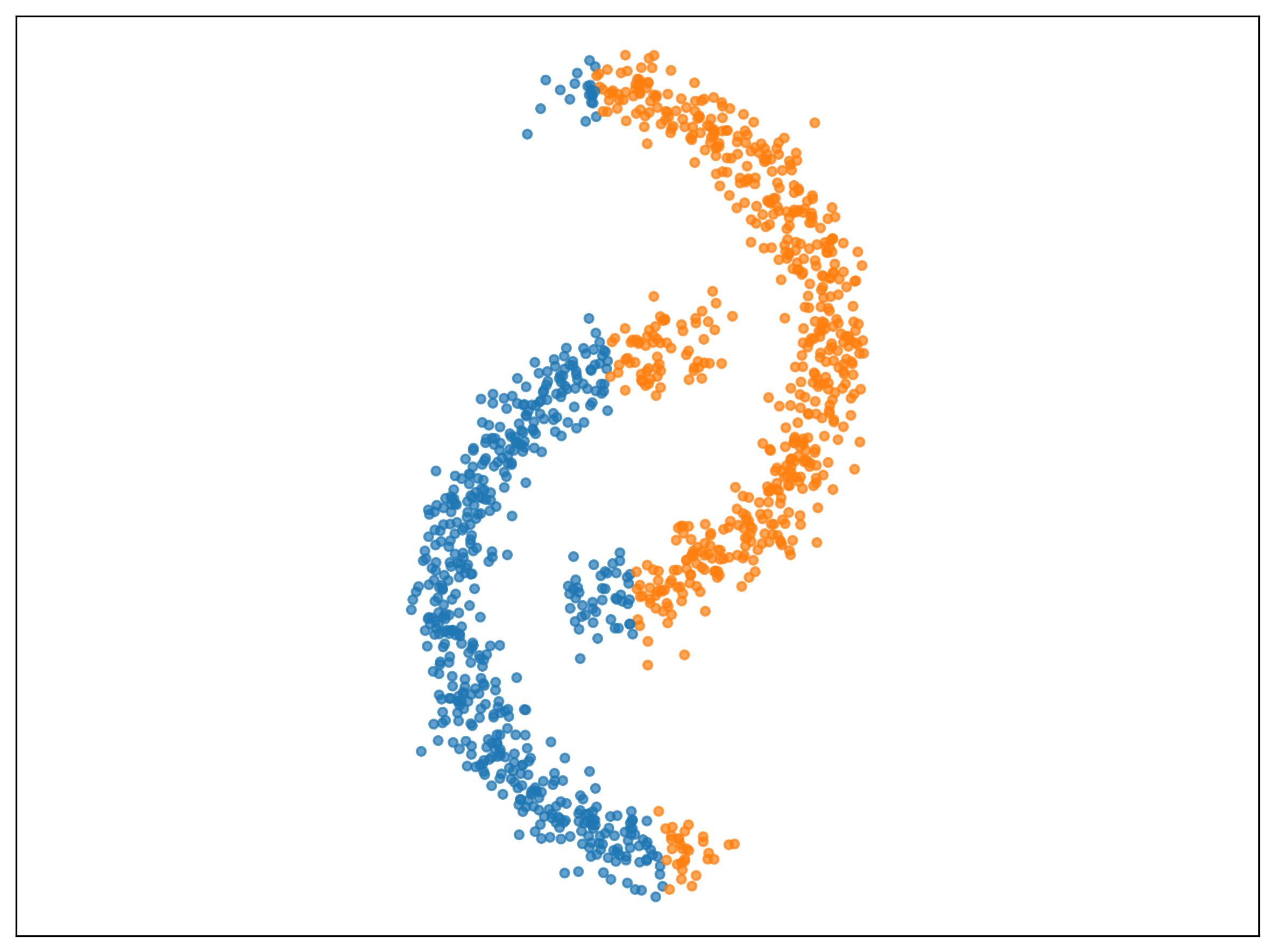} & 
    \includegraphics[width=0.16\textwidth]{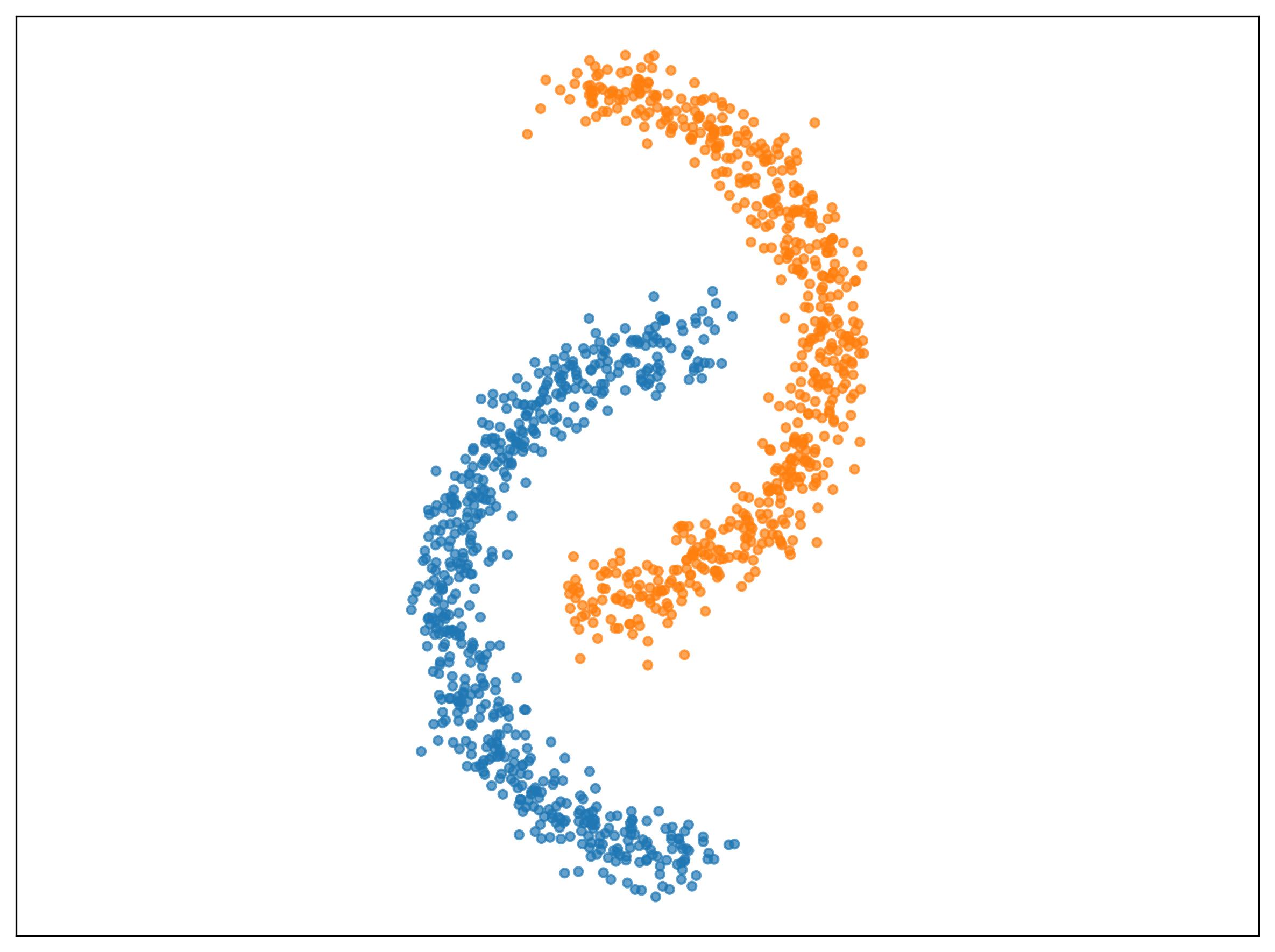} \\
    & & \footnotesize NMI=0.42 & \footnotesize NMI=0.49 & \footnotesize NMI=0.53 & \footnotesize NMI=0.42 & \footnotesize NMI=1.00 \\
    \addlinespace[4pt]
    
    \rotatebox{90}{\footnotesize Diff-Sizes} & 
    \includegraphics[width=0.16\textwidth]{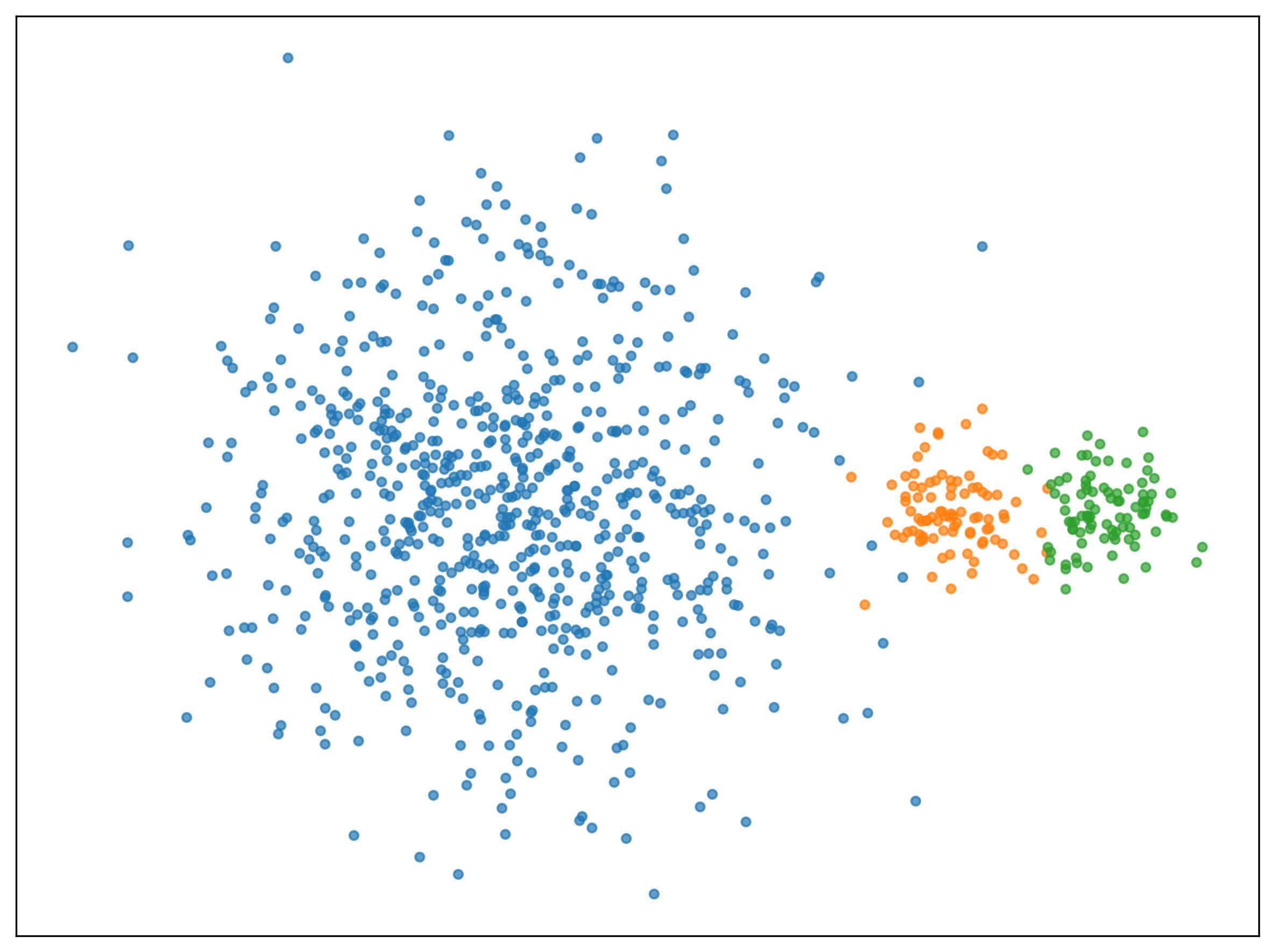} &
    \includegraphics[width=0.16\textwidth]{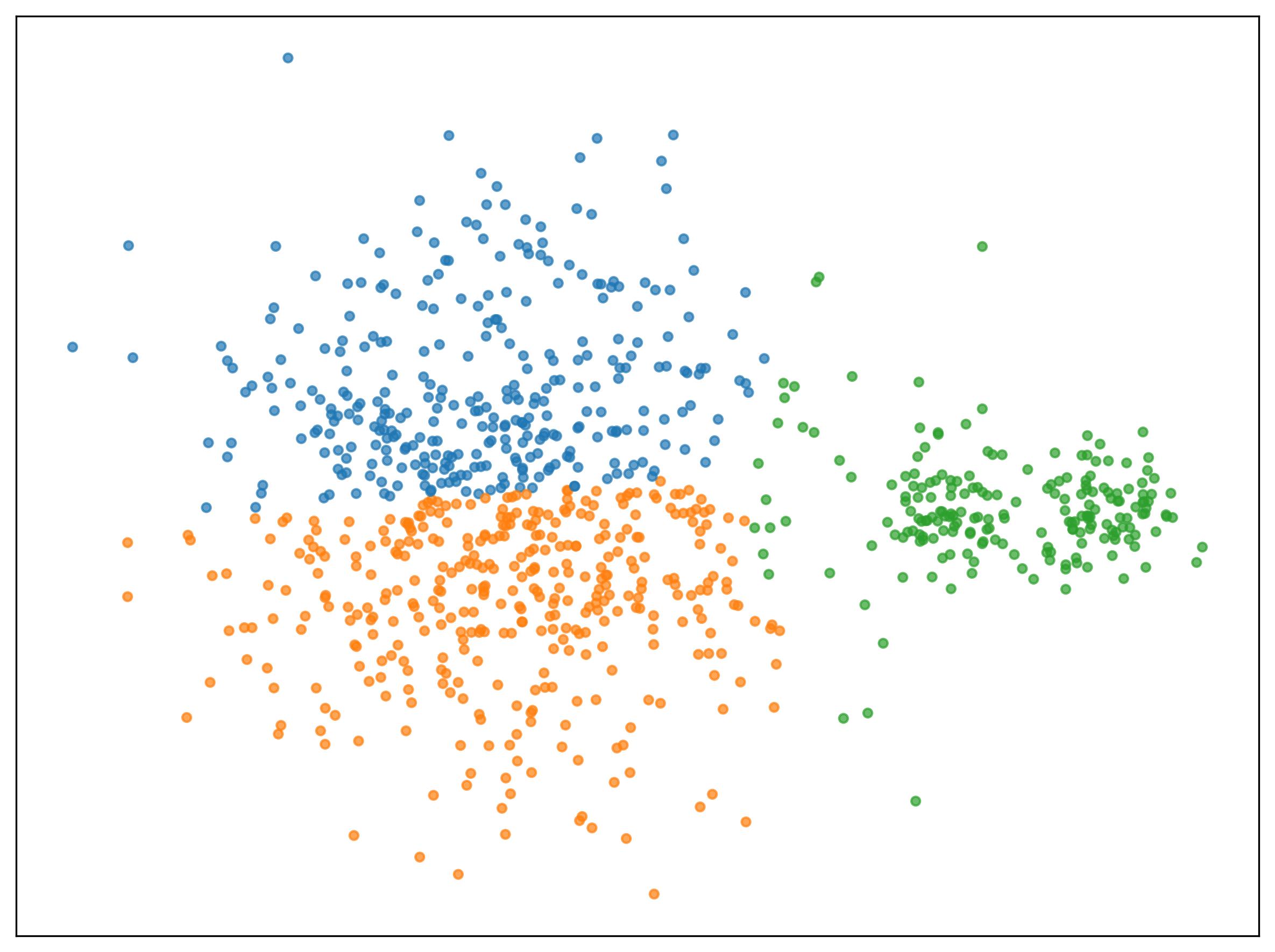} &
    \includegraphics[width=0.16\textwidth]{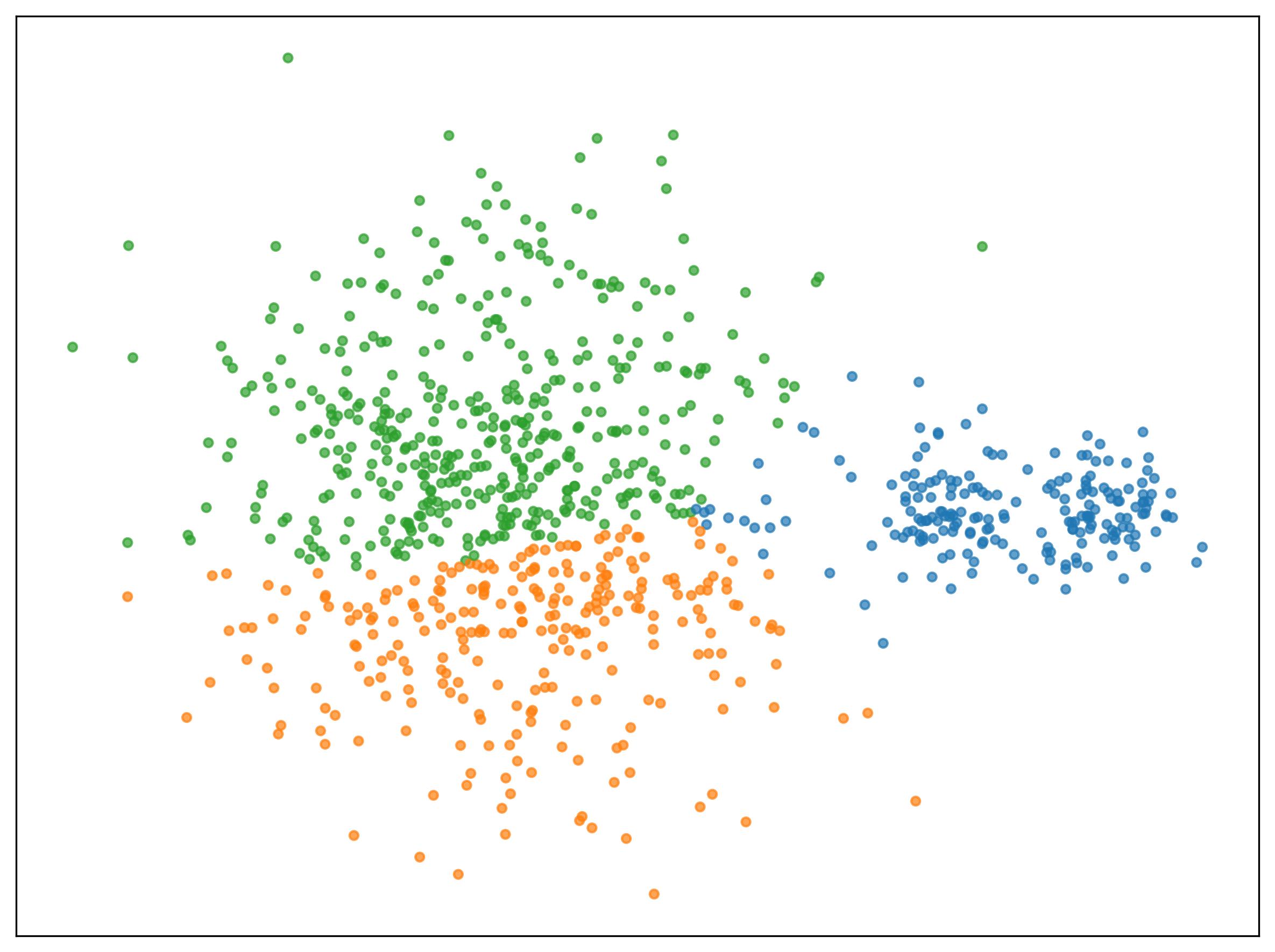} &
    \includegraphics[width=0.16\textwidth]{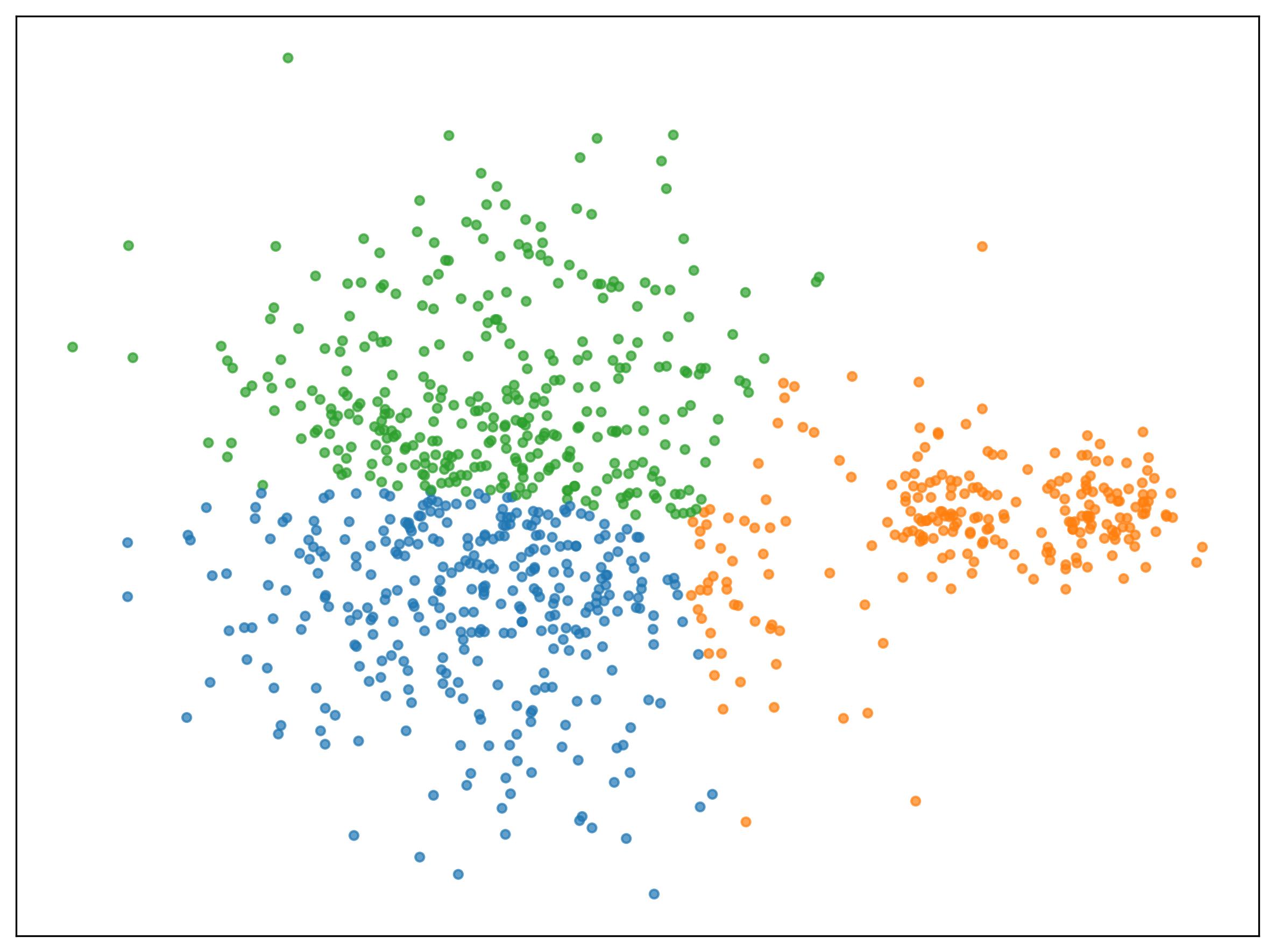} &
    \includegraphics[width=0.16\textwidth]{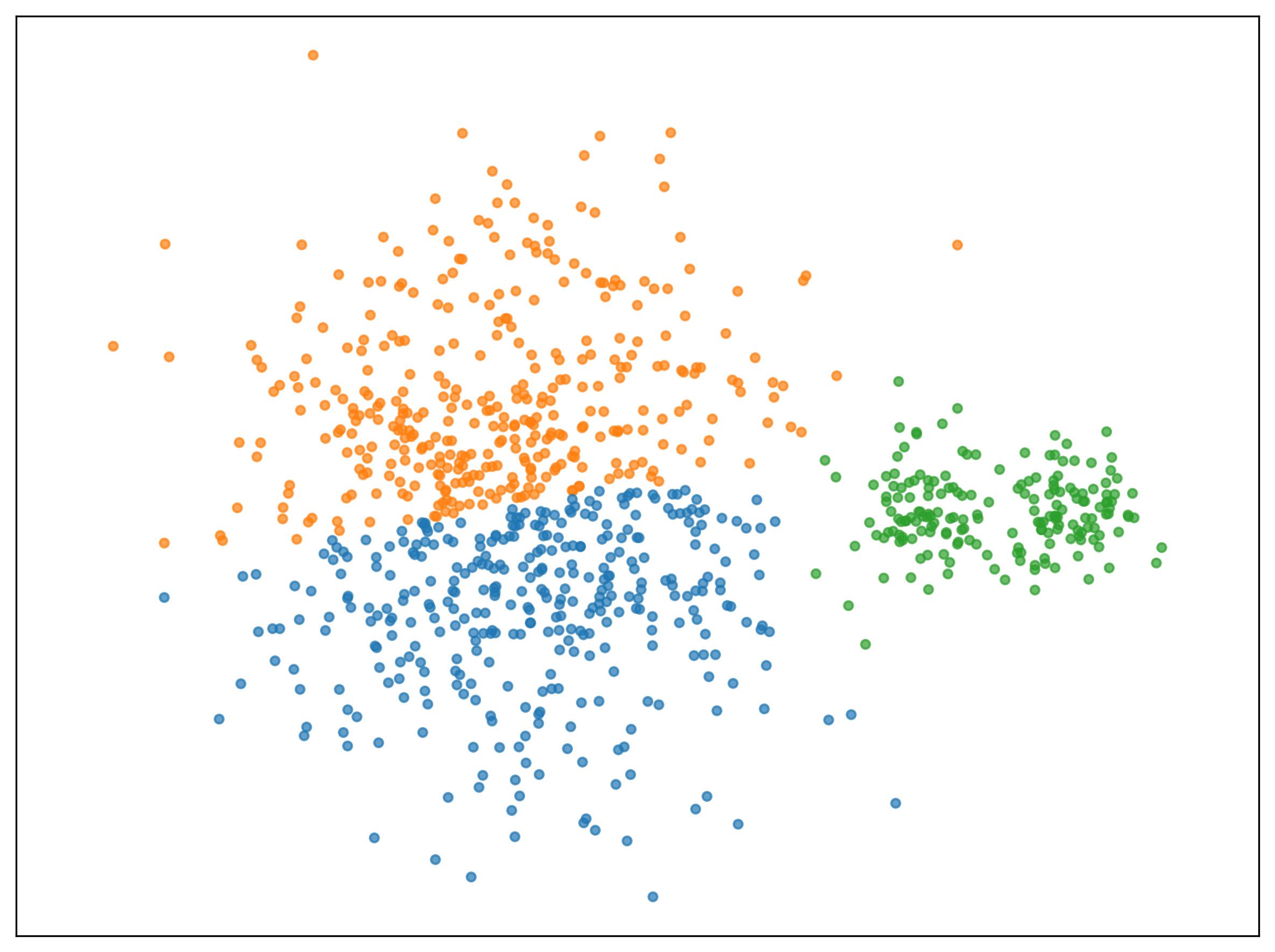} & 
    \includegraphics[width=0.16\textwidth]{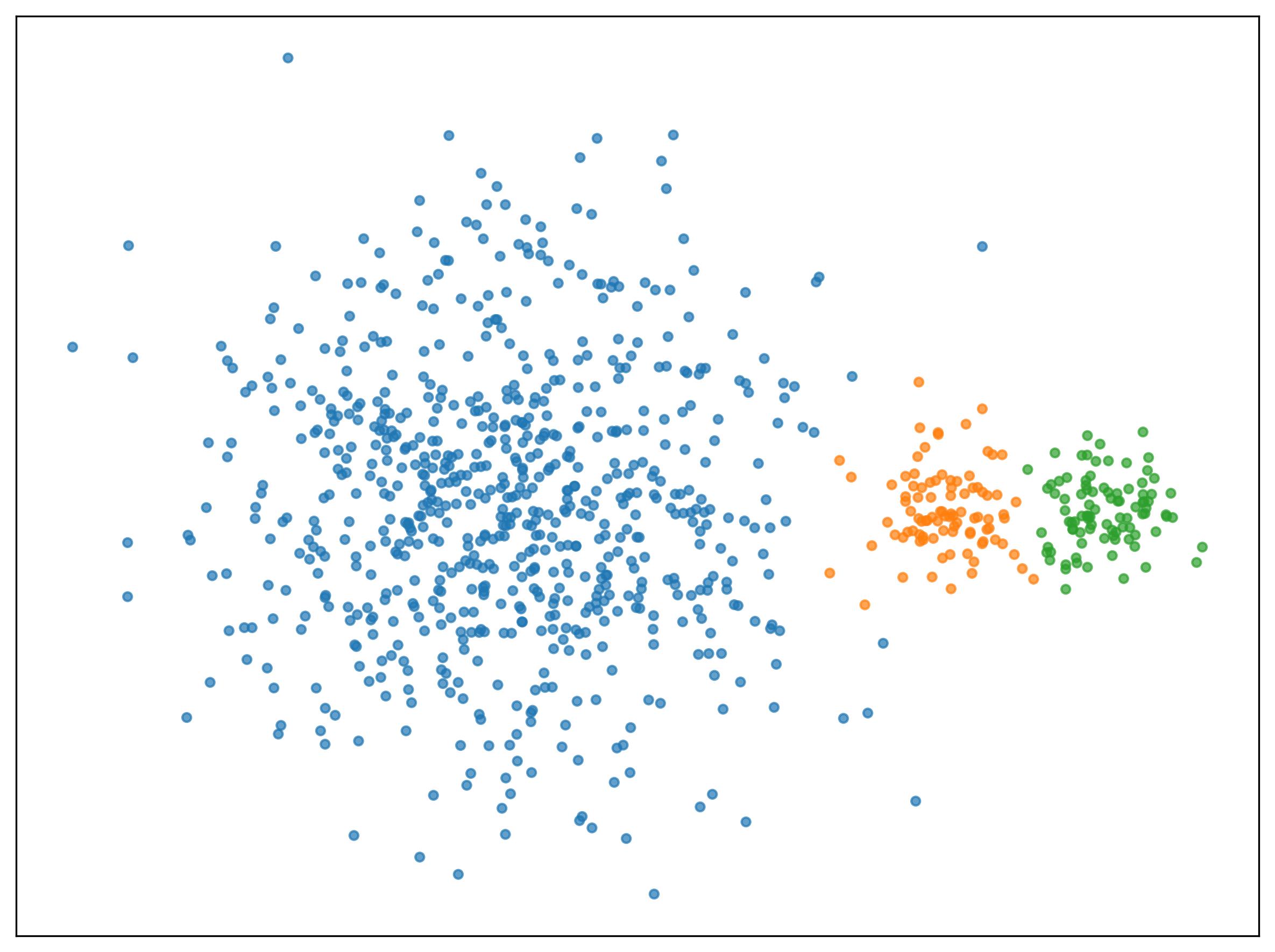} \\
    & & \footnotesize NMI=0.48 & \footnotesize NMI=0.52 & \footnotesize NMI=0.41 & \footnotesize NMI=0.55 & \footnotesize NMI=0.92 \\
    \addlinespace[4pt]

    \rotatebox{90}{\footnotesize AC} & 
    \includegraphics[width=0.16\textwidth]{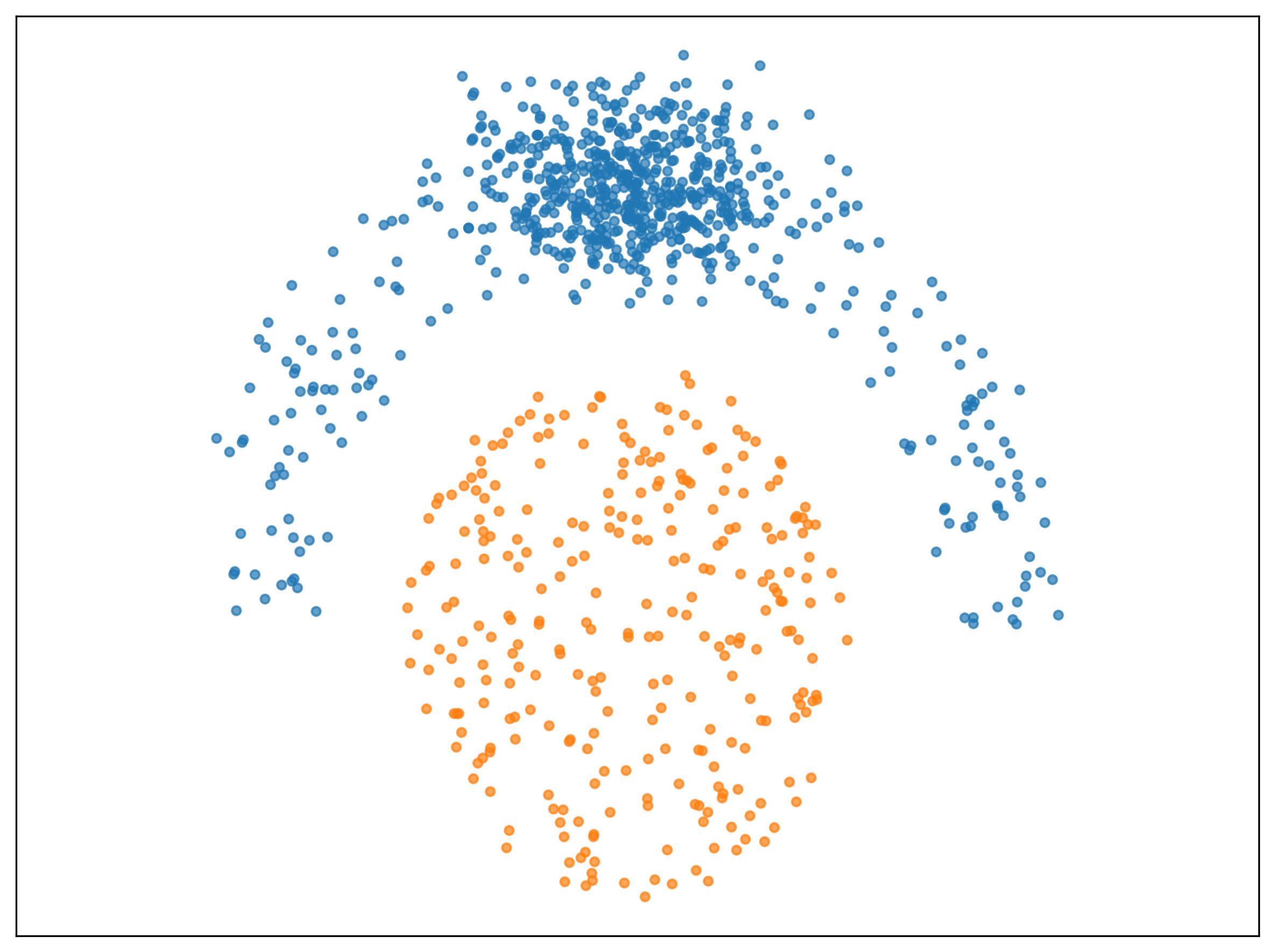} &
    \includegraphics[width=0.16\textwidth]{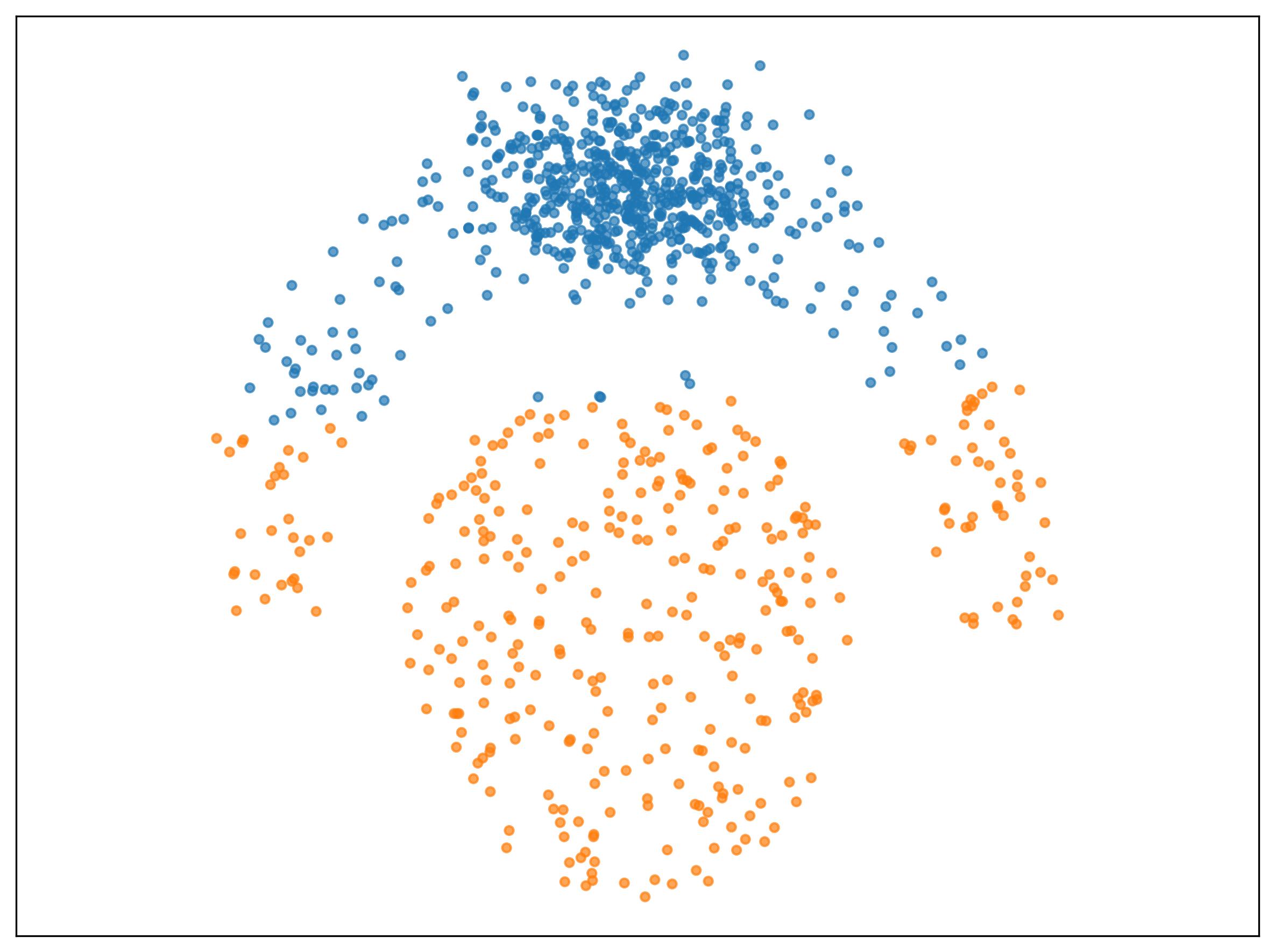} &
    \includegraphics[width=0.16\textwidth]{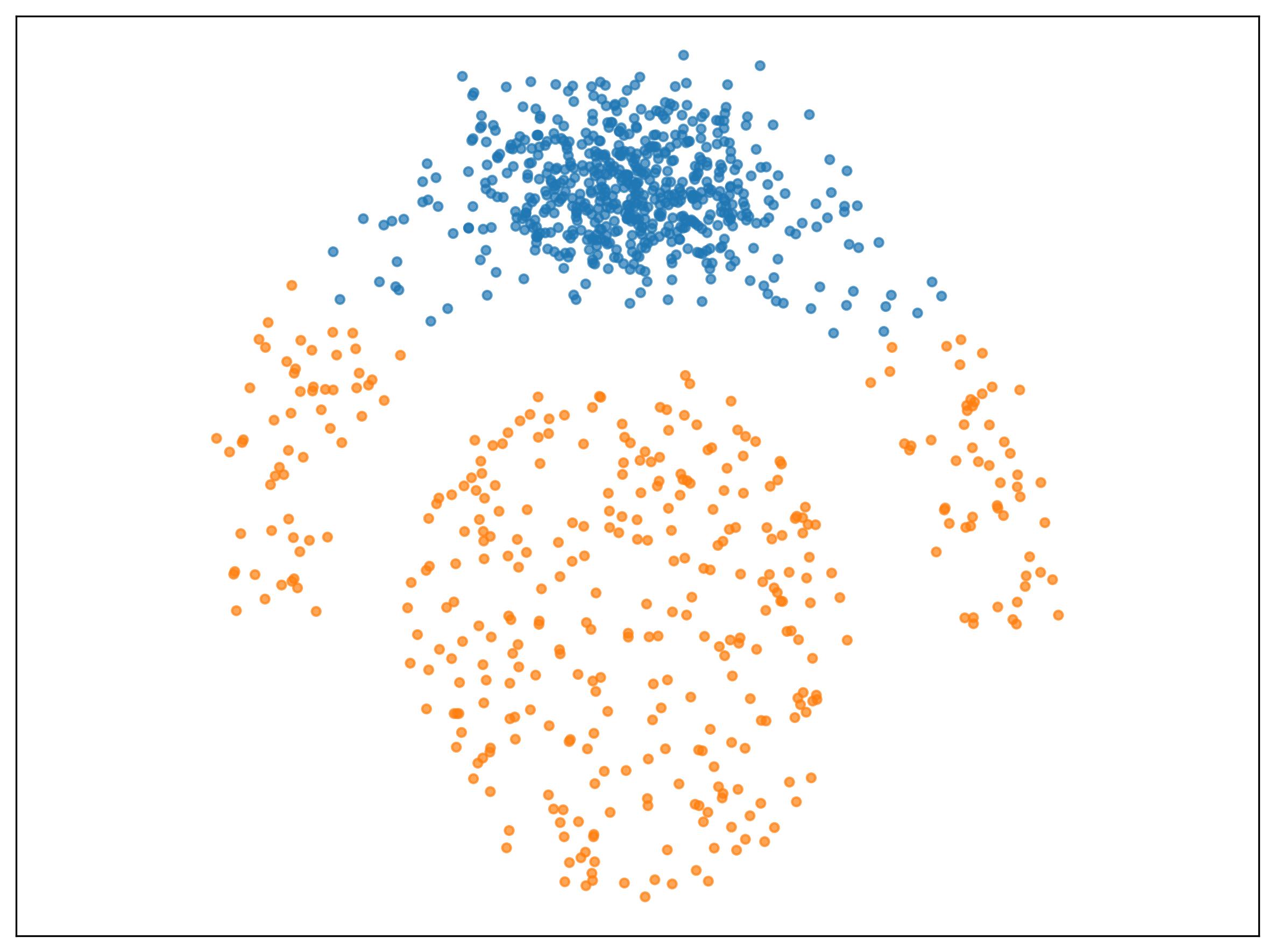} &
    \includegraphics[width=0.16\textwidth]{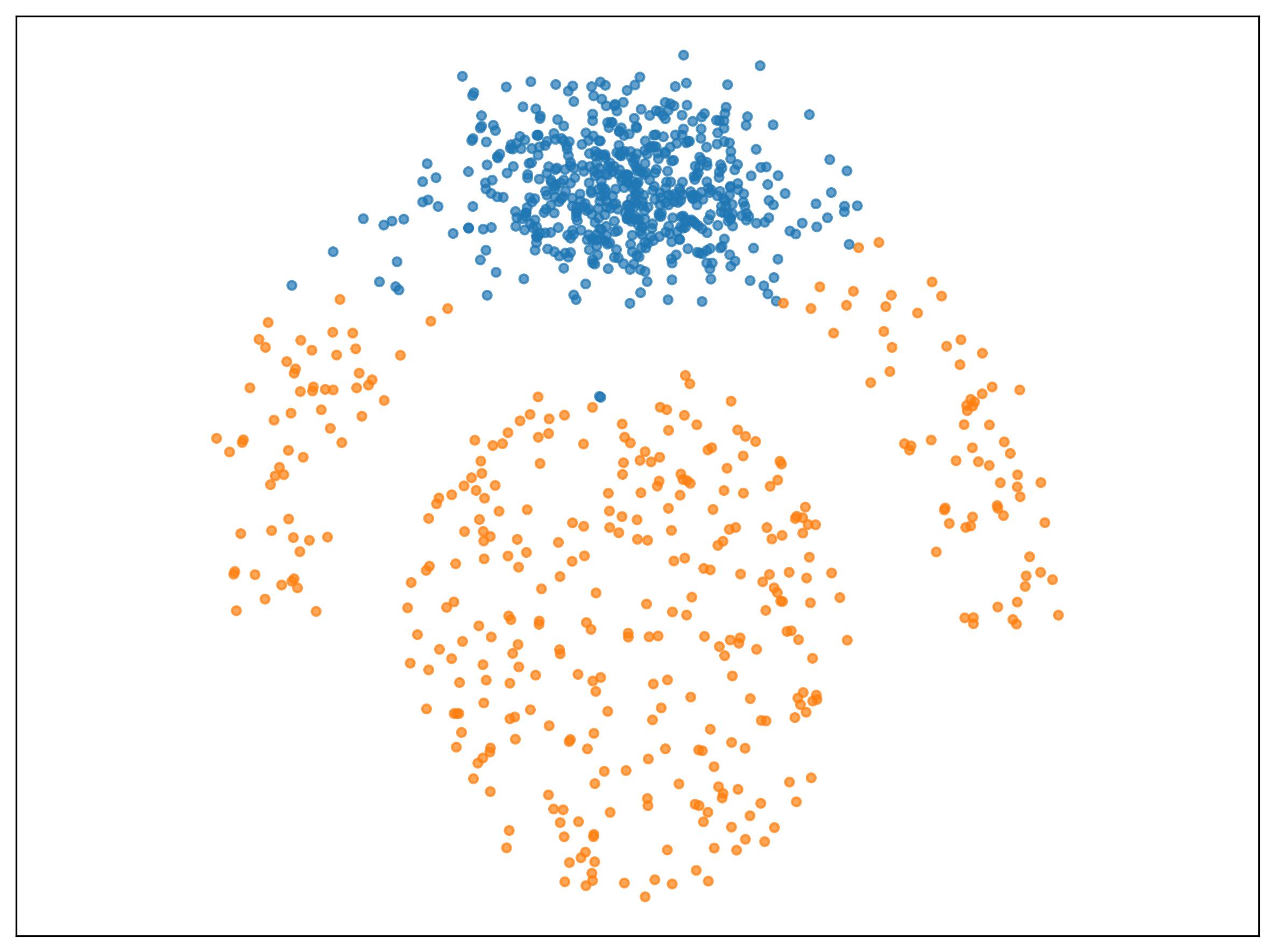} &
    \includegraphics[width=0.16\textwidth]{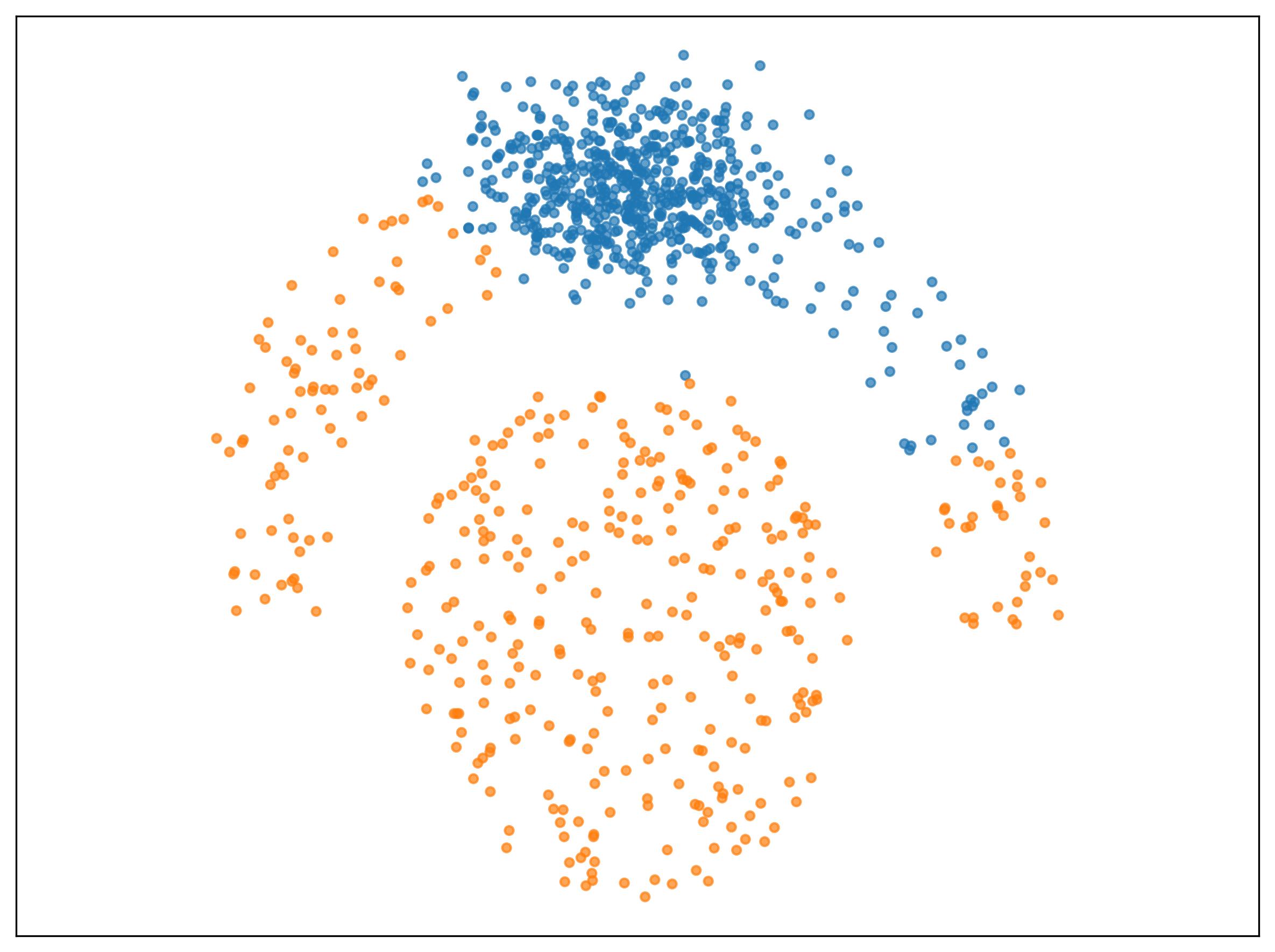} &
    \includegraphics[width=0.16\textwidth]{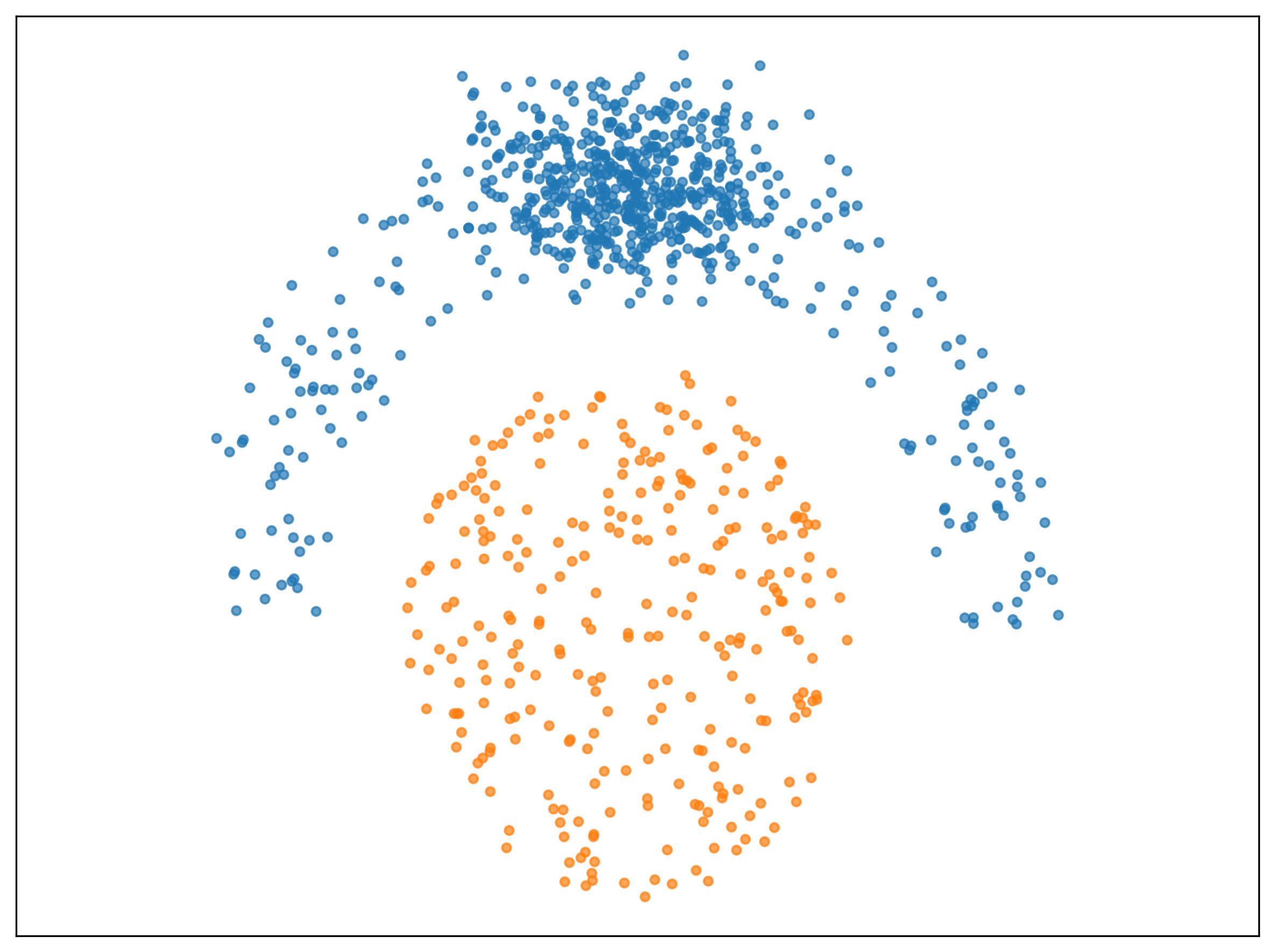} \\
    & & \footnotesize NMI=0.61 & \footnotesize NMI=0.56 & \footnotesize NMI=0.51 & \footnotesize NMI=0.56 & \footnotesize NMI=1.00 \\
    
    \bottomrule
\end{tabular}}
\end{table*}

\subsection{Fundamental Limitations of DEC and IDEC}

Deep Embedded Clustering (DEC) \cite{xie2016unsupervised} is an influential deep clustering method. It is the first method to introduce the Kullback–Leibler (KL) divergence \cite{kullback1997information} into unsupervised deep clustering which has been employed in later developments of deep clustering. Like $k$-means clustering, DEC and its improved version IDEC \cite{ijcai2017p243} employ a centroid  to represent a cluster. Unlike $k$-means clustering, the clusters discovered by DEC/IDEC cannot be defined because of its learned latent representation. This is the key reason why DEC, so as many deep clustering methods, is hard to analyze.

We first establish whether DEC/IDEC has the same fundamental limitations of $k$-means clustering, and then provide the reason why this is so.

This requires to fine-tune Definition \ref{def-clustering-aim} to make it explicit the role the learned latent representation plays to yield centroid-based clusters to represent realistic clusters in a dataset which also illustrated in Figure \ref{fig:definition_shift}.

\begin{definition}
    To achieve the aim in Definition \ref{def-clustering-aim}, the learned latent representation of DEC must transform the clusters of arbitrary shapes, varied sizes and densities in input space into ones that can be represented by centroids in the latent space.
    \label{def-requirements}
\end{definition}

While Definition \ref{def-requirements} clarifies the necessary properties of an ideal latent representation, it remains an open question whether ae-based frameworks like DEC and IDEC can theoretically satisfy these criteria. Existing literature offers little insight into this fundamental capability, despite the widespread adoption of these methods to cluster images.

To empirically investigate this, we conduct a comparative analysis between IDEC and $k$-means on a series of synthetic datasets designed to expose the known limitations of $k$-means. The clustering results are summarized in Table \ref{tab:clustering_results_examples}, with extended examples and experimental results provided in Appendix~\ref{sec:more_experiments} (Table \ref{tab:clustering_results_more_examples}).

These results show that IDEC shares many similar limitations of $k$-means. Specifically, IDEC has difficulty identifying clusters of \textbf{arbitrary shapes, varied sizes and densities}. Furthermore, by visualizing the learned latent representations in Table \ref{tab:idec_hidden_vis}, we observe that the distribution of the embedded space does not allow IDEC to achieve the aim stated in Definition \ref{def-requirements}.

\subsection{Limitations of Representation-Based Clustering}

Our finding on DEC/IDEC has a wider implication. It reveals the need to have a clear definition of clustering so that one can judge whether a method employed can achieve the intended aim. The lack of a clear definition, specifically about the requirement of the problem, makes the assessment of any proposed method difficult. We have identified the requirements for the learned representation in Definition \ref{def-requirements} (i.e., the role it must play) for a clustering method that relies on centroid-based clusters.

A variant of Definition \ref{def-requirements} is as follows:

\begin{definition}
    To achieve the clustering aim as stated in Definition \ref{def-clustering-aim}, a method that conflates the representation learning and clustering \textbf{must transform all data points in each cluster (of arbitrary shape, having data size and density differ from those in other clusters) in input space into the same cluster in the mapped space}.
    \label{def-requirements-v2}
\end{definition}

For a clustering method relying on learned representation as stated in the last two definitions, the central question remains, i.e., 
\begin{quote}
    Does the representation learning enable clusters of arbitrary shapes, varied sizes and densities in input space to be mapped into an intended form in the mapped space? 
\end{quote}

Table \ref{tab:clustering_results_examples} shows that neither DEC/IDEC nor Contrastive Clustering (CC) \cite{li2021contrastive, wang2022chaos} can achieve the aim stated in either Definition \ref{def-requirements} or Definition~\ref{def-requirements-v2}. More analyses can be found in Appendix \ref{sec:limitations_other_methods}.

\section{Cluster as Distribution: A New Definition}

\subsection{Rethinking Clustering Paradigms}

Note that all clustering methods discussed above, including deep clustering, rely on a point-to-point similarity/distance function.
Instead of defining clusters based on a point-to-point similarity function, we have a better ready tool to define clusters as distributions as follows:

`\textbf{Data points in a cluster $C$ are independent and identically distributed (i.i.d.) samples generated from a distribution $P_C$. Different clusters are a result of different distributions.}'

This cluster definition is better than the earlier ones (paraphrased in Definition~\ref{def-typical}) because (a) it does not demand a point-to-point similarity measure; and (b)  the `cluster as a distribution' definition provides the means to perform clustering using a distributional kernel, without a learned representation.

Having the above cluster definition, the clustering task can be re-expressed as follows:
\begin{definition}
    Given a dataset of data points in $\mathbb{R}^d$ without cluster labels, the aim of clustering is to discover each cluster as generated from a distribution, where the clusters may have arbitrary shapes, varied sizes and densities.
    \label{def-distribution-clustering}
\end{definition}

Note three key points in this definition:
\begin{itemize}
    \item Cluster-as-distribution is emphasized, rather than the kind of clusters as in Definition \ref{def-clustering-aim}. 
    
    \item None of the previous definitions have clusters defined in terms of distributions (despite the fact that cluster-as-distribution is a well-established concept).

    \item The target clusters have arbitrary shapes, varied sizes and densities are the natural consequence of i.i.d. samples from distributions. As such, there is no need to state that each cluster consists of similar points (which was the key ingredient in all the previous definitions).
\end{itemize}

To intuitively illustrate the evolution from Definition \ref{def-typical} to Definition \ref{def-distribution-clustering}, we present a schematic diagram in Figure \ref{fig:definition_shift}. The key advantage of assuming each cluster as a distribution is that \emph{each discovered cluster matches the distribution in which the cluster is generated}; and the clusters can be any arbitrary shapes, sizes and densities.  Without assuming each cluster as a distribution, a clustering method  would need to adopt a cluster definition, specific to the method, which defines the type of clusters it can discover. 

For example, a connected component is often used to define a cluster (as in the case of density-connected clusters in DBSCAN, or subgraphs as clusters in Spectral Clustering, a centroid to present each cluster in $k$-means clustering that can be viewed as all points in the cluster is connected to its centroid). The connected components created by each clustering method must satisfy some explicit or implicit criterion that  restricts them to certain type of clusters only. This prevents the clustering method from discovering clusters of arbitrary shapes, densities and sizes which match the distributions that generate the clusters. The fundamental limitations of each of these clustering methods are now well-understood \cite{nadler2006fundamental,zhu2016density,ting2022pskc}. In other words, none of these clustering methods can achieve the aim of clustering stated in Definition \ref{def-clustering-aim}. The finding in this paper adds DEC/IDEC to this category of clustering methods.\\
The difficulty in analyzing the fundamental limitations of DEC is that the kind of clusters, that DEC can discover, is undefined. Like in the case of Spectral Clustering \cite{von2007tutorial}, this is because the learned mapping function is a `black box'. 
The above analyses have prompted us to make the following hypothesis:
\begin{hypothesis}
Distributional information of (unknown) clusters in a dataset is a sufficient ingredient in a clustering method to achieve the aim of clustering stated in Definitions~\ref{def-clustering-aim} and \ref{def-distribution-clustering}.
\label{hypothesis-dist}
\end{hypothesis}

\begin{table*}[htbp]
    \centering
    \caption{A conceptual comparison between CaD Clustering and Deep Clustering.} 
    \label{tab:conceptual_comparison}
    \renewcommand{\arraystretch}{1.0} 
    \setlength{\tabcolsep}{6pt}       
    
    \begin{tabularx}{\textwidth}{@{} l >{\raggedright\arraybackslash}X >{\raggedright\arraybackslash}X @{}}
    \toprule
    \textbf{Attribute} & \textbf{CaD Clustering} & \textbf{Deep Clustering} \\ 
    \midrule
    
    \textbf{Assumption} & Cluster $C$ is a set of i.i.d. points from \hspace*{2mm} Distribution $\mathcal{P}_C$ & A cluster is a set of points, and \hspace*{2mm} a~centroid represents a cluster\\ 
    \textbf{Clustering Definition} & Def. \ref{def-clustering-aim} or \ref{def-distribution-clustering} & Def. \ref{def-typical} \\
    \textbf{Cluster Definition} & Def. \ref{def_NSS-clusters} & Cannot be defined \\
    \textbf{Clusters discovered} & Match Def. \ref{def-distribution-clustering} & Match Def. \ref{def-typical}; Not Match Def. \ref{def-distribution-clustering} \\
    \textbf{Objective/Loss Function} & $\displaystyle \sum_{C \in \mathbb{C}}K(\mathcal{P}_C,\mathcal{P}_C) \cdot |C|$ & $\displaystyle L_{Representation} + \gamma \cdot L_{Clustering}$ \\
    \textbf{Optimization} & Simple greedy search & Must use Deep Learning \\
    \textbf{Clustering process} & With Distributional Kernel & Conflate with representation learning \\
    \textbf{Algorithmic focus} & Clustering only & Representation learning \\
    \textbf{Information used in Clustering} & Cluster Distribution & Info. other than Cluster Distribution \\
    \textbf{Aim of the mapping} & Def. \ref{def-requirements2} & Def. \ref{def-requirements2} \\
    \textbf{Meaning of centroids} & $\widehat{\phi}(\mathcal{P}_{C})$: clear meaning & $\varphi(C)$: unclear meaning \\
    \textbf{Prior knowledge \cite{lu2024survey}} & Not Required & Required \\
    \textbf{Algorithmic complexity} & Linear time & At least Quadratic time \\
    
    \bottomrule
    \end{tabularx}
\end{table*}

Indeed, there are now clustering methods which assume that each cluster is generated from a distribution or cluster-as-distribution (CaD), e.g., psKC, IDKC and KBC \cite{ting2022pskc,zhu2023idkc,zhang2025kbc}, and they successfully discover the clusters in all three datasets shown in Table \ref{tab:clustering_results_examples}. They discover clusters as distributions via a distributional kernel using a simple greedy search only, and have linear time complexity. Therefore, they run significantly faster than the optimization-based kernel $k$-means clustering,  Spectral Clustering and Deep Clustering.

Though having the same optimization objective, kernel $k$-means clustering must use an Expectation-Maximization optimization to perform the optimization via a point-to-point kernel, while KBC or psKC \cite{ting2022pskc,zhang2025kbc} uses a greedy search to achieve a similar objective via a distributional kernel $\mathcal{K}$. The latter produces better clustering outcomes in terms of NMI in the experiments \cite{zhang2025kbc,ting2022pskc} because KBC or psKC make full use of the distributional information of clusters in the dataset. As a result, its optimized solution $C_j = \{ x \in \mathcal{X} | \argmax_{i \in [1,K]} \mathcal{K}(\delta(x), P_{B_i}) = j \}$ can be derived from the same objective function of the optimal solution, i.e., $\max \sum\nolimits_{k=1}^K \mathcal{K}(P_{C_k}, P_{C_k}) \times |C_k|$, where $P_{B_i}$ is a proxy to the optimal $P_{C_i}$ (see Eq (\ref{eqn_IKBC}) and the algorithmic details of KBC provided in Appendix \ref{sec:kbc_alg}). 

Because the distributional information in a dataset is ignored and a cluster is defined as a set of points (be it in the input space or the feature space of the kernel), kernel $k$-means clustering resorts to using a centroid to represent a cluster without treating the centroid as a distribution---this is the root cause of its inability to achieve the aim stipulated in Definition \ref{def-clustering-aim}. 
The empirical support for the hypothesis has been provided in \cite{zhang2025kbc}. 


Hypothesis \ref{hypothesis-dist} stems from the fact that it is difficult to achieve the aim stated in Definition \ref{def-clustering-aim} by using any of the existing methods, including deep clustering, that ignore the distributional information of clusters, and they all attempt to find each cluster based on the similarity of points (recall Definition~\ref{def-typical}). The differences between Deep Clustering or DC (that do not employ the distributional information of clusters) and Cluster-as-Distribution (CaD) clustering are provided in the next section.

\subsection{The differences between DC and CaD Clustering}

\begin{figure}[htbp]
    \centering
    \includegraphics[width=\linewidth]{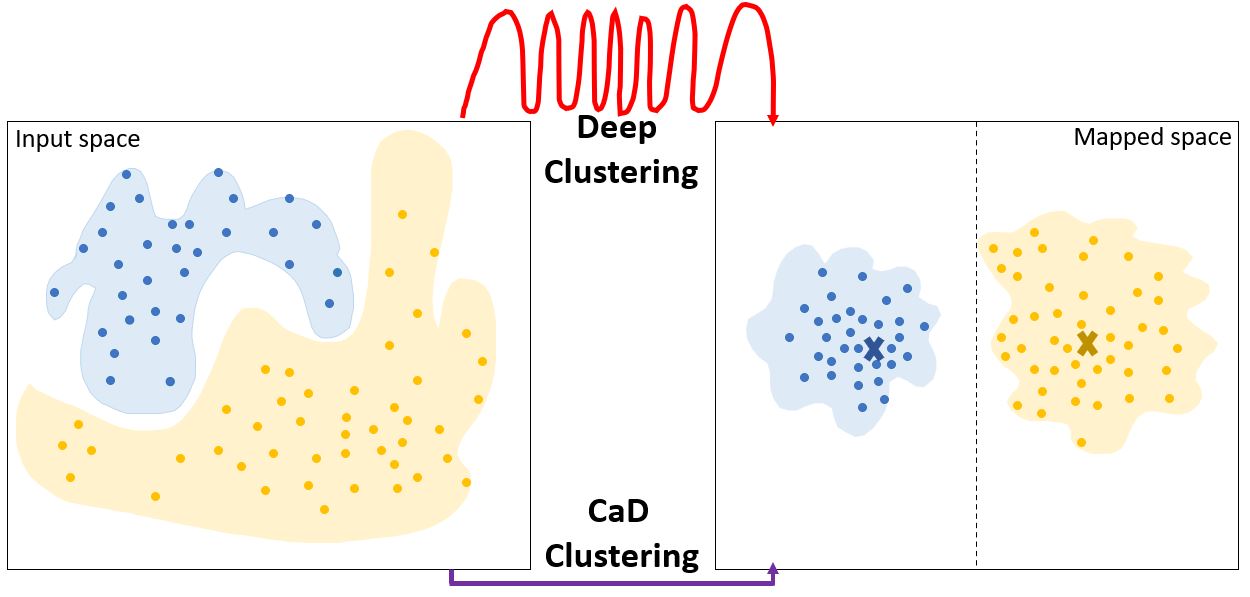}
    \caption{An illustration of Deep Clustering (DC) and CaD Clustering via different means (deep learning versus  simple greedy search with distributional kernel, respectively) to achieve the same intended aims of clustering via mapping from clusters in the input space to centroids in the mapped space.}
    \label{fig:main_diff}
\end{figure}

We denote the clustering methods, that assume cluster-as-distribution (CaD) as the means to perform clustering, as CaD Clustering. 
Figure \ref{fig:main_diff} shows an illustration of one of the key differences between Deep Clustering and CaD Clustering, i.e., deep learning versus simple greedy search with distributional kernel. Although they both intend to achieve the same aim of clustering via mapping from complex clusters in the input space to simple centroids in the mapped space, Deep Clustering fails thus far but CaD Clustering has already succeeded.

The current comparison results (e.g., \cite{ting2022pskc,zhu2023idkc,zhang2025kbc}) as well as those shown in Table~\ref{tab:clustering_results_examples} provide evidence of the advantage of the direct approach in terms of clustering outcome---CaD Clustering achieves Definition \ref{def-clustering-aim}, but Deep Clustering cannot.

We also provide a comprehensive comparison in Table \ref{tab:conceptual_comparison} and a more analysis can be found in Appendix \ref{sec:difference_dc_cad}.


\subsection{High Dimensional Datasets}
Revisiting the questions outlined in Section \ref{sec:definition_of_clustering}, we now examine the performance of deep clustering in high-dimensional scenarios. Experimental results in Table \ref{tab:image_dataset_run} indicate that while deep clustering approaches exhibit a certain degree of efficacy on high-dimensional datasets, they do not yield a fundamental performance breakthrough, notably, CaD achieves performance comparable to, or even surpassing, that of deep learning-based approaches on these datasets. As shown in Table \ref{tab:image_dataset_run}, Cluster-as-Distribution (CaD) methods demonstrate superior performance across the vast majority of high-dimensional single-cell and spatial transcriptomics datasets. Notably, on the Tutorial dataset presented in Table \ref{tab:clustering_results_tutorial}, KBC outperforms other baselines by a substantial margin. While the visualization of $k$-means appears promising (depicting its best-case scenario), the average results reveal that $k$-means remains limited by a well-known deficiency: its high sensitivity to initialization. Furthermore, on high-dimensional image benchmarks, CaD methods achieve highly competitive results compared to deep clustering approaches, with particularly notable advantages observed on the COIL20 dataset. A more comprehensive analysis regarding these high-dimensional scenarios is provided in Appendix \ref{sec:analysis_on_high_dimension}.

\begin{table*}[!ht]
\centering
\caption{Clustering performance (NMI) on high-dimensional benchmarks. Each reported Normalized Mutual Information (NMI) \cite{strehl2002nmi} value is averaged over 10 runs. An example clustering visualization result in Table \ref{tab:clustering_results_tutorial} is obtained for the Tutorial dataset. The data characteristics of all datasets are provided in Table \ref{tab:image_dataset_information} in the Appendix.}
\label{tab:image_dataset_run}
\begin{tabular*}{\textwidth}{@{\extracolsep{\fill}}l r r c c c c c c}
\toprule
\textbf{Datasets} & \textbf{Points} & \textbf{Dim} & \textbf{Clusters} & \textbf{$k$-means} & \textbf{IDEC} & \textbf{CC} & \textbf{SpectralNet} & \textbf{KBC} \\
\midrule
Tutorial      & 1,556  & 2,000 & 2  & 0.31           & 0.01            & 0.02 & 0.04 & \textbf{0.87} \\
Tonsil        & 5,778  & 2,000 & 13 & 0.56           & \textbf{0.63}   & 0.52 & 0.60 & 0.52 \\
Airway        & 7,193  & 2,000 & 7  & 0.53           & 0.46            & 0.34 & \textbf{0.63} & 0.62 \\
Crohn         & 39,563 & 2,000 & 27 & \textbf{0.62}  & 0.55            & 0.54 & 0.59 & \textbf{0.62} \\
DLPFC         & 4,221  & 400   & 7  & 0.56           & 0.54            & 0.50 & 0.42 & \textbf{0.64} \\
\addlinespace[4pt]
USPS          & 11,000 & 256  & 10 & 0.45 & 0.69           & 0.49 & 0.78 & \textbf{0.82} \\
STL-10        & 13,000 & 128  & 10 & 0.61 & 0.67           & \textbf{0.76} & 0.67 & 0.67 \\
CIFAR-10      & 60,000 & 128  & 10 & 0.65 & \textbf{0.76}  & 0.67 & 0.73 & 0.74 \\
ImageNet-10   & 13,000 & 128  & 10 & 0.73 & 0.80           & 0.85 & 0.80 & \textbf{0.88} \\
ImageNet-Dogs & 19,500 & 128  & 15 & 0.43 & 0.39           & 0.45 & 0.50 & \textbf{0.51} \\
MNIST         & 70,000 & 784  & 10 & 0.50 & \textbf{0.86}  & 0.68 & 0.85 & 0.82 \\
COIL-20       & 1,440  & 1024 & 20 & 0.76 & 0.77           & 0.39 & 0.81 & \textbf{0.98} \\
\bottomrule
\end{tabular*}
\end{table*}

\begin{table*}[!ht]
\centering
\caption{Clustering results comparison on Tutorial dataset. Each reported Normalized Mutual Information (NMI) \cite{strehl2002nmi} value is averaged over 10 runs; the clustering visualization result is obtained from the run having the highest NMI.}
\label{tab:clustering_results_tutorial}
\setlength{\tabcolsep}{1pt}
\renewcommand{\arraystretch}{0.4}
\resizebox{\textwidth}{!}{
\begin{tabular}{c c c c c c c}
    \toprule
    & \small \textbf{Original Data} & \small \textbf{$k$-means} & \small \textbf{IDEC} & \small \textbf{CC} & \small \textbf{SpectralNet} & \small \textbf{KBC} \\
    \midrule
    
    \rotatebox{90}{\footnotesize tutorial} & 
    \includegraphics[width=0.16\textwidth]{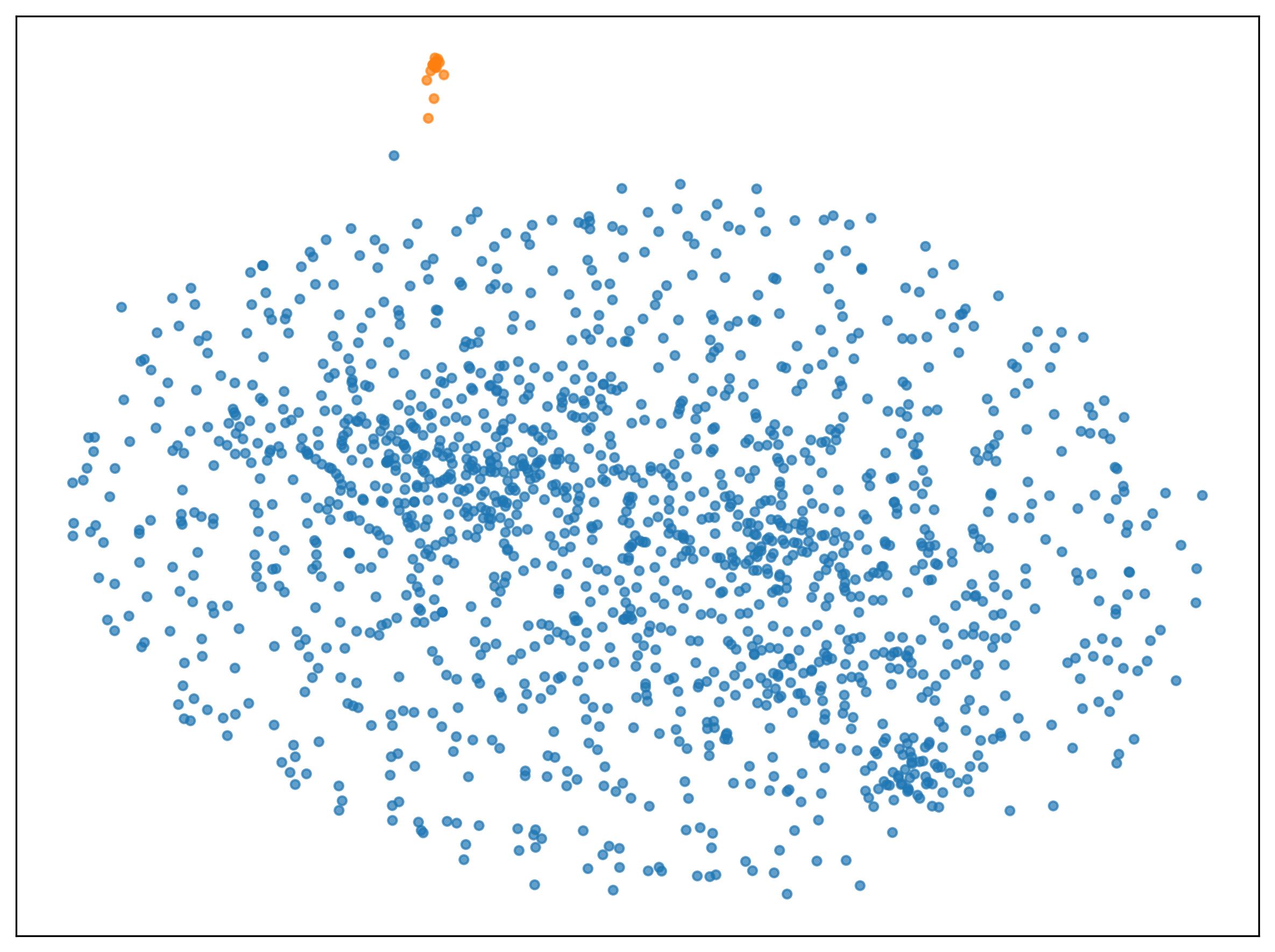} &
    \includegraphics[width=0.16\textwidth]{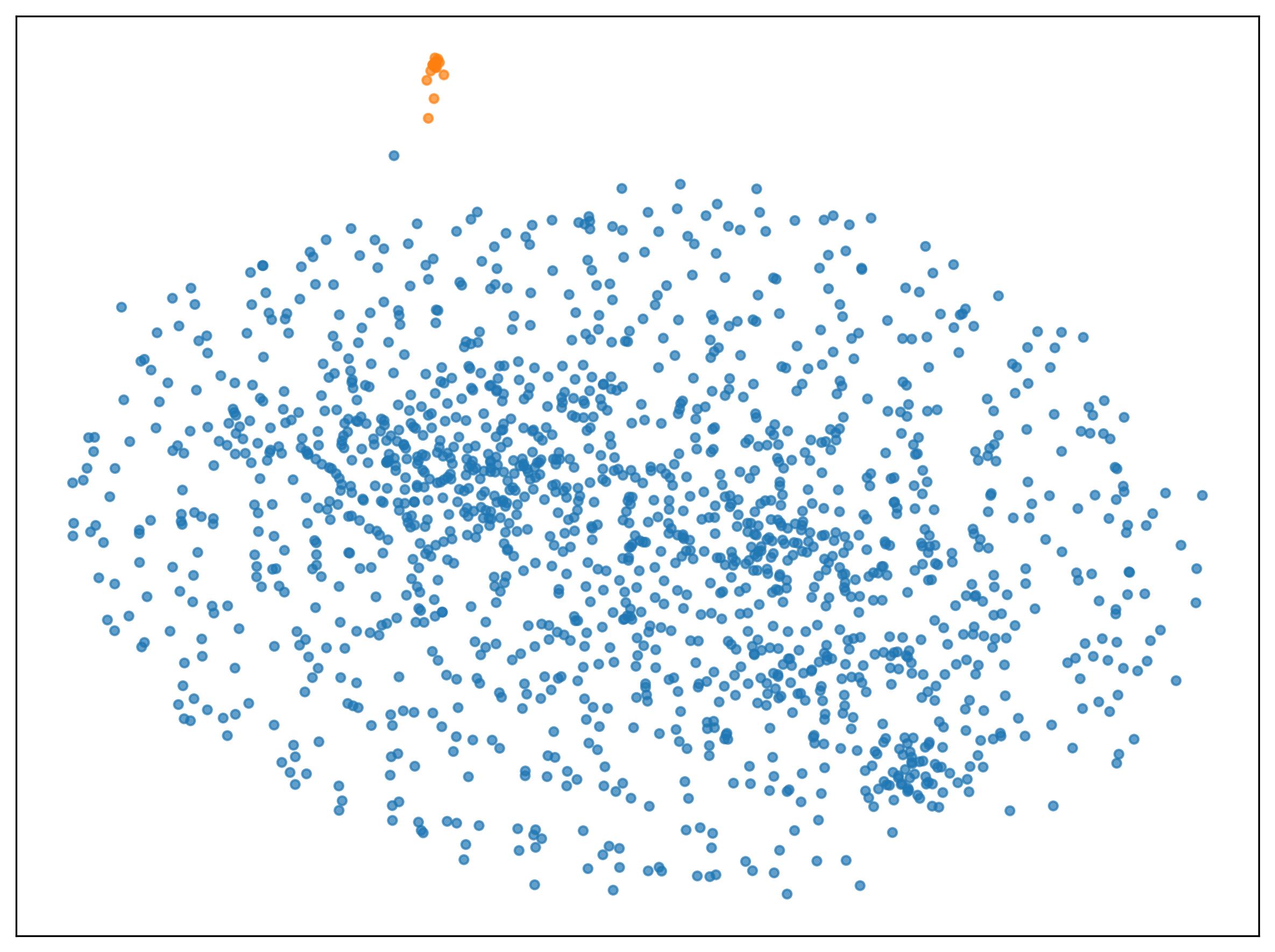} &
    \includegraphics[width=0.16\textwidth]{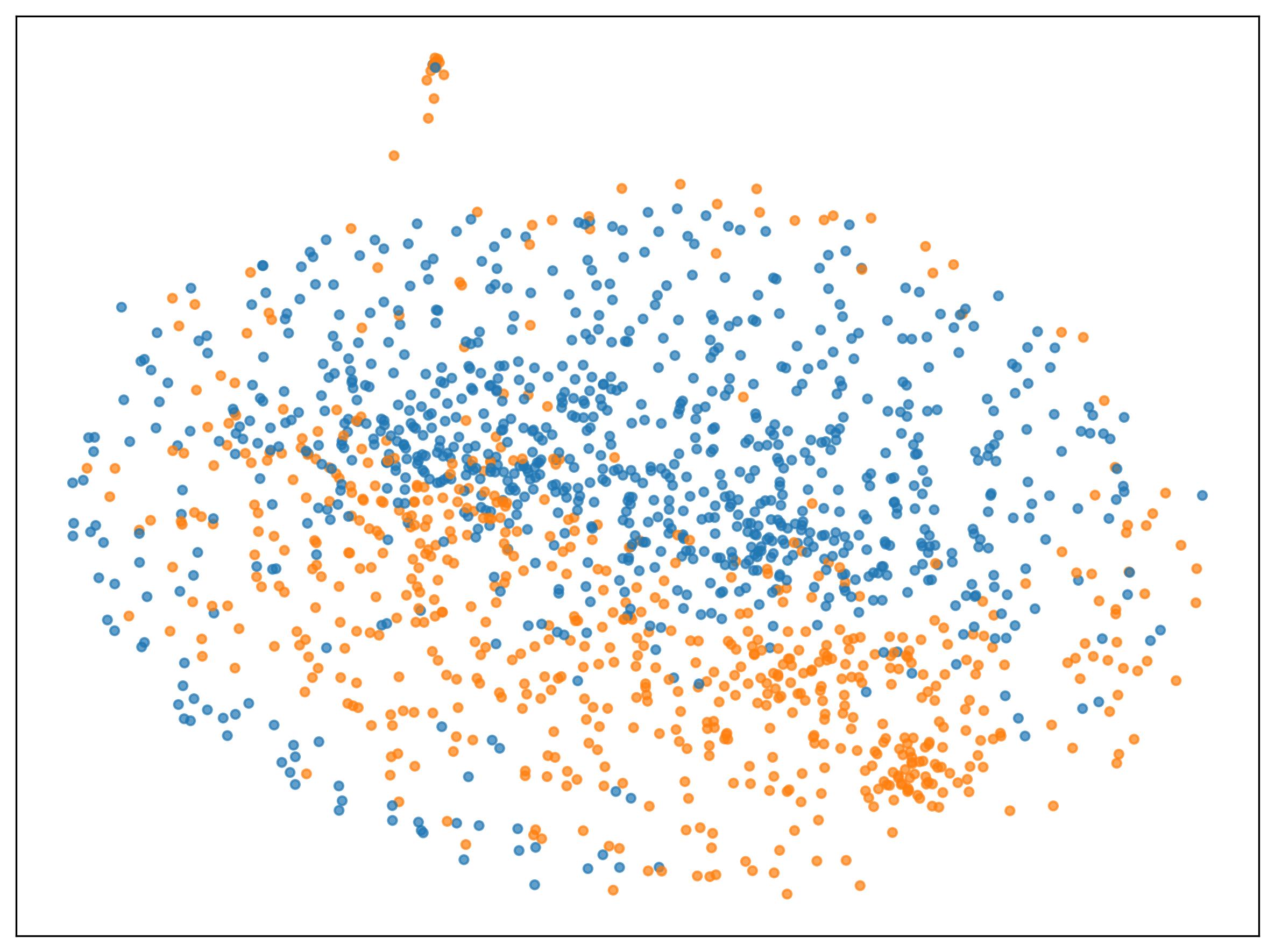} &
    \includegraphics[width=0.16\textwidth]{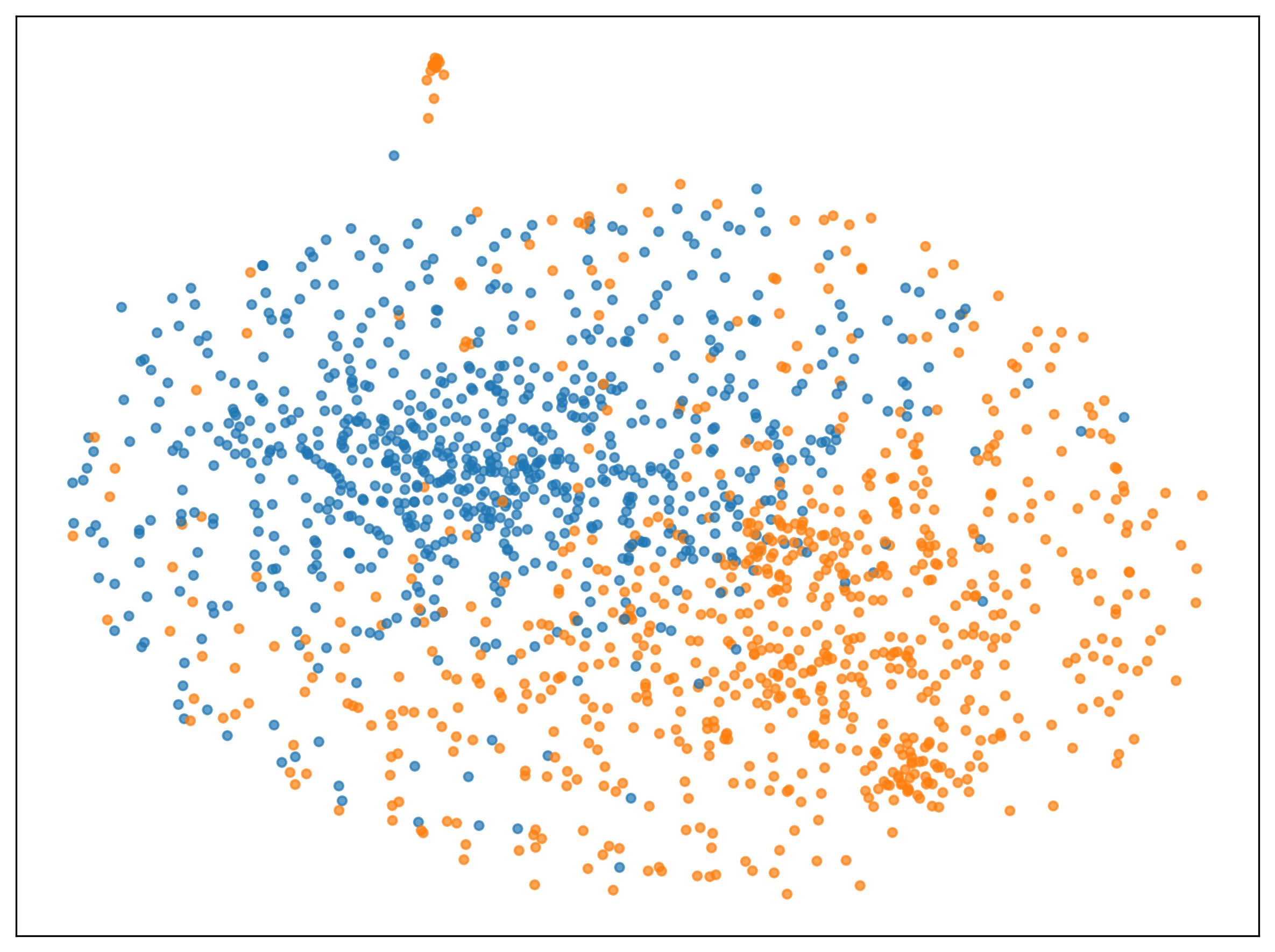} &
    \includegraphics[width=0.16\textwidth]{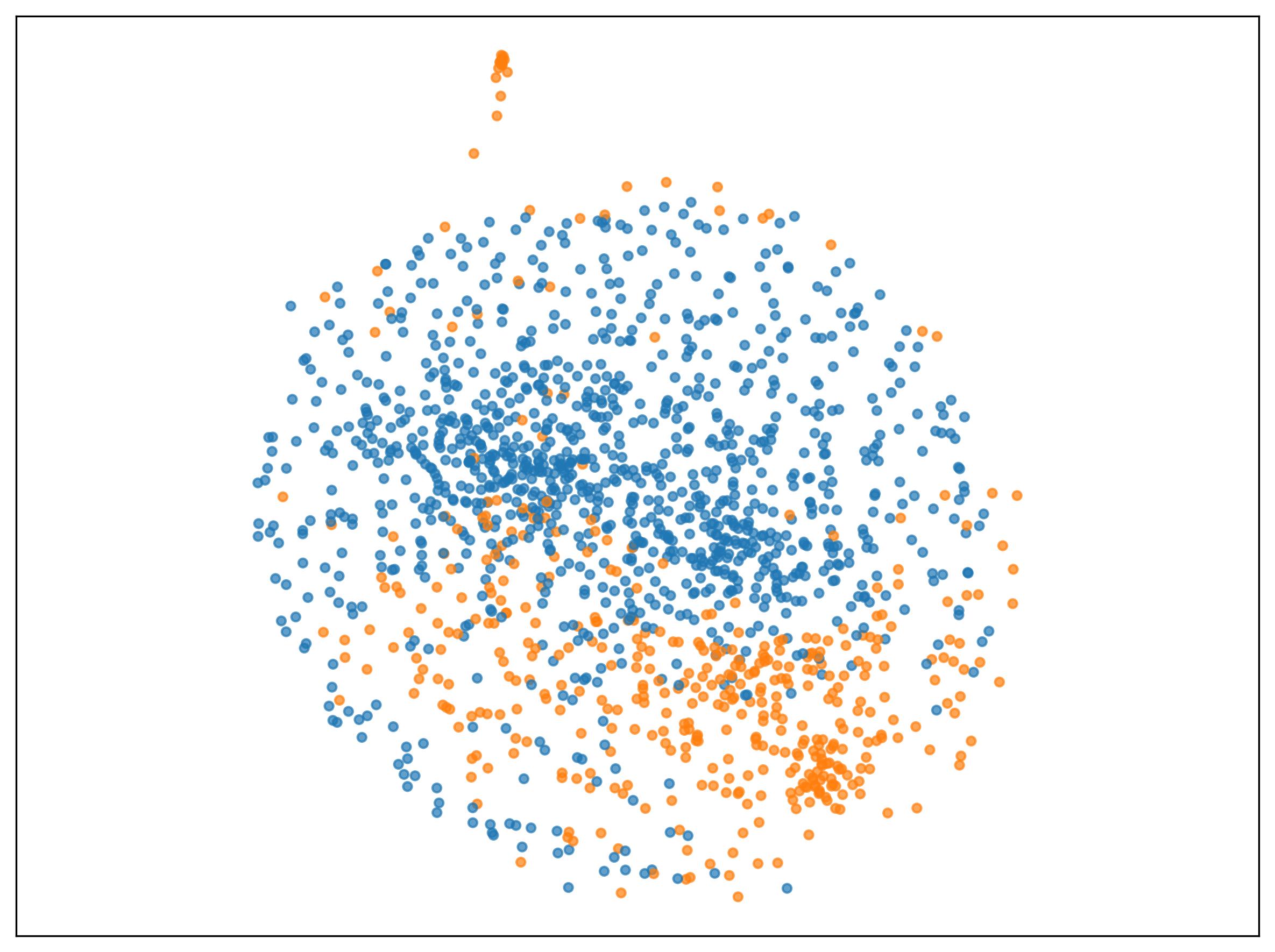} &
    \includegraphics[width=0.16\textwidth]{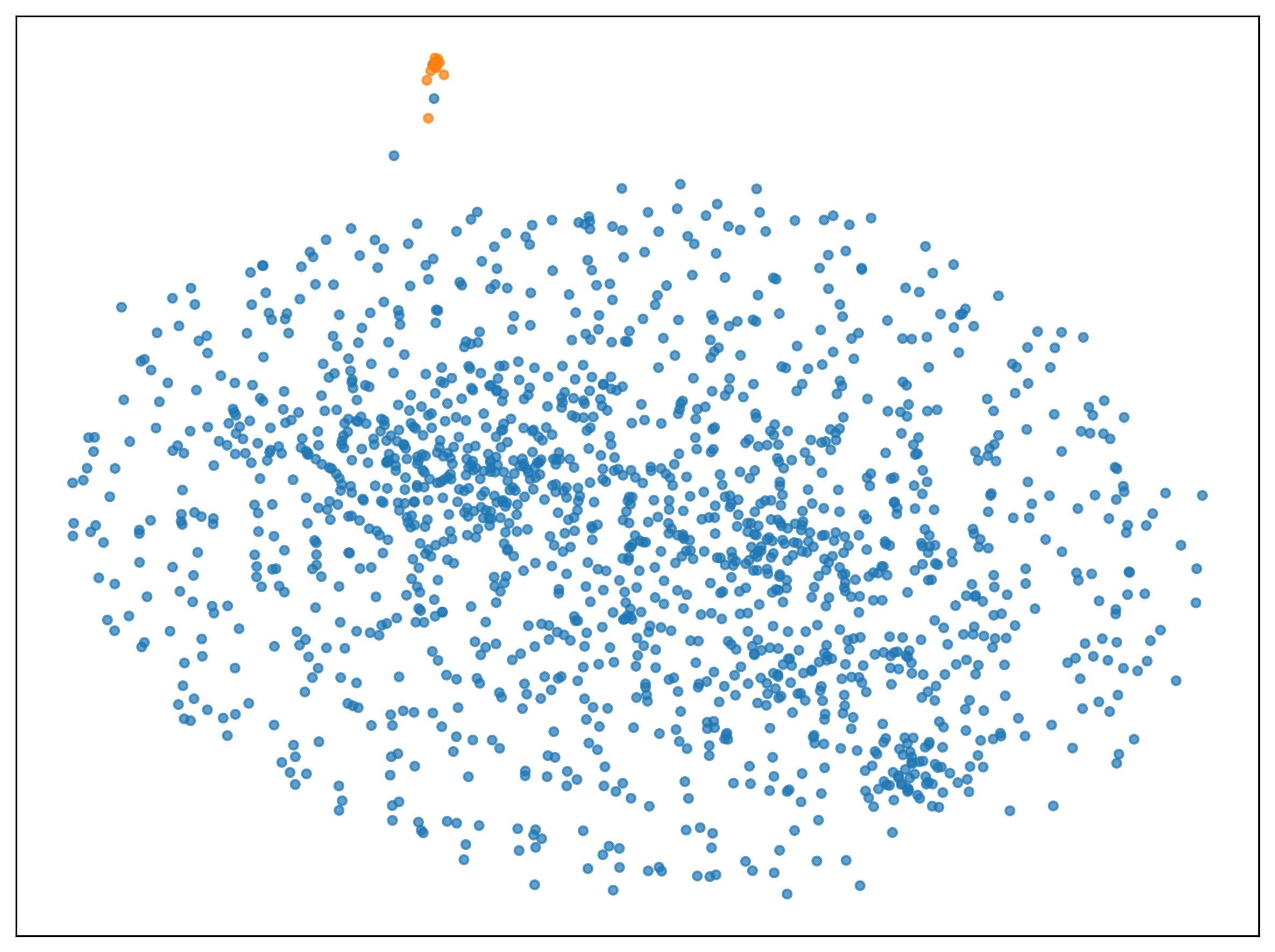} \\
    & 
    & \footnotesize NMI=0.31 & \footnotesize NMI=0.01 & \footnotesize NMI=0.02 & \footnotesize NMI=0.04 & \footnotesize NMI=0.87 \\
    
    \bottomrule
\end{tabular}}
\end{table*}

\section{Recommendations}

To progress in the field of clustering, we have the following recommendations:
\begin{itemize}[noitemsep, topsep=0pt] 
    \item A deep clustering method must have the first two of the following, if a theory could not be established:
    \begin{itemize}[noitemsep] 
        \item[a)] Demonstrate empirically that it performs better than $k$-means clustering such that it does not have the commonly known fundamental limitations. This can be easily assessed using the same or similar simple examples we have used here. Assessment on image datasets only is not sufficient.
        
        \item[b)] Be clear about the added value of the latent representation learning by comparing deep clustering with the baseline CaD Clustering that employs a simple greedy search with distributional kernel.
        \item[c)] Provide a theoretical backing that the learned latent representation is suitable for clustering w.r.t. any one of Definitions \ref{def-requirements} \& \ref{def-requirements-v2}.
    \end{itemize}
    \item For a CaD clustering method, investigate whether a sophisticated optimization can add value to the simple greedy search version.
    \item Replace Definition \ref{def-typical} with Definition \ref{def-distribution-clustering} as the revised definition of clustering in the future. This would avoid any costly future undertakings, in terms of time and effort, in ignoring the distributional information of clusters which can be easily extracted from a given dataset.
\end{itemize}

\section{Conclusions}
As far as we know, we are first to stipulate the intended aim of deep clustering (i.e., Definition \ref{def-requirements} or \ref{def-requirements-v2}).  Through empirical assessments, we have uncovered its fundamental limitations that fail to meet its intended aim because the cluster definition relies on a point-to-point similarity function, ignoring the distributional information clusters in a given dataset. By defining cluster as distribution (CaD), we show that CaD clustering via a distributional kernel can achieve the intended aim of deep clustering, without using deep learning. This occurs in high dimensional datasets as well as image datasets, contrary to the conventional wisdom that deep learning  has an advantage over non-deep learning methods in these domains.




\section{Impact Statements}
This paper presents work whose goal is to advance the field of machine learning. There are many potential societal consequences of our work, none of which we feel must be specifically highlighted here.

\clearpage

\bibliography{02_ICML2026/reference}
\bibliographystyle{icml2026}

\newpage
\appendix
\onecolumn

\section{Existing Definitions of Clustering}

Example definitions of clusters and clustering are given as follows:
\begin{enumerate}
    \item `Clustering is the grouping of similar objects' \cite{hartigan1975clustering}
    \item `Clustering is a process of grouping data items based on a measure of similarity.' \cite{jain1999dataclustering}
    \item `Objects within the same cluster are highly similar to one another. Objects in different clusters are very dissimilar to one another.' \cite{aggarwal2014dataclustering}
    \item `Cluster analysis or simply clustering is the process of partitioning a set of data objects into subsets. Each subset is a cluster, such that objects in a cluster are similar to one another, yet dissimilar to objects in other clusters.' \cite{han2022data}
    \item `Clustering aims at grouping data instances into several clusters, where instances from the same cluster share similar semantics and instances from different clusters are dissimilar.' \cite{lu2024survey}
\end{enumerate}

These definitions are fine for a general understanding of what a cluster or clustering is. But they have two key deficiencies. First, two data points at different ends of an (elongated) cluster may be less similar compared with their individual closest points from other clusters. 

Second, all the above definitions provide no information about the clusters that enables one to assess the `goodness' of the clusters produced from a clustering method, except the need to have a point-to-point similarity measure. The relative density of clusters has been recognized to be a key cluster characteristic that poses a key difficulty for many existing clustering methods, i.e., they have difficulty identifying clusters of varied densities \cite{nadler2006fundamental, zhu2016density}.

\section{KBC Algorithm}
\label{sec:kbc_alg}
For clarity and reproducibility, we provide the pseudocode of the KBC algorithm below. 
\begin{algorithm*}[htbp]
   \caption{Kernel Bounded Clustering}
   \label{alg:IKBC}
   \begin{algorithmic}[1]

   \STATE \textbf{Input:} $D$ - dataset, $k$ - number of clusters, $s$ - sample size, $\tau$ - similarity threshold
   \STATE \textbf{Output:} $\mathbb{C} = \{ C_1, \dots, C_k \}$
   
   \STATE \textbf{Step 1:} Find the $k$ initial clusters
   
   \STATE From $D_{s} \subset D$, find the largest $k$ initial clusters $G_{j}$ via kernel $\kappa$ as follows:
   
   \STATE \hspace{5mm} $G_{j} = \{\mathbf{x},\mathbf{y}\in D\ |\ \mbox{there exists a chain: } \mathbf{z}_1,\mathbf{z}_2,\cdots,\mathbf{z}_q; \mathbf{z}_1= \mathbf{x}, \mathbf{z}_q=\mathbf{y}, \forall i\ \kappa(\mathbf{z}_i, \mathbf{z}_{i+1}) > \tau \}, \forall {j\in [1,k]}$ \label{marker}
   
   \IF{the number of clusters in $\{G_1, G_2, \dots, G_j\}<k$}
      \STATE Assert `Parameter $\tau$ is set too small !' and Exit
   \ENDIF
   
   \STATE \textbf{Step 2:} Assignment of data points to clusters
   
   \STATE \hspace{5mm} $C_j=\{\mathbf{x}\in D\ |\ \mathop{\argmax}\limits_{i \in [1,k]} K(\delta (\mathbf{x}), \mathcal{P}_{G_i}) =j\}, \forall_{j\in [1,k]} $
   
   \STATE \textbf{Step 3:} Refine $\mathbb{C} = \{ C_1, \dots, C_k \}$ to improve the objective: $\max_\mathbb{C} \sum_{C \in \mathbb{C}} \sum_{\mathbf{x} \in C} K(\delta(\mathbf{x}), \mathcal{P}_C ).$
   
   \STATE \textbf{Return} $\mathbb{C}$
   
   \end{algorithmic}
\end{algorithm*}

\section{Fundamental limitations of other clustering methods that rely on a learned representation}
\label{sec:limitations_other_methods}
Our discussion and experimental analysis primarily focus on DEC/IDEC and CC, motivated by the following considerations. DEC/IDEC is among the earliest methods to employ deep learning especially autoencoders for representation learning in clustering tasks. Moreover, it is the first to propose a unified objective that jointly optimizes representation learning and clustering objective by combining a representation reconstruction loss with a clustering loss. This formulation has since become the basis for many subsequent deep clustering methods. Image data have also inspired alternative algorithmic designs such as CC that introduces a self-supervised contrastive learning framework tailored for image clustering. It has influenced a broad line of later deep clustering \cite{lin2021cc2, pan2021cc4, chen2023cc3}.

While there are attempts to improve Gaussian Mixture Model (GMM) clustering methods \cite{jiang2016vade} via VAE \cite{kingma2013auto} (e.g., VaDE \cite{jiang2016vade}), the approach is similar to DEC/IDEC, except that centroid-based clusters are replaced with clusters having Gaussian distributions. But there is no evidence that VaDE can achieve Definition~\ref{def-requirements-v2}, neither empirically nor in theory. While GMM may be viewed as an improvement over $k$-means clustering, but the same fundamental limitations remain that restrict GMM to discovering spherical shaped clusters of Gaussian distributions only. Note that both GMM and $k$-means clustering employ the same Expectation-Maximization optimization \cite{dempster1977maximum}. 

More specifically, although VaDE incorporates data distribution into the clustering process, its core idea is to enforce a specific distributional assumption—most notably a Gaussian mixture—within the latent encoding space. While this strategy accounts for certain distributional characteristics of the data, it does not approach the clustering problem from the perspective of the data’s intrinsic distribution in the original space.

In addition to the fundamental limitations of deep clustering, there are other issues related specifically to the specific deep learning methods employed. For example, variational autoencoder \cite{kingma2013auto} is known to have unresolved issues such as unstable training,  and the gradient vanishes \cite{basodi2020gradient}, which result in the inability to learn desirable embedded features.  To address these issues, the KL divergence measure \cite{kullback1997information} (used in DEC/IDEC to formulate the clustering loss) is replaced with Wasserstein distance \cite{villani2009wasserstein} to produce Wasserstein Embedding Clustering (WEC) \cite{cai2024wec}. Yet, WEC employed the same centroid-based clusters as DEC/IDEC to formulate the clustering loss. The evaluation, as many other deep clustering methods, is performed on image datasets only \cite{cai2024wec}. The question in relation to the fundamental limitations we raised here is not answered.

An advantage of deep clustering over traditional clustering, puts forward by a recent survey paper \cite{lu2024survey}, is that deep clustering has enabled some additional prior knowledge to be incorporated into the learning process, whereas $k$-means clustering could not employ this knowledge. The prior knowledge includes  distribution prior (e.g., VaDE and ClusterGAN \cite{jiang2016vade,mukherjee2019clustergan}, augmentation invariance (e.g., CC \cite{li2021contrastive, wang2022chaos}) and pseudo-labeling (e.g., DEC/IDEC \cite{xie2016unsupervised,ijcai2017p243}); and WEC \cite{cai2024wec} incorporates both distribution prior and pseudo-labeling. However, none of these methods have been shown to have achieved either Definition \ref{def-requirements} or Definition \ref{def-requirements-v2}.

\section{The differences between Deep Clustering and CaD Clustering}
\label{sec:difference_dc_cad}
As the intended outcome, i.e., \textbf{representing clusters in input space as centroids in mapped space}, is the same for both clustering approaches, the clustering definition can be re-expressed as follows:

\begin{definition}
    To achieve the aim of clustering as stated in Definition \ref{def-clustering-aim}, a mapping $\Phi$ must transform clusters $C$ of arbitrary shapes, varied sizes and densities in input space into ones that can be represented by centroids $\mathbf{z}_c$ in the mapped space.
    \label{def-requirements2}
\end{definition}

In terms of distribution, there is a ready tool to represent each cluster as a distribution, i.e., Kernel Mean Embedding \cite{KernelMeanEmbedding2017}. By using a kernel $\kappa$ with its feature map $\phi$, a distribution $\mathcal{P}_C$ of a cluster $C$ can be represented as $\widehat{\phi}(\mathcal{P}_{C}) = \frac{1}{|C|} \sum_{\mathbf{y} \in C} \phi(\mathbf{y})$. This is the feature map of a distributional kernel $K$, which is derived from $\kappa$ as follows \cite{KernelMeanEmbedding2017}:
\begin{eqnarray}
{{K}}(\mathcal{P}_X,\mathcal{P}_Y) 
 & = & \frac{1}{|X||Y|}\sum_{x\in X} \sum_{y\in Y} \kappa(x,y) \label{eqn_KME2}\\
 & = &  \left< \widehat{\phi}(\mathcal{P}_X), \widehat{\phi}(\mathcal{P}_Y) \right> \label{eqn_KME}
\end{eqnarray}

The distributional kernel $K$ computes the similarity between two distributions $\mathcal{P}_X,\mathcal{P}_Y$.

In other words, using a distributional kernel, the mapping of each cluster $C$ is: $\Phi(C) = \widehat{\phi}(\mathcal{P}_{C})$ which is a centroid in the mapped space, stated in Definition \ref{def-requirements2}. This centroid is a result of assuming each cluster as distribution via a distributional kernel.

Using the learned latent representation $\varphi$ in deep clustering, the mapping of each cluster $C$ is:  $\Phi(C) = \varphi(C) = \mathbf{z}_c$, where  $\mathbf{z}_c$ is the centroid in the mapped space representing cluster $C$ in input space. While this is the intended aim of deep clustering, it is unclear that (i) the representation mapping $\varphi$ can achieve the intended aim; and (ii) what the formal cluster definition is by having the centroid $\mathbf{z}_c$.

A more complete conceptual comparison between CaD Clustering and Deep Clustering is summarized in Table~\ref{tab:conceptual_comparison}, and the detailed descriptions are provided in the rest of this section.

A common way to formulate the loss function $L$ of deep clustering is to decompose into two components as follows \cite{ijcai2017p243} \footnote{In fact, the optimization objectives of most deep clustering methods consist of a combination of representation learning and clustering-related losses. The primary distinction among them lies in the order and manner in which these objectives are optimized.}:
\[
L = L_{R} + \gamma \cdot L_C
\]
where $L_R$ is the representation learning loss which aims to learn a good representation of input data and the representation learning module can be either auto-encoder based module \cite{bank2023autoencoders}, generative module \cite{kingma2013auto}, mutual information based module \cite{kinney2014mi} or contrastive based module \cite{chopra2005learning};  $L_C$ is the clustering loss and $\gamma$ is a user-set parameter regularizing the two components \cite{long2015learning, caron2018deep, chang2017deep, guo2018deep}.

For example, Wasserstein Embedding Clustering (WEC) \cite{cai2024wec} has $L_R$ based on Wasserstein distance that has a regularization term which penalizes the discrepancy between the distribution of mapped points and a prior distribution (where this term is measured using another distribution measure, e.g., Maximum Mean Discrepancy (MMD) \cite{smola2006maximum}); and $L_C$ is measured using KL-divergence between the distributions of soft label assignments based on Student's t-distribution \cite{student1908probable} and its enhanced version which emphasizes labels with high probability. In summary, WEC employs three distributional measures (i.e., Wasserstein distance, MMD and KL-divergence) in formulating the joint learning of latent representation and clustering. Also note that WEC has used the identical $L_C$ as used in DEC/IDEC---assuming that \emph{each cluster can be represented using a centroid}---the same assumption as in $k$-means clustering, albeit the latter performs the centroid-based representation in input space, and the deep clustering methods in the mapped space.

In contrast, the CaD clustering employs an Isolation Distributional Kernel\footnote{It is derived from Kernel Mean Embedding that uses Isolation Kernel.} (IDK) \cite{ting2020isolation} that requires a simple greedy search only to optimize the clustering objective. In terms of using distributional information, WEC has employed three types of distributional information but none of them is about cluster distribution. CaD clustering uses cluster distribution only to perform the clustering, where each distribution is represented using IDK without optimization.

It is interesting to note the objective function of CaD clustering for a given dataset $D$ in order to produce a set of clusters $\mathbb{C}$ using a distributional kernel $K$:
\begin{equation}  \max_\mathbb{C} \sum_{C\in\mathbb{C}}\sum_{x\in C}K(\delta(x),\mathcal{P}_C)
    = \max_\mathbb{C} \sum_{C \in \mathbb{C}}K(\mathcal{P}_C,\mathcal{P}_C) \cdot |C|.
    \label{eqn_IKBC}
\end{equation}
where $\mathcal{P}_C$ is the distribution of cluster $C$ and $\delta(x)$ is a Dirac measure which converts a data point into a distribution.

The clustering process is clear in CaD clustering, i.e., it maximizes the similarity of each data point w.r.t. the distribution of a cluster, in order to produce a set of clusters which has the maximum self similarity for all clusters, while each has a large cluster set size. In this process, only the distributional kernel $K$ is required to compute the similarity. In addition, the type of clusters it discovered can be formally defined \cite{zhang2025kbc} (see Definition \ref{def_NSS-clusters} in the Appendix). On the other hand, the complexity of deep clustering, e.g., WEC uses three different distributional measures for different purposes. This has made the entire process very opaque. As a result, the type of clusters a deep clustering discovered is hard to define formally. As far as we know, no cluster definitions have been provided in the literature on deep clustering.

The objective, specified in Equation (\ref{eqn_IKBC}), can be achieved in three different ways by psKC, IDKC and KBC without a sophisticated optimization. 

The advantage of deep clustering over CaD clustering is representation learning which allows the former to deal with domains such as images directly. However, its clustering performance has yet to be proven in domains outside images. The added value, due to the complexity, the additional learning and prior knowledge of deep clustering, needs to be justified in the face of the simpler, greedy search and no-prior-knowledge-required CaD clustering.

In conclusion, two approaches of deep clustering and CaD clustering have the identical aim of mapping, as stated in Definition \ref{def-requirements2}, yet the approaches differ as follows:
\begin{itemize}
    \item Deep Clustering emphasizes on its unique ability to learn a latent representation, assuming that the learned representation would enable clustering to produce simple clusters in the form of centroids in the mapped space to represent complex clusters in input space. Deep clustering employs a complicated deep learning method to learn the latent representation mapping. As a result, it is unclear (from the loss function) how the learned mapping can achieve the aim of mapping and what each centroid means.
    
    The emphasis can be construed as misplaced because the learned representation has not been shown to be suitable for clustering. This can be viewed as a legacy of Definition \ref{def-typical} which focuses on similarity between points without understanding the nature of clusters (as distributions) in the input space. 
    \item CaD Clustering focuses on the most direct way to represent each cluster as a distribution using a distributional kernel without optimization, and employs a simple greedy search to optimize the clustering objective, without representation learning or additional prior knowledge. The distributional kernel is used to represent each cluster in order to achieve the stated aim of mapping---each centroid in the mapped space represents a cluster in the input space. The entire CaD clustering process is easy to understand from the objective function,  and it can be completed in linear time.
    
    This is a direct outcome of Definition \ref{def-distribution-clustering} which stresses that points in a cluster are i.i.d. samples from a distribution.
\end{itemize}

\section{Performance Analysis on High-Dimensional Datasets}
\label{sec:analysis_on_high_dimension}

Many deep clustering approaches \cite{ijcai2017p243, shah2018deepcontinuousclustering, zhou2024comprehensive, ren2024deep} are built on the hypothesis that traditional clustering algorithms fail in high-dimensional settings due to unreliable similarity measures. To mitigate this, they employ neural networks, such as autoencoders, to learn low-dimensional embeddings. While these methods show promise on certain benchmarks like MNIST, our broader analysis across diverse high-dimensional domains challenges the universal necessity of such representation learning.

As presented in Table \ref{tab:image_dataset_run}, we evaluate representative algorithms on various high-dimensional image datasets. The results reveal that deep clustering methods, such as IDEC and CC, do not consistently outperform $k$-means; in fact, CC's performance collapses on datasets like ImageNet-10 and ImageNet-Dogs. In contrast, our Cluster-as-Distribution (CaD) method, KBC, achieves superior results across nearly all cases, particularly on the COIL-20 dataset (1024 dimensions), where it reaches a near-perfect NMI of 0.98.

This trend of CaD superiority is even more evident in the biological domain. Table \ref{tab:image_dataset_run} summarizes the performance on single-cell and spatial transcriptomics datasets, which are characterized by high dimensionality (400 to 2,000 attributes) and complex noise patterns. In these scenarios, IDEC and CC often fail to surpass, or even underperform, basic $k$-means. For instance, on the \textit{tutorial} dataset, both IDEC and CC yield near-zero NMI, whereas KBC achieves a robust 0.87.

Furthermore, the robustness of KBC is validated on the \textit{w100Gaussians} synthetic dataset (Table \ref{tab:clustering_results_w100Gaussians}). This dataset consists of two 100-dimensional subspace clusters in a 200-dimensional space that only overlap at the origin. While $k$-means and deep clustering methods fail to correctly identify these subspace structures, KBC achieves perfect clustering.

In summary, these empirical results suggest that current deep clustering methods are not yet robust enough for high-dimensional data across different domains. Our CaD-based KBC demonstrates that, with an appropriate distribution-based clustering mechanism, high-dimensional challenges can be effectively addressed without relying on complex and often unstable representation learning components.

\begin{table*}[htbp]
\centering
\caption{Clustering results on high-dimensional scenarios. Each reported Normalized Mutual Information (NMI) \cite{strehl2002nmi} value is averaged over 10 runs; the clustering visualization result using t-SNE \cite{maaten2008tsne} is obtained from the run having the highest NMI.}
\label{tab:clustering_results_w100Gaussians}
\setlength{\tabcolsep}{1pt} 
\renewcommand{\arraystretch}{0.5}
\resizebox{\textwidth}{!}{
\begin{tabular}{c c c c c c c}
    \toprule
    & \small \textbf{Original Data} & \small \textbf{$k$-means} & \small \textbf{IDEC} & \small \textbf{CC} & \small \textbf{SpectralNet} & \small \textbf{KBC} \\
    \midrule
    
    \rotatebox{90}{\footnotesize w100Gaussians} & 
    \includegraphics[width=0.16\textwidth]{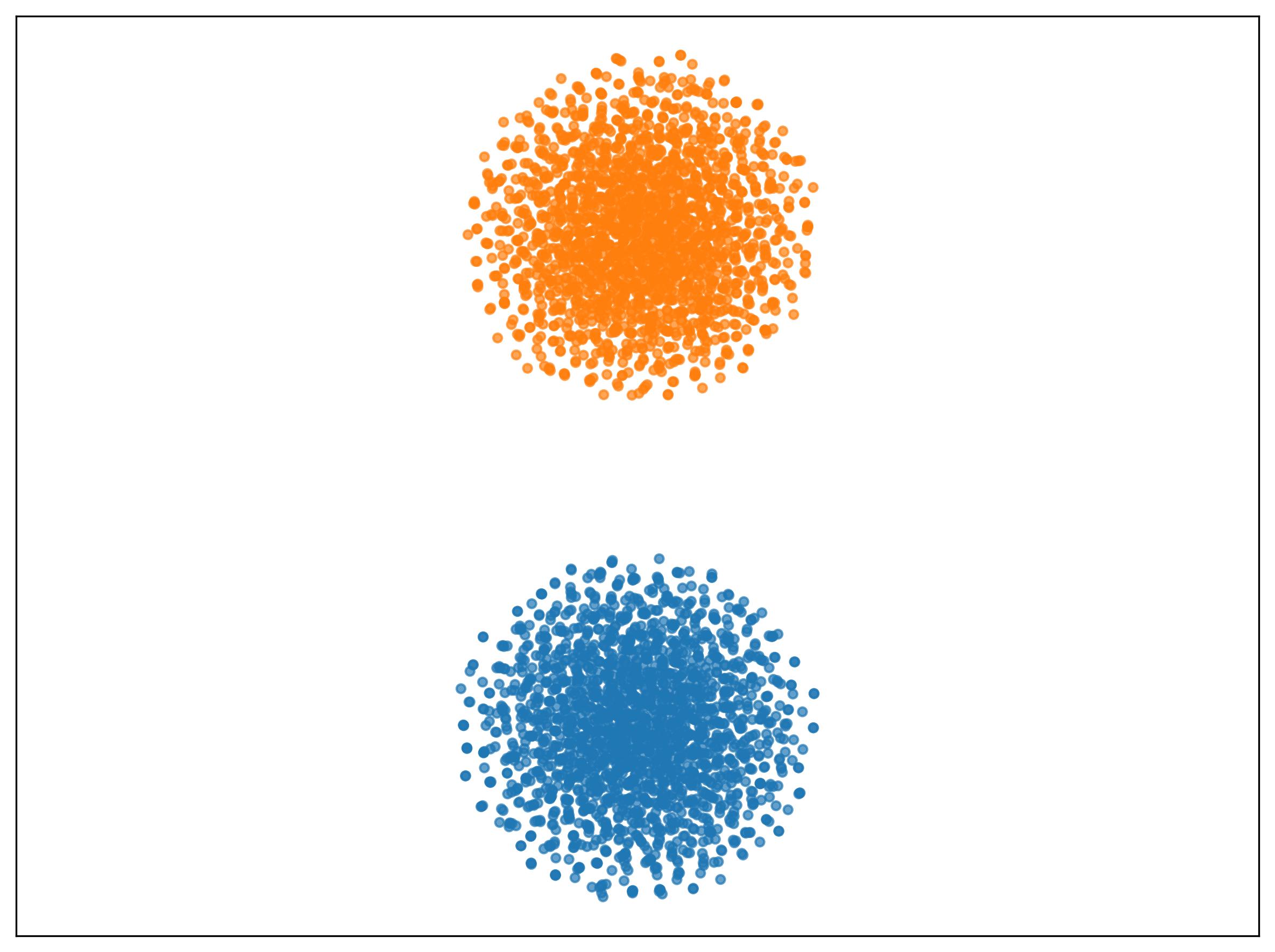} &
    \includegraphics[width=0.16\textwidth]{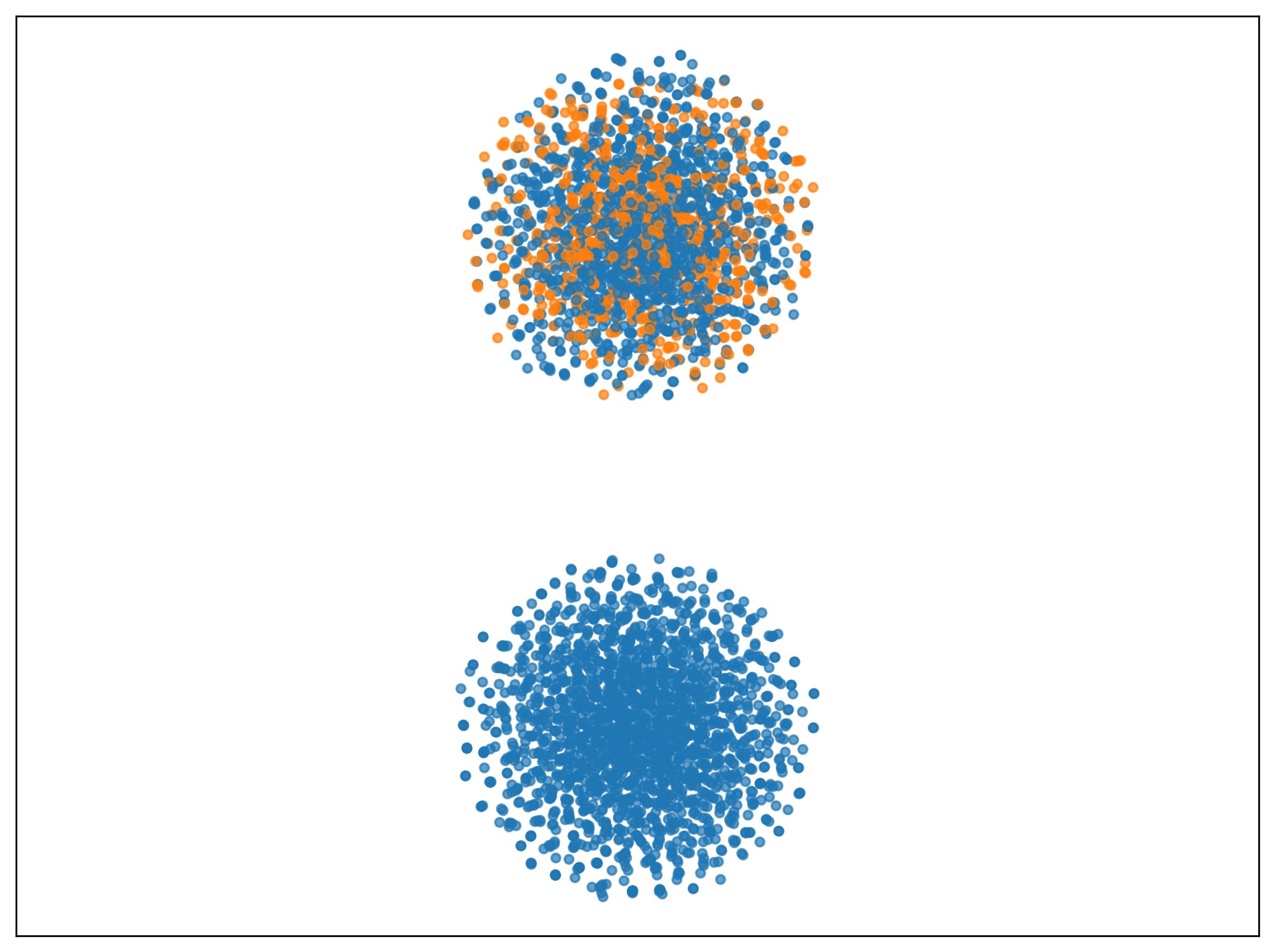} &
    \includegraphics[width=0.16\textwidth]{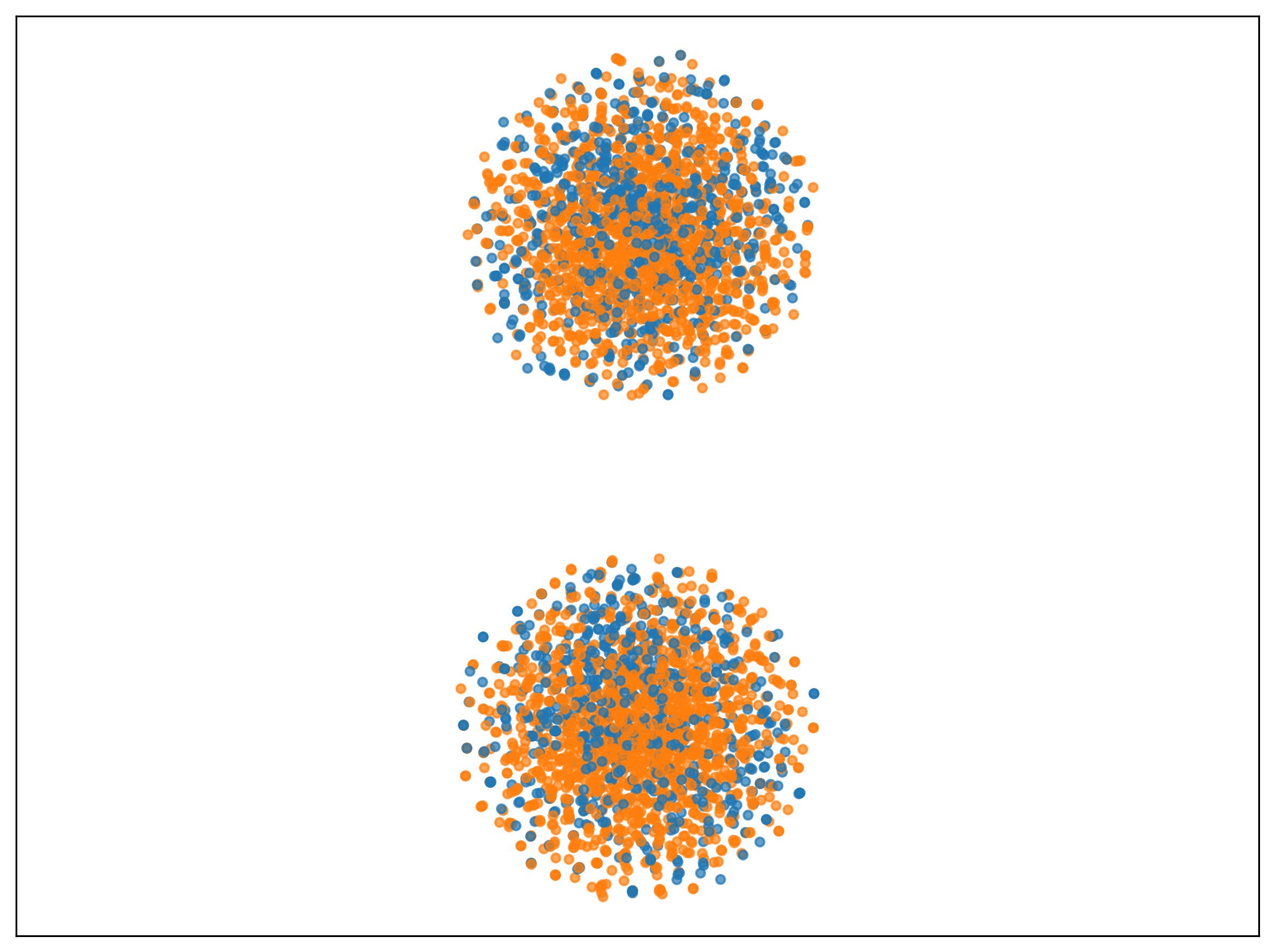} &
    \includegraphics[width=0.16\textwidth]{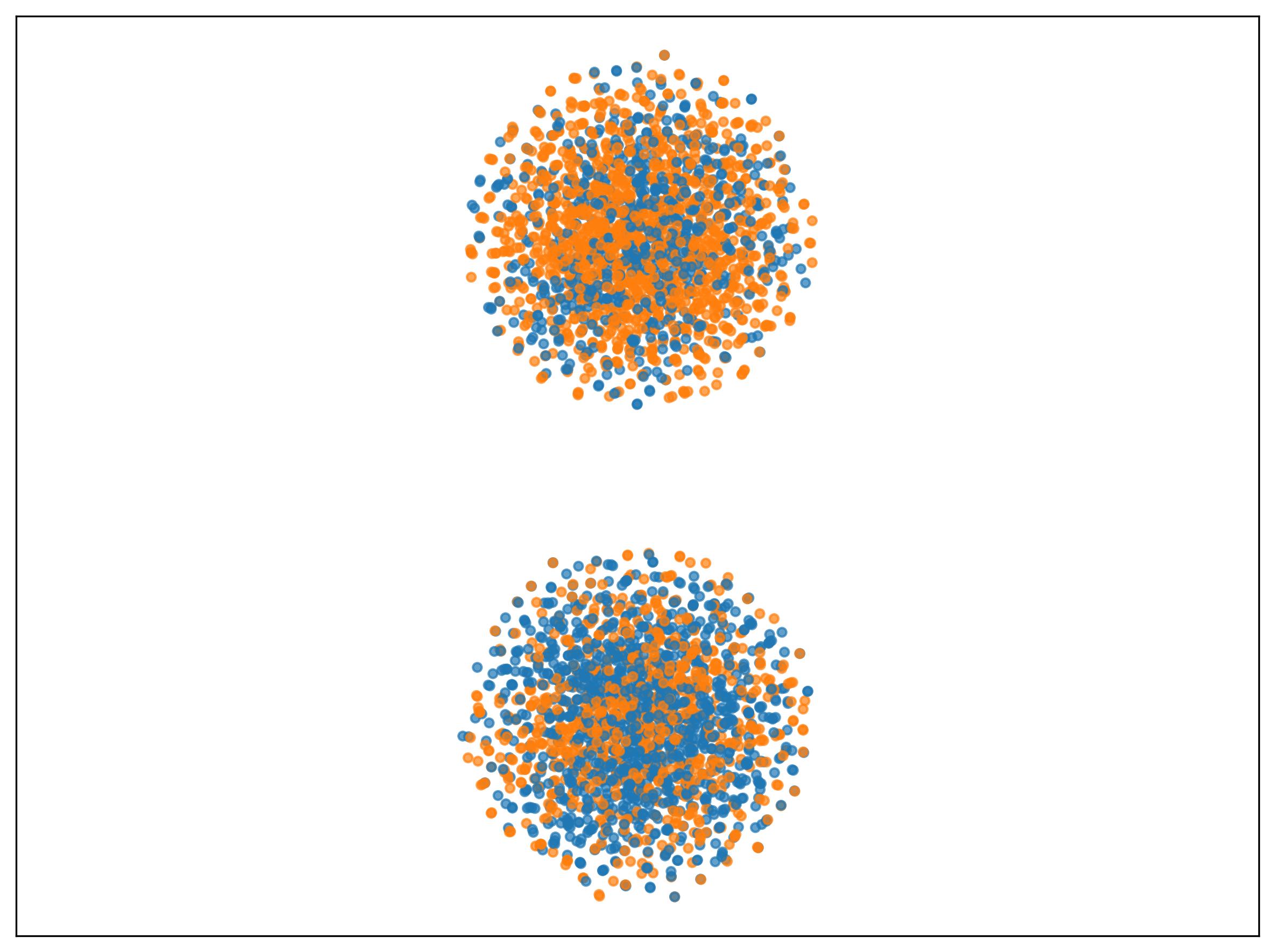} &
    \includegraphics[width=0.16\textwidth]{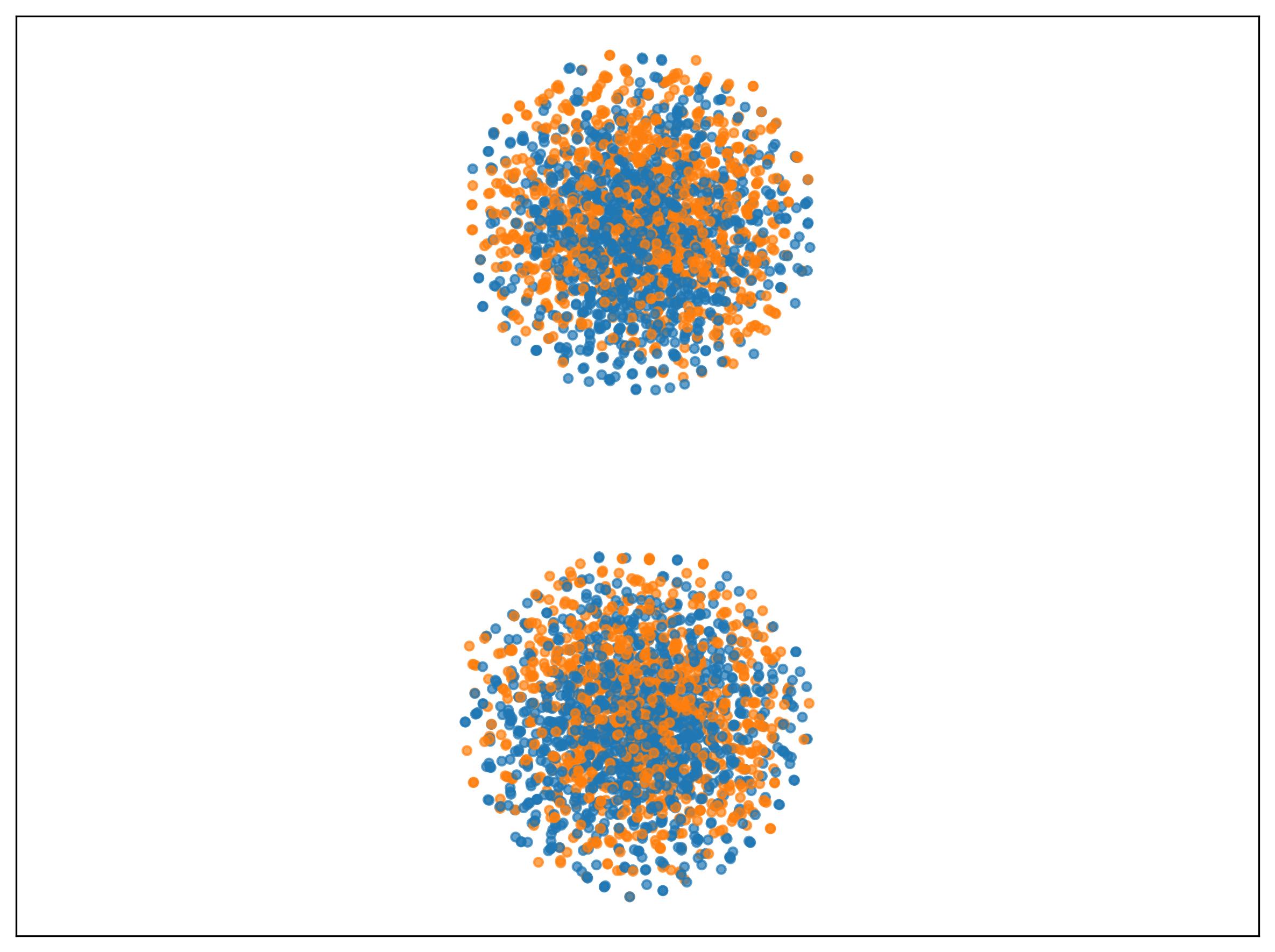} & 
    \includegraphics[width=0.16\textwidth]{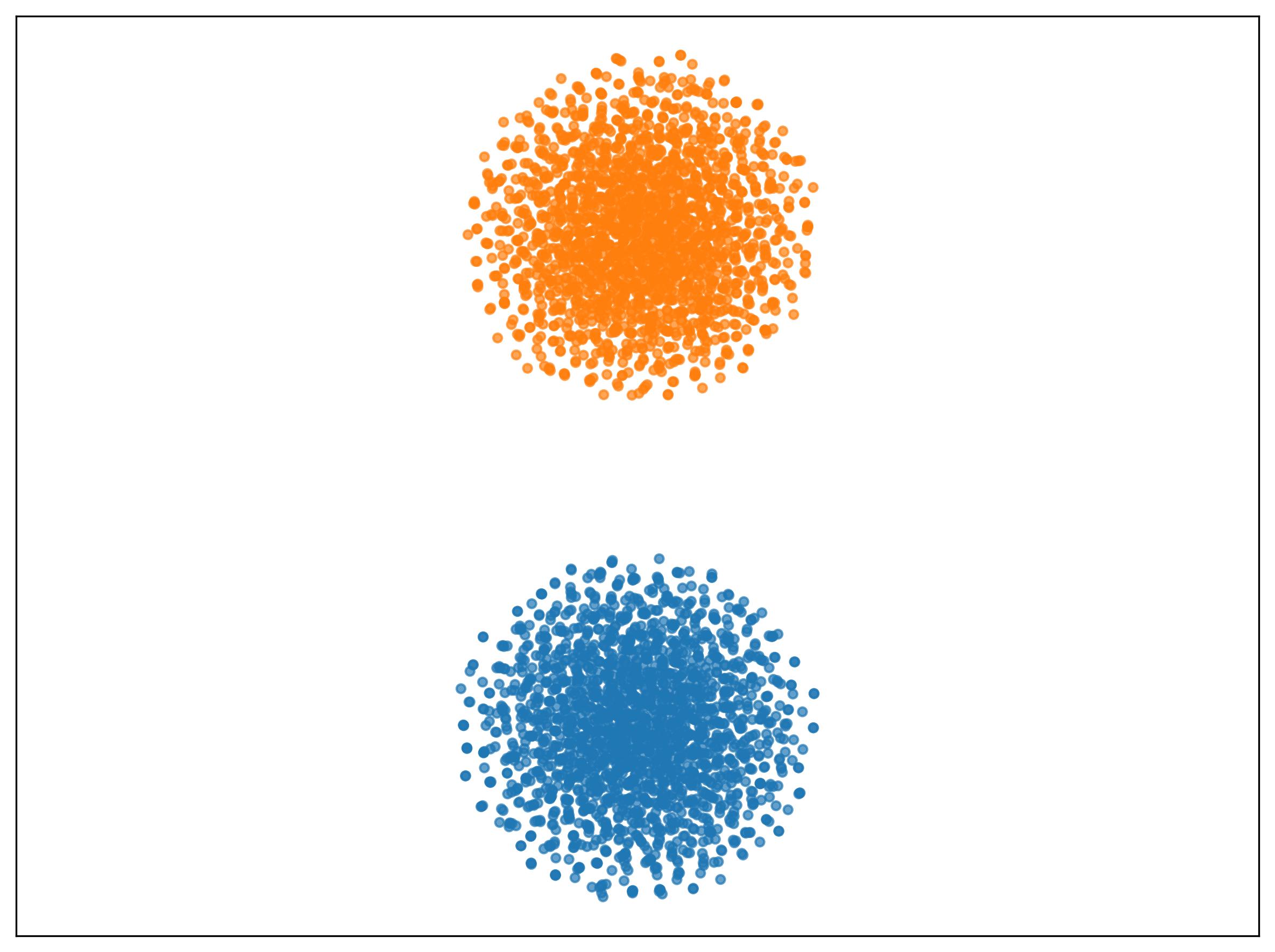} \\
    & & \footnotesize NMI=0.16 & \footnotesize NMI=0.00 & \footnotesize NMI=0.04 & \footnotesize NMI=0.00 & \footnotesize NMI=1.00 \\

    \rotatebox{90}{\footnotesize tutorial} & 
    \includegraphics[width=0.16\textwidth]{img/datasets/tutorial.jpg} &
    \includegraphics[width=0.16\textwidth]{img/kmeans/tutorial.jpg} &
    \includegraphics[width=0.16\textwidth]{img/idec/tutorial/tutorial.jpg} &
    \includegraphics[width=0.16\textwidth]{img/cc/tutorial/tutorial.jpg} &
    \includegraphics[width=0.16\textwidth]{img/spectralnet/fig/tutorial/tutorial.jpg} & 
    \includegraphics[width=0.16\textwidth]{img/kbc/tutorial/tutorial.jpg} \\
    & & \footnotesize NMI=0.31 & \footnotesize NMI=0.01 & \footnotesize NMI=0.02 & \footnotesize NMI=0.04 & \footnotesize NMI=0.87 \\

    \rotatebox{90}{\footnotesize tonsil} & 
    \includegraphics[width=0.16\textwidth]{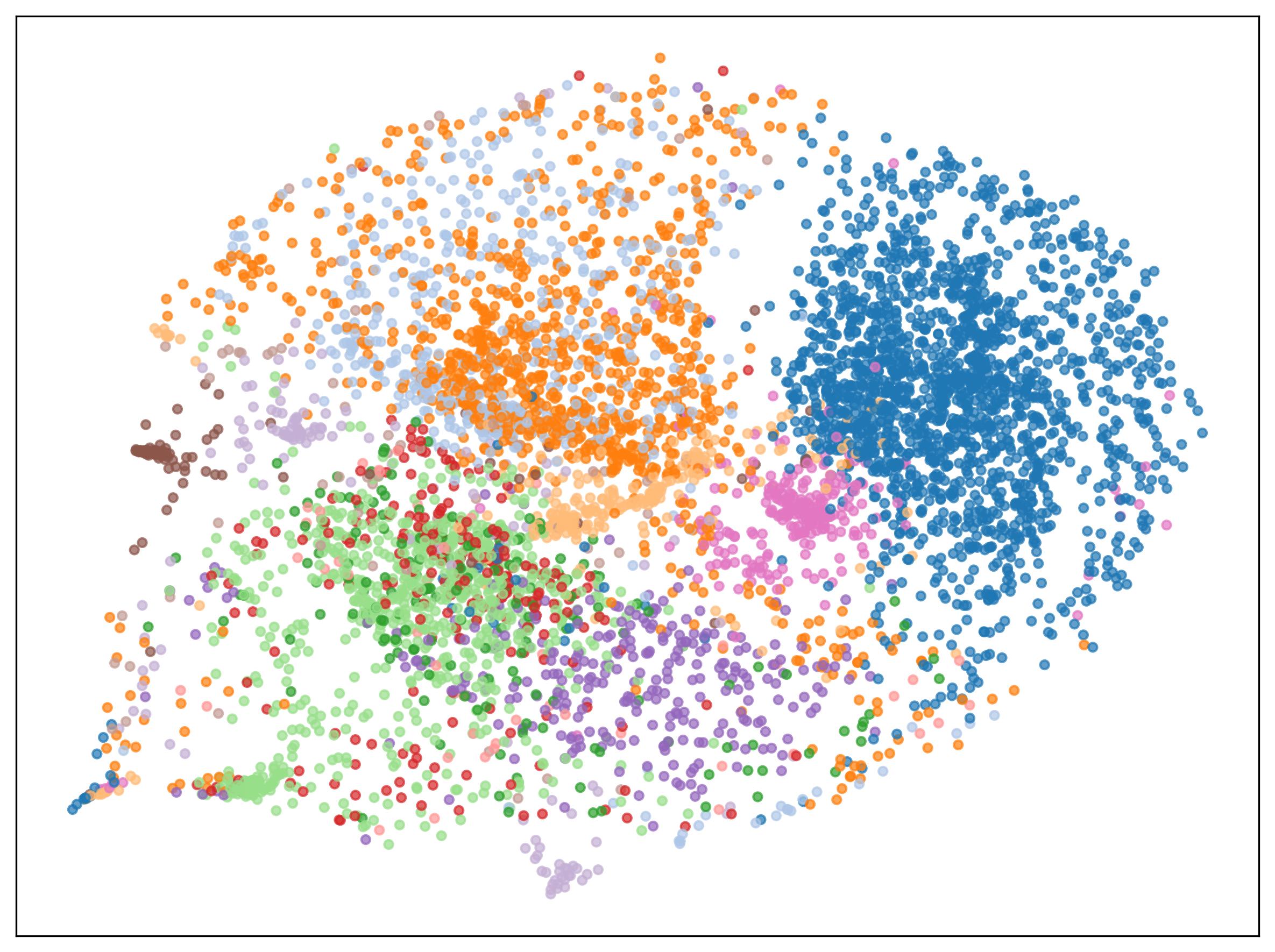} &
    \includegraphics[width=0.16\textwidth]{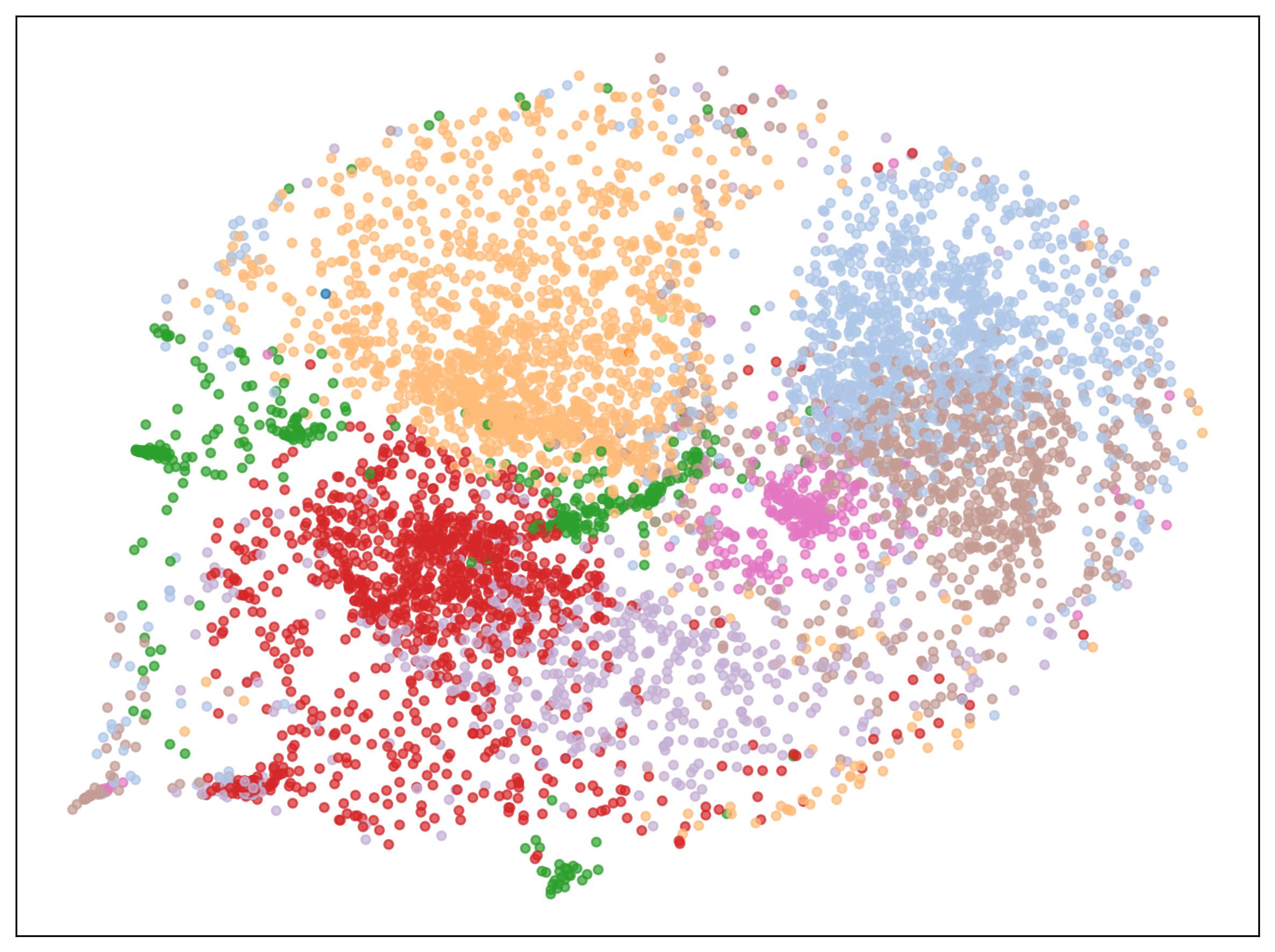} &
    \includegraphics[width=0.16\textwidth]{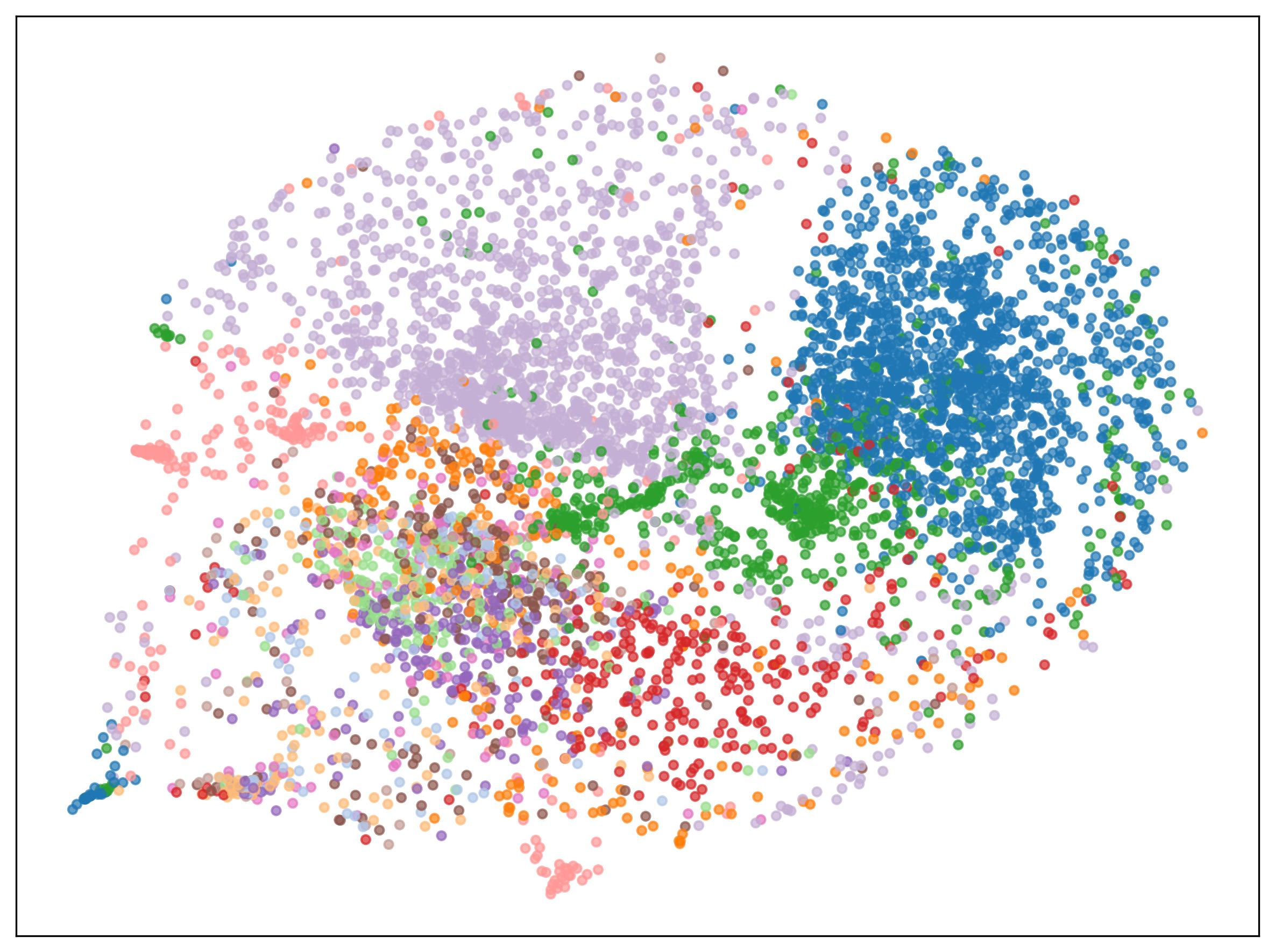} &
    \includegraphics[width=0.16\textwidth]{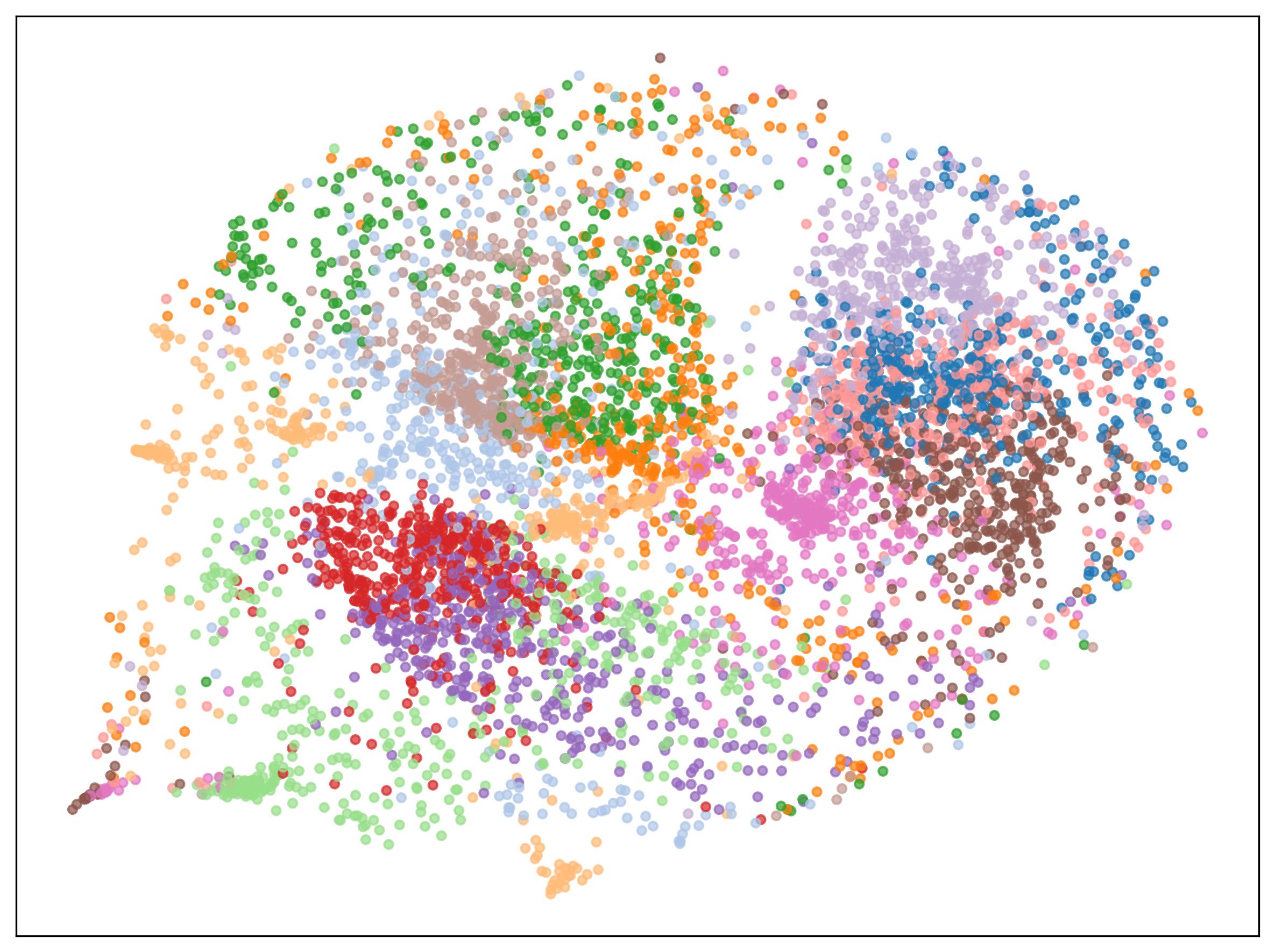} &
    \includegraphics[width=0.16\textwidth]{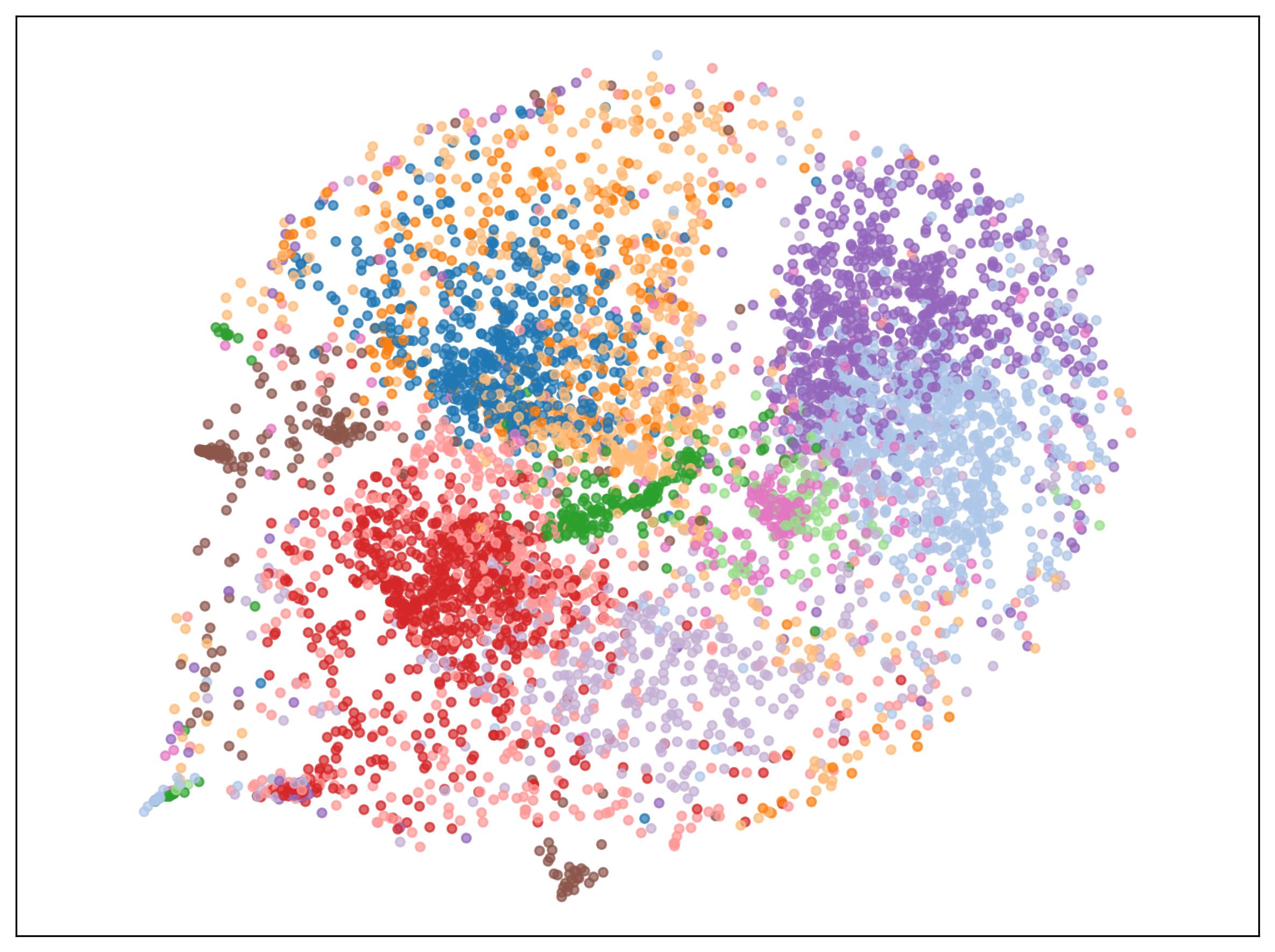} & 
    \includegraphics[width=0.16\textwidth]{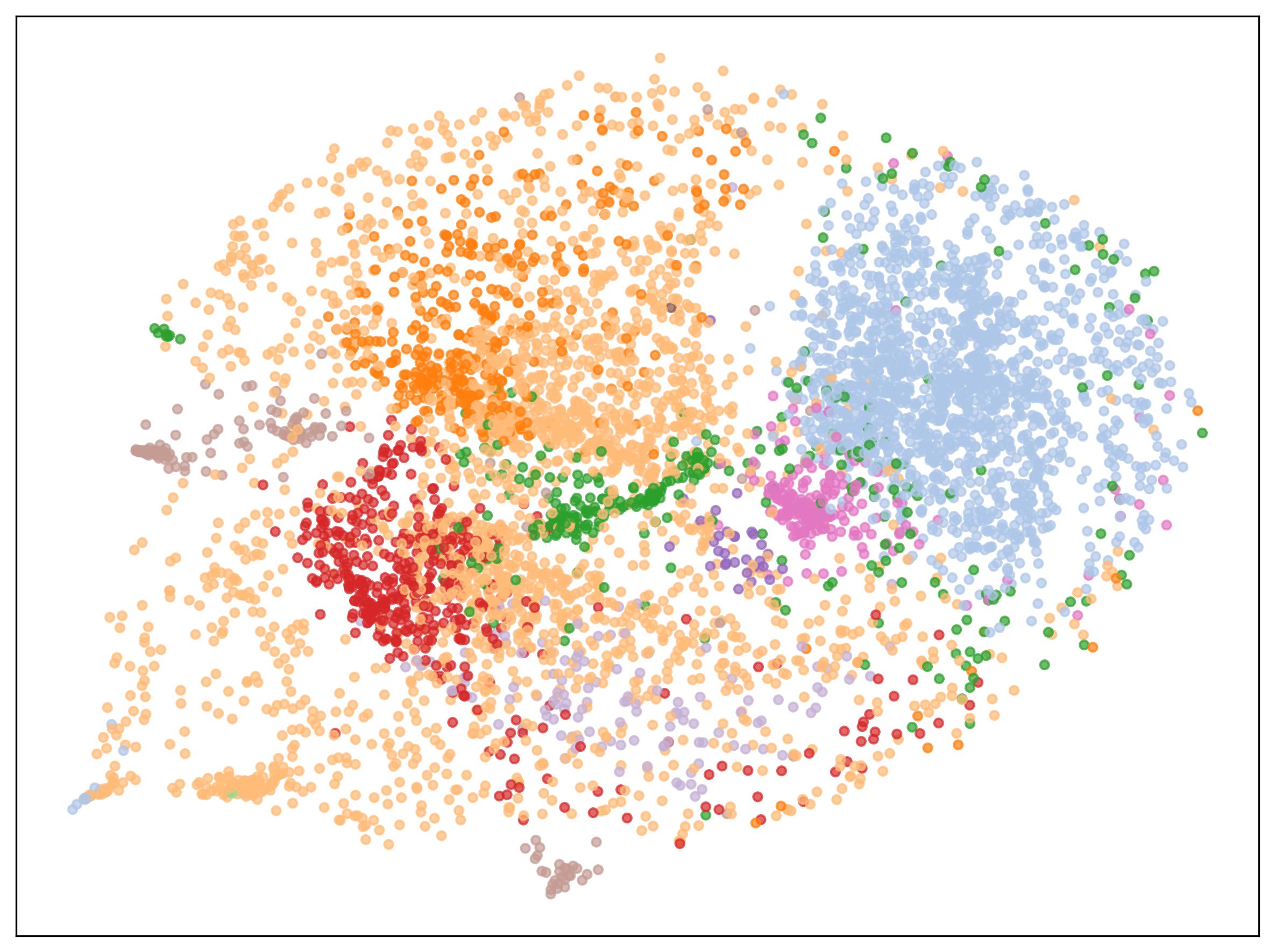} \\
    & & \footnotesize NMI=0.56 & \footnotesize NMI=0.63 & \footnotesize NMI=0.52 & \footnotesize NMI=0.60 & \footnotesize NMI=0.52 \\

    \rotatebox{90}{\footnotesize airway} & 
    \includegraphics[width=0.16\textwidth]{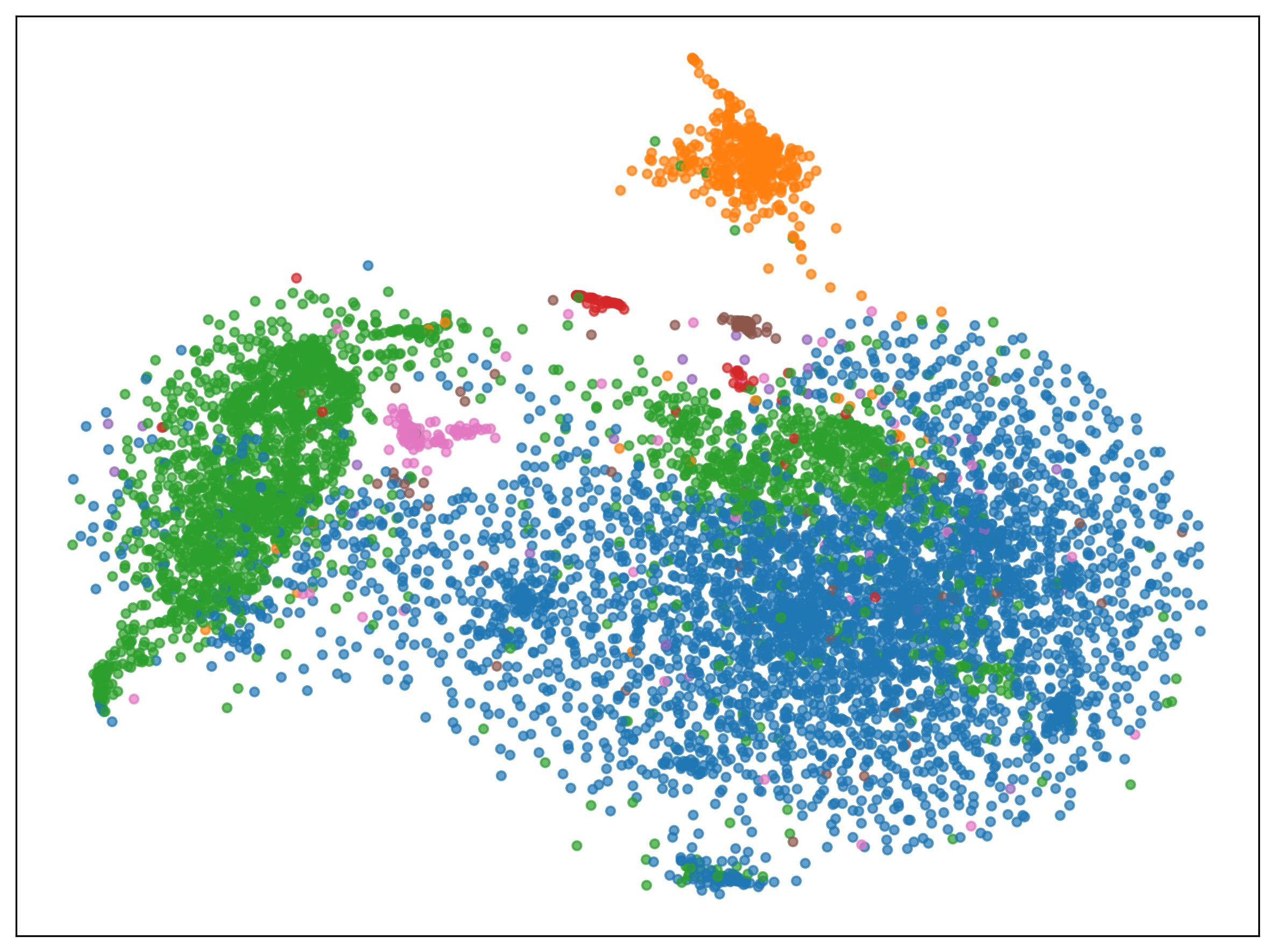} &
    \includegraphics[width=0.16\textwidth]{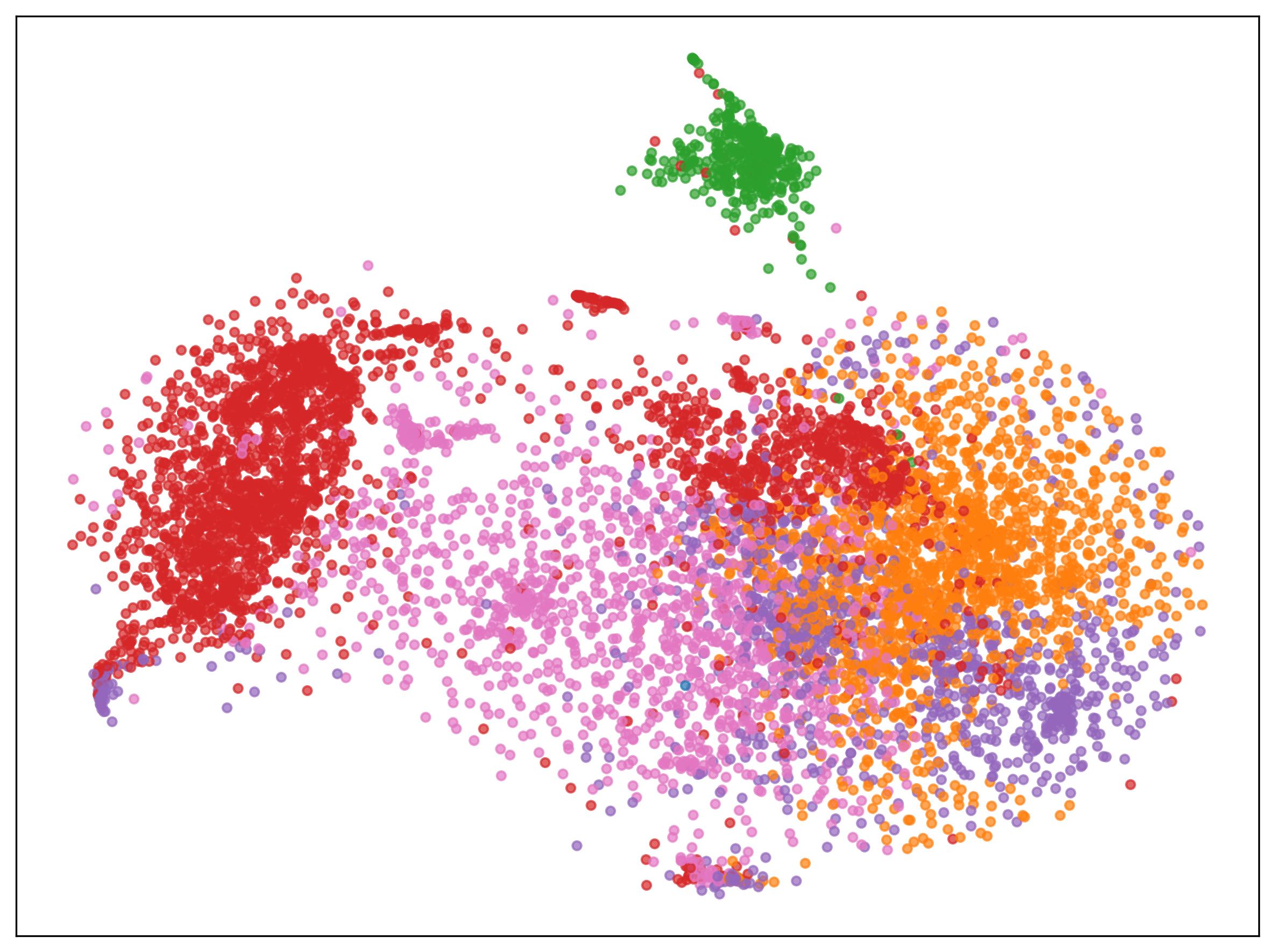} &
    \includegraphics[width=0.16\textwidth]{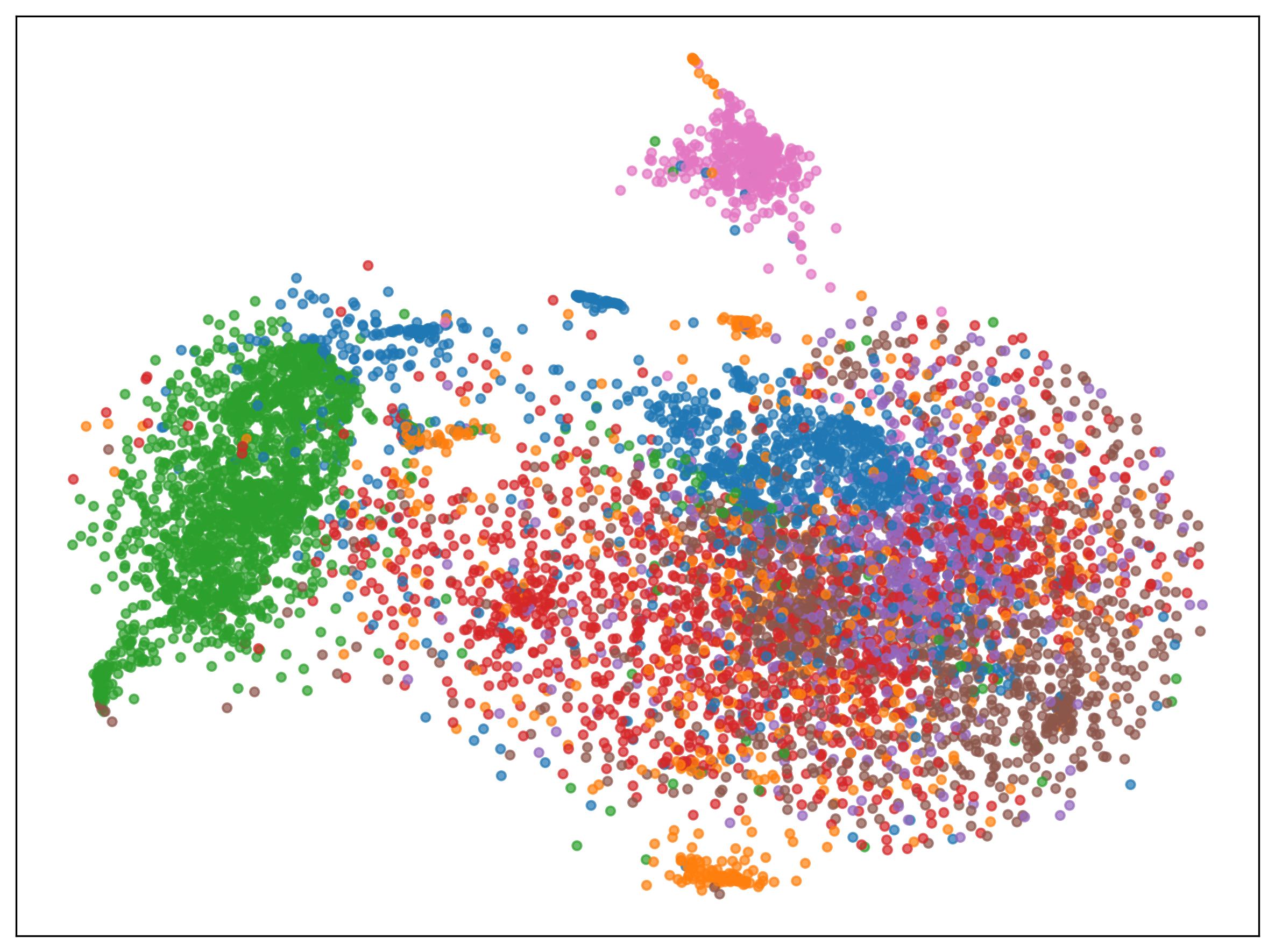} &
    \includegraphics[width=0.16\textwidth]{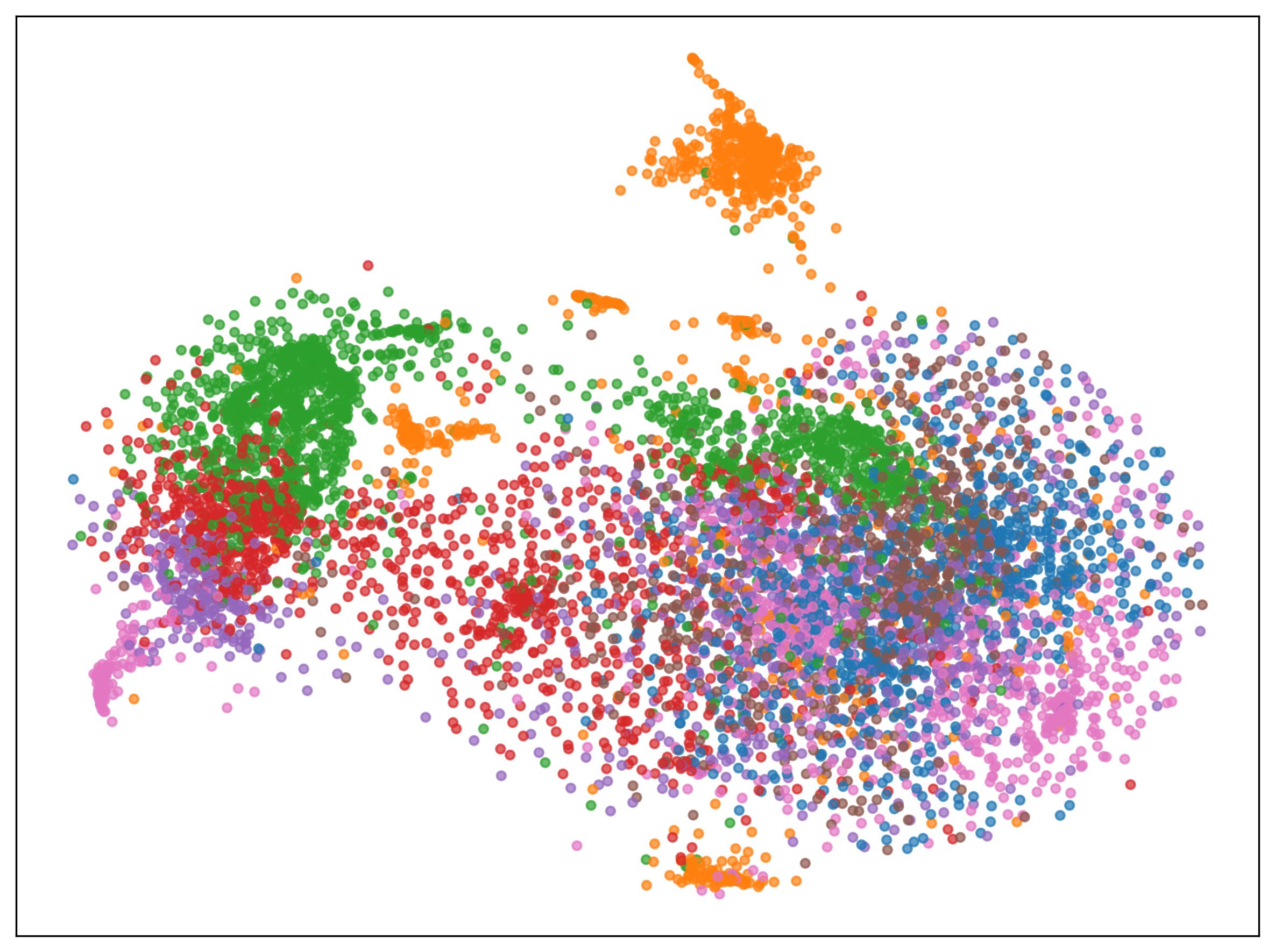} &
    \includegraphics[width=0.16\textwidth]{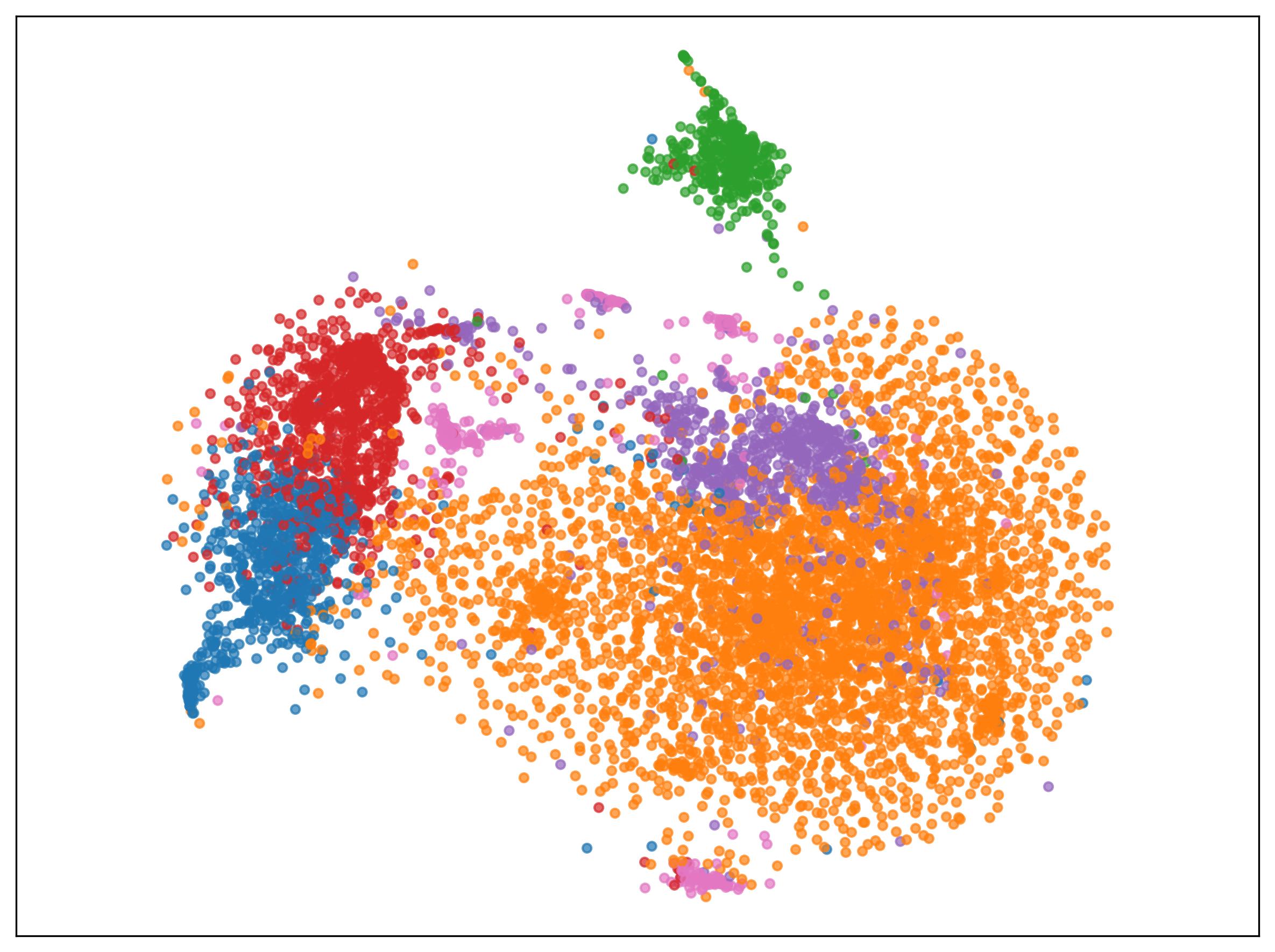} & 
    \includegraphics[width=0.16\textwidth]{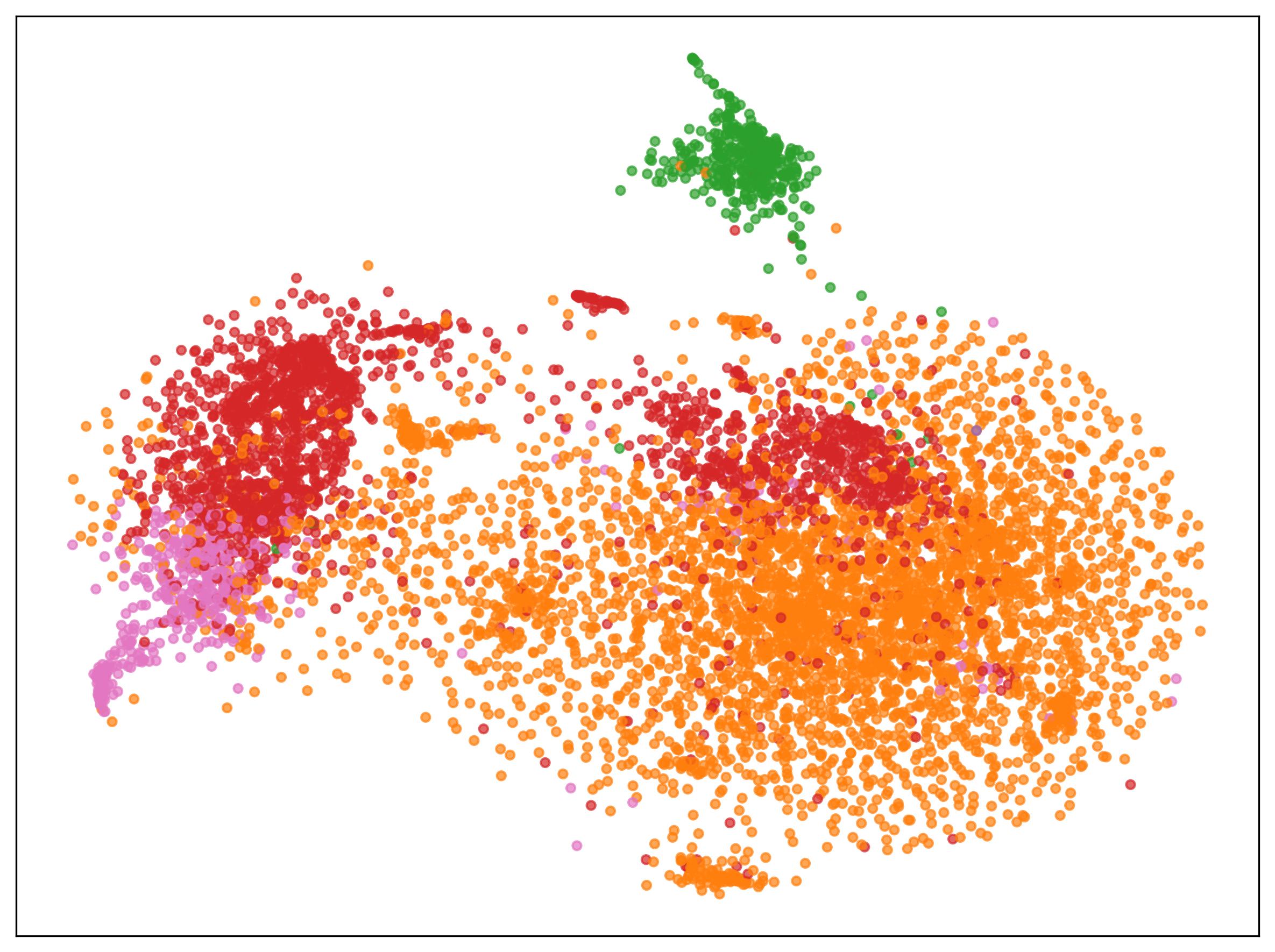} \\
    & & \footnotesize NMI=0.53 & \footnotesize NMI=0.46 & \footnotesize NMI=0.34 & \footnotesize NMI=0.63 & \footnotesize NMI=0.62 \\

    \rotatebox{90}{\footnotesize crohn} & 
    \includegraphics[width=0.16\textwidth]{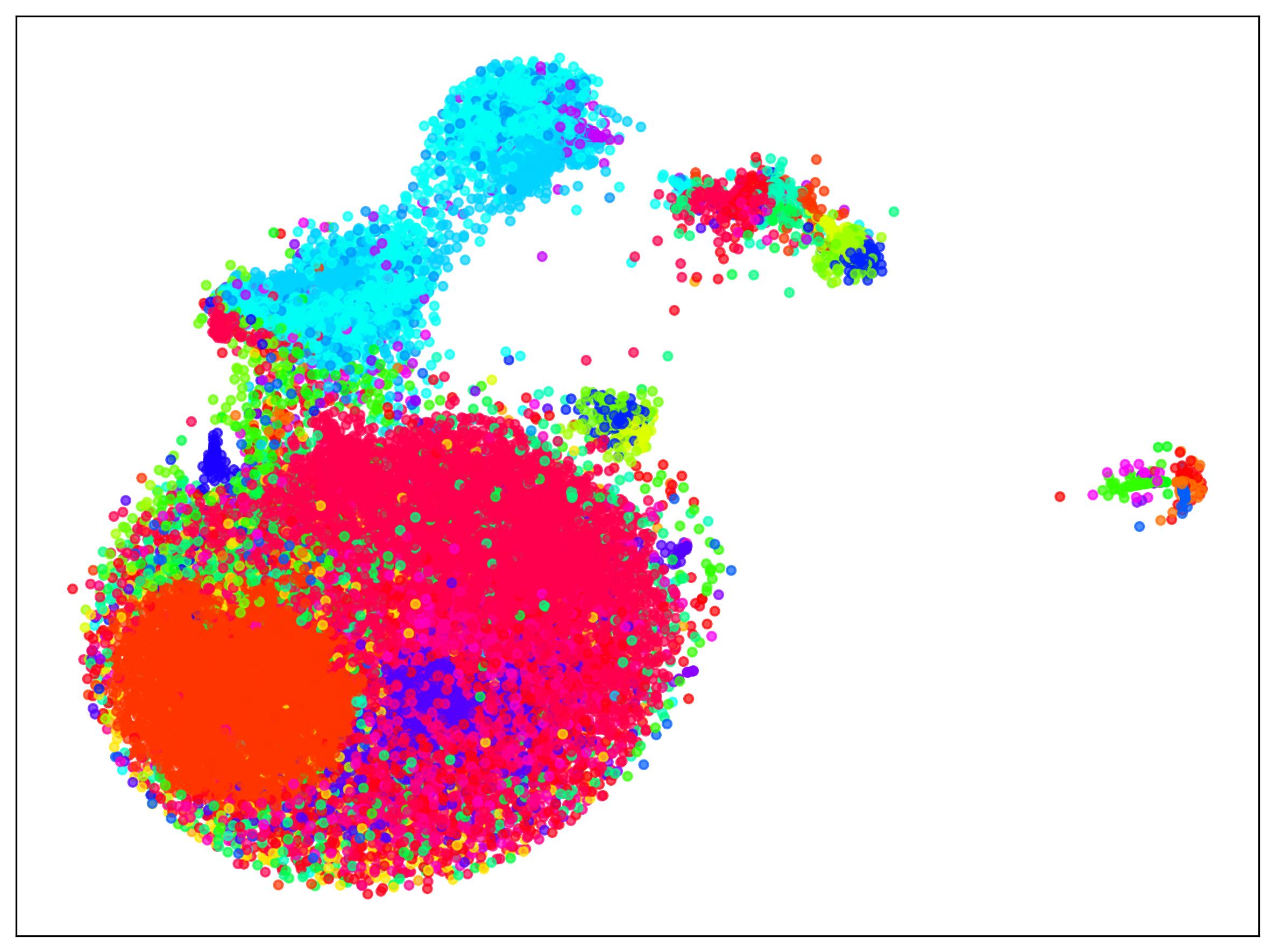} &
    \includegraphics[width=0.16\textwidth]{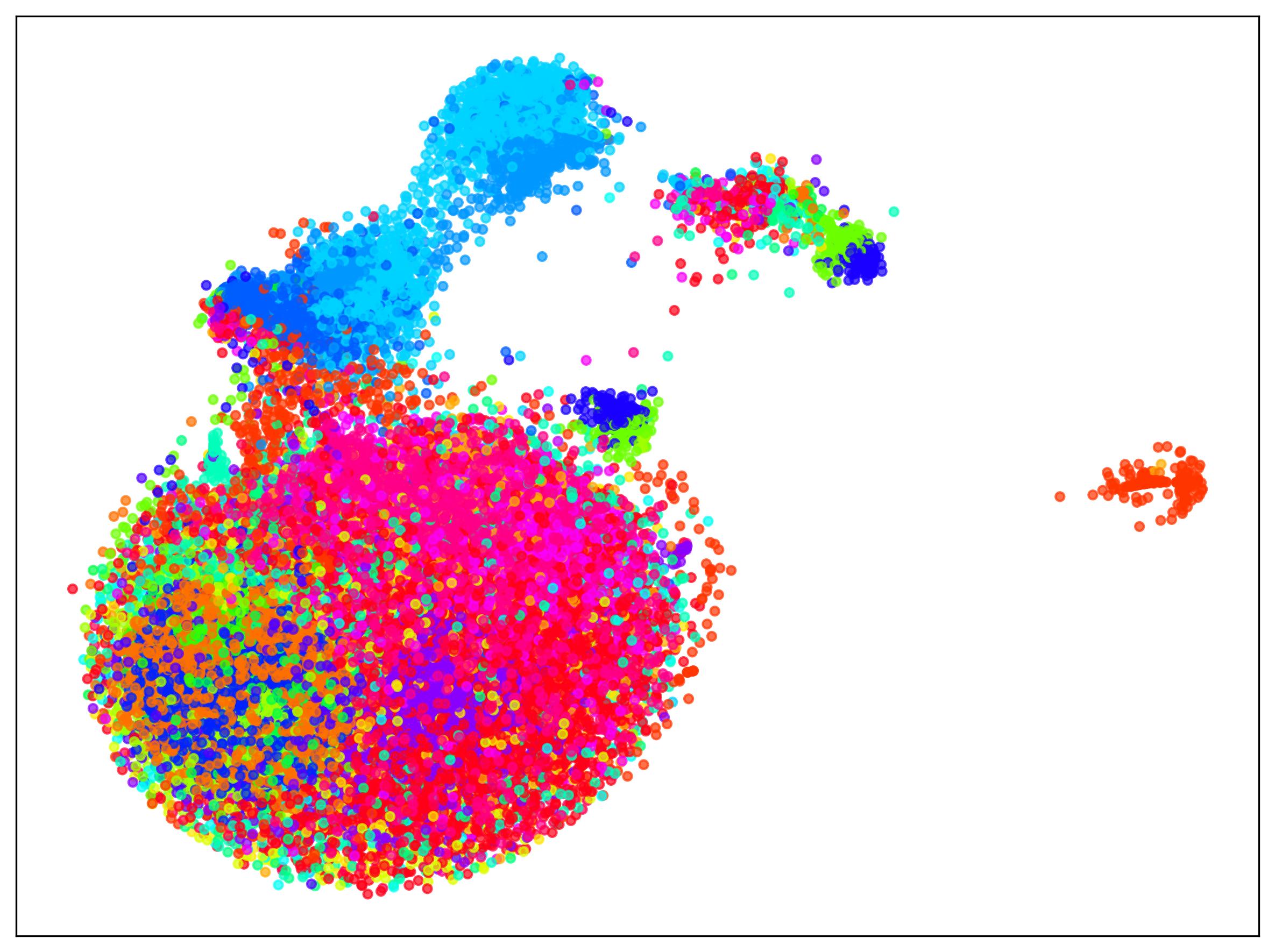} &
    \includegraphics[width=0.16\textwidth]{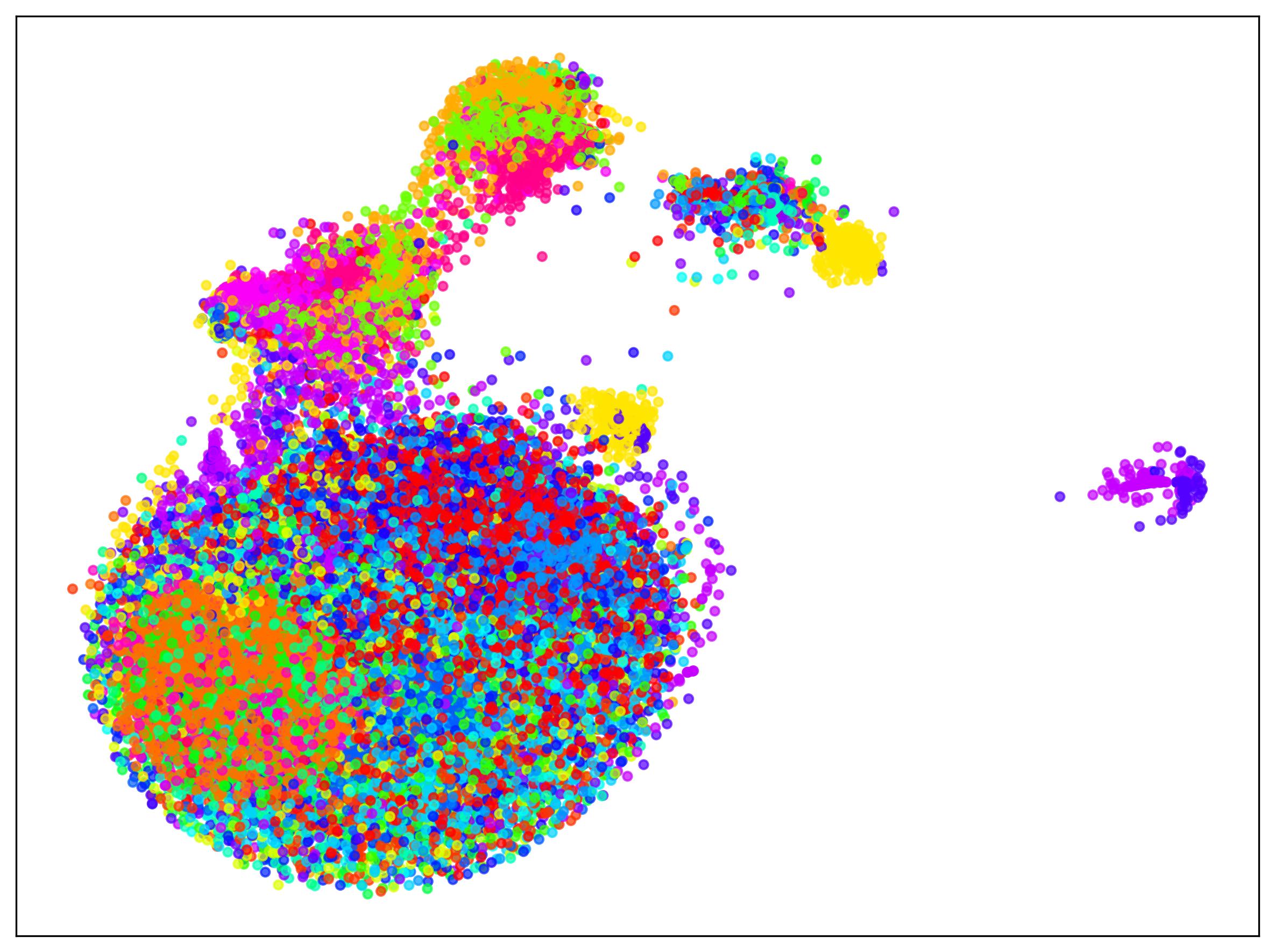} &
    \includegraphics[width=0.16\textwidth]{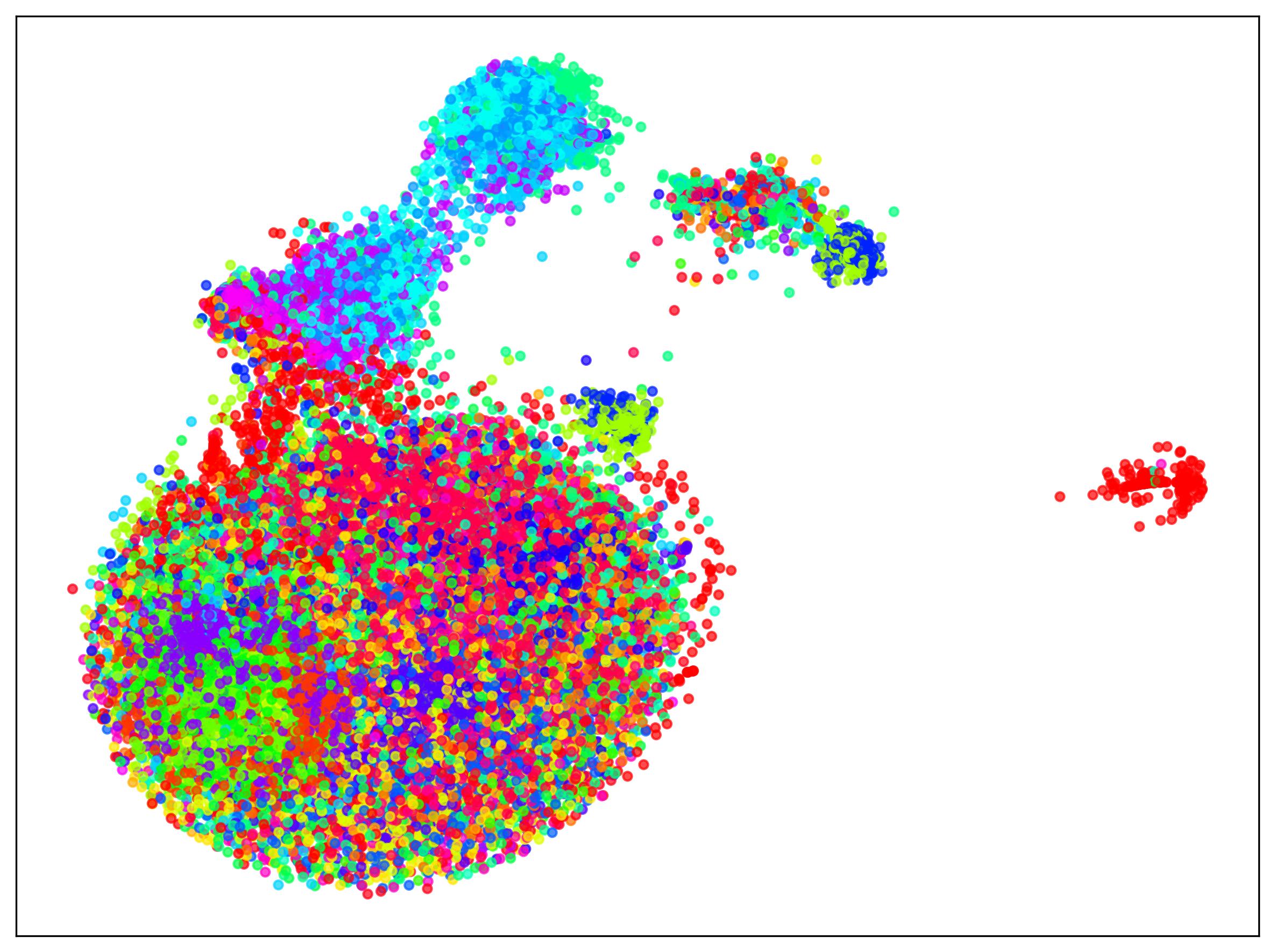} &
    \includegraphics[width=0.16\textwidth]{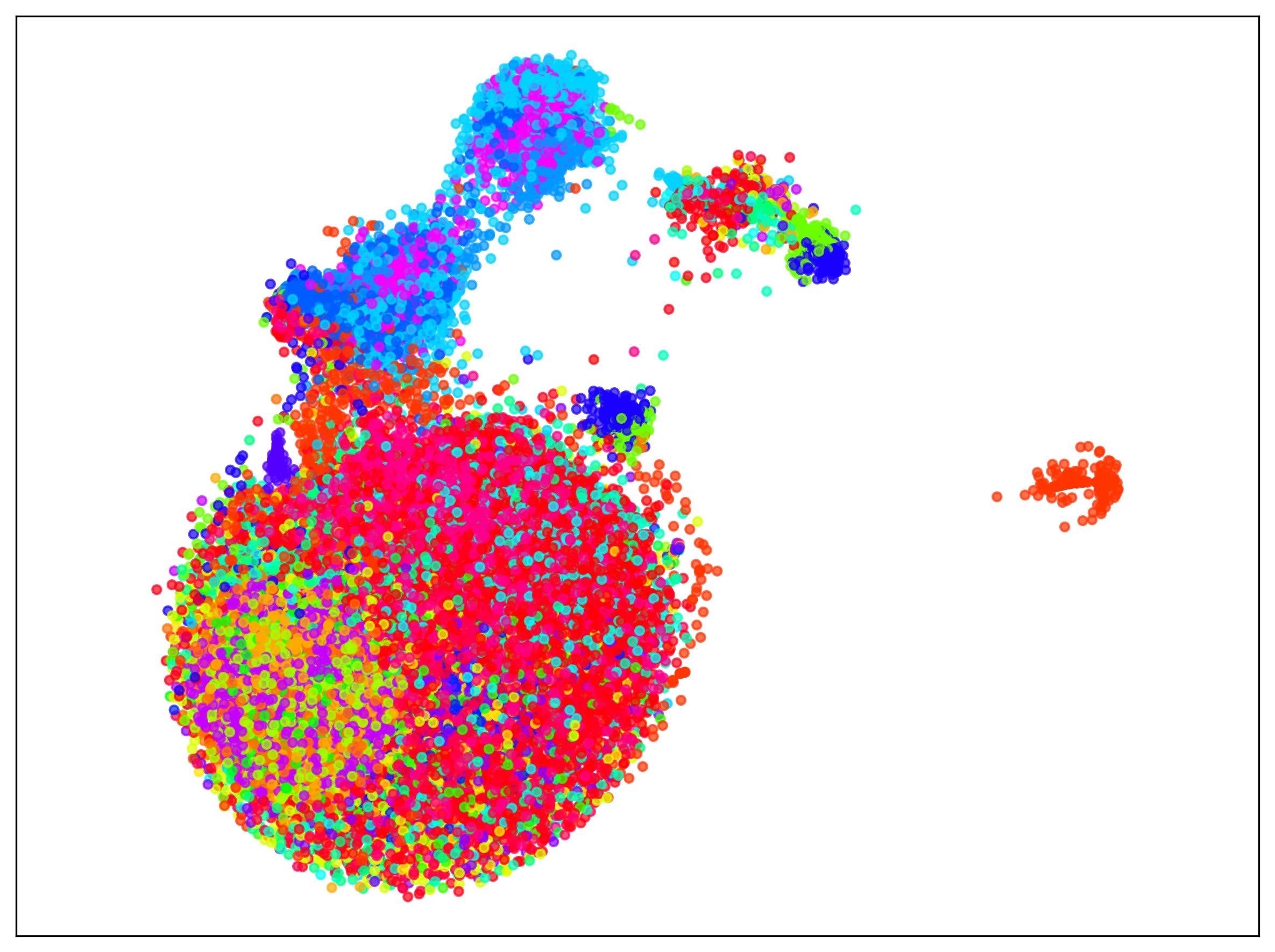} & 
    \includegraphics[width=0.16\textwidth]{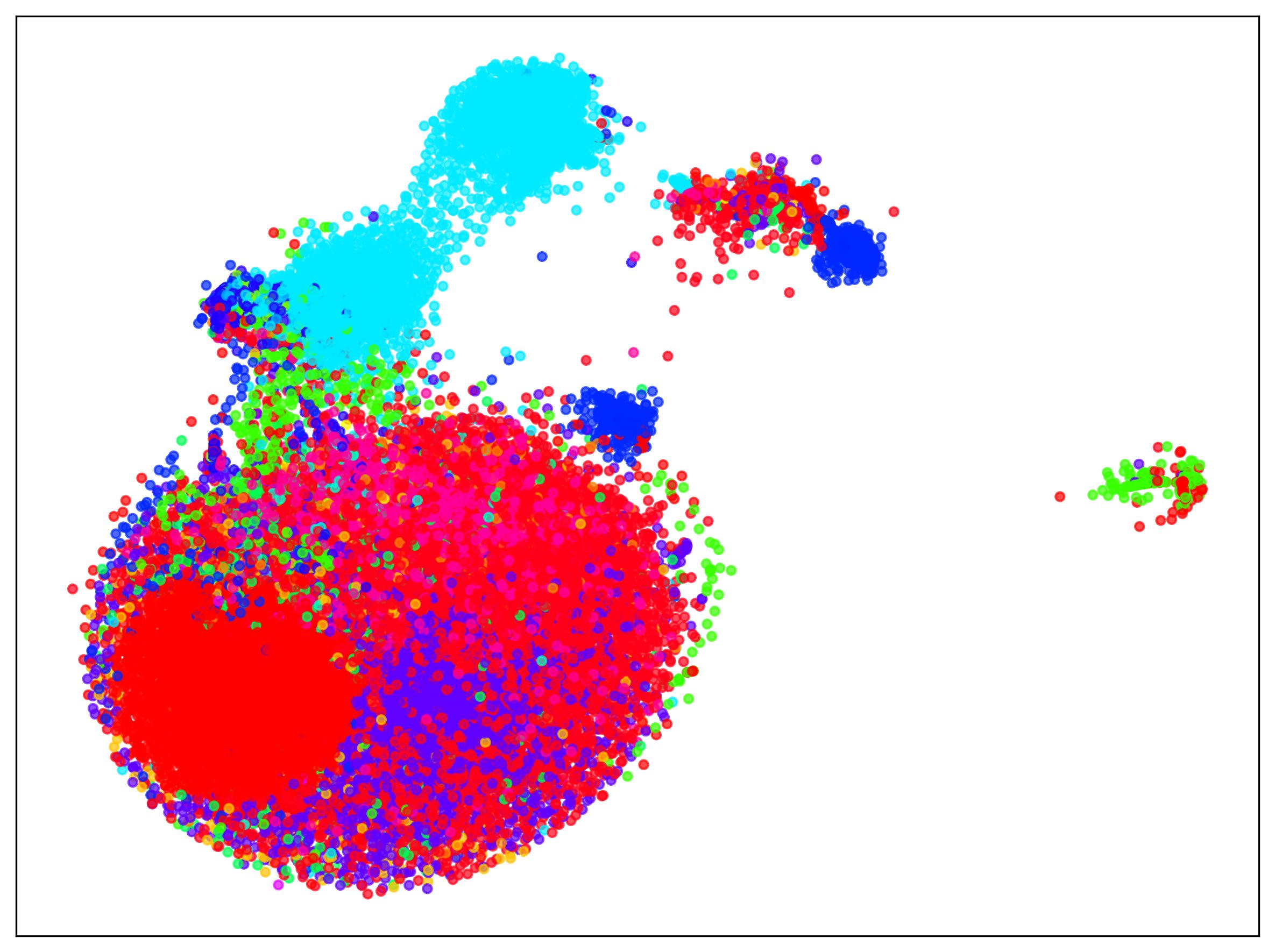} \\
    & & \footnotesize NMI=0.62 & \footnotesize NMI=0.55 & \footnotesize NMI=0.54 & \footnotesize NMI=0.59 & \footnotesize NMI=0.62 \\
    
    \bottomrule
\end{tabular}}
\end{table*}

\begin{table*}[htbp]
\centering
\caption{The clustering results on the 2Crescents dataset with different gap sizes. The proposed method (KBC) demonstrates consistent performance across varying data distributions.}
\label{tab:non_spherical_with_gap}
\setlength{\tabcolsep}{1pt} 
\renewcommand{\arraystretch}{0.2} 
\resizebox{\textwidth}{!}{
\begin{tabular}{c c c c c c c}
    \toprule
    & \small \textbf{Dataset} & \small \textbf{$k$-means} & \small \textbf{IDEC} & \small \textbf{CC} & \small \textbf{SpectralNet} & \small \textbf{KBC} \\
    \midrule

    \rotatebox{90}{\footnotesize Gap 0.5} & 
    \includegraphics[width=0.16\textwidth]{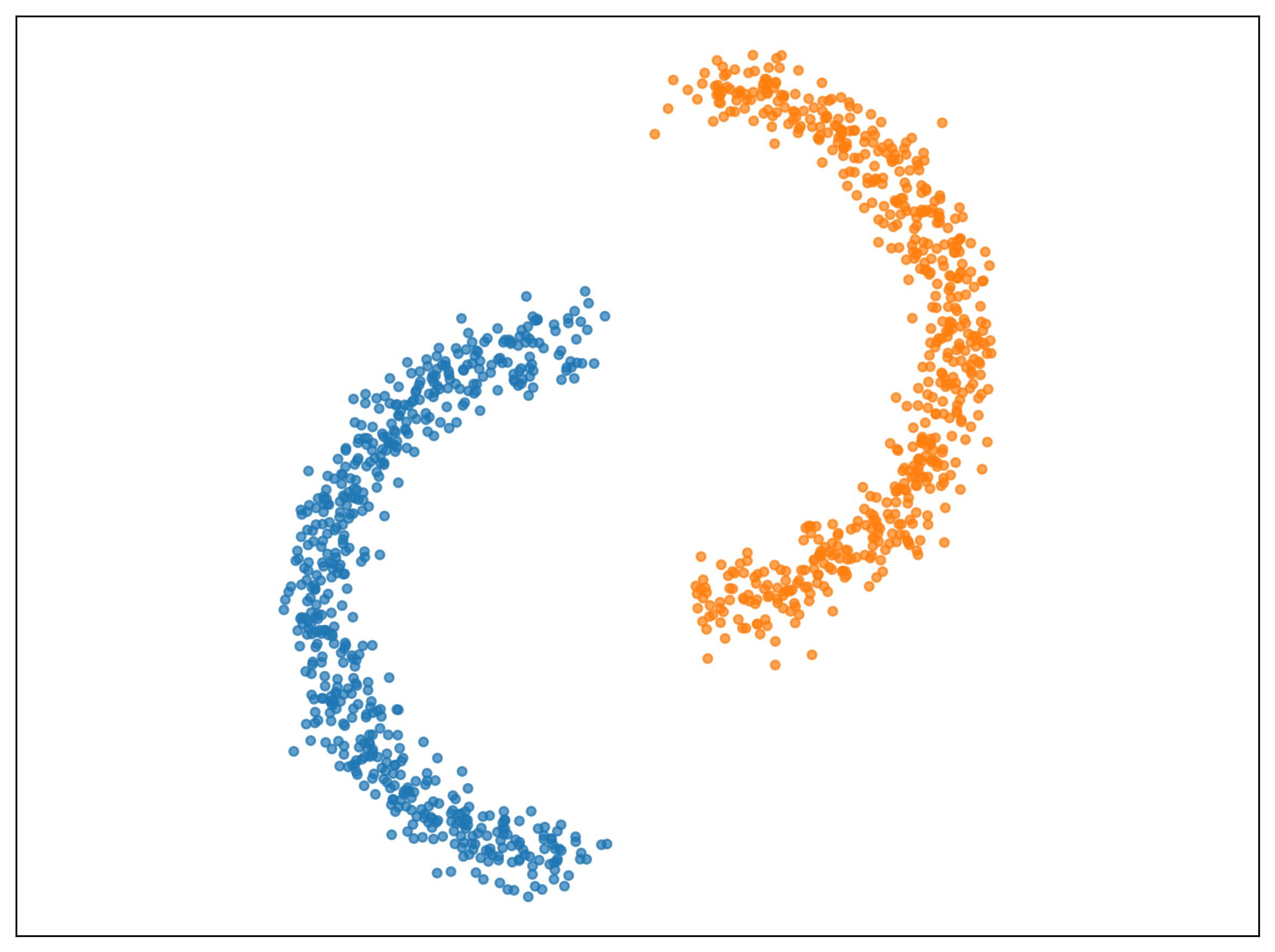} &
    \includegraphics[width=0.16\textwidth]{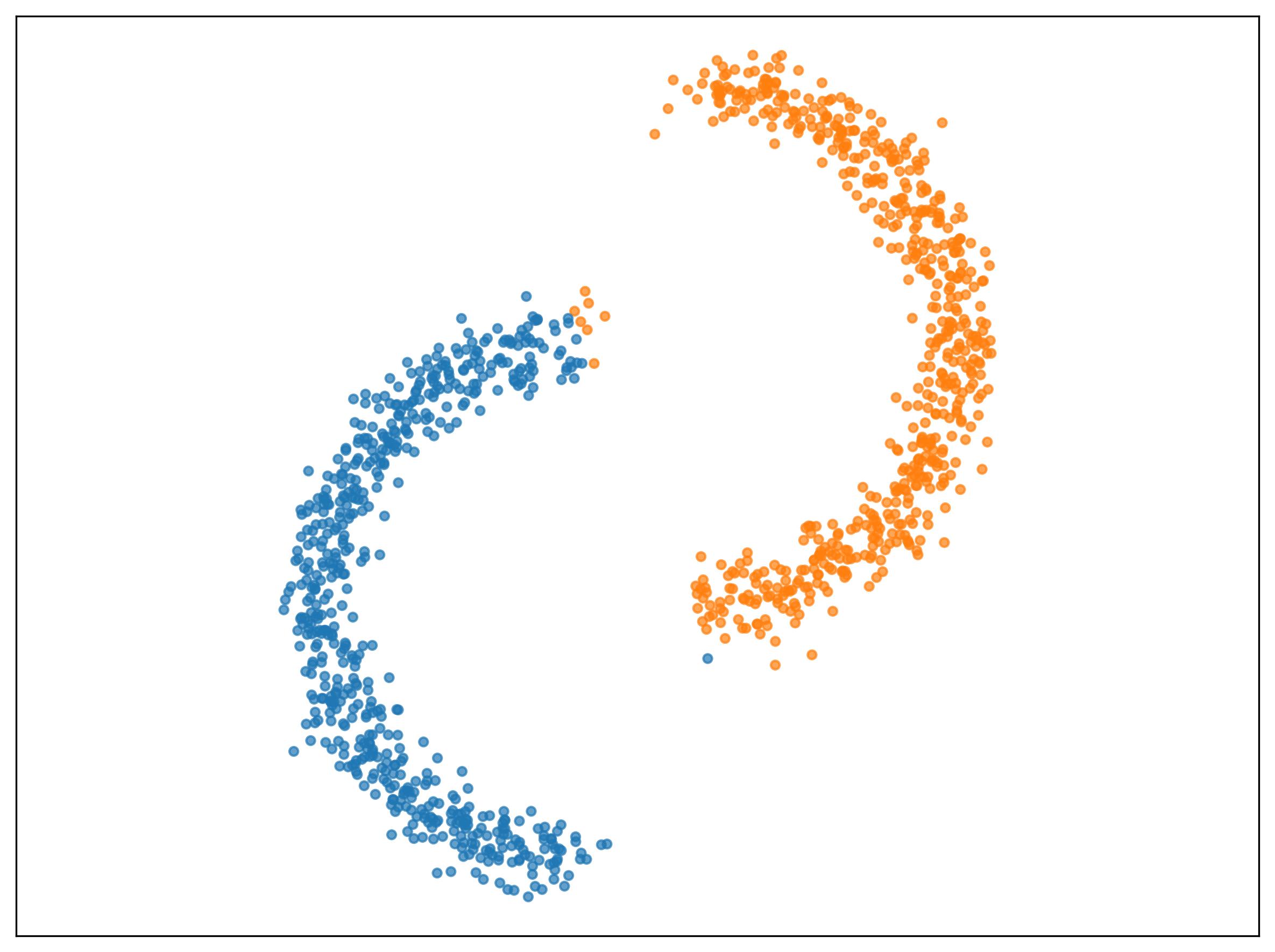} &
    \includegraphics[width=0.16\textwidth]{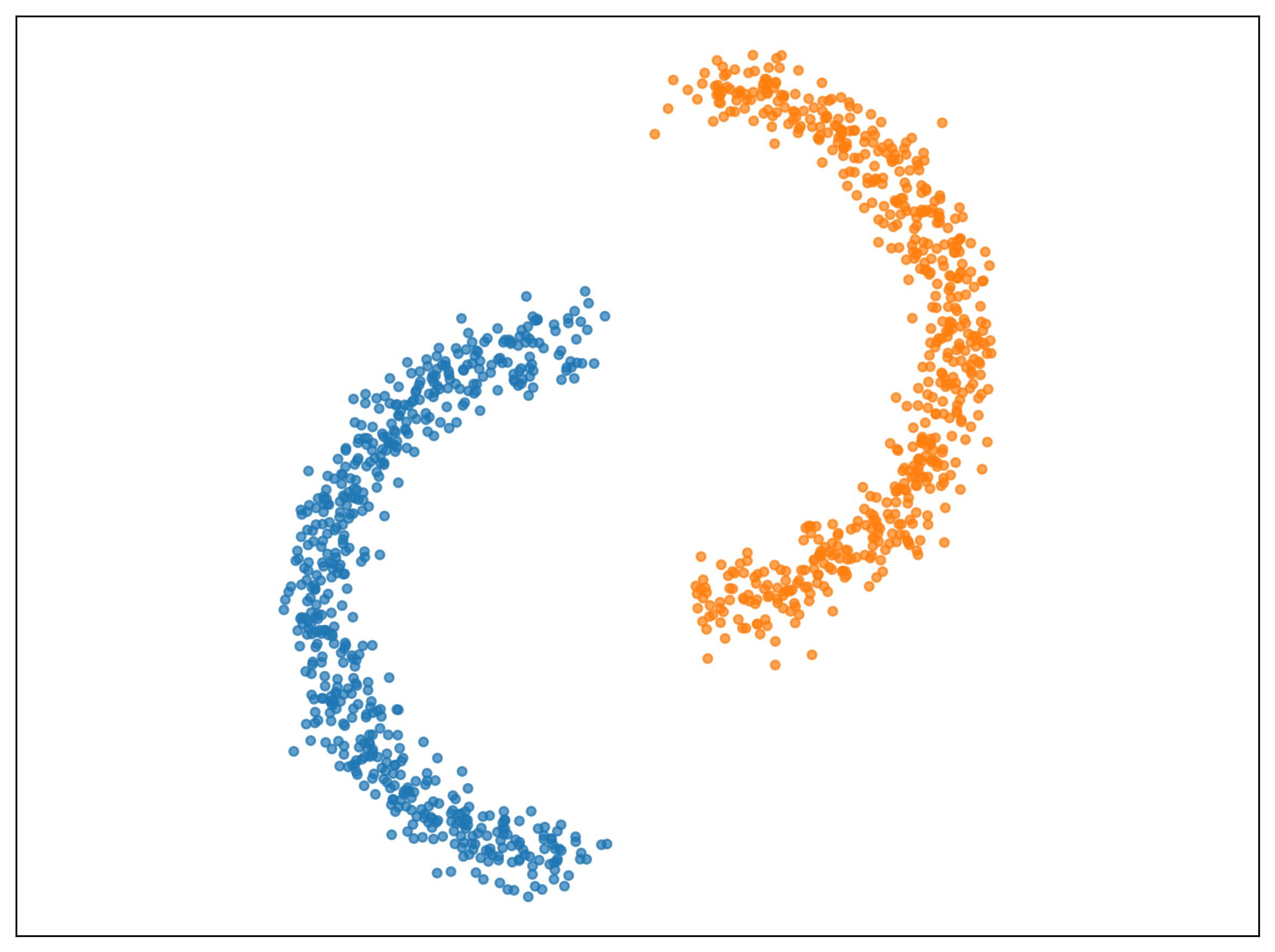} &
    \includegraphics[width=0.16\textwidth]{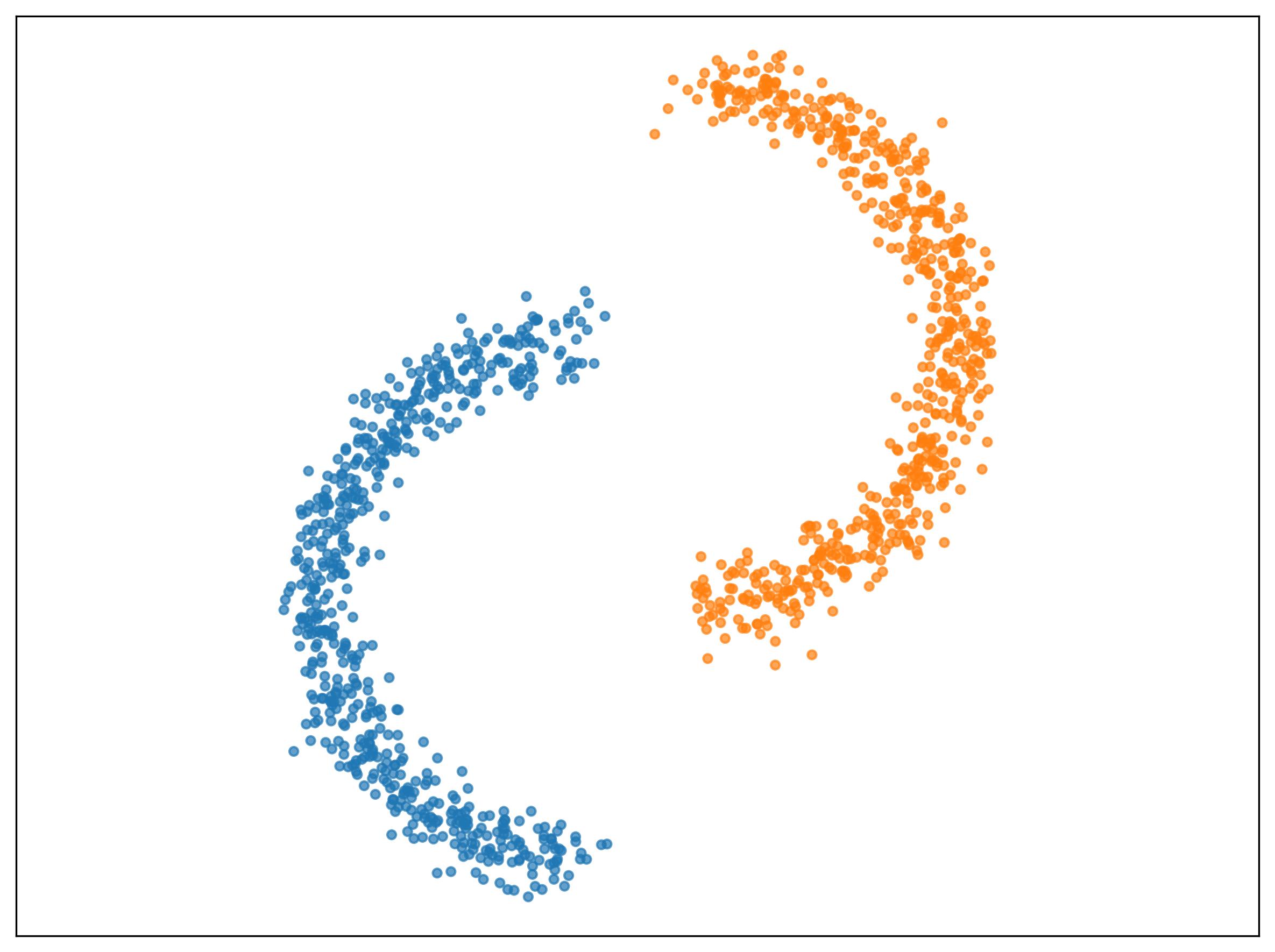} &
    \includegraphics[width=0.16\textwidth]{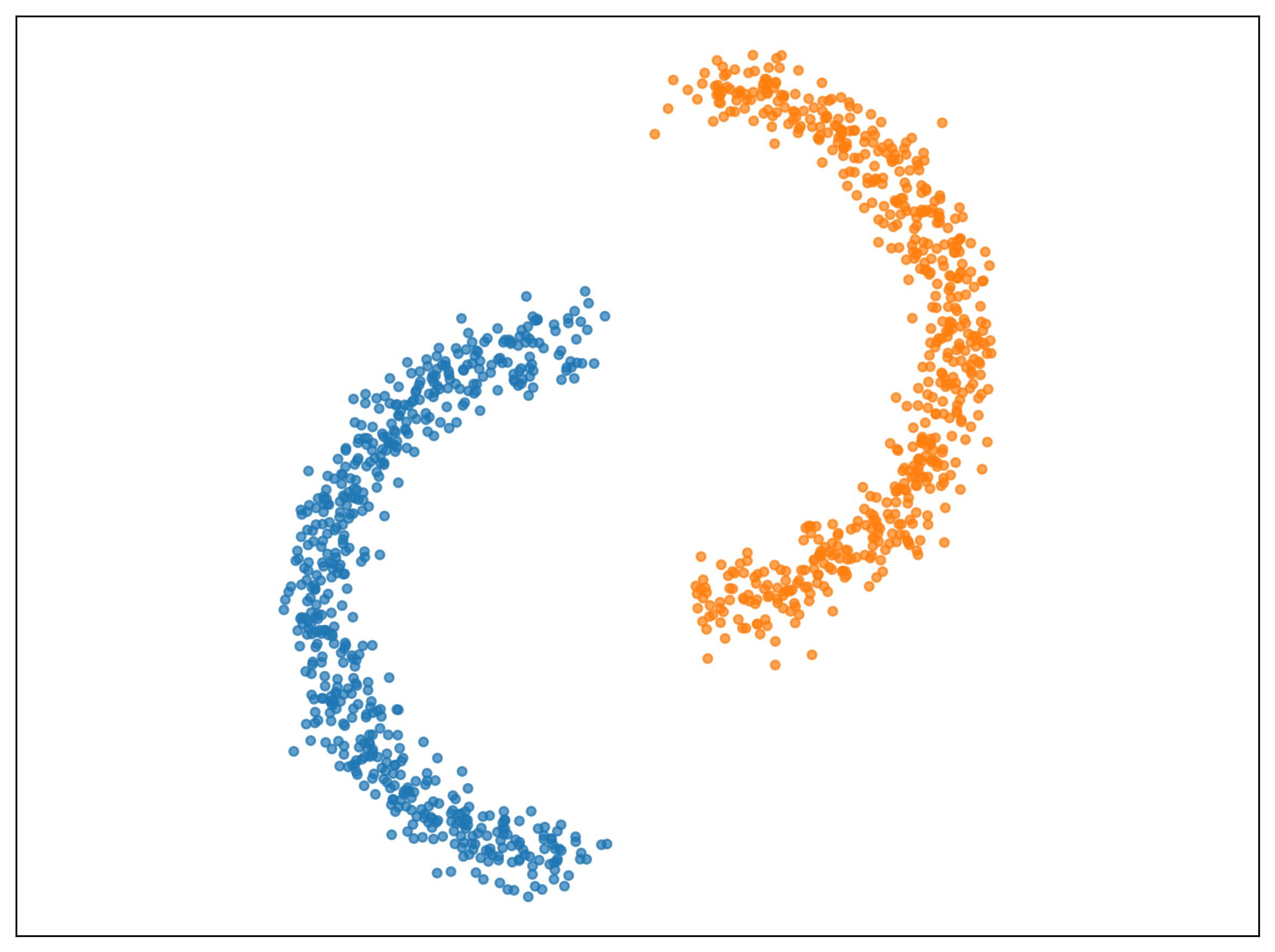} &
    \includegraphics[width=0.16\textwidth]{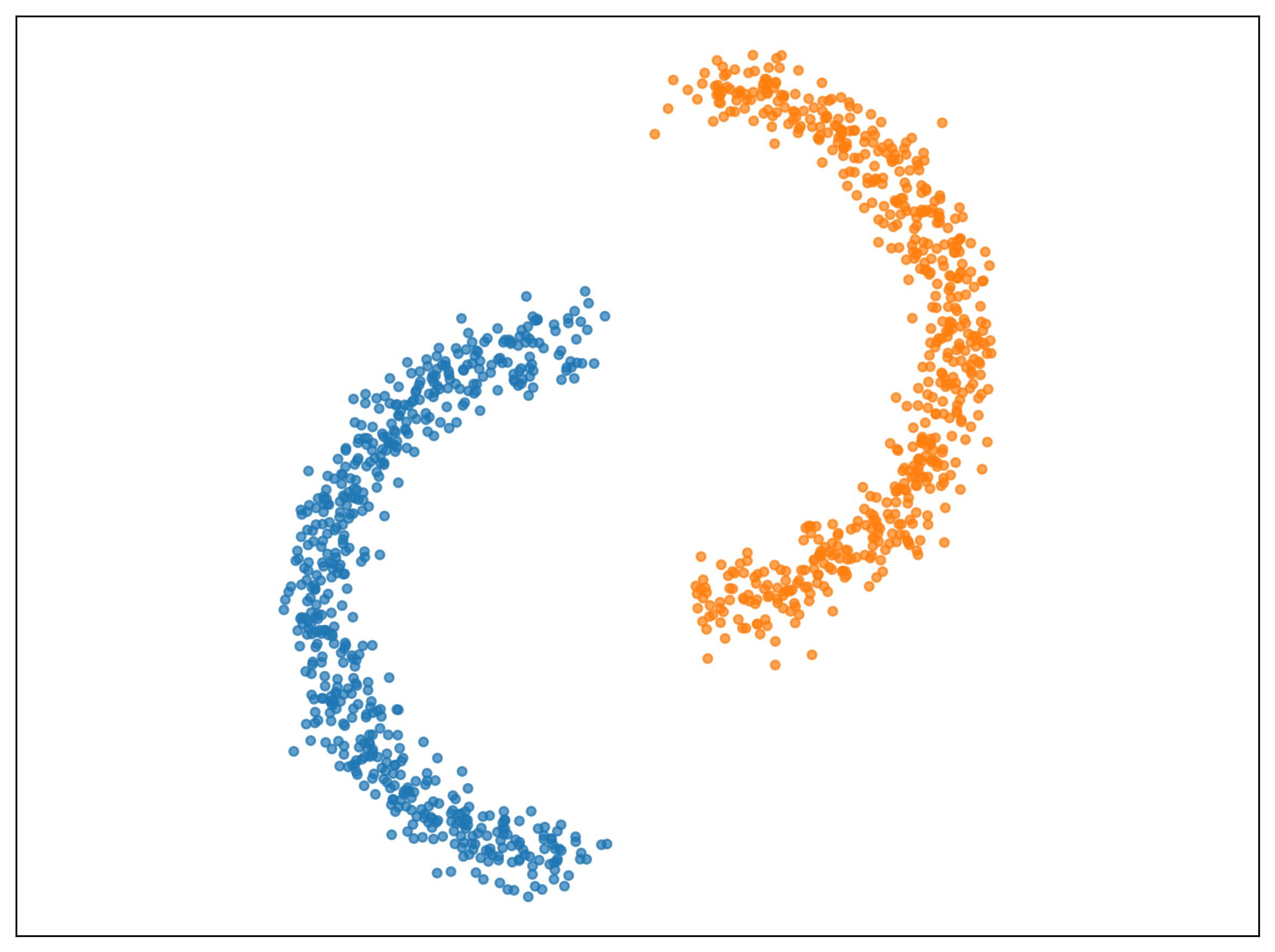} \\
    & & \footnotesize NMI=0.95 & \footnotesize NMI=1.00 & \footnotesize NMI=1.00 & \footnotesize NMI=1.00 & \footnotesize NMI=1.00 \\
    
    \addlinespace[4pt]
    \rotatebox{90}{\footnotesize Gap 0.3} & 
    \includegraphics[width=0.16\textwidth]{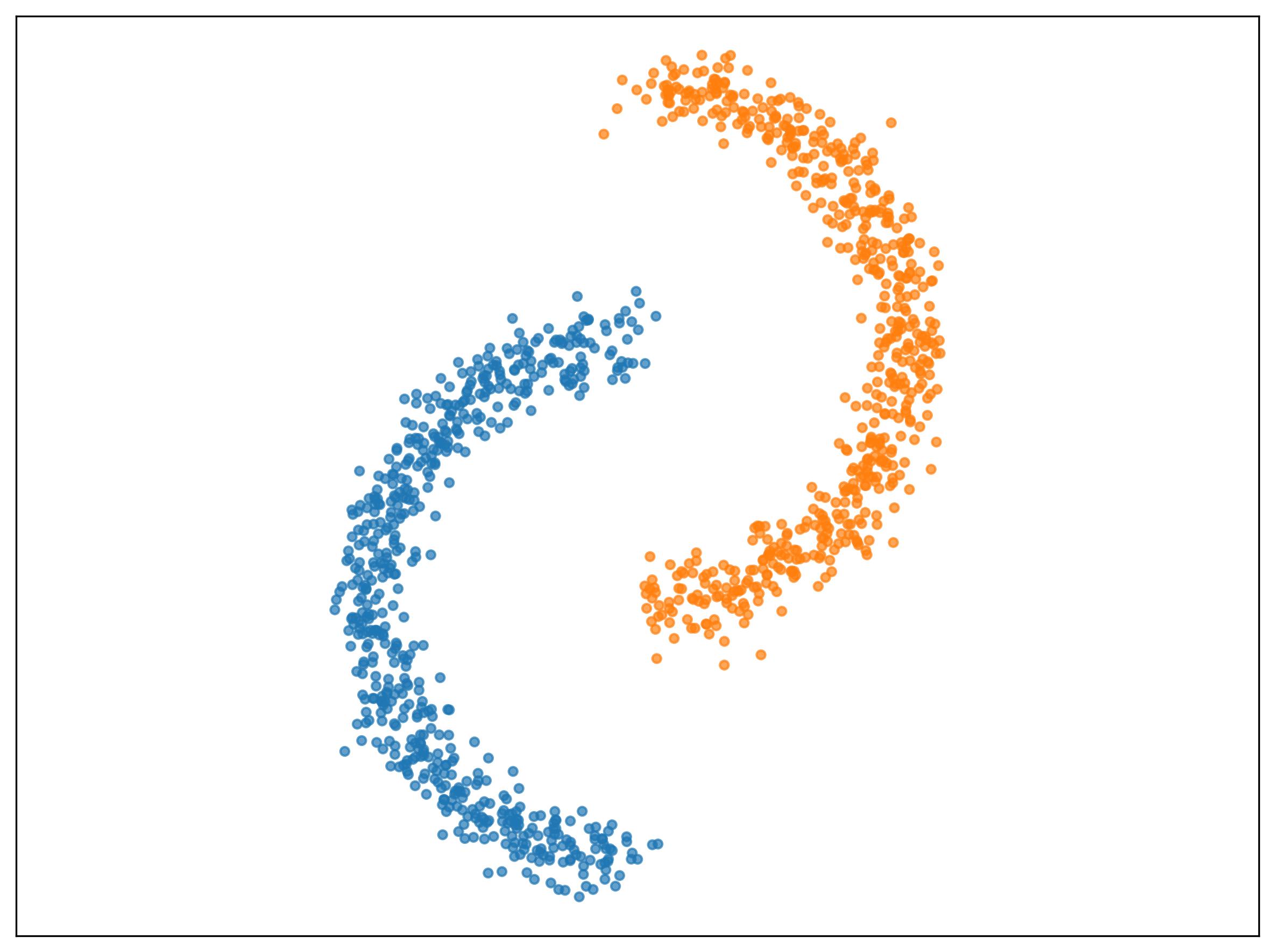} &
    \includegraphics[width=0.16\textwidth]{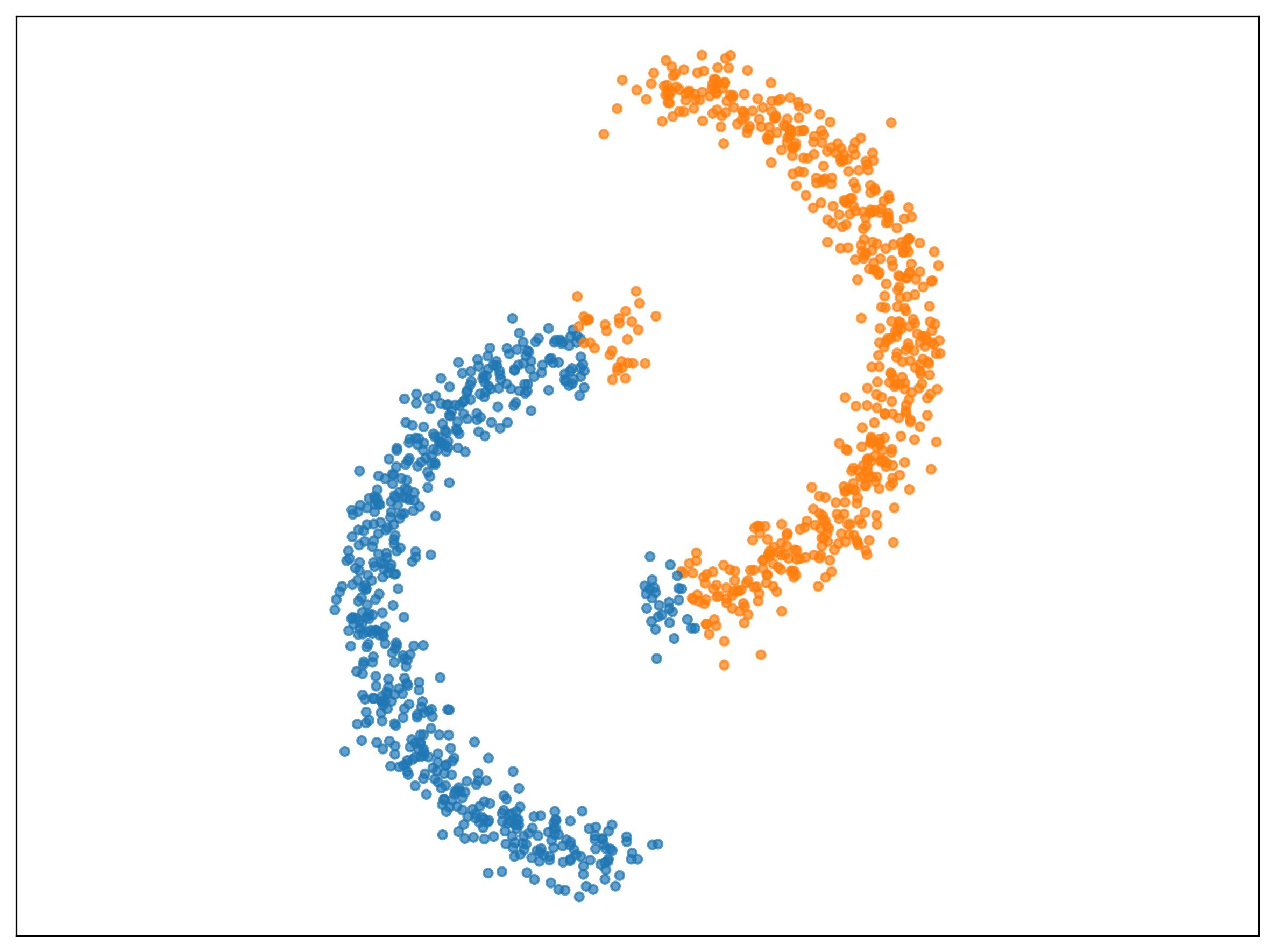} &
    \includegraphics[width=0.16\textwidth]{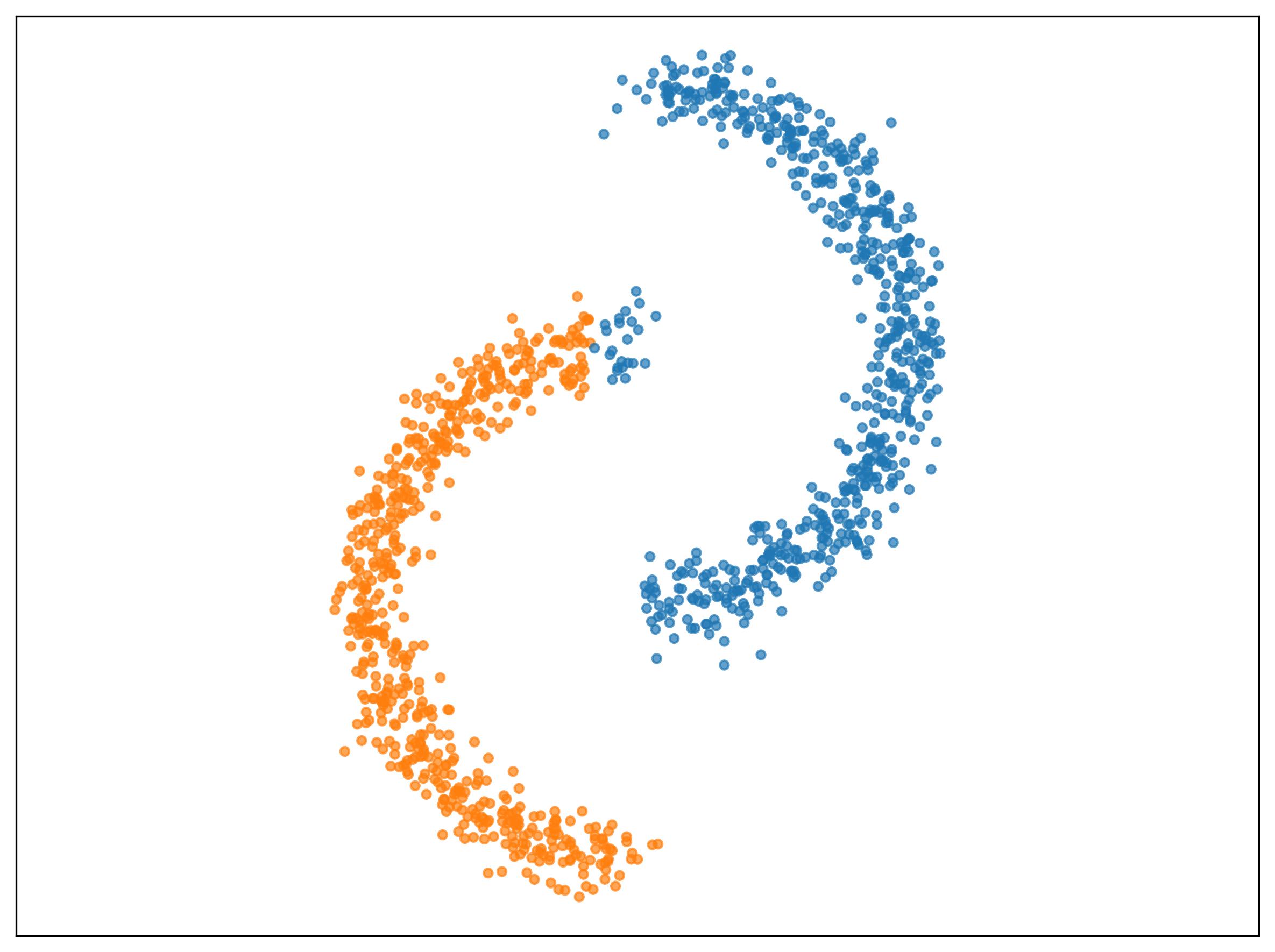} &
    \includegraphics[width=0.16\textwidth]{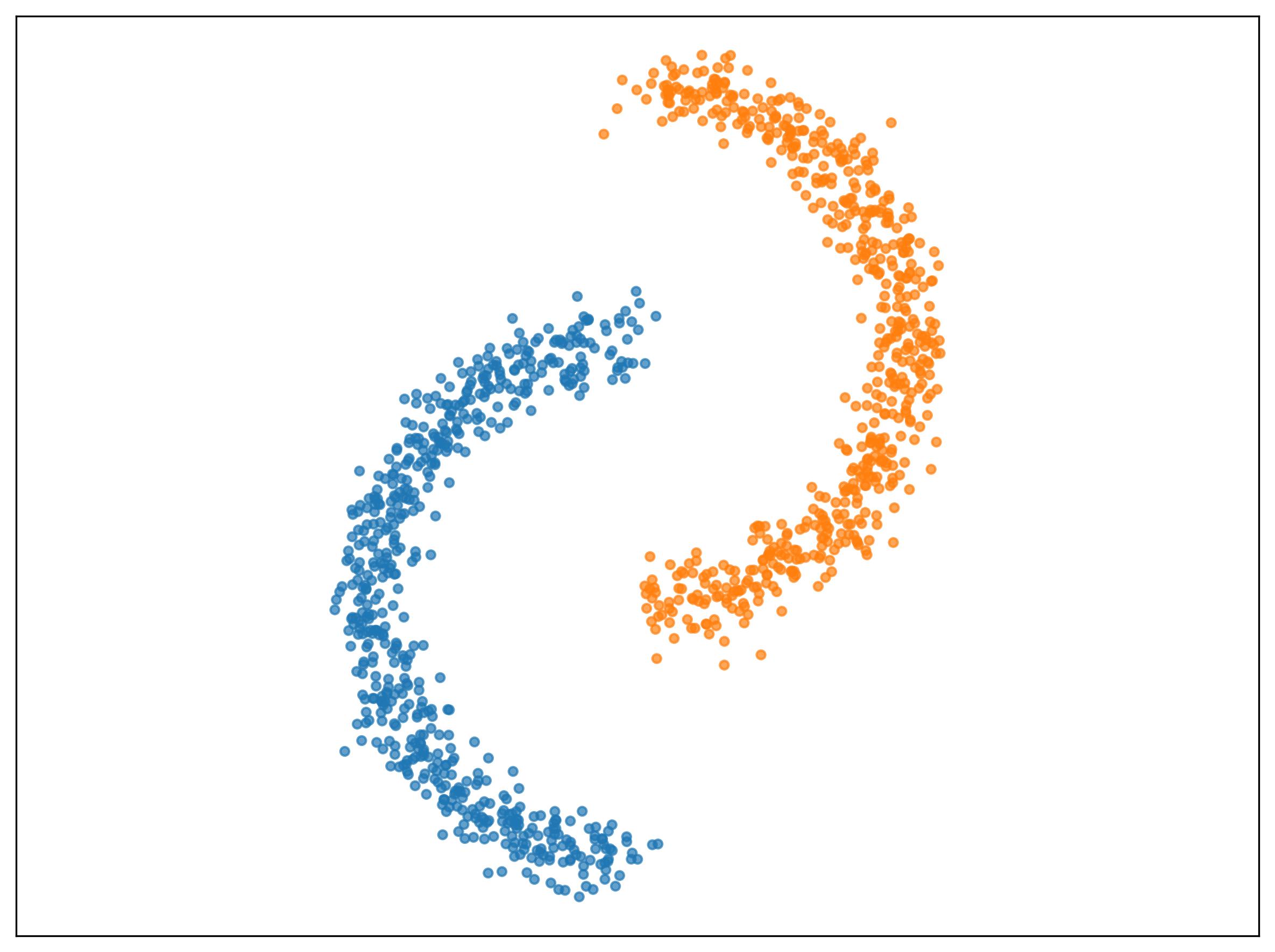} &
    \includegraphics[width=0.16\textwidth]{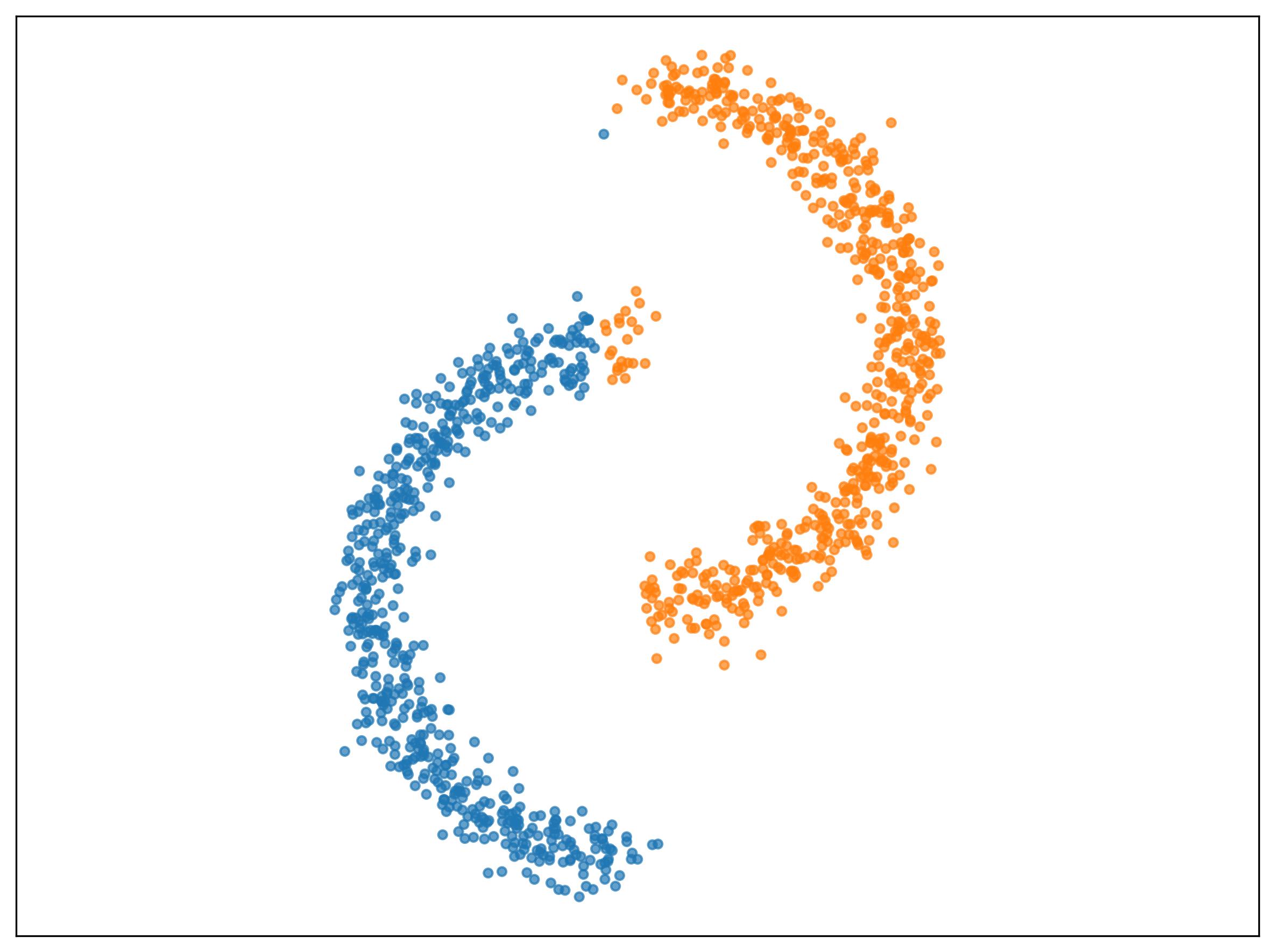} &
    \includegraphics[width=0.16\textwidth]{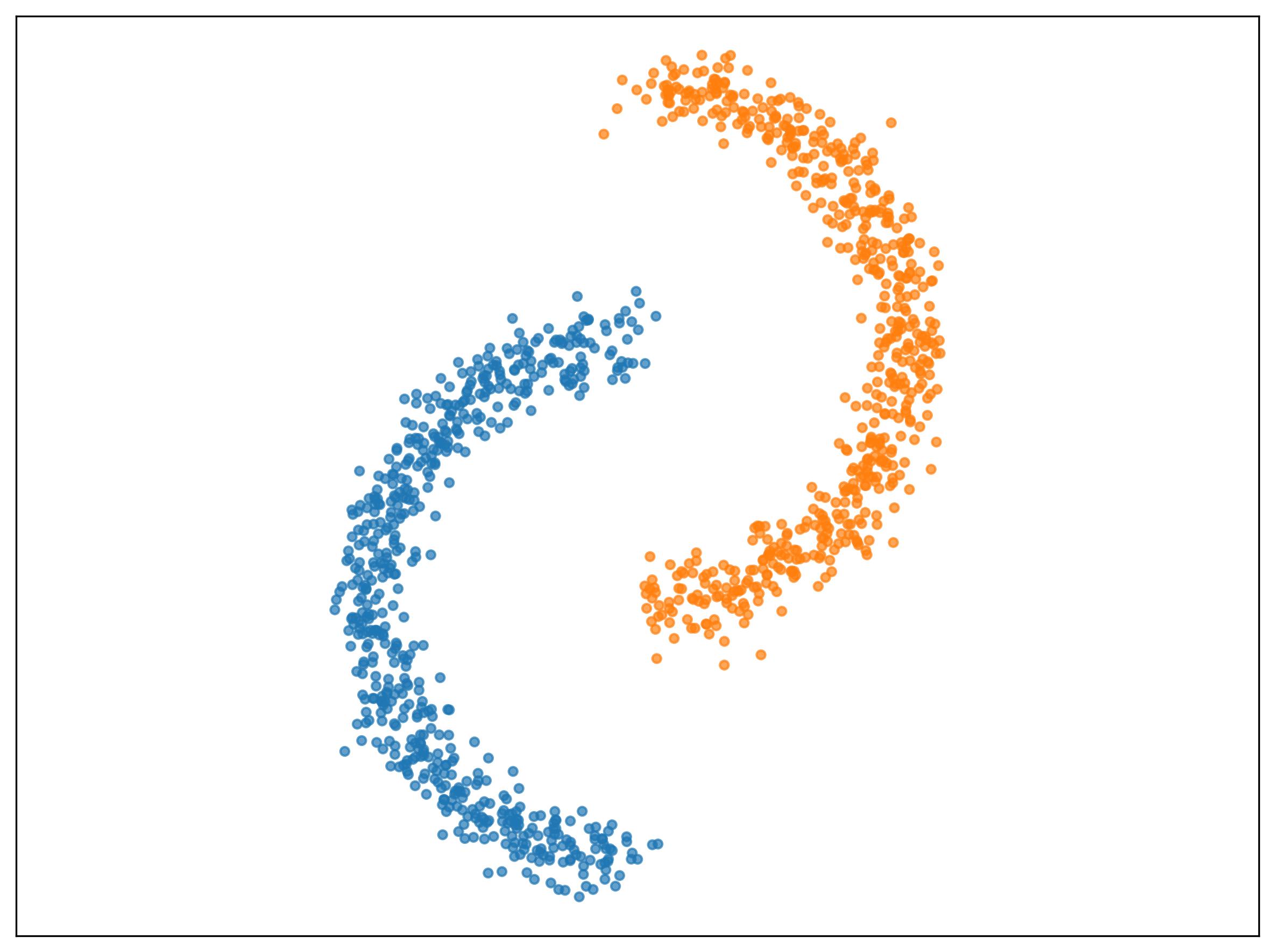} \\
    & & \footnotesize NMI=0.71 & \footnotesize NMI=0.88 & \footnotesize NMI=1.00 & \footnotesize NMI=0.88 & \footnotesize NMI=1.00 \\
    
    \addlinespace[4pt]
    \rotatebox{90}{\footnotesize Original} & 
    \includegraphics[width=0.16\textwidth]{img/datasets/nonSpherical.jpg} &
    \includegraphics[width=0.16\textwidth]{img/kmeans/nonSpherical.jpg} &
    \includegraphics[width=0.16\textwidth]{img/idec/nonSpherical/nonSpherical.jpg} &
    \includegraphics[width=0.16\textwidth]{img/cc/nonSpherical/nonSpherical.jpg} &
    \includegraphics[width=0.16\textwidth]{img/spectralnet/fig/nonSpherical/nonSpherical.jpg} &
    \includegraphics[width=0.16\textwidth]{img/kbc/nonSpherical/nonSpherical.jpg} \\
    & & \footnotesize NMI=0.42 & \footnotesize NMI=0.49 & \footnotesize NMI=0.53 & \footnotesize NMI=0.42 & \footnotesize NMI=1.00 \\
    
    \bottomrule
\end{tabular}}
\end{table*}

\section{Cluster Definition of Cluster-as-Distribution Clustering}
The following definition is extracted from \cite{zhang2025kbc}, and it is the same for psKC \cite{ting2022pskc} and IDKC \cite{zhu2023idkc}:
\begin{definition}
\label{def_NSS-clusters}
The set of clusters $\mathbb{C} = \{ C_1, \dots, C_k \}$ constitutes a Voronoi diagram in the feature space $\widehat{\phi}$ of $K$, and the boundary between any two of the $k$ Voronoi cells, representing the $k$ clusters, is defined for $\mathbf{x} \in \mathbb{R}^d$ as follows:
\begin{align*}
& K(\delta(\mathbf{x}),\mathcal{P}_{C_i})=K(\delta(\mathbf{x}),\mathcal{P}_{C_j})
\Leftrightarrow  \\
& \left< \phi(\mathbf{x}),\widehat{\phi}(\mathcal{P}_{C_i}) \right> = \left< \phi(\mathbf{x}),\widehat{\phi}(\mathcal{P}_{C_j}) \right>,\ \forall i \ne j
\end{align*}

where the kernel mean map of $K$ is $\widehat{\phi}(\mathcal{P}_{C}) = \frac{1}{|C|} \sum_{\mathbf{y} \in C} \phi(\mathbf{y})$;  $\phi(\cdot)$ is the feature map of a point-to-point kernel $\kappa$;  and $\widehat{\phi}(\delta(\mathbf{x})) = \phi(\mathbf{x})$.
\end{definition}

Note that $\widehat{\phi}(\mathcal{P}_{C})$ is a mean vector in the feature space of distributional kernel $K$. It corresponds to a centroid in a mapped space of the latent representation learned by deep clustering. A question is raised: Why does a centroid work in CaD clustering, but not in Deep Clustering. The answer is that $\widehat{\phi}(\mathcal{P}_{C})$ represents a cluster distribution $\mathcal{P}_{C}$; but what each centroid in Deep Clustering represents is unclear (recall the issue of unclear cluster definition stated in Table \ref{tab:conceptual_comparison}).

Note that  Gaussian Mixture Model (GMM) clustering methods \cite{jiang2016vade} are not considered to be a CaD clustering because they are limited to discover clusters having Gaussian distributions only.  As a result, they cannot discover clusters of arbitrary shapes, stated in Definition \ref{def-distribution-clustering}.

Table \ref{tab:non_spherical_with_gap} shows that on the original 2Crescents dataset, neither $k$-means nor deep clustering methods achieve satisfactory performance, whereas the KBC algorithm successfully captures the underlying structure and delivers strong results. As the margin between the two classes increases, all methods exhibit performance improvements; however, the qualitative behavior remains unchanged - KBC consistently outperforms the baselines. Moreover, the performance of deep clustering algorithms improves at a faster rate than that of $k$-means, suggesting that deep methods benefit more from enhanced class separability.

\section{More Experiments}
\label{sec:more_experiments}

\subsection{Image Dataset Information}

In this section, we present the detailed descriptions of the image datasets used in our experiments.

\begin{table}[htbp]
\centering
\caption{Data Characteristics of Image Datasets. For the CC \cite{li2021contrastive} algorithm, all datasets except MNIST \cite{lecun2002gradient} and COIL-20 \cite{nene1996coil20} are preprocessed into 512-dimensional vectors through ResNet \cite{he2016deep} and subsequently projected into 128-dimensional embeddings during clustering. In contrast, KBC \cite{zhang2025kbc} directly consumes flattened raw images for MNIST, and for COIL-20 the images are first downsampled to $32 \times 32$ resolution and then flattened into vectors.}
\label{tab:image_dataset_information}
\vskip 0.15in
\begin{small}
\begin{tabular}{lrrcc}
\toprule
\textbf{Dataset} & \textbf{Points} & \textbf{Image Size} & \textbf{Dimensions} & \textbf{Clusters} \\ 
\midrule
STL-10 \cite{coates2011stl-10}         & 13,000  & $96 \times 96 \times 3$ & 128   & 10 \\
CIFAR-10 \cite{krizhevsky2009learning} & 60,000  & $32 \times 32 \times 3$ & 128   & 10 \\
ImageNet-10 \cite{deng2009imagenet}    & 13,000  & $96 \times 96 \times 3$ & 128   & 10 \\
ImageNet-Dogs \cite{deng2009imagenet}  & 19,500  & $96 \times 96 \times 3$ & 128   & 15 \\
ImageNet-Tiny \cite{deng2009imagenet}  & 100,000 & $64 \times 64 \times 3$ & 128   & 200 \\
\midrule
MNIST \cite{lecun2002gradient}         & 70,000  & $28 \times 28$          & 784   & 10 \\
COIL-20 \cite{nene1996coil20}          & 1,440   & $128 \times 128$        & 1,024 & 20 \\
\bottomrule
\end{tabular}
\end{small}
\vskip -0.1in
\end{table}

\subsection{Single-Cell Transcriptomics Data availability}
The single-cell transcriptomics datasets analyzed in this study are publicly available. The Tutorial dataset (1\% Jurkat cells) is available from the scCAD GitHub repository (\url{https://github.com/xuyp-csu/scCAD/blob/main/1%25Jurkat.h5}. The human Airway dataset is available from the Gene Expression Omnibus (GEO) under accession number GSE103354. The preprocessed human Tonsil dataset (study SCP2169) and the Crohn's disease dataset (study SCP359) are available from the Broad Institute Single Cell Portal at (\url{https://singlecell.broadinstitute.org/single_cell/study/SCP2169/slide-tags-snrna-seq-on-human-tonsil}) and (\url{https://singlecell.broadinstitute.org/single_cell/study/SCP359/ica-ileum-lamina-propria-immunocytes-sinai}).

\subsection{More results}
In this section, we present the corresponding Adjusted Rand Index (ARI) \cite{hubert1985comparing} results for all algorithms across all datasets used in our analysis.

\begin{table*}[htbp]
\centering
\caption{Clustering results comparison. Each reported Adjusted Rand Index (ARI) \cite{hubert1985comparing} value is averaged over 10 runs.}
\label{tab:main_ari}
\begin{tabular*}{\textwidth}{@{\extracolsep{\fill}}l r r c c c c c c}
\toprule
\textbf{Datasets} & \textbf{Points} & \textbf{Dim} & \textbf{Clusters} & \textbf{$k$-means} & \textbf{IDEC} & \textbf{CC} & \textbf{SpectralNet} & \textbf{KBC} \\
\midrule
2Crescents    & 1,200  & 2     & 2  & 0.52 & 0.59 & 0.64 & 0.52 & \textbf{1.00} \\
Diff-Sizes    & 900    & 2     & 3  & 0.32 & 0.36 & 0.25 & 0.36 & \textbf{0.97} \\
AC            & 1,004  & 2     & 2  & 0.69 & 0.59 & 0.54 & 0.60 & \textbf{1.00} \\
Tutorial      & 1,556  & 2,000 & 2  & 0.31 & 0.01 & 0.00 & 0.03 & \textbf{0.93} \\
Tonsil        & 5,778  & 2,000 & 13 & 0.44 & \textbf{0.62} & 0.28 & 0.44 & 0.42 \\
Airway        & 7,193  & 2,000 & 7  & 0.41 & 0.31 & 0.21 & 0.68 & \textbf{0.68} \\
Crohn         & 39,563 & 2,000 & 27 & 0.32 & 0.26 & 0.24 & 0.31 & \textbf{0.55} \\
DLPFC         & 4,221  & 400   & 7  & 0.44 & 0.42 & 0.35 & 0.33 & \textbf{0.54} \\
USPS          & 11,000 & 256   & 10 & 0.29 & 0.52 & 0.34 & 0.64 & \textbf{0.72} \\
STL-10        & 13,000 & 128   & 10 & 0.47 & 0.58 & \textbf{0.73} & 0.49 & 0.61 \\
CIFAR-10      & 60,000 & 128   & 10 & 0.53 & \textbf{0.72} & 0.59 & 0.64 & 0.67 \\
ImageNet-10   & 13,000 & 128   & 10 & 0.57 & 0.72 & 0.81 & 0.68 & \textbf{0.87} \\
ImageNet-Dogs & 19,500 & 128   & 15 & 0.25 & 0.24 & 0.29 & 0.28 & \textbf{0.38} \\
MNIST         & 70,000 & 784   & 10 & 0.38 & \textbf{0.80} & 0.62 & 0.77 & 0.77 \\
COIL-20       & 1,440  & 1024  & 20 & 0.54 & 0.59 & 0.14 & 0.66 & \textbf{0.95} \\
spiral        & 312    & 2     & 3  & 0.00 & 0.00 & 0.00 & 0.07 & \textbf{1.00} \\
4C            & 1,000  & 2     & 4  & 0.27 & 0.21 & 0.01 & 0.40 & \textbf{1.00} \\
RingG         & 1,536  & 2     & 4  & 0.43 & 0.57 & 0.43 & 0.74 & \textbf{0.99} \\
complex9      & 3,031  & 2     & 9  & 0.38 & 0.34 & 0.26 & 0.39 & \textbf{1.00} \\
S3D3D3        & 900    & 2     & 3  & 0.68 & 0.51 & 0.48 & 0.64 & \textbf{0.96} \\
OGOL          & 1,400  & 2     & 2  & 0.42 & 0.59 & 0.00 & 0.80 & \textbf{0.92} \\
\bottomrule
\end{tabular*}
\end{table*}

\subsection{Additional examples demonstrating the fundamental limitations of the $k$-means algorithm}
We provide additional examples illustrating the fundamental limitations of $k$-means in Table \ref{tab:clustering_results_more_examples}.

\begin{table*}[htbp]
\centering
\caption{Additional examples demonstrating the fundamental limitations of the $k$-means algorithm}
\label{tab:clustering_results_more_examples}

\setlength{\tabcolsep}{1pt} 
\renewcommand{\arraystretch}{0.2} 

\resizebox{\textwidth}{!}{ 
\begin{tabular}{c c c c c c c} 
    \toprule
    & \small \textbf{Original Data} & \small \textbf{$k$-means} & \small \textbf{IDEC} & \small \textbf{Contrastive Clustering} & \small \textbf{SpectralNet} & \small \textbf{KBC} \\
    \midrule
    
    \rotatebox{90}{\footnotesize spiral} & 
    \includegraphics[width=0.16\textwidth]{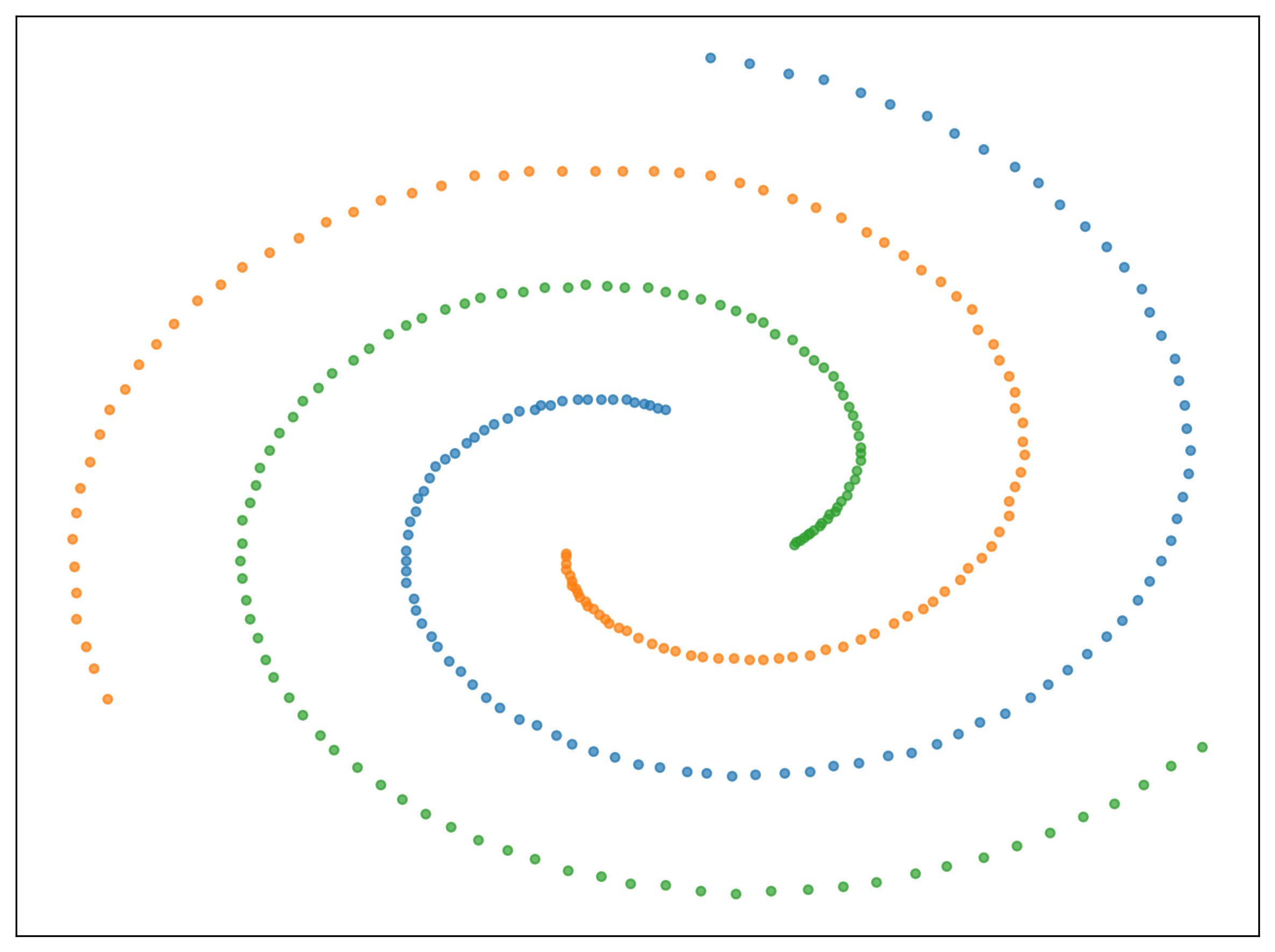} &
    \includegraphics[width=0.16\textwidth]{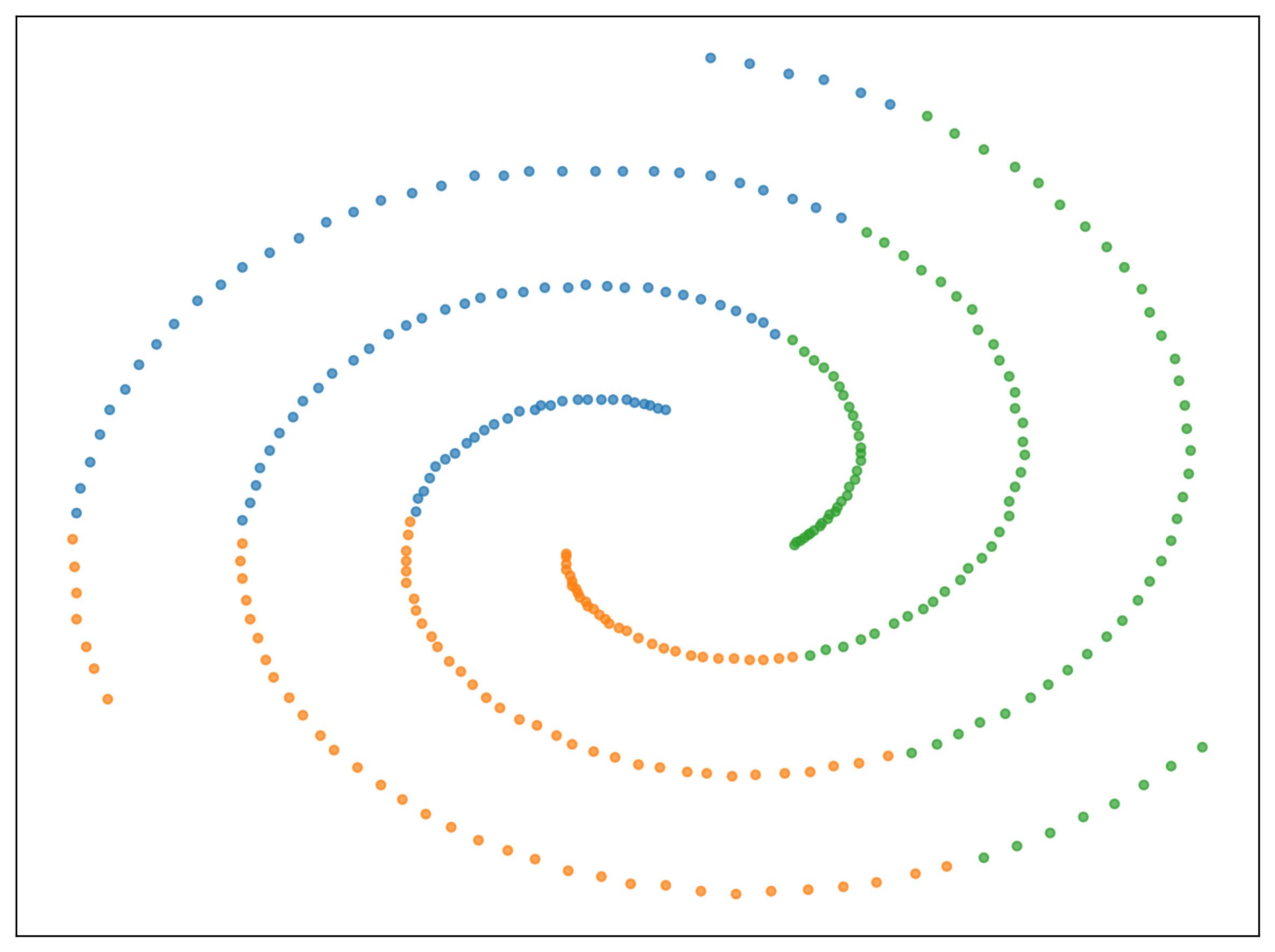} &
    \includegraphics[width=0.16\textwidth]{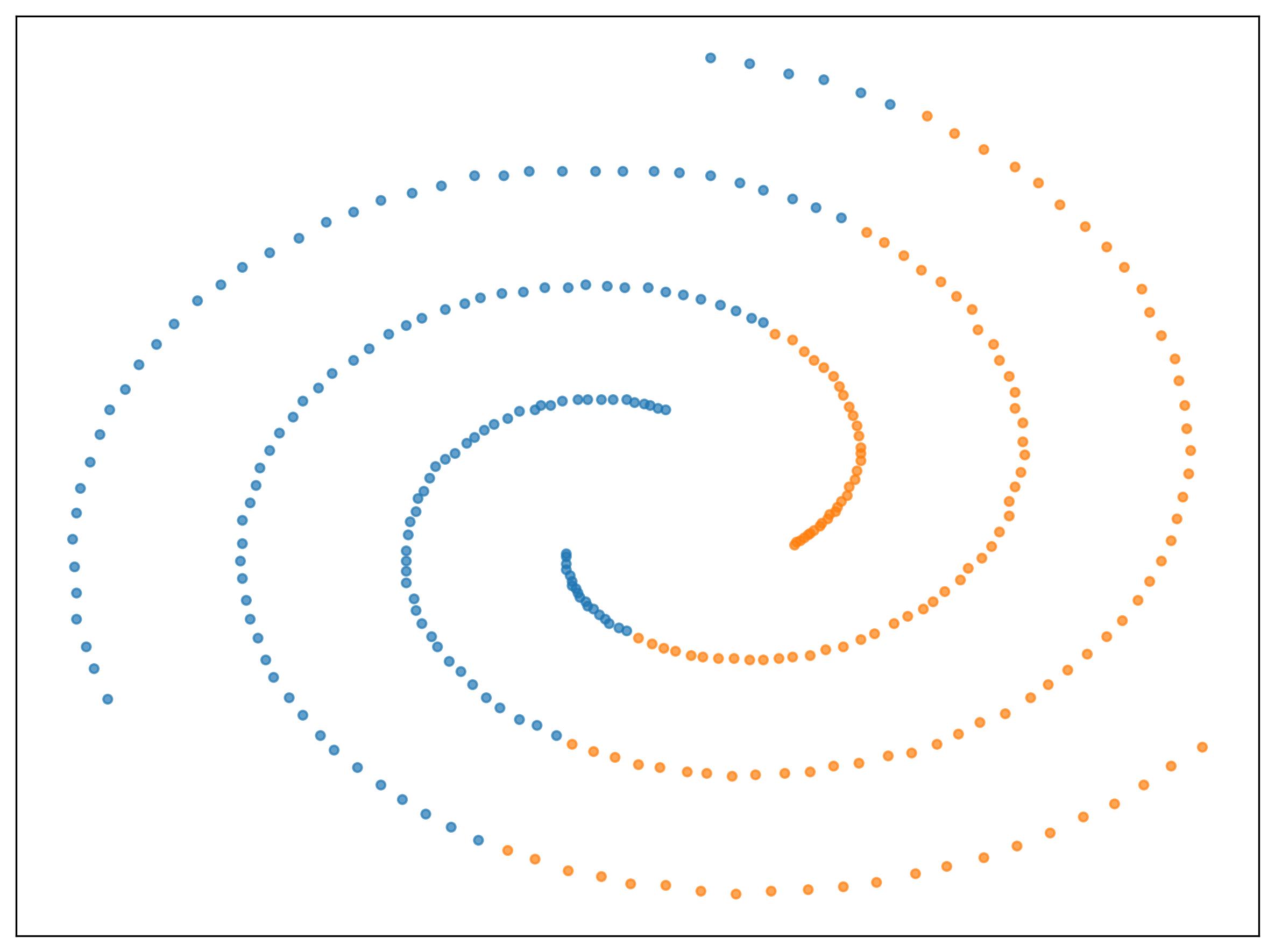} &
    \includegraphics[width=0.16\textwidth]{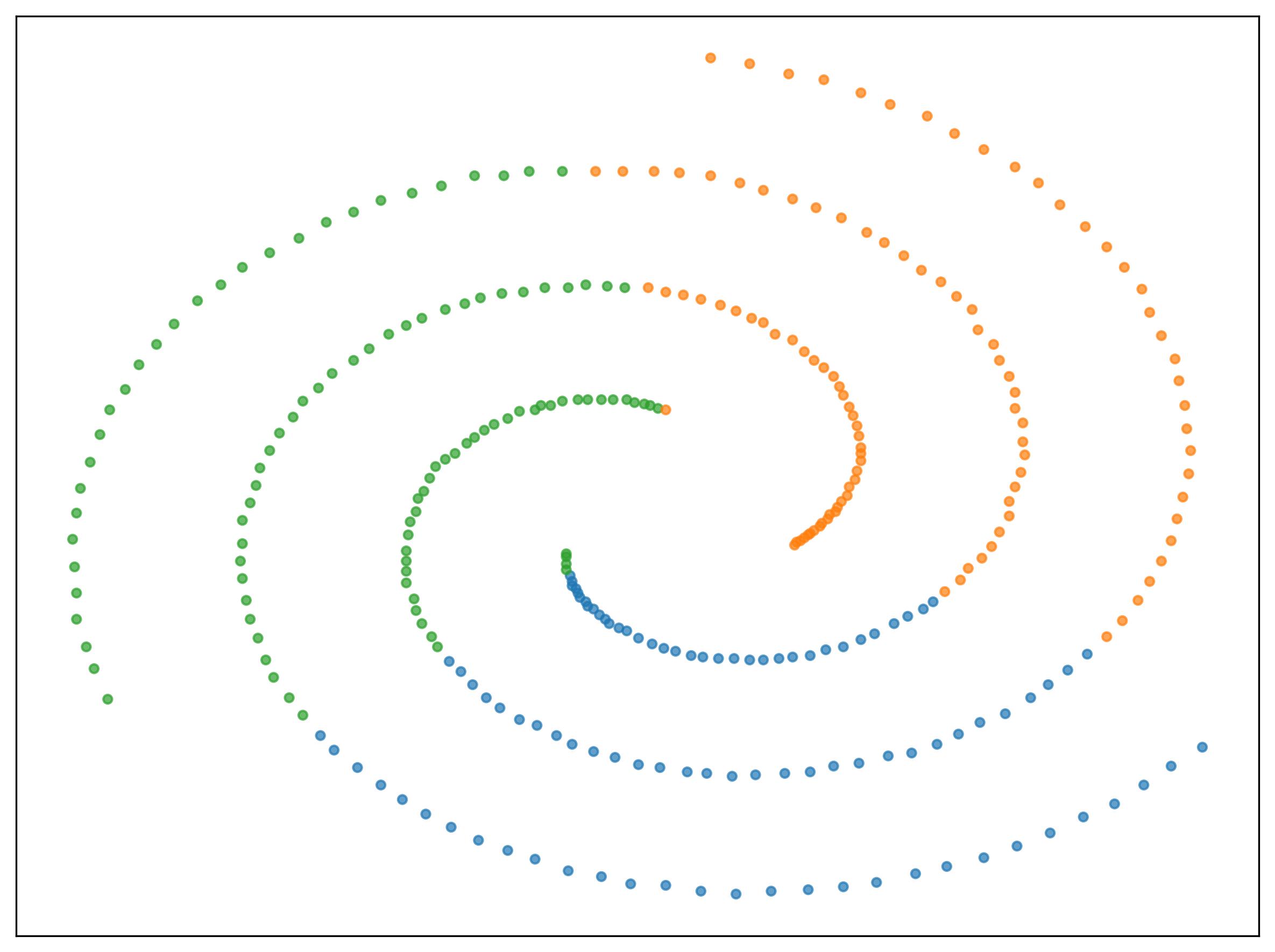} &
    \includegraphics[width=0.16\textwidth]{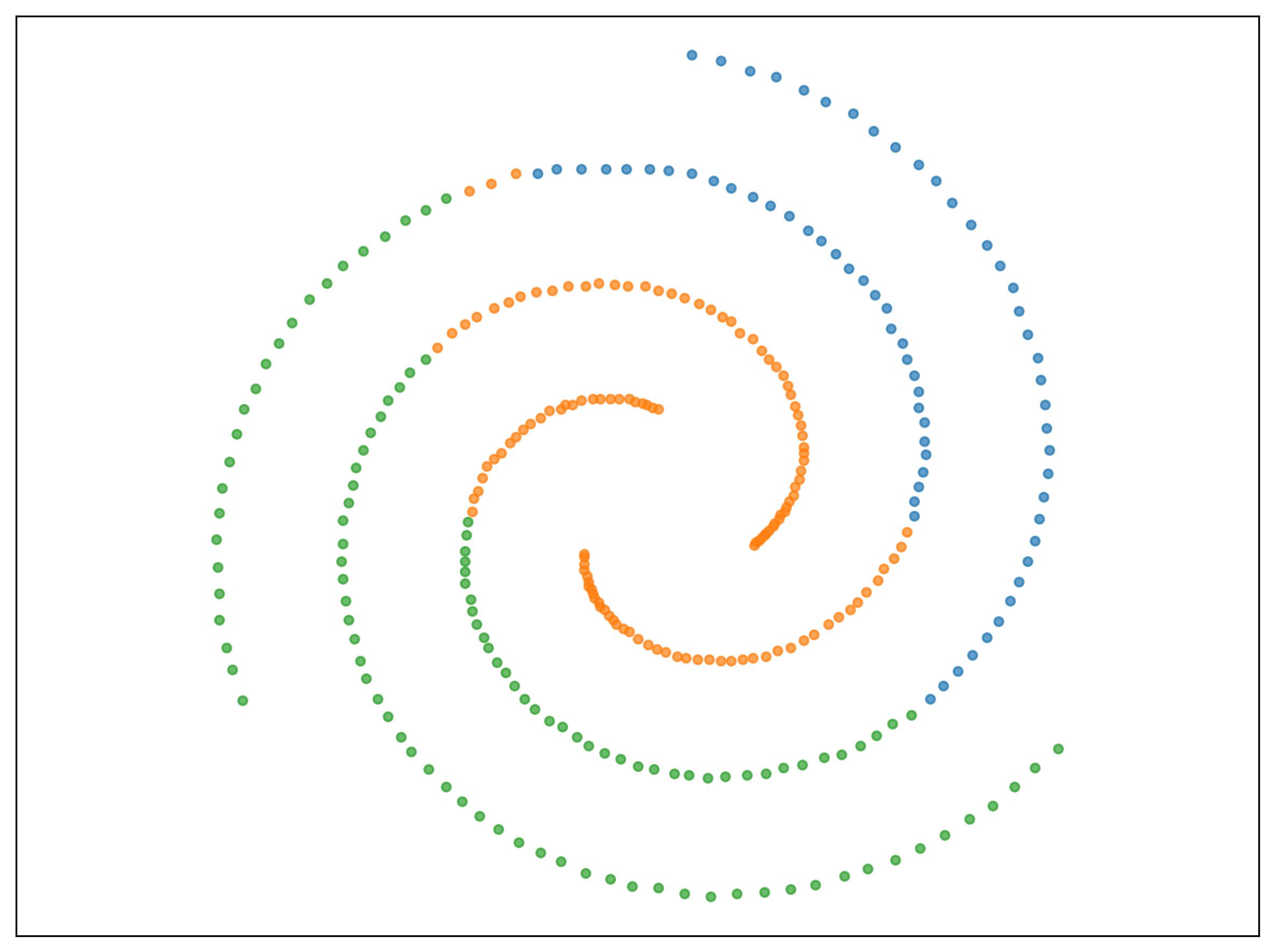} & 
    \includegraphics[width=0.16\textwidth]{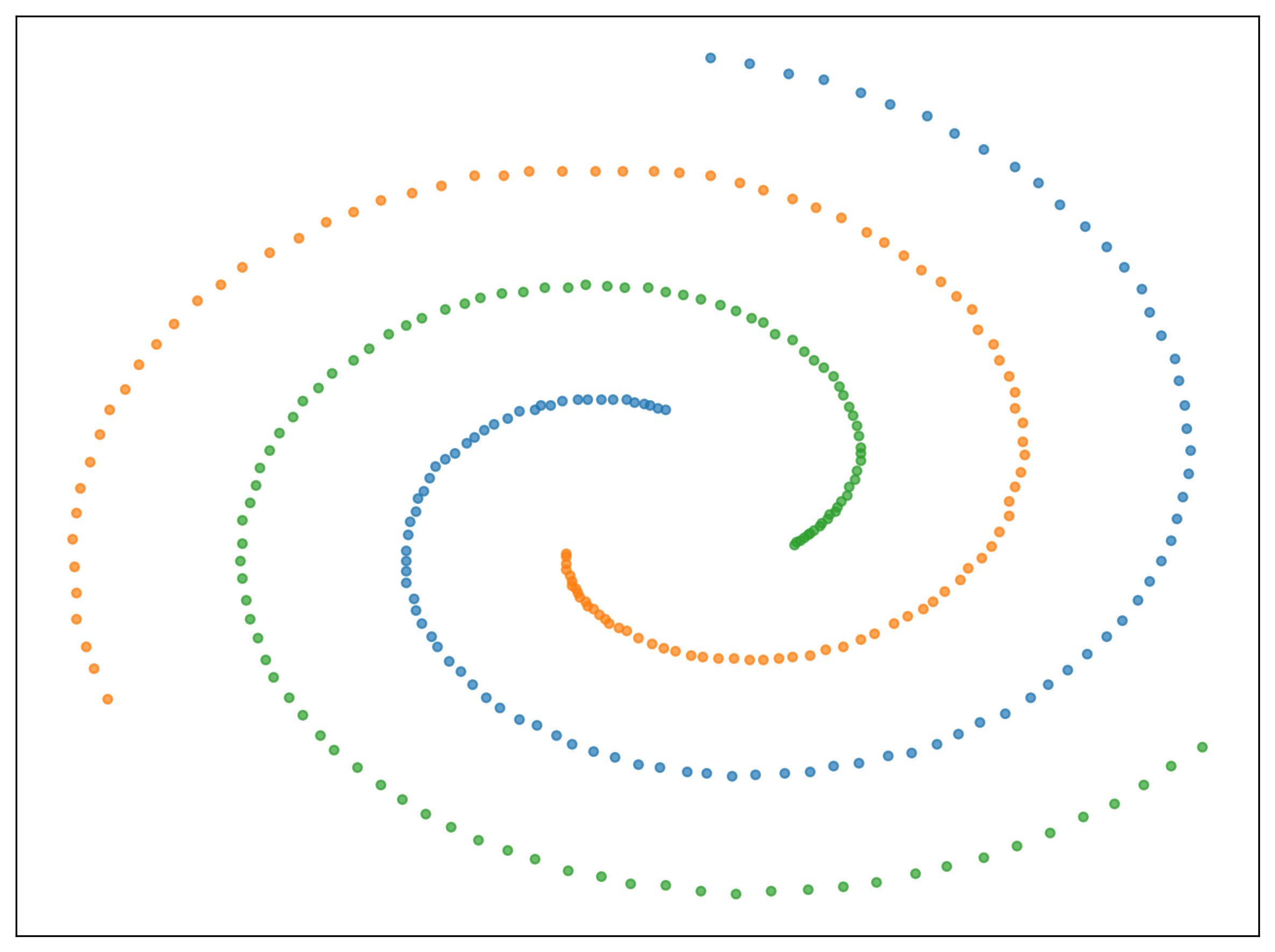} \\
    & & \footnotesize NMI=0.00 & \footnotesize NMI=0.00 & \footnotesize NMI=0.00 & \footnotesize NMI=0.11 & \footnotesize NMI=1.00 \\
    \addlinespace[4pt]
    
    \rotatebox{90}{\footnotesize 4C} & 
    \includegraphics[width=0.16\textwidth]{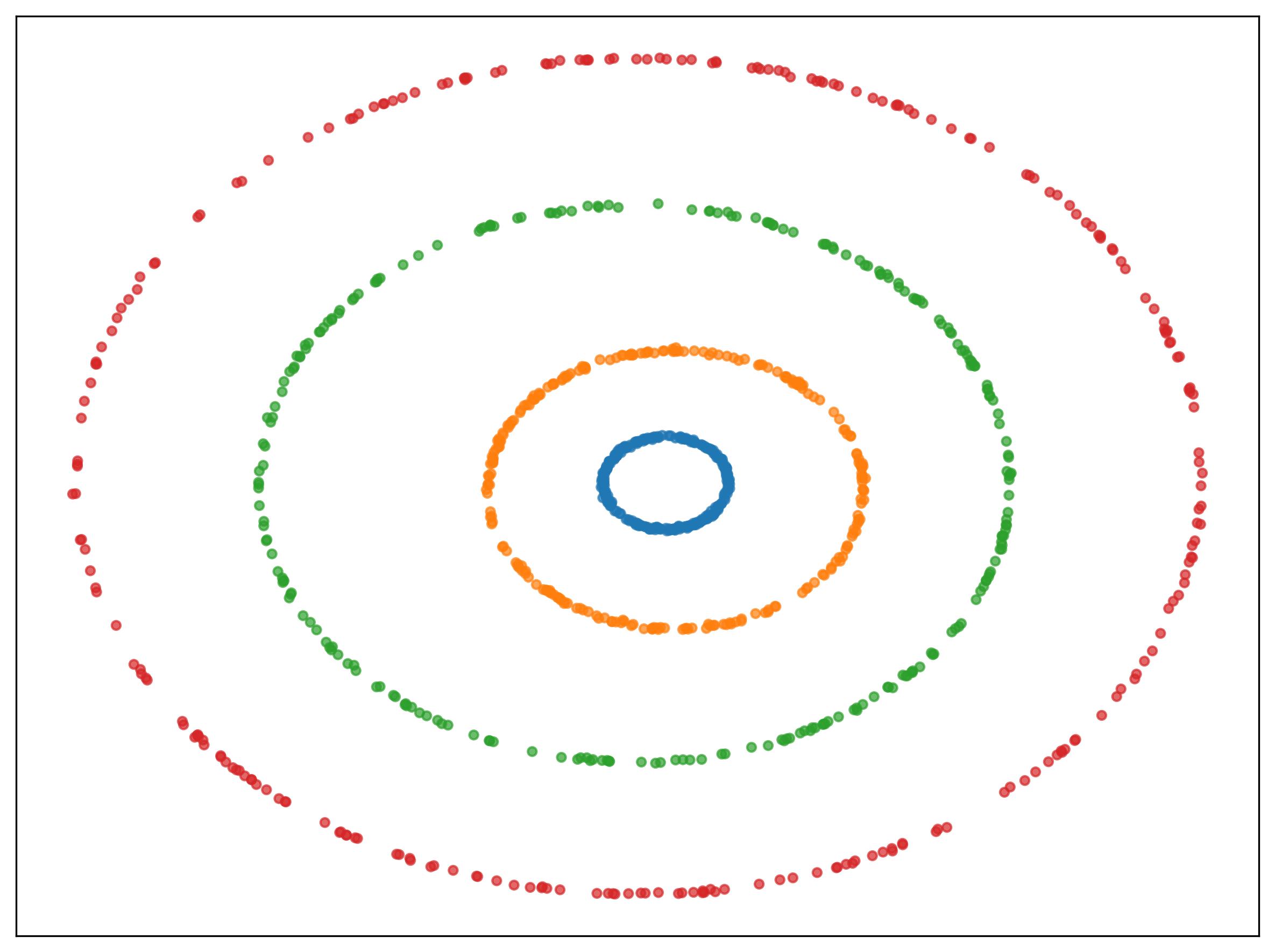} &
    \includegraphics[width=0.16\textwidth]{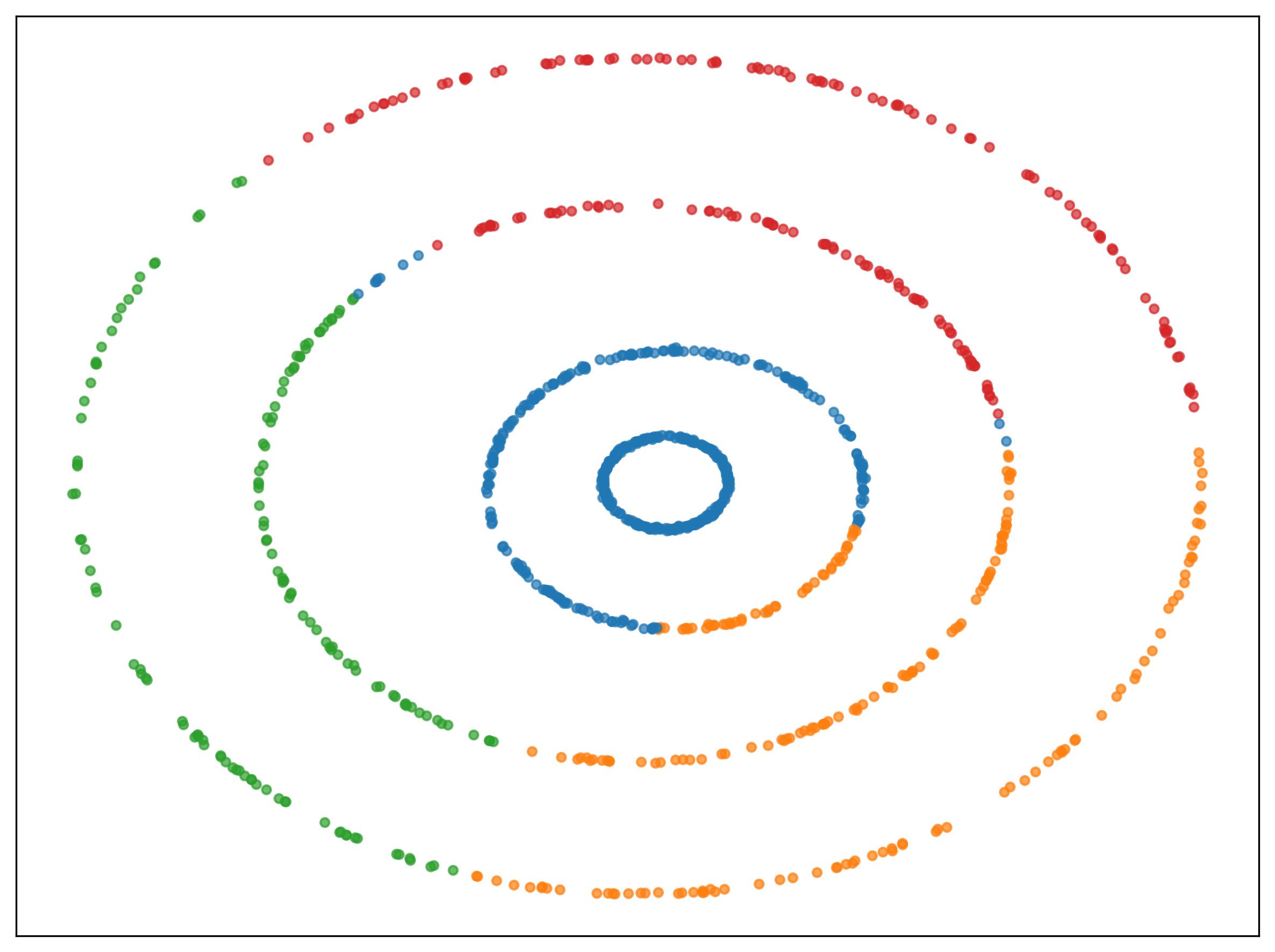} &
    \includegraphics[width=0.16\textwidth]{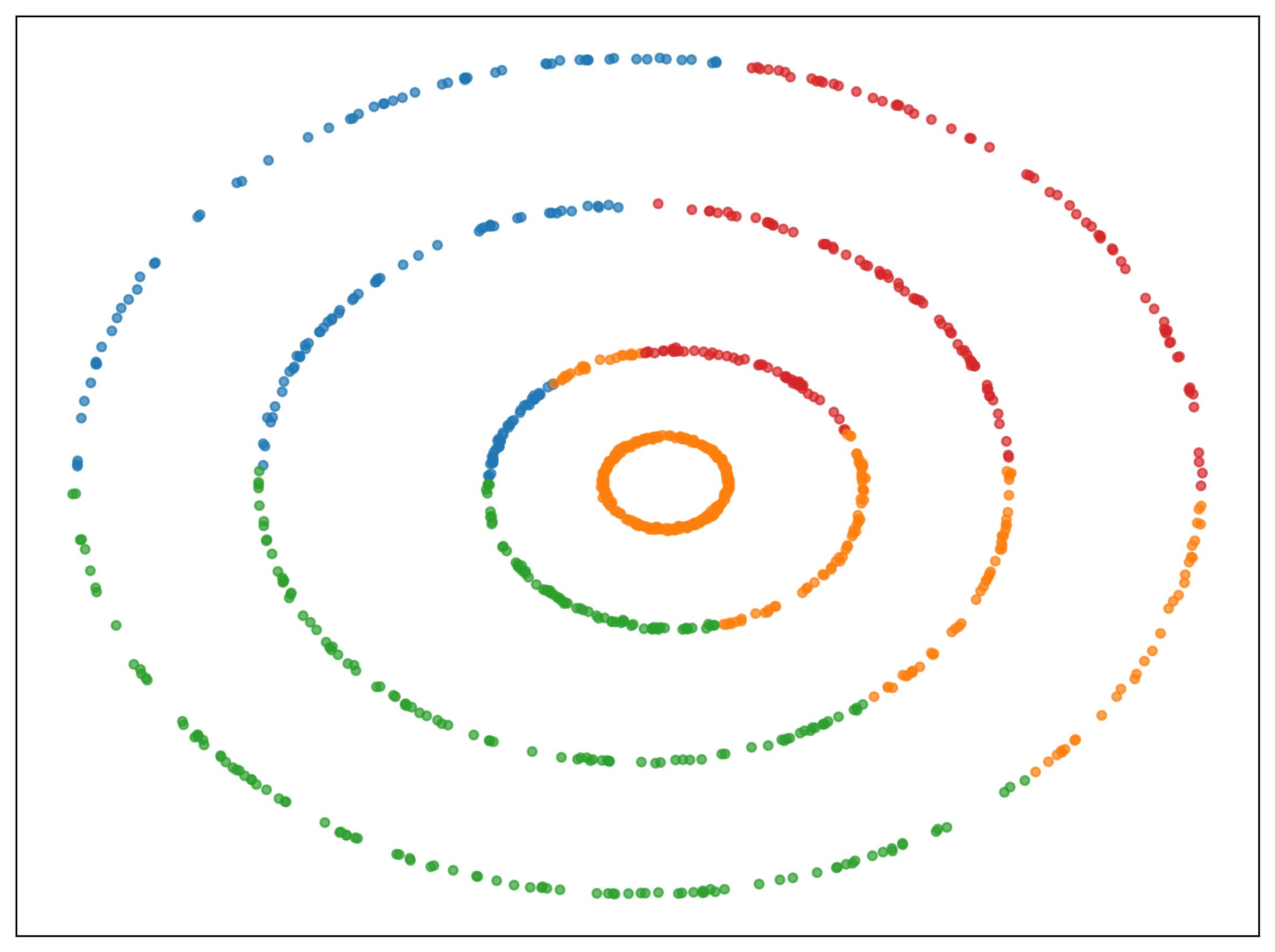} &
    \includegraphics[width=0.16\textwidth]{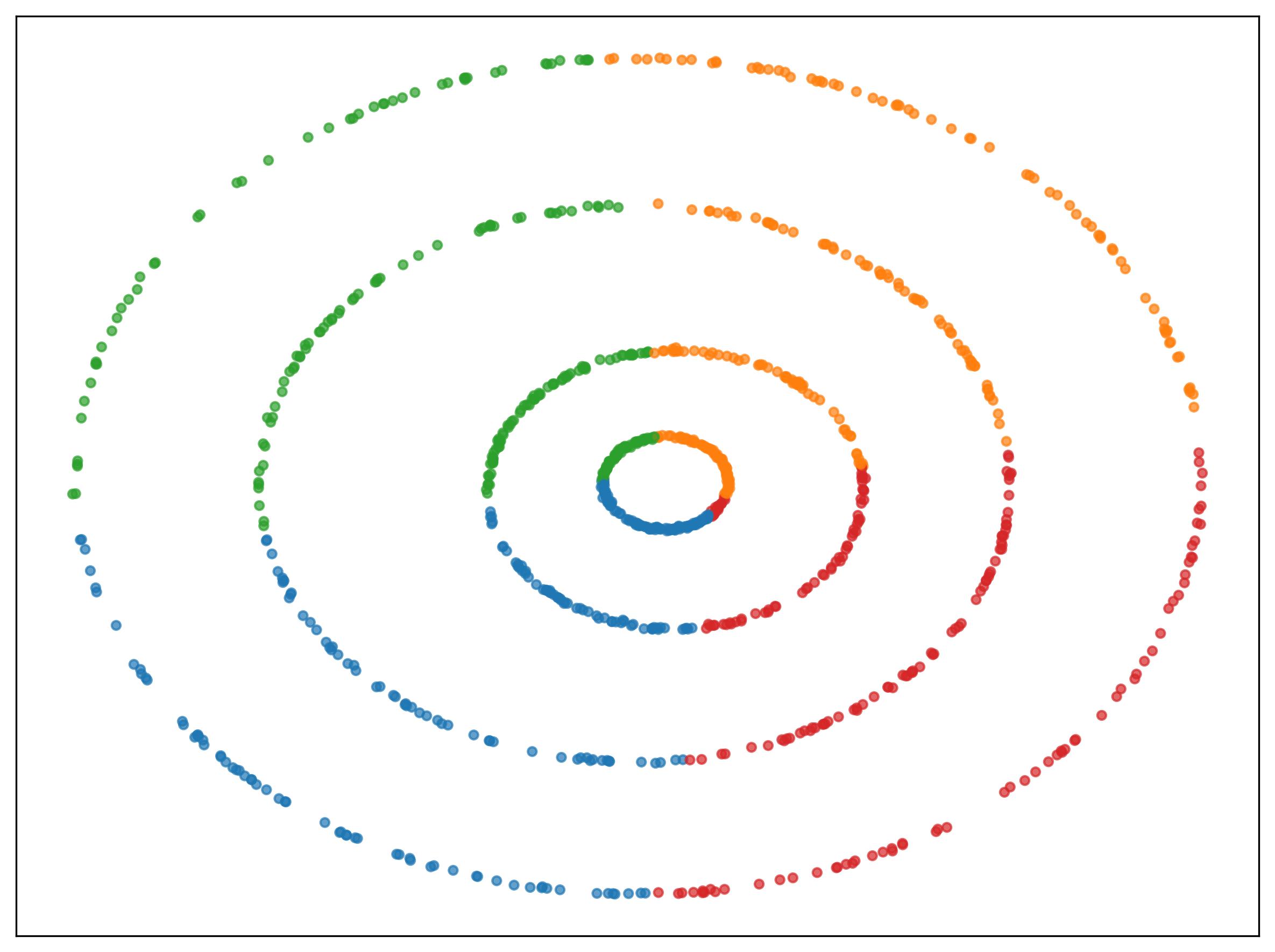} &
    \includegraphics[width=0.16\textwidth]{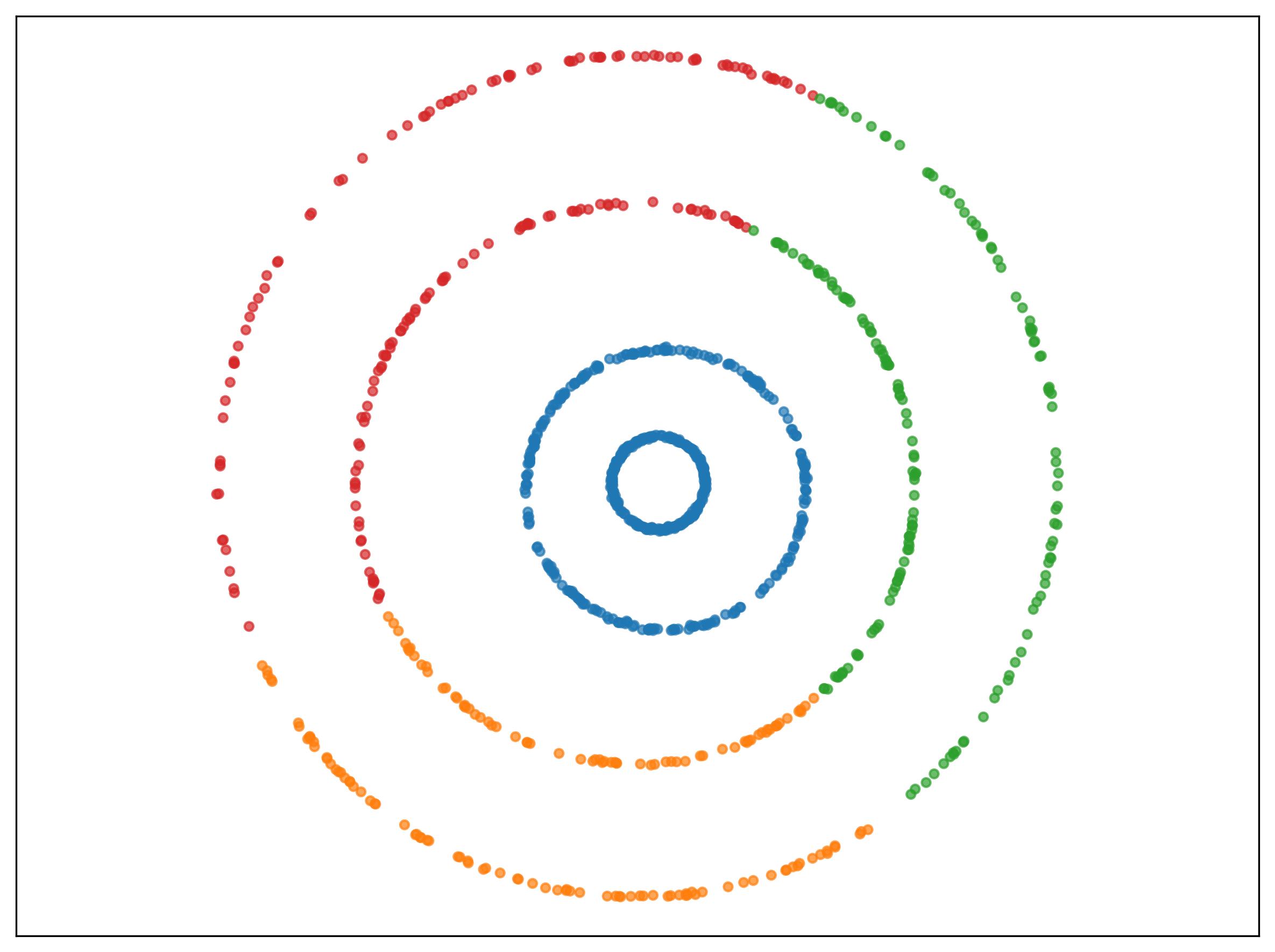} & 
    \includegraphics[width=0.16\textwidth]{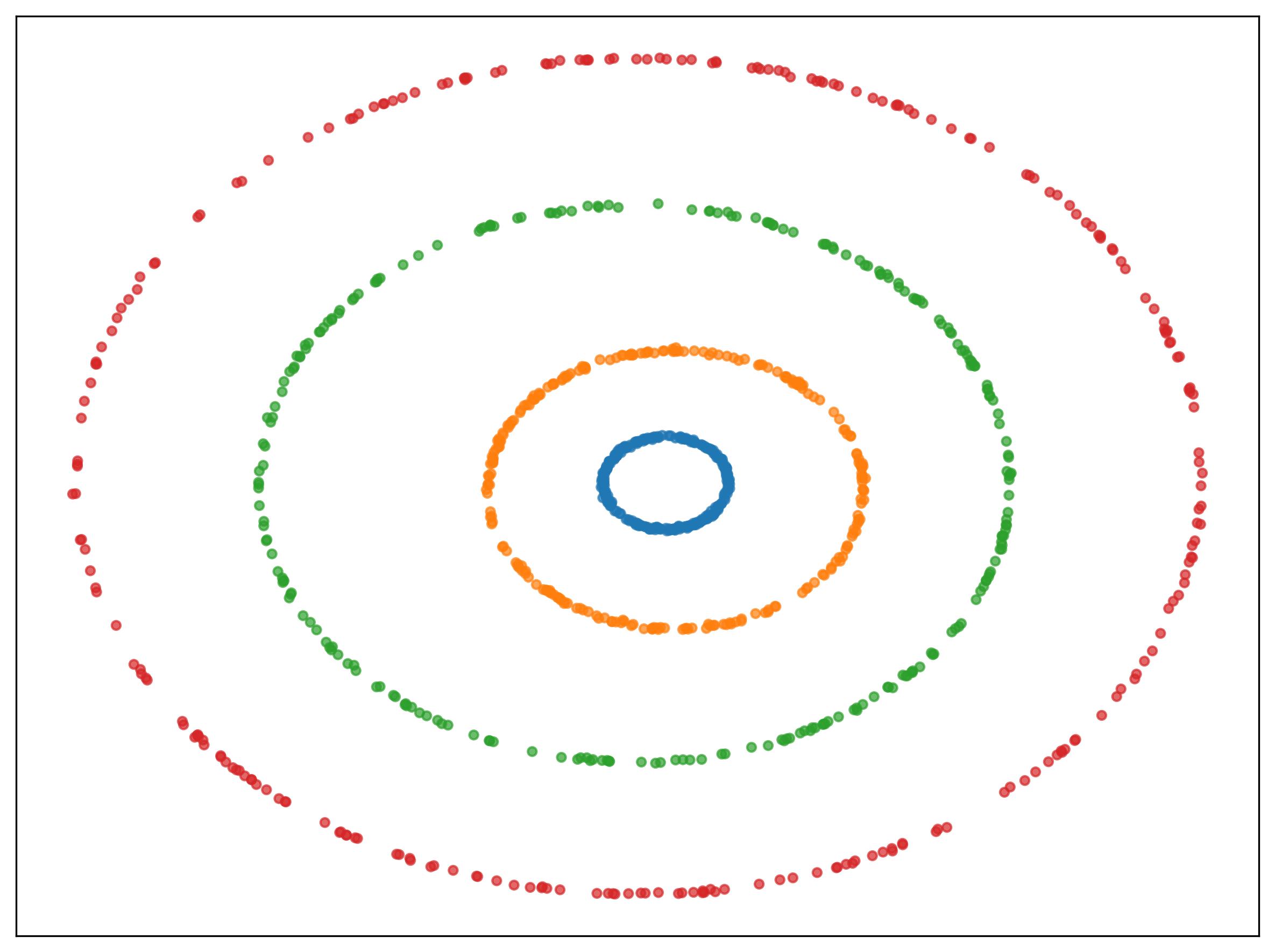} \\
    & & \footnotesize NMI=0.35 & \footnotesize NMI=0.23 & \footnotesize NMI=0.02 & \footnotesize NMI=0.53 & \footnotesize NMI=1.00 \\
    \addlinespace[4pt]

    \rotatebox{90}{\footnotesize RingG} & 
    \includegraphics[width=0.16\textwidth]{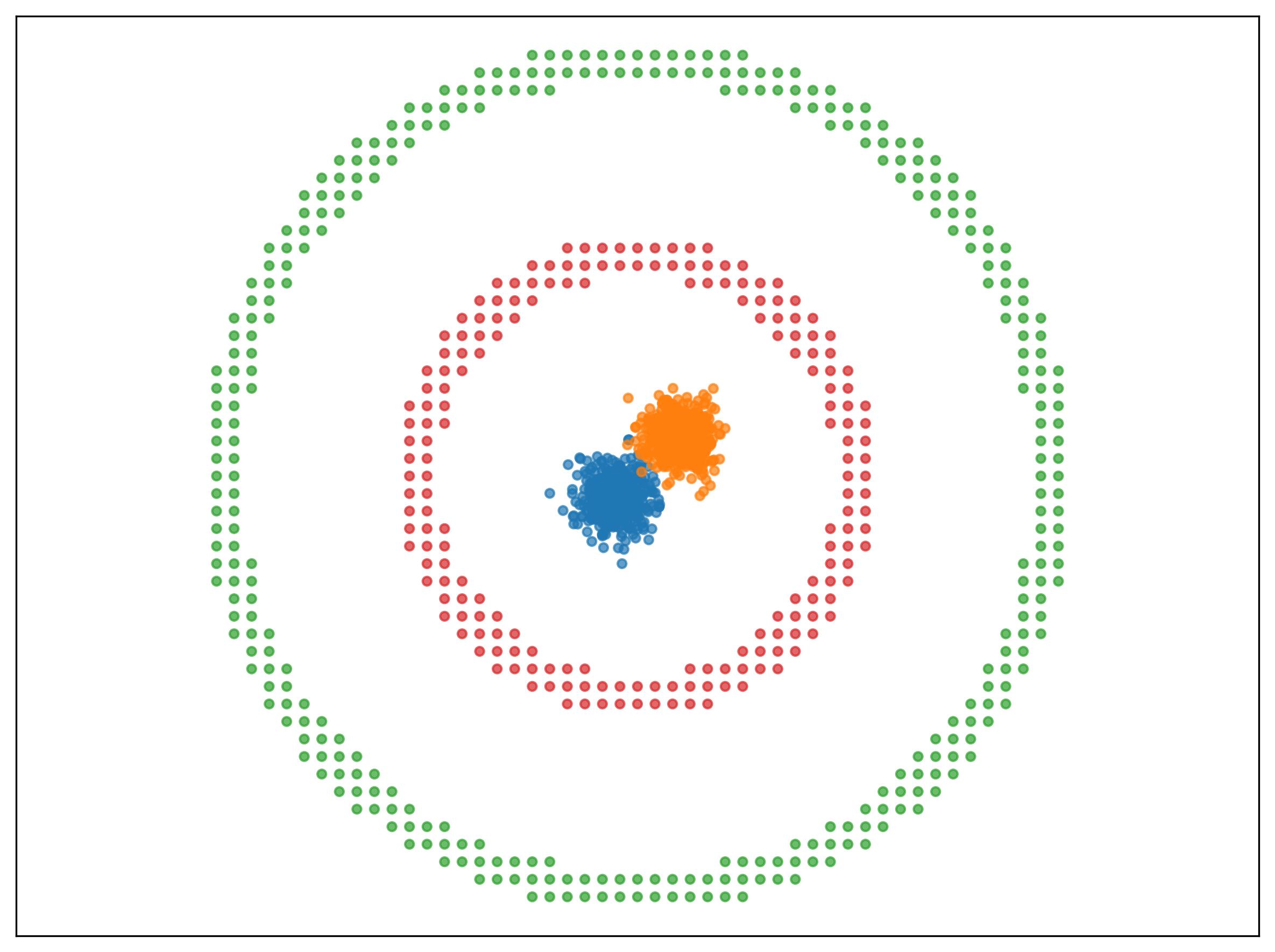} &
    \includegraphics[width=0.16\textwidth]{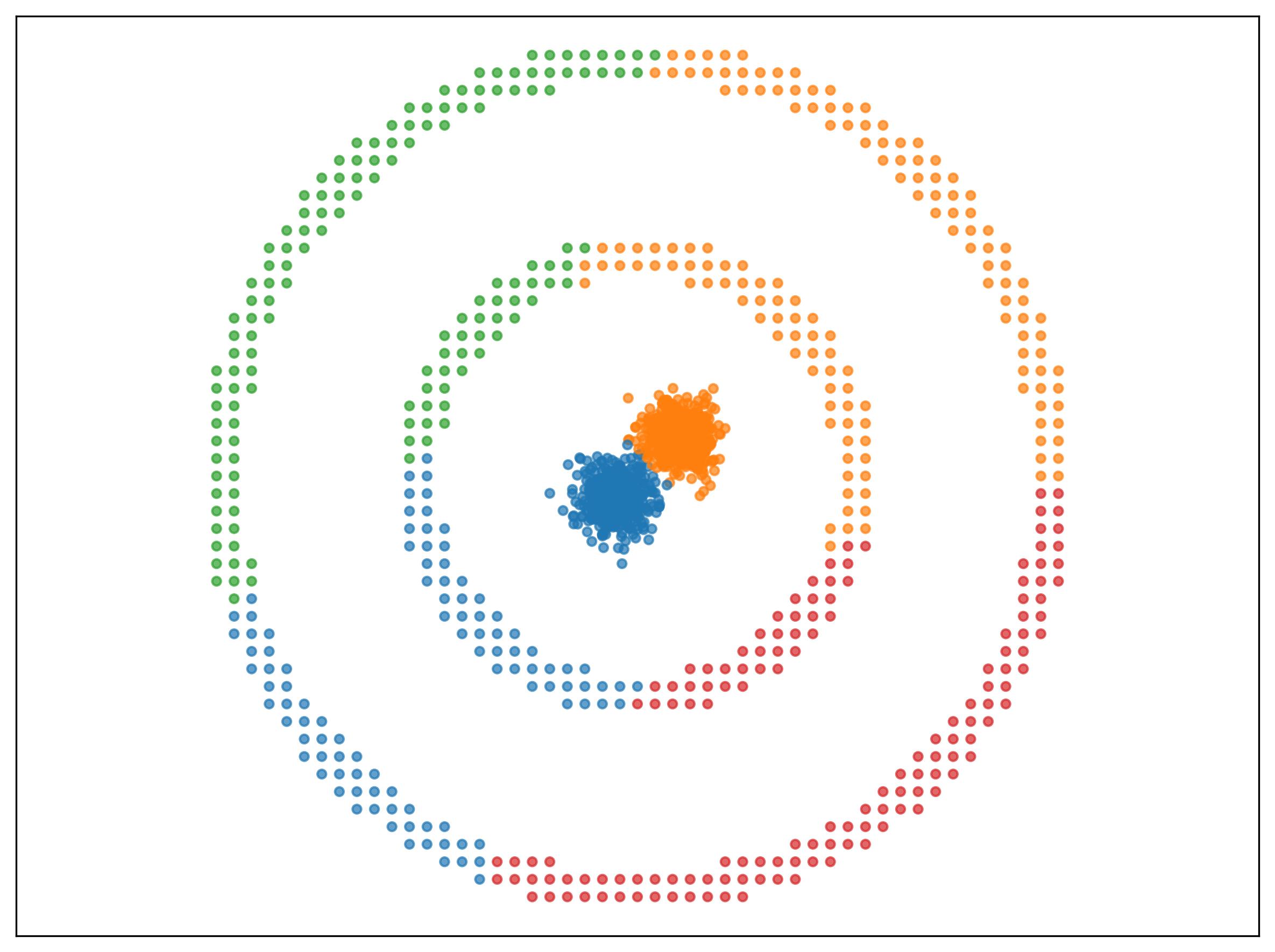} &
    \includegraphics[width=0.16\textwidth]{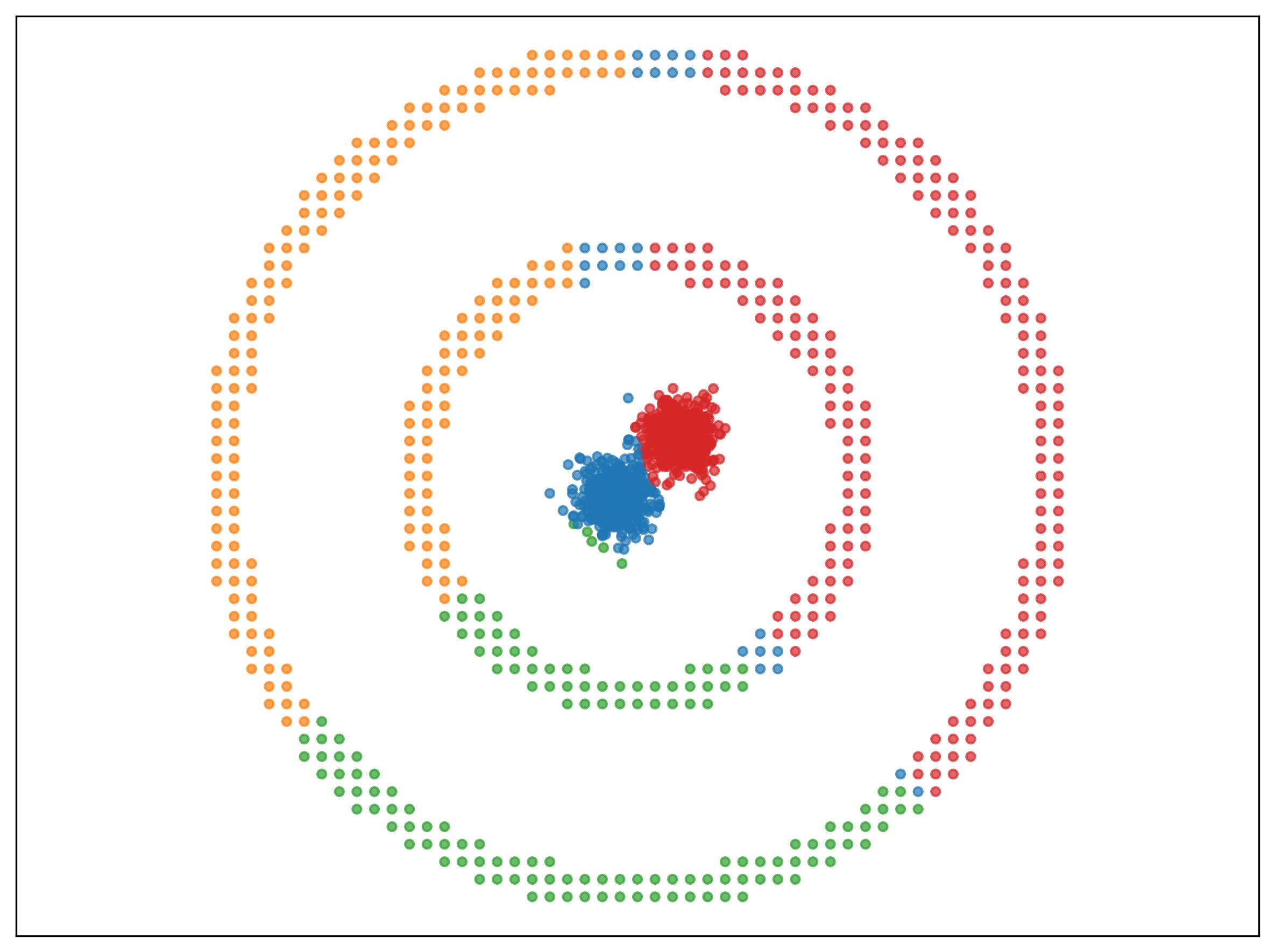} &
    \includegraphics[width=0.16\textwidth]{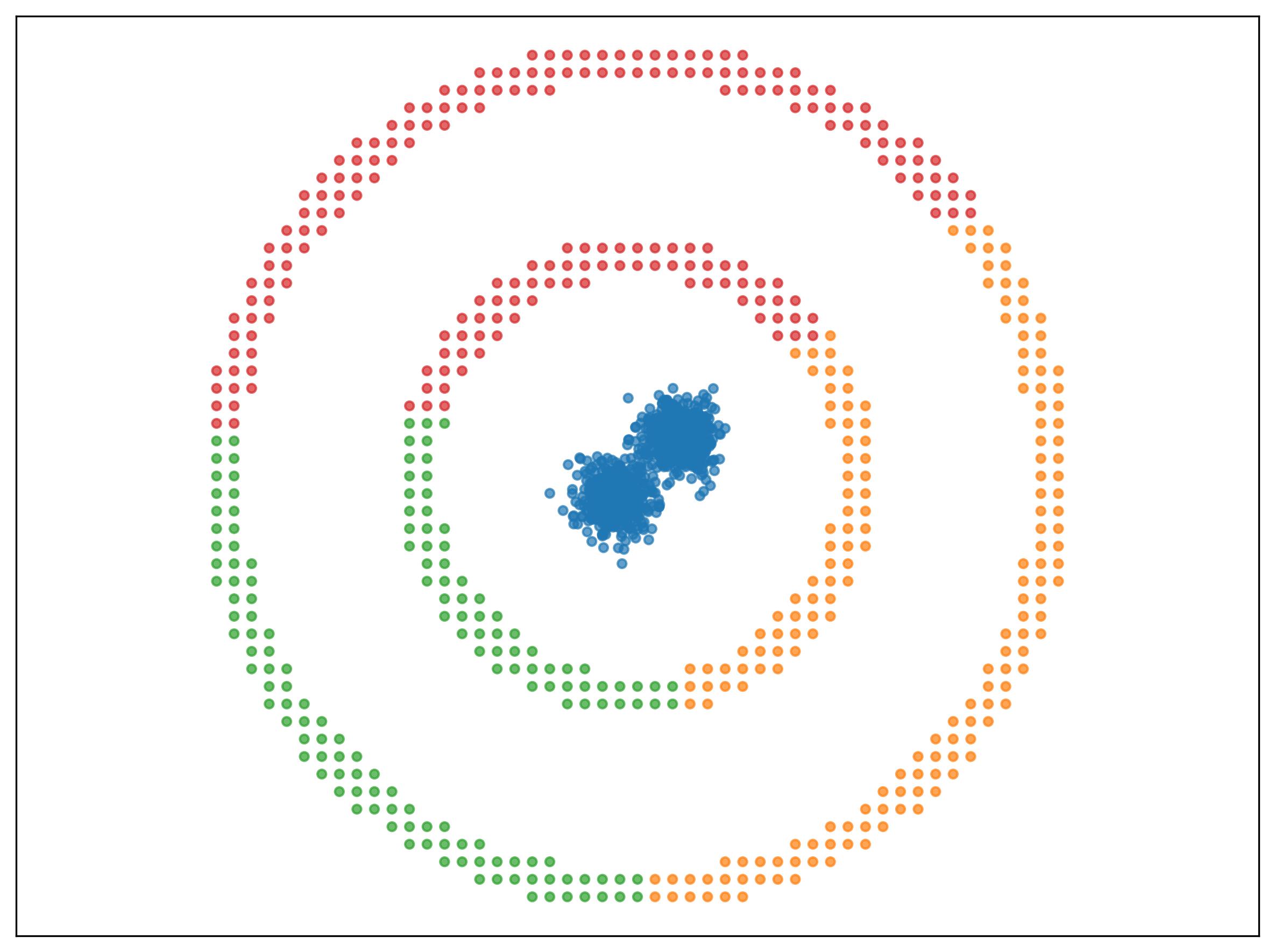} &
    \includegraphics[width=0.16\textwidth]{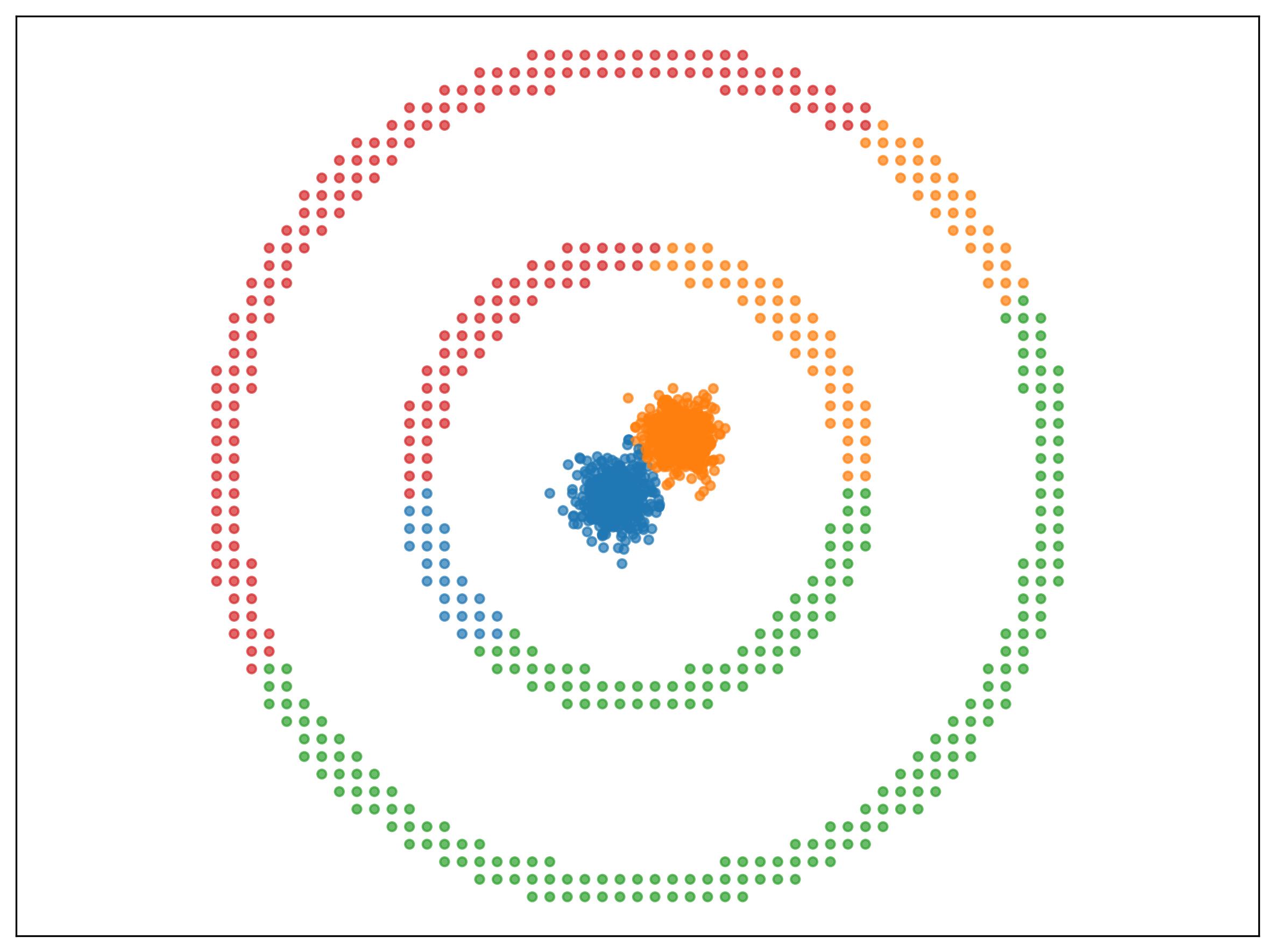} & 
    \includegraphics[width=0.16\textwidth]{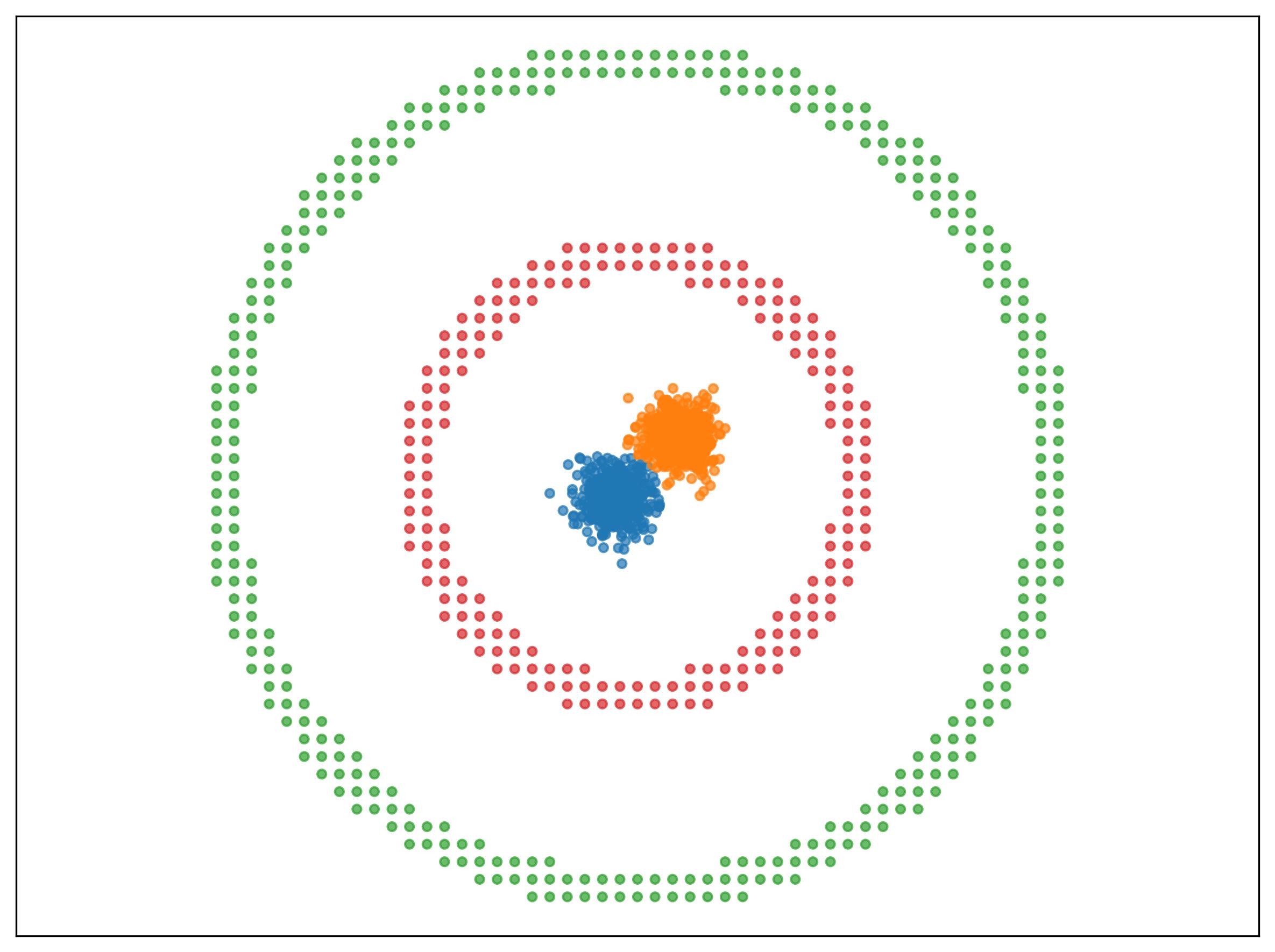} \\
    & & \footnotesize NMI=0.50 & \footnotesize NMI=0.54 & \footnotesize NMI=0.55 & \footnotesize NMI=0.69 & \footnotesize NMI=0.98 \\
    \addlinespace[4pt]

    \rotatebox{90}{\footnotesize complex9} & 
    \includegraphics[width=0.16\textwidth]{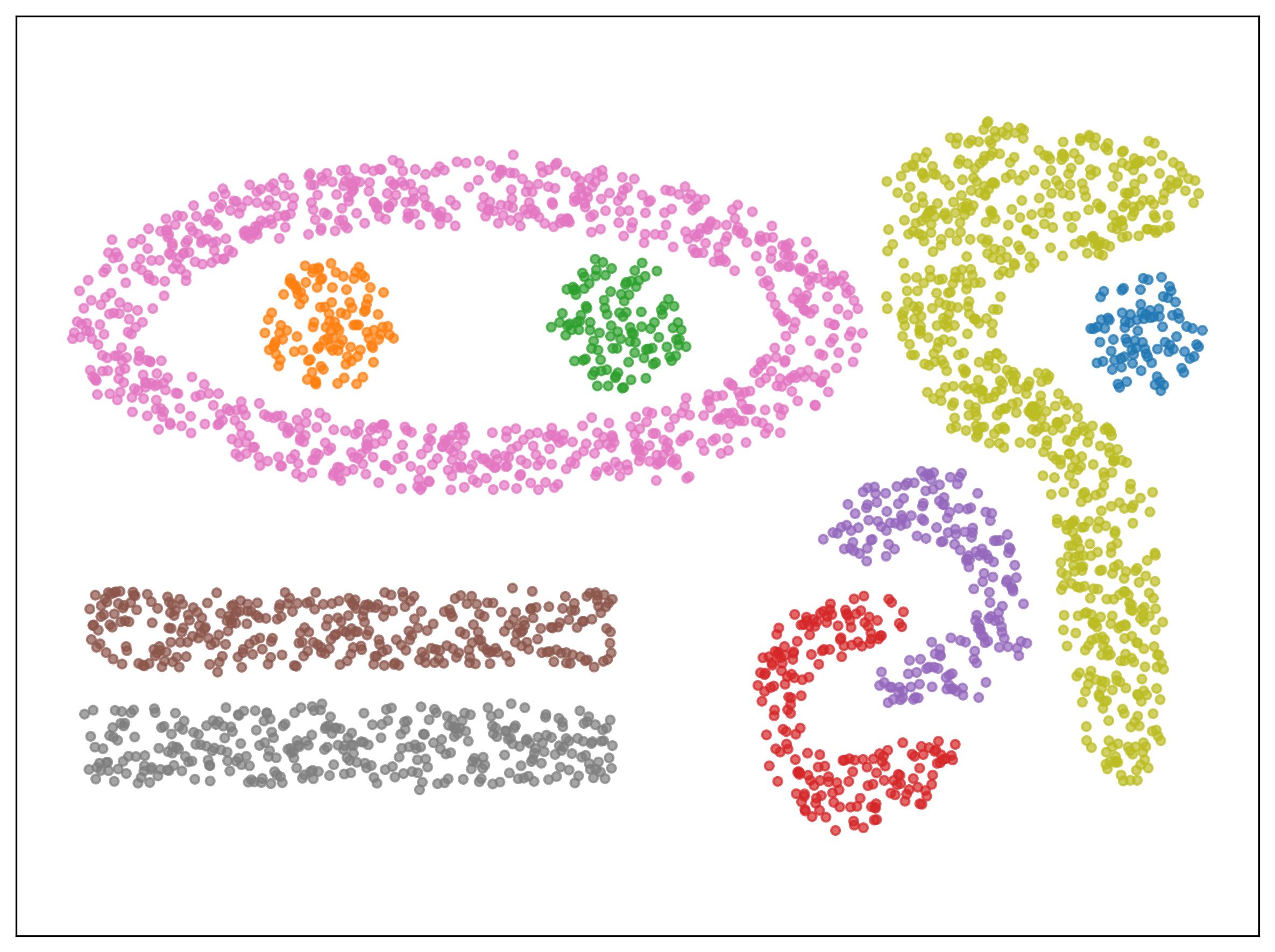} &
    \includegraphics[width=0.16\textwidth]{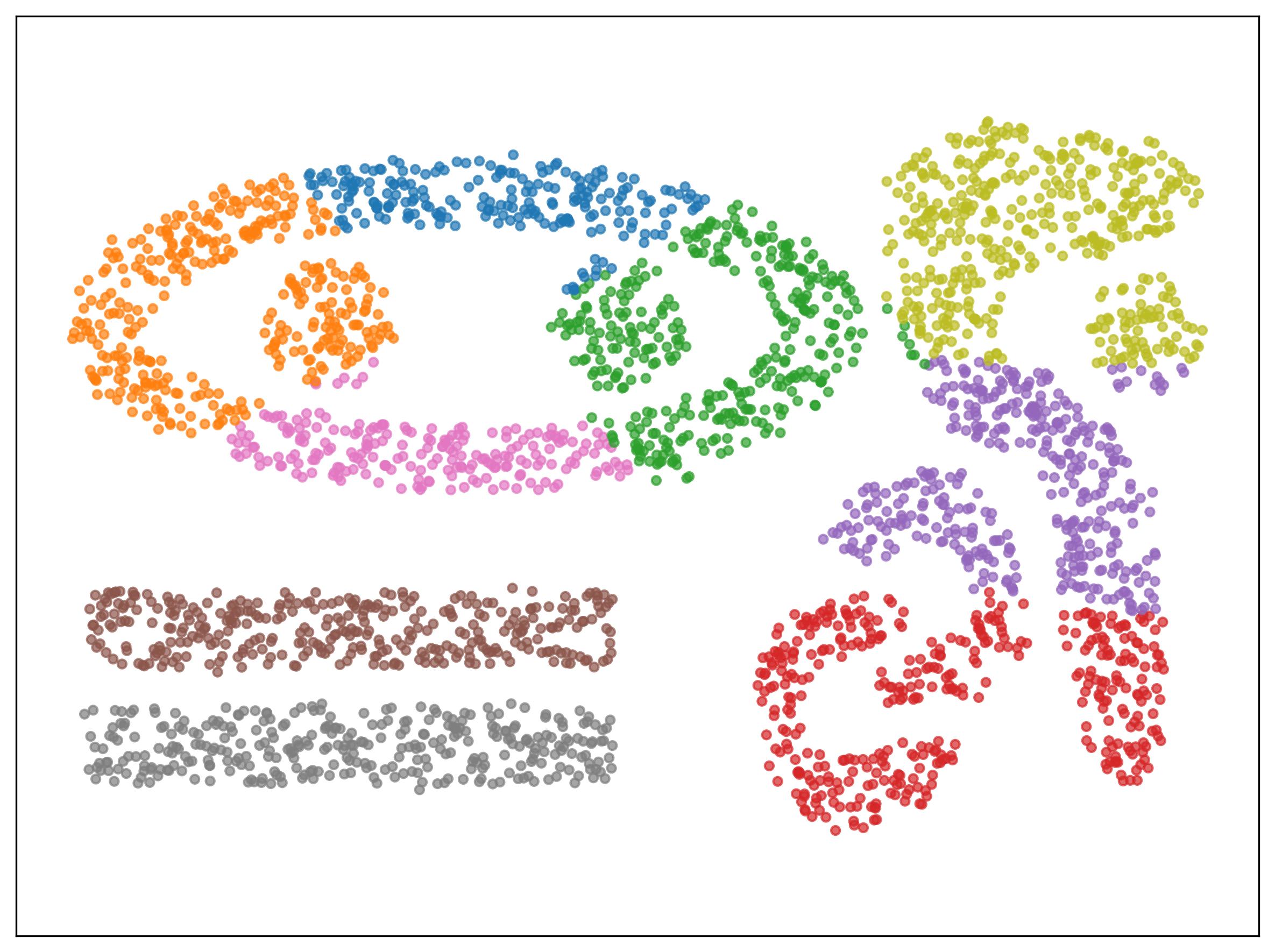} &
    \includegraphics[width=0.16\textwidth]{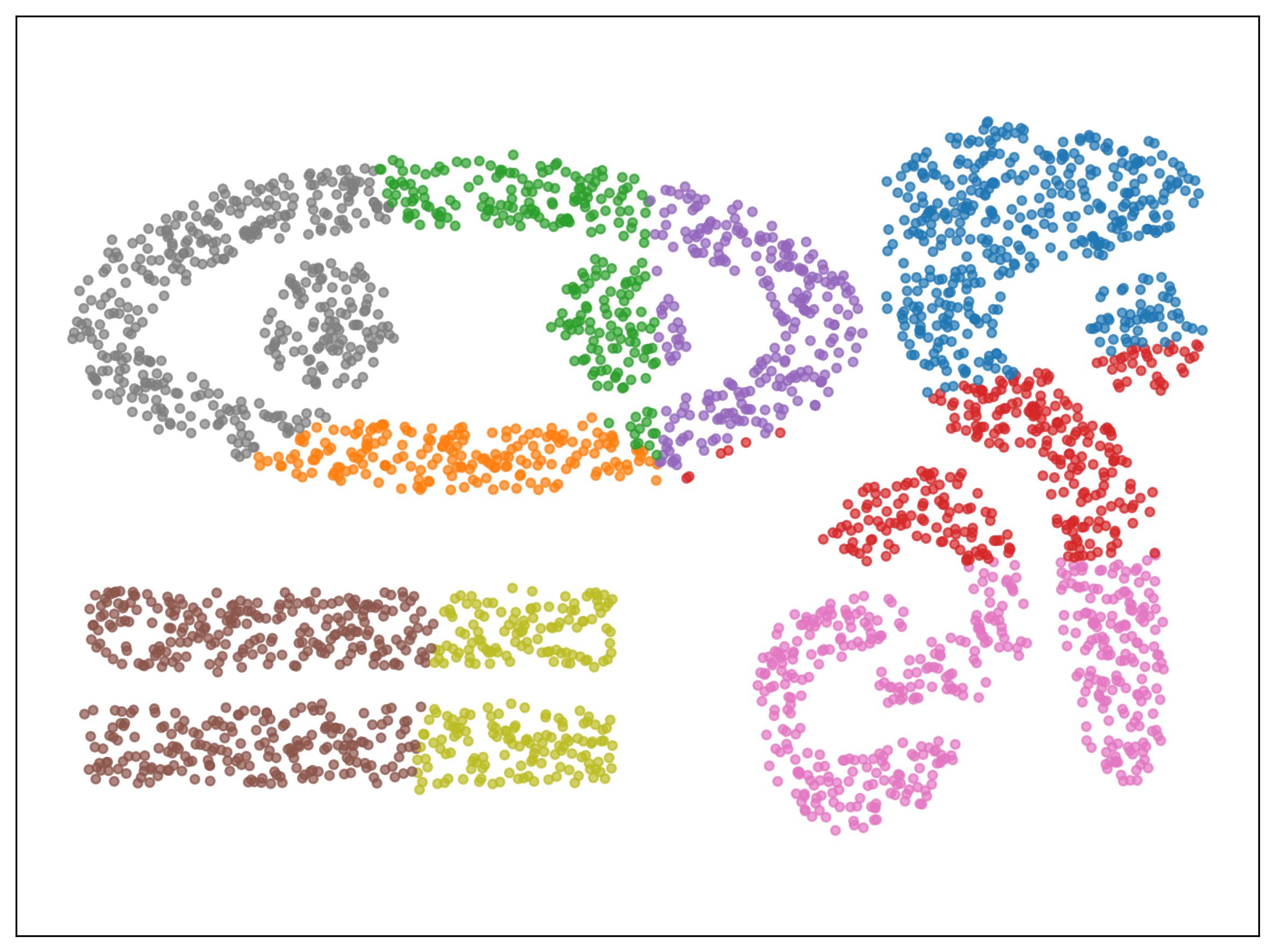} &
    \includegraphics[width=0.16\textwidth]{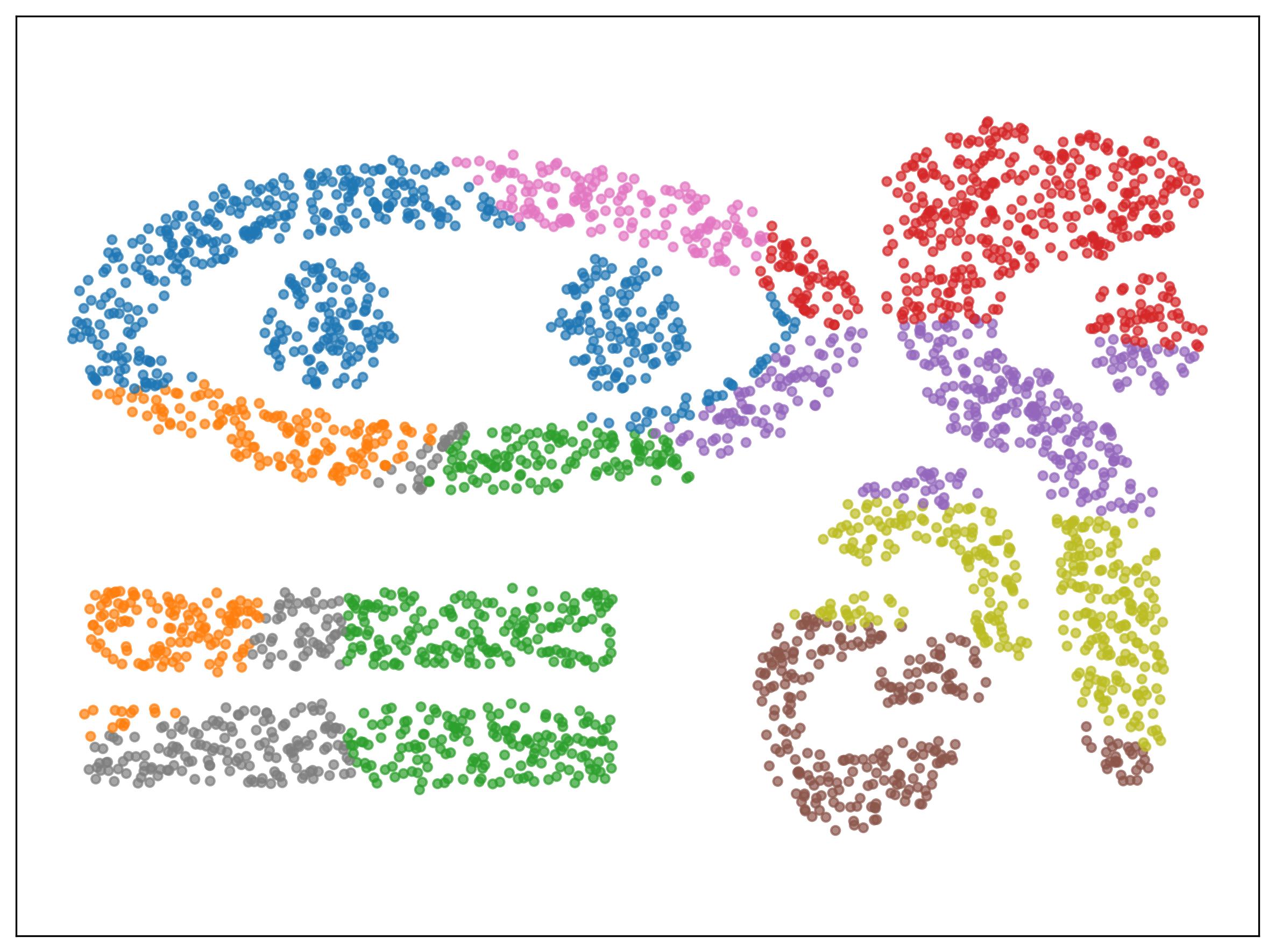} &
    \includegraphics[width=0.16\textwidth]{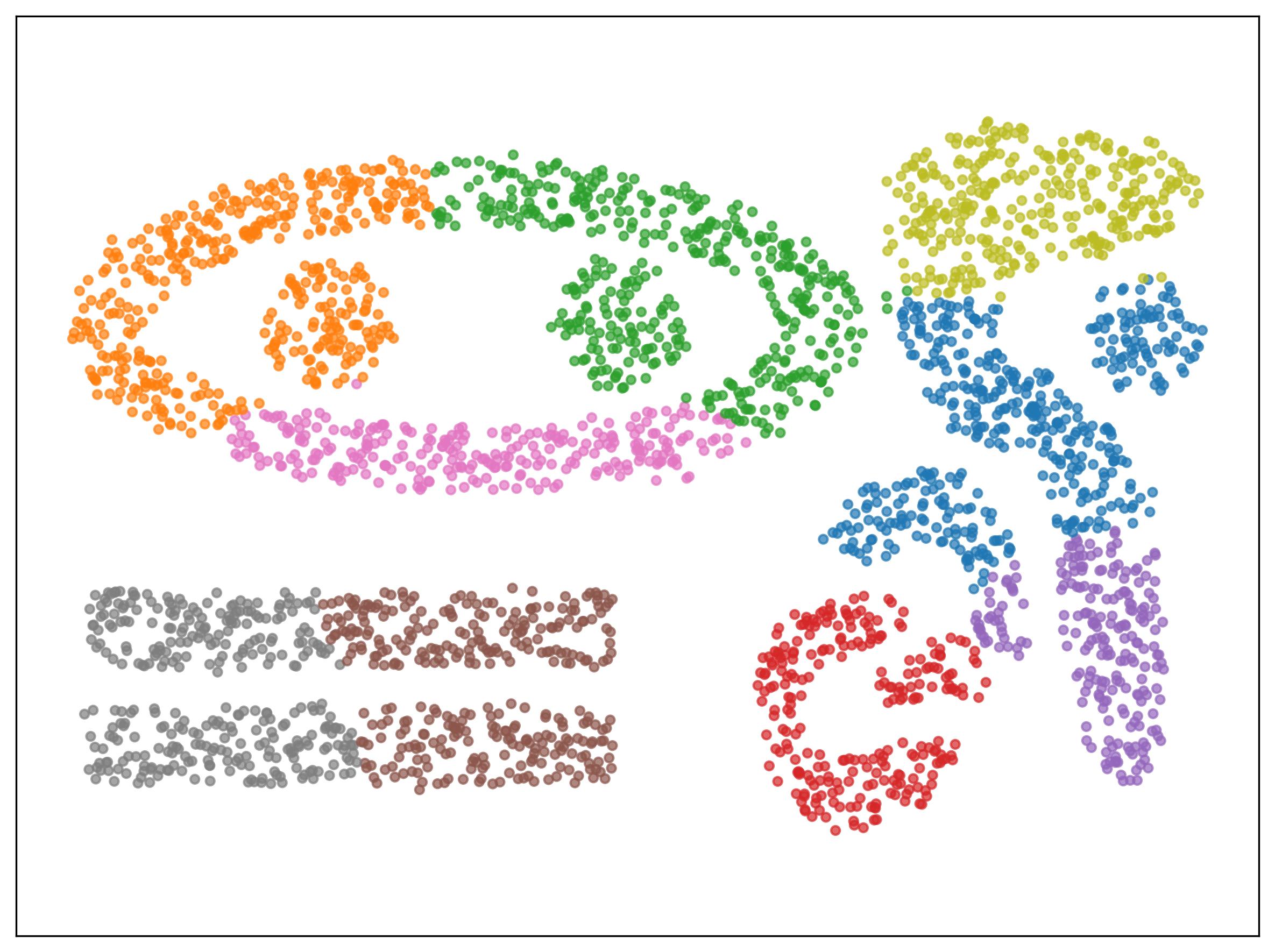} & 
    \includegraphics[width=0.16\textwidth]{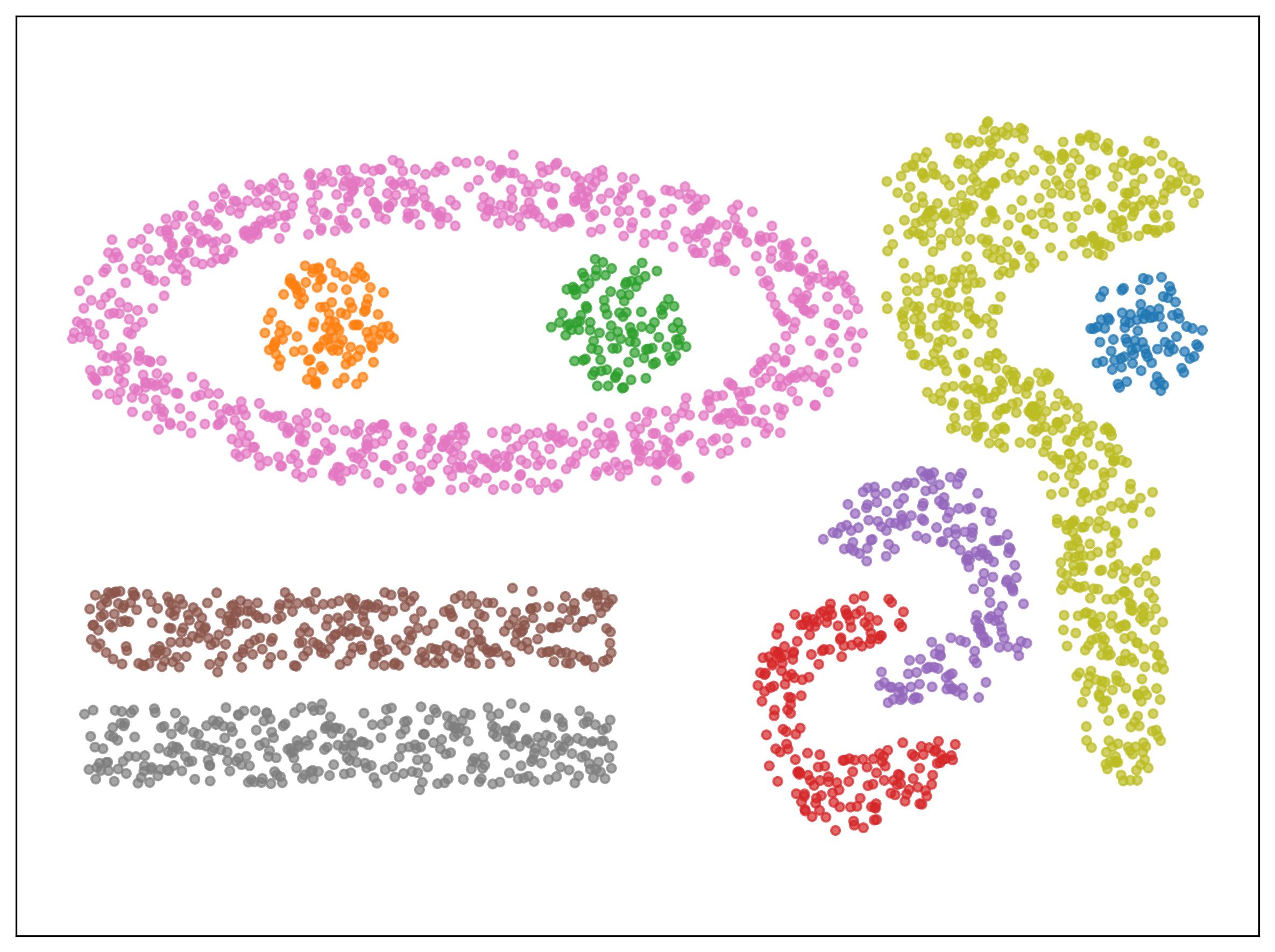} \\
    & & \footnotesize NMI=0.65 & \footnotesize NMI=0.61 & \footnotesize NMI=0.51 & \footnotesize NMI=0.66 & \footnotesize NMI=1.00    \\
    \addlinespace[4pt]

    \rotatebox{90}{\footnotesize S3D3D3} & 
    \includegraphics[width=0.16\textwidth]{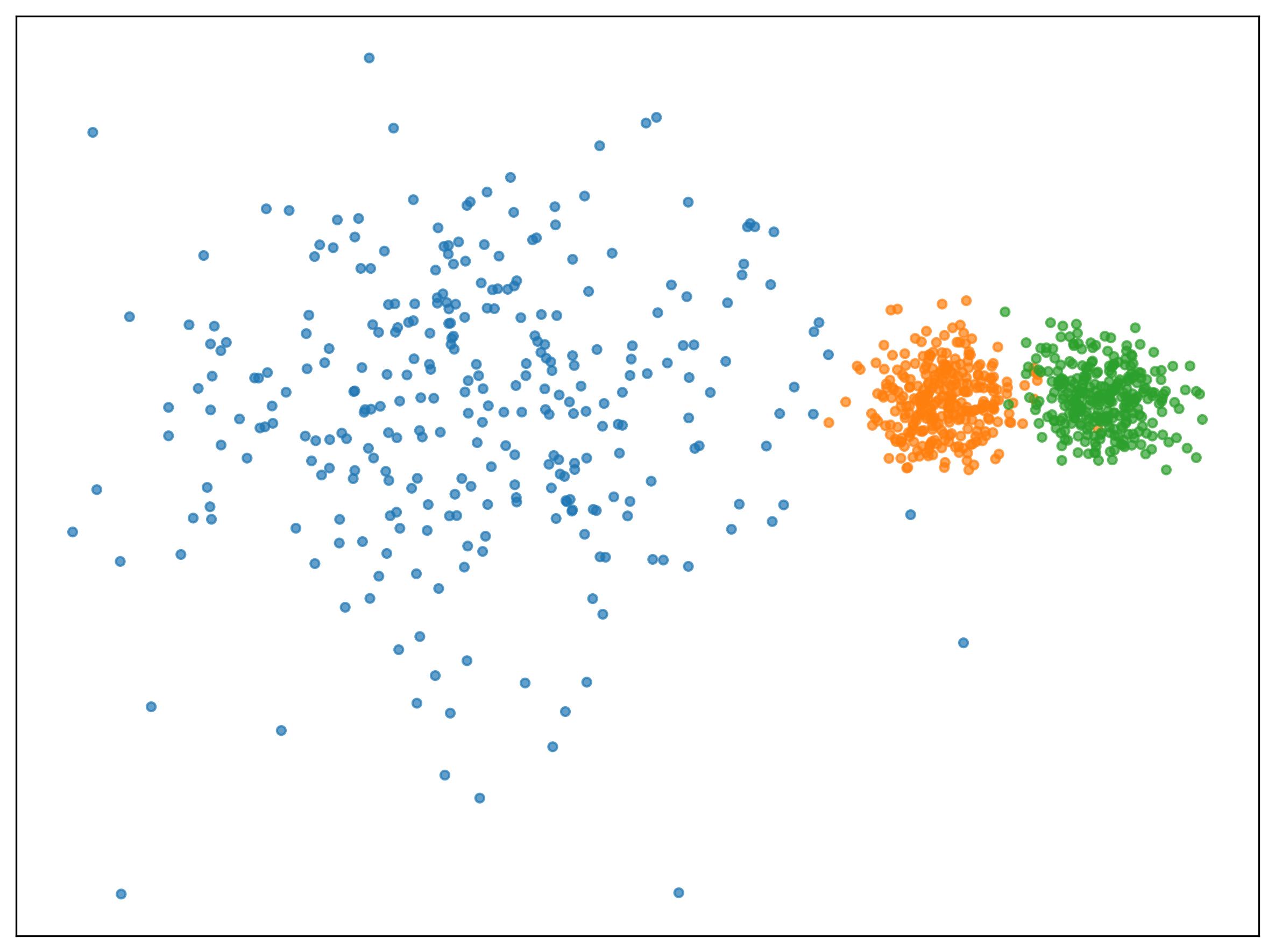} &
    \includegraphics[width=0.16\textwidth]{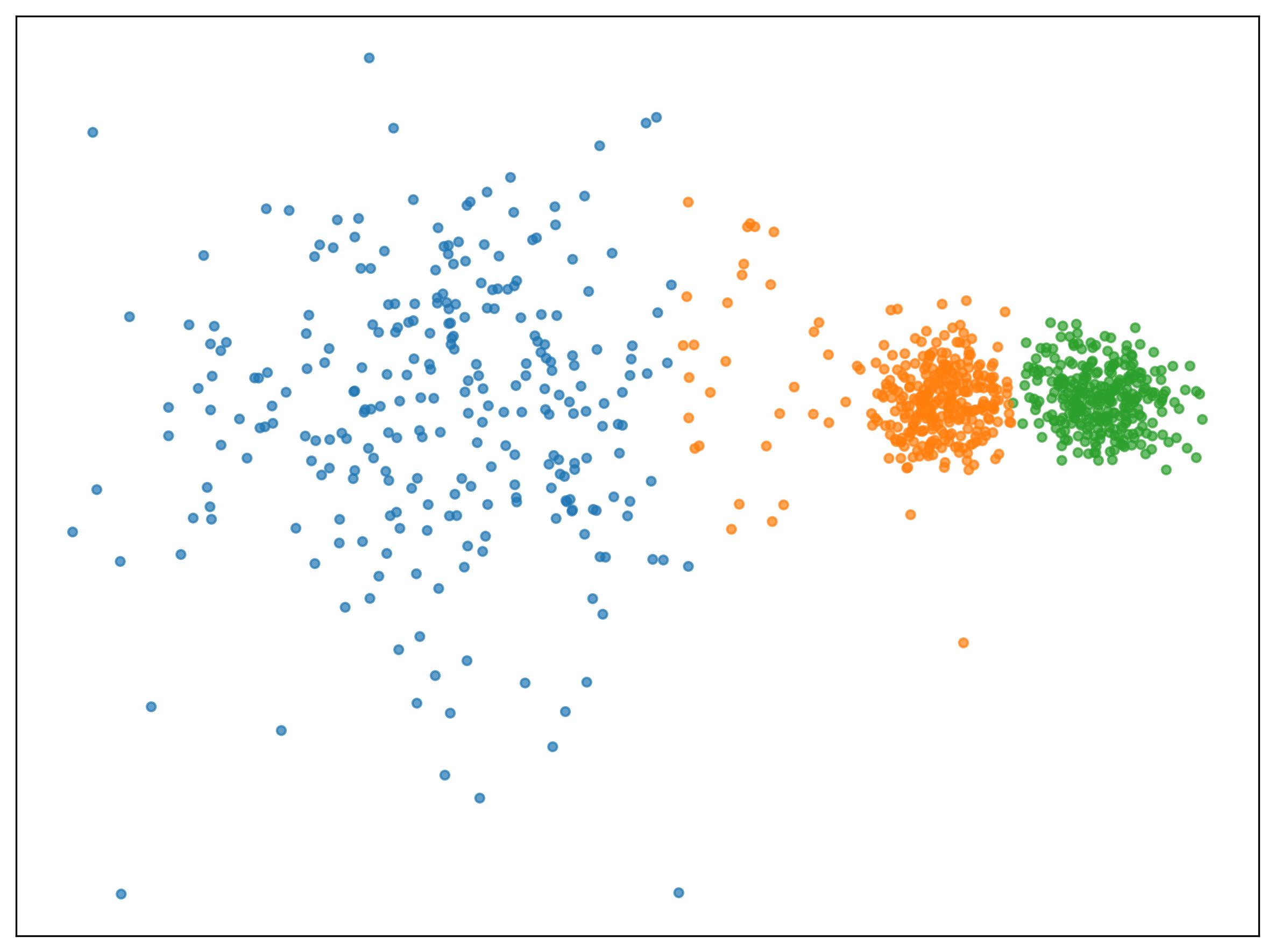} &
    \includegraphics[width=0.16\textwidth]{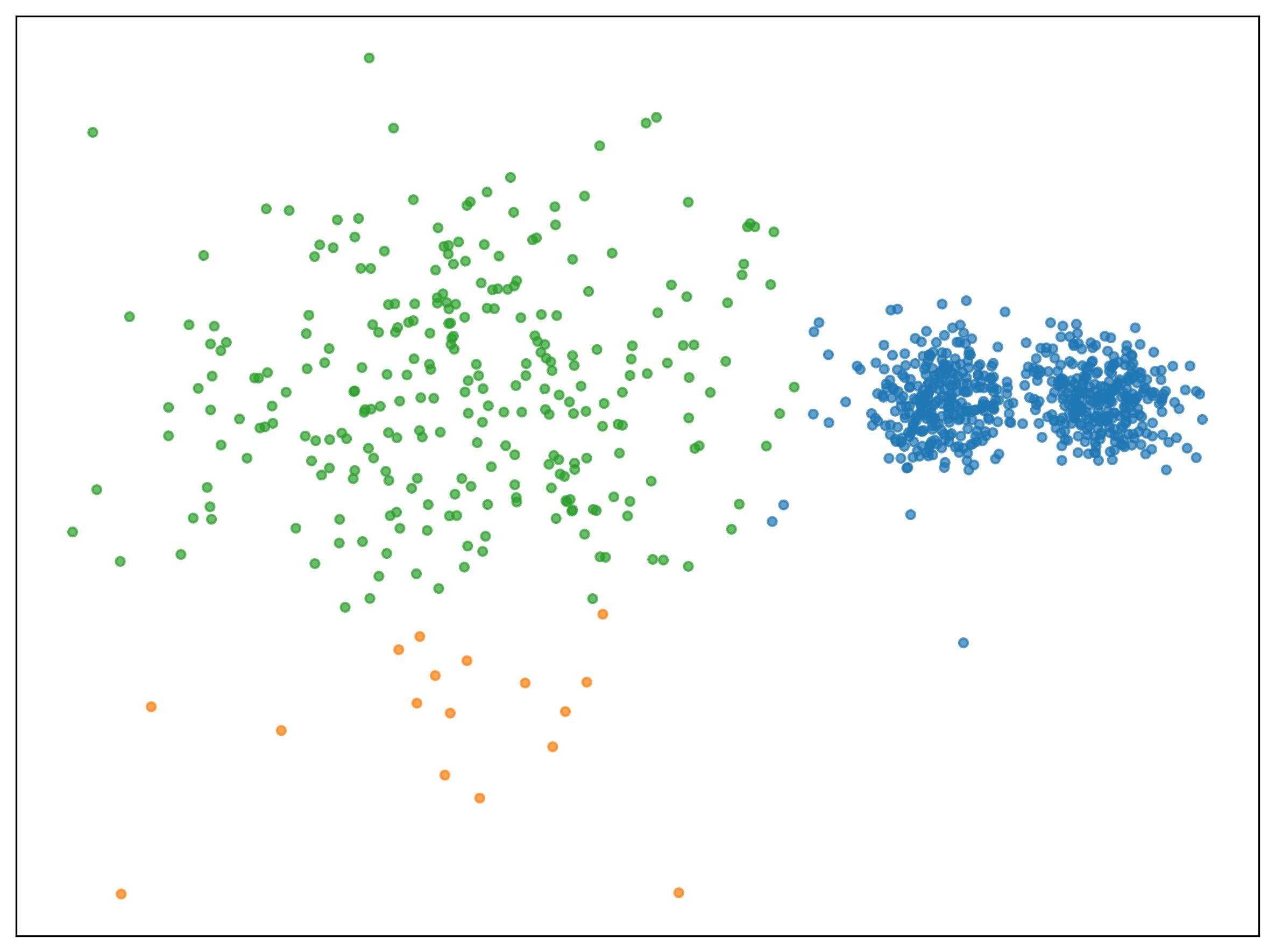} &
    \includegraphics[width=0.16\textwidth]{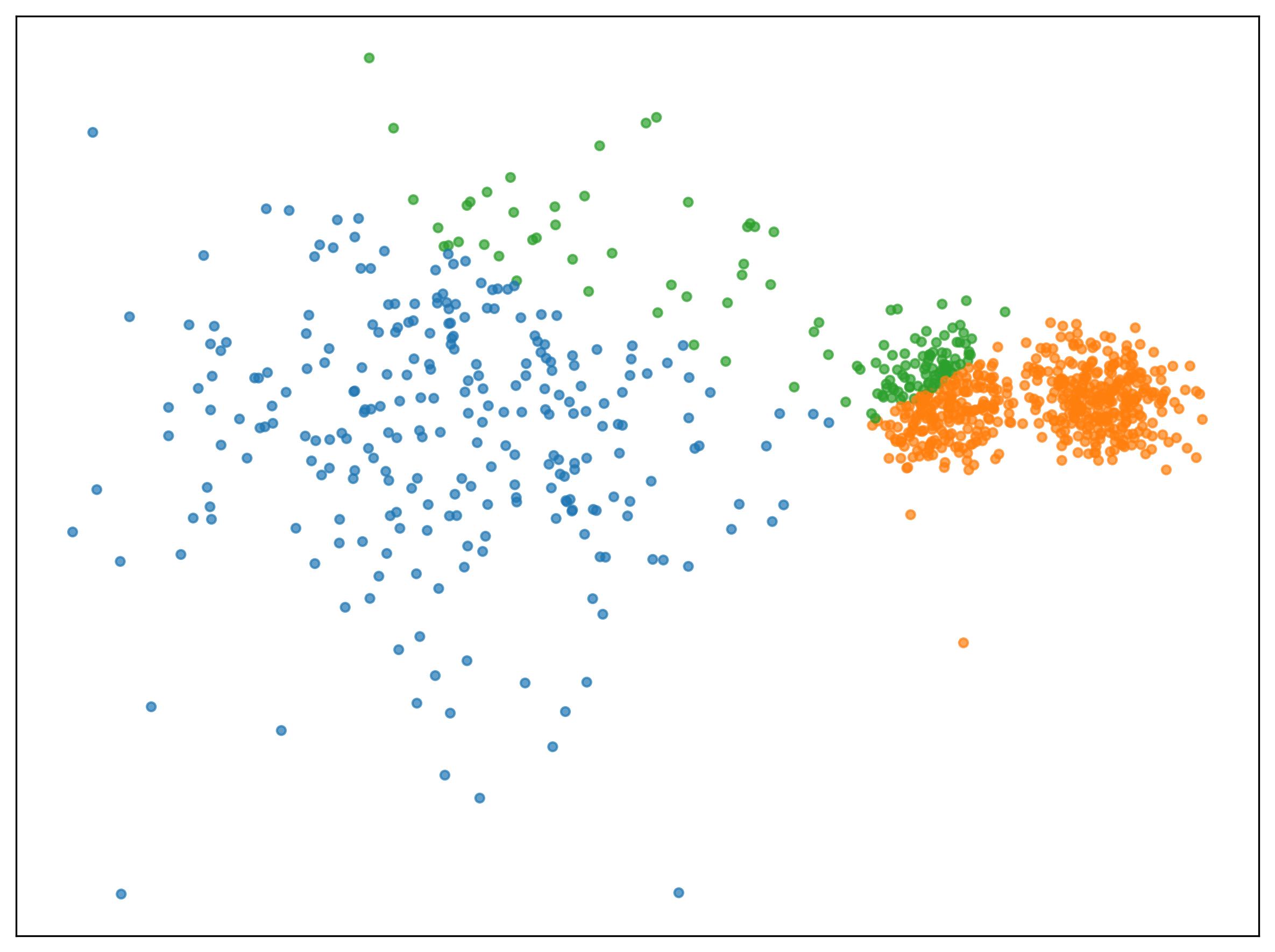} &
    \includegraphics[width=0.16\textwidth]{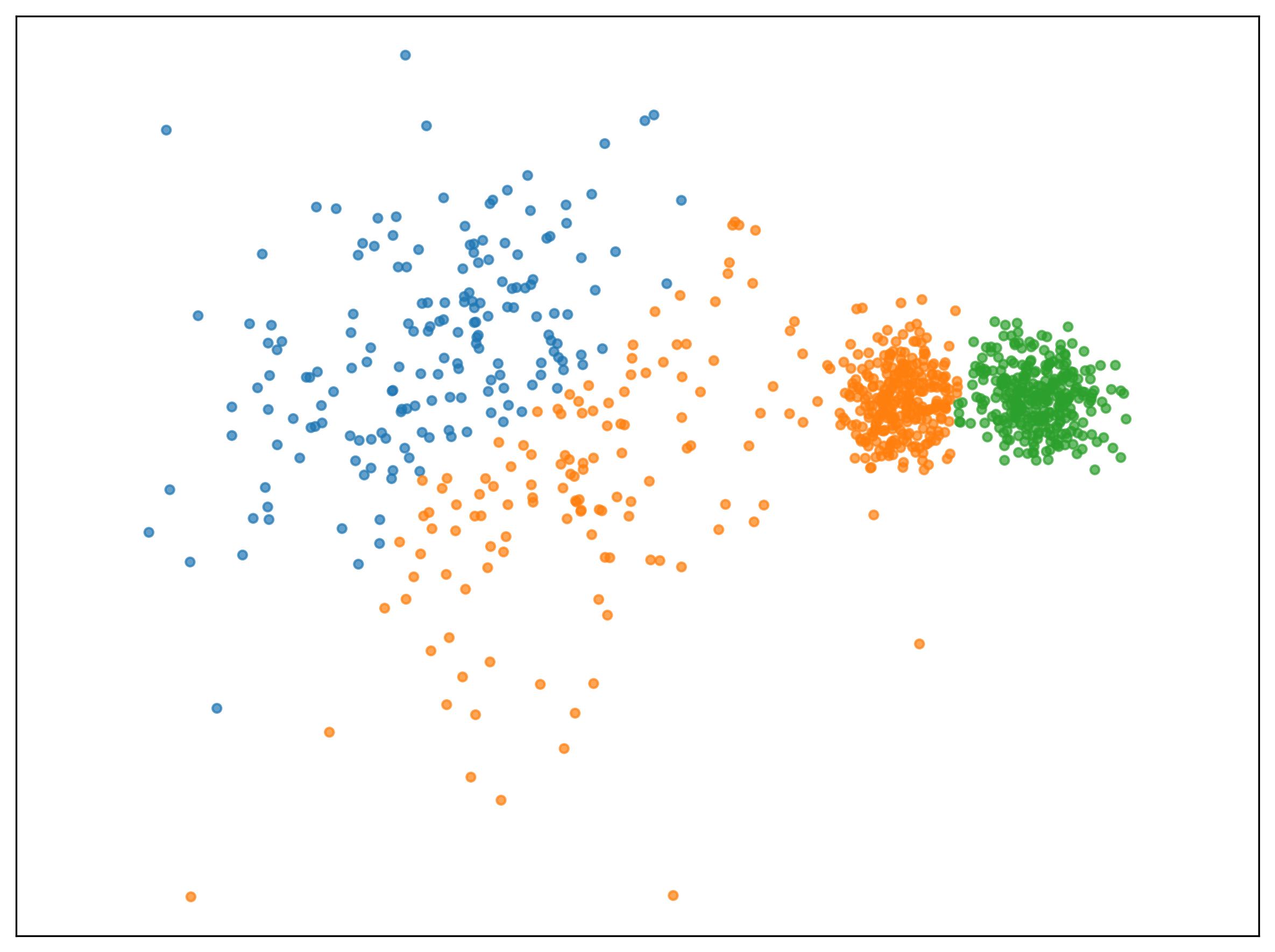} & 
    \includegraphics[width=0.16\textwidth]{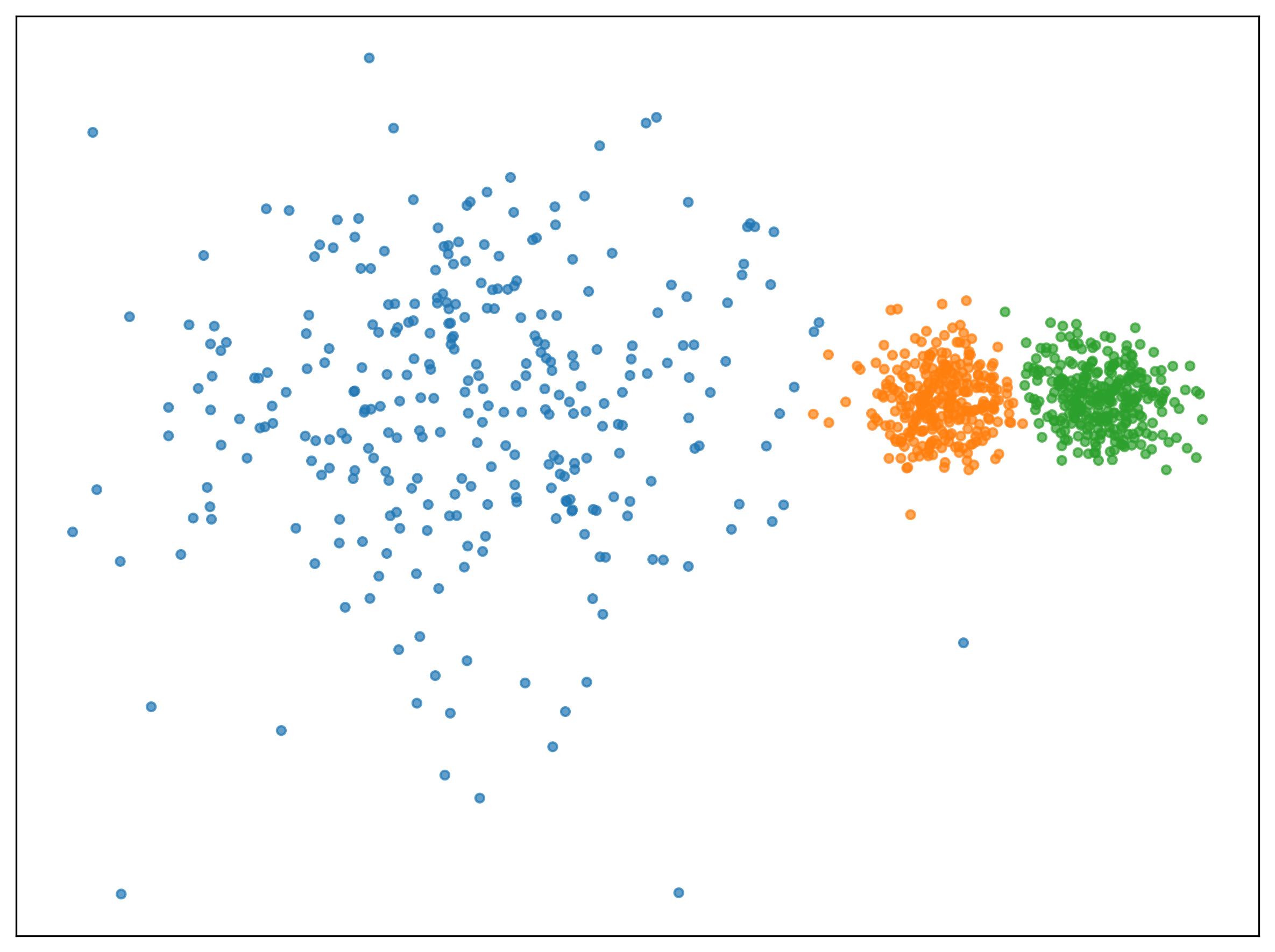} \\
    & & \footnotesize NMI=0.74 & \footnotesize NMI=0.65 & \footnotesize NMI=0.58 & \footnotesize NMI=0.70 & \footnotesize NMI=0.94 \\
    \addlinespace[4pt]

    \rotatebox{90}{\footnotesize OGOL} & 
    \includegraphics[width=0.16\textwidth]{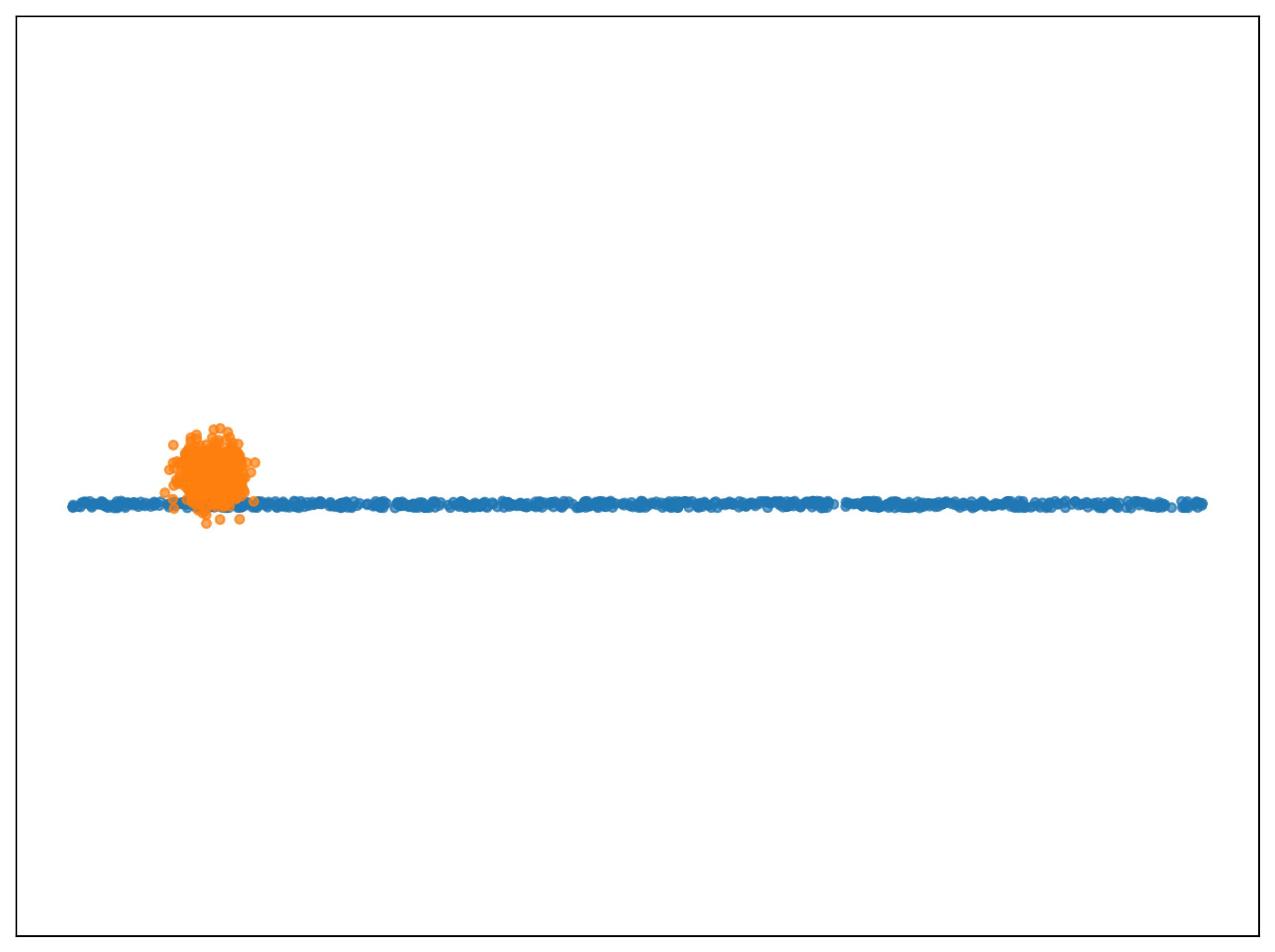} &
    \includegraphics[width=0.16\textwidth]{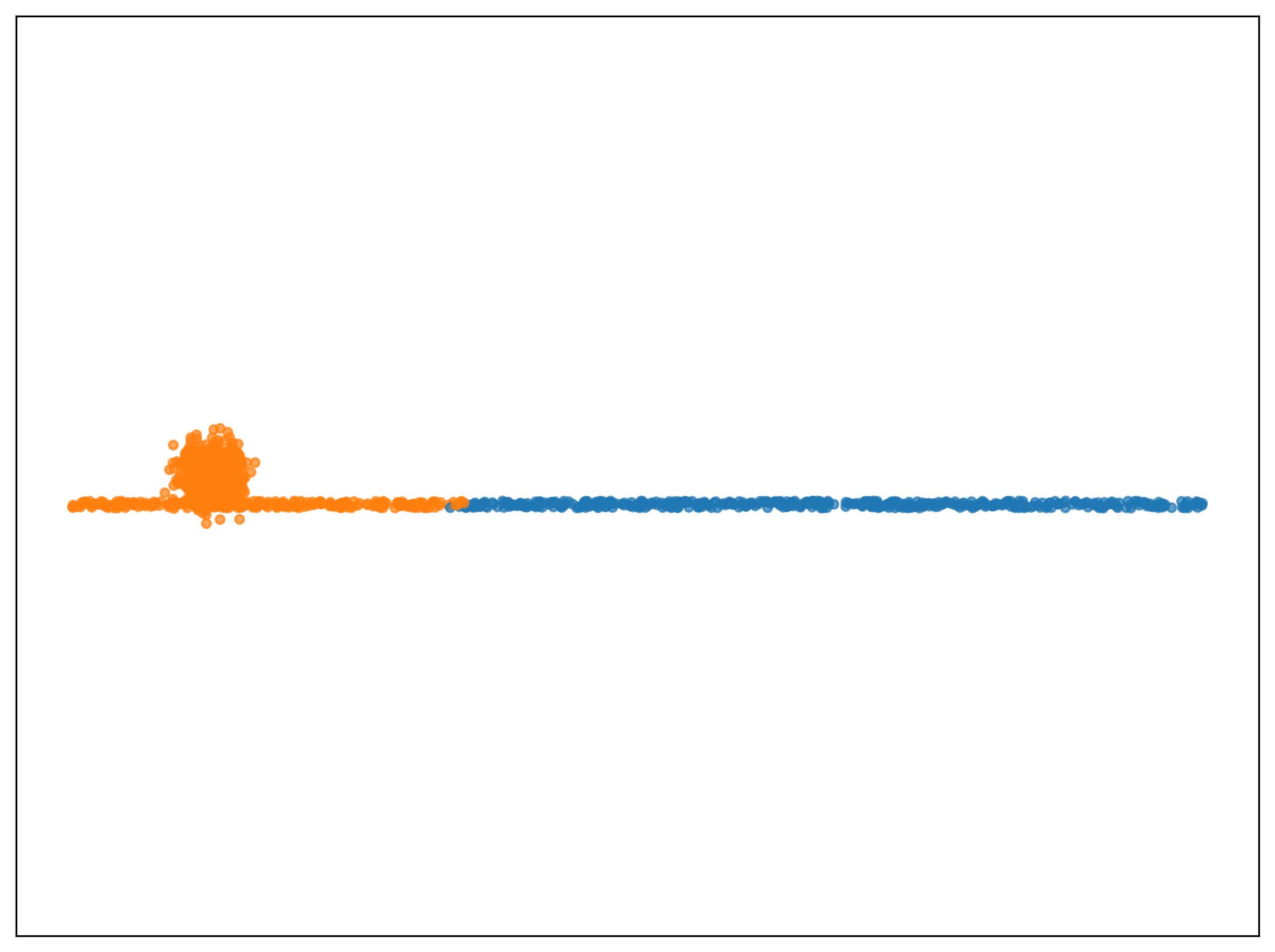} &
    \includegraphics[width=0.16\textwidth]{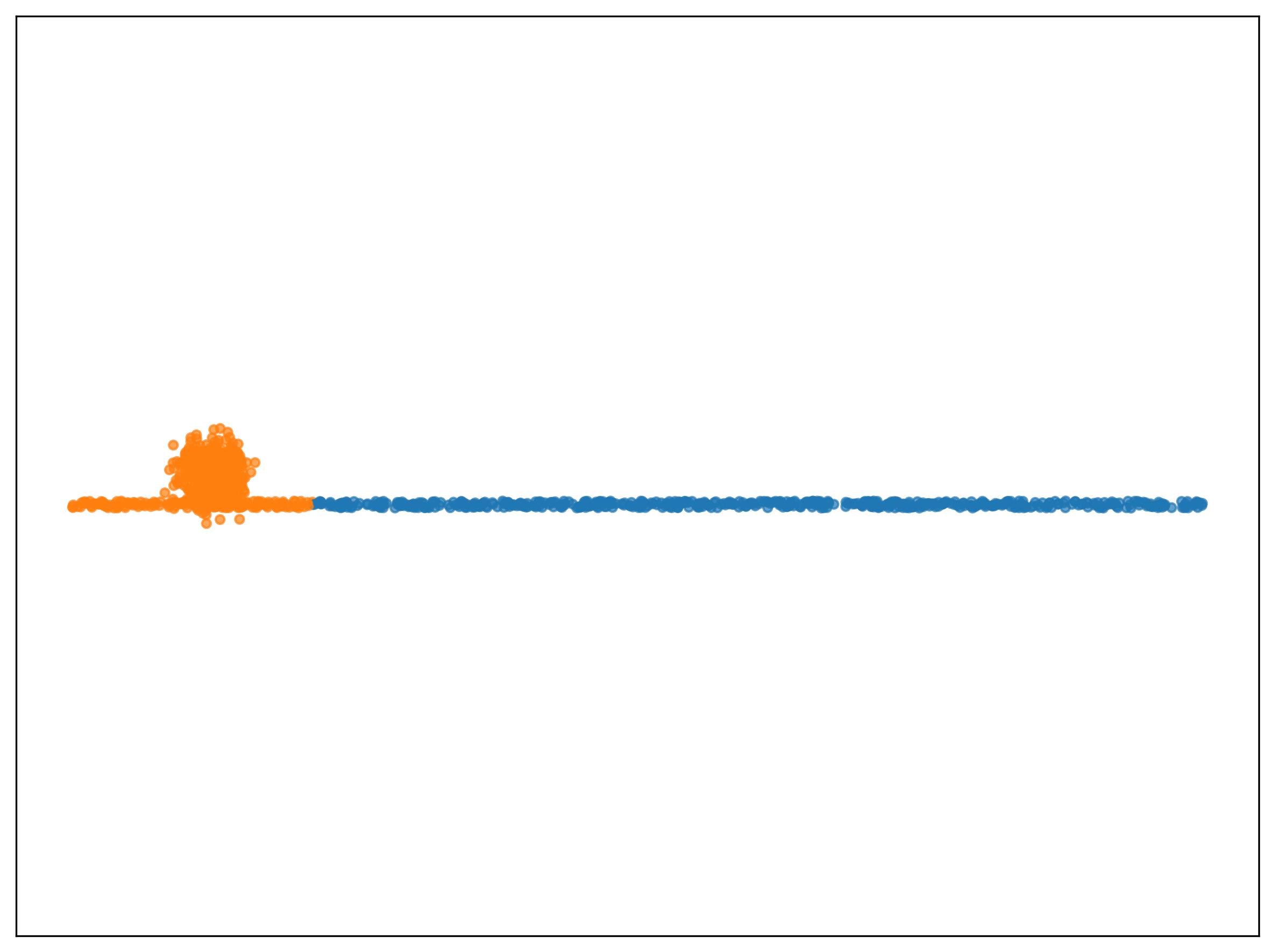} &
    \includegraphics[width=0.16\textwidth]{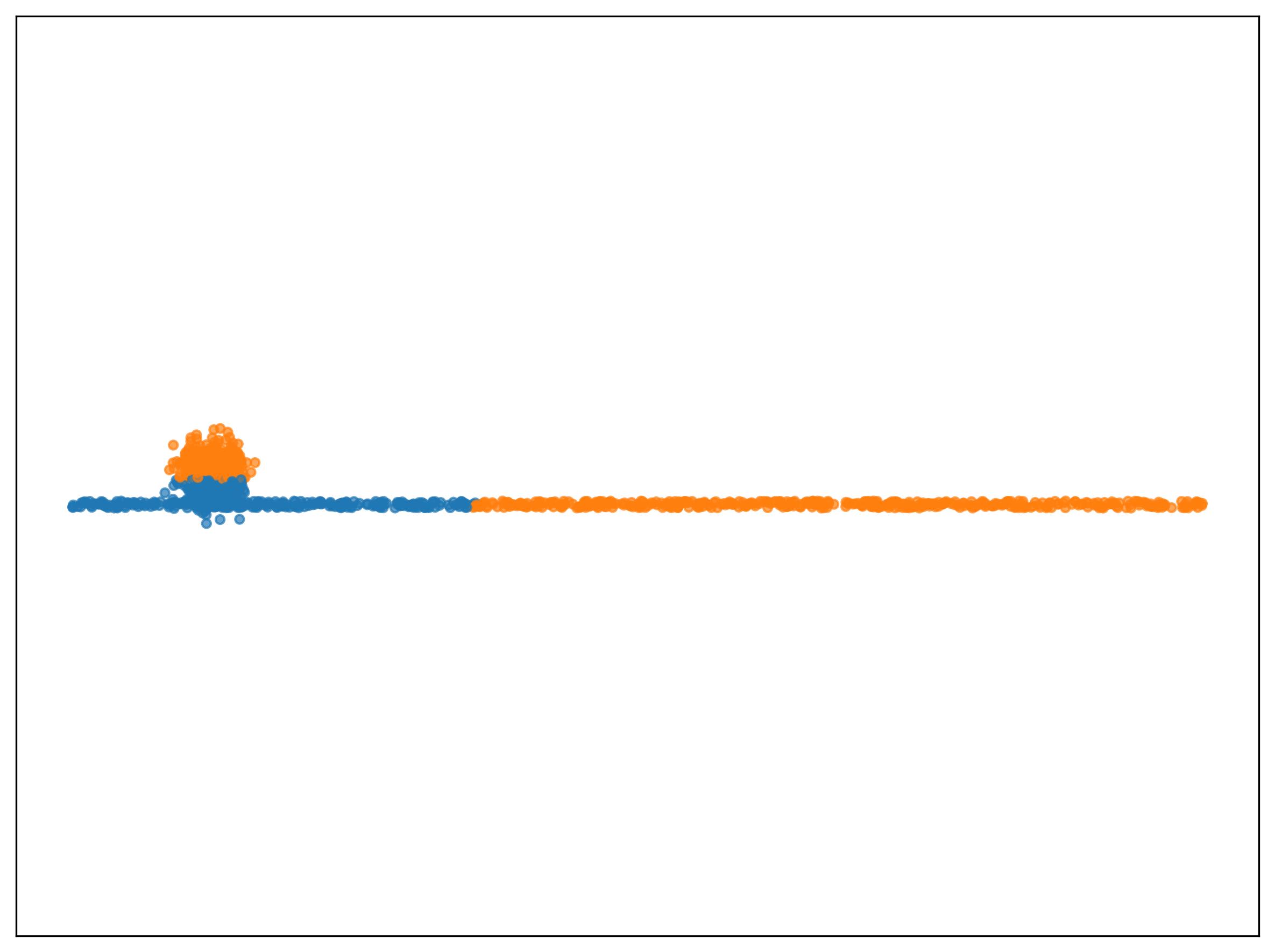} &
    \includegraphics[width=0.16\textwidth]{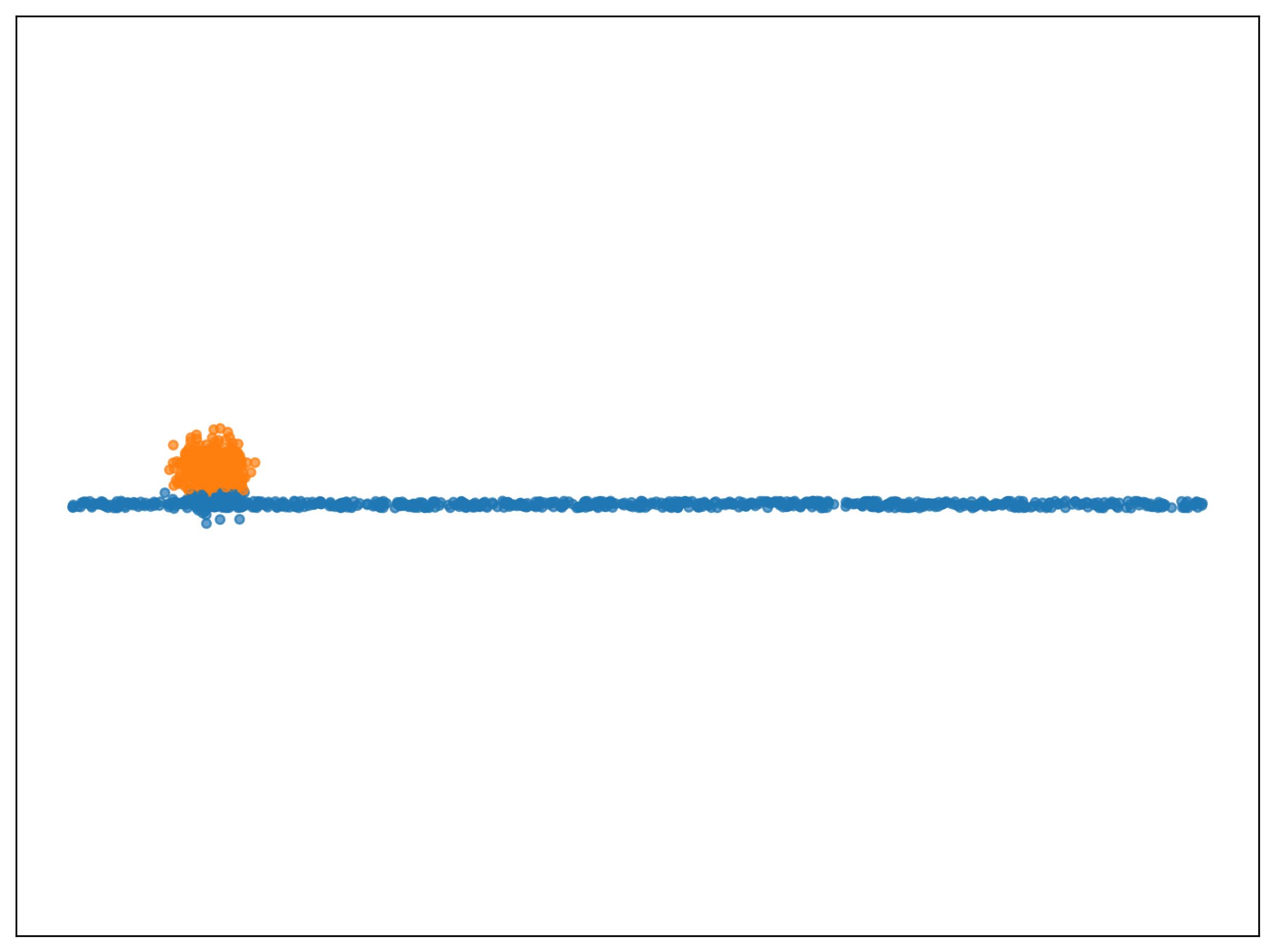} & 
    \includegraphics[width=0.16\textwidth]{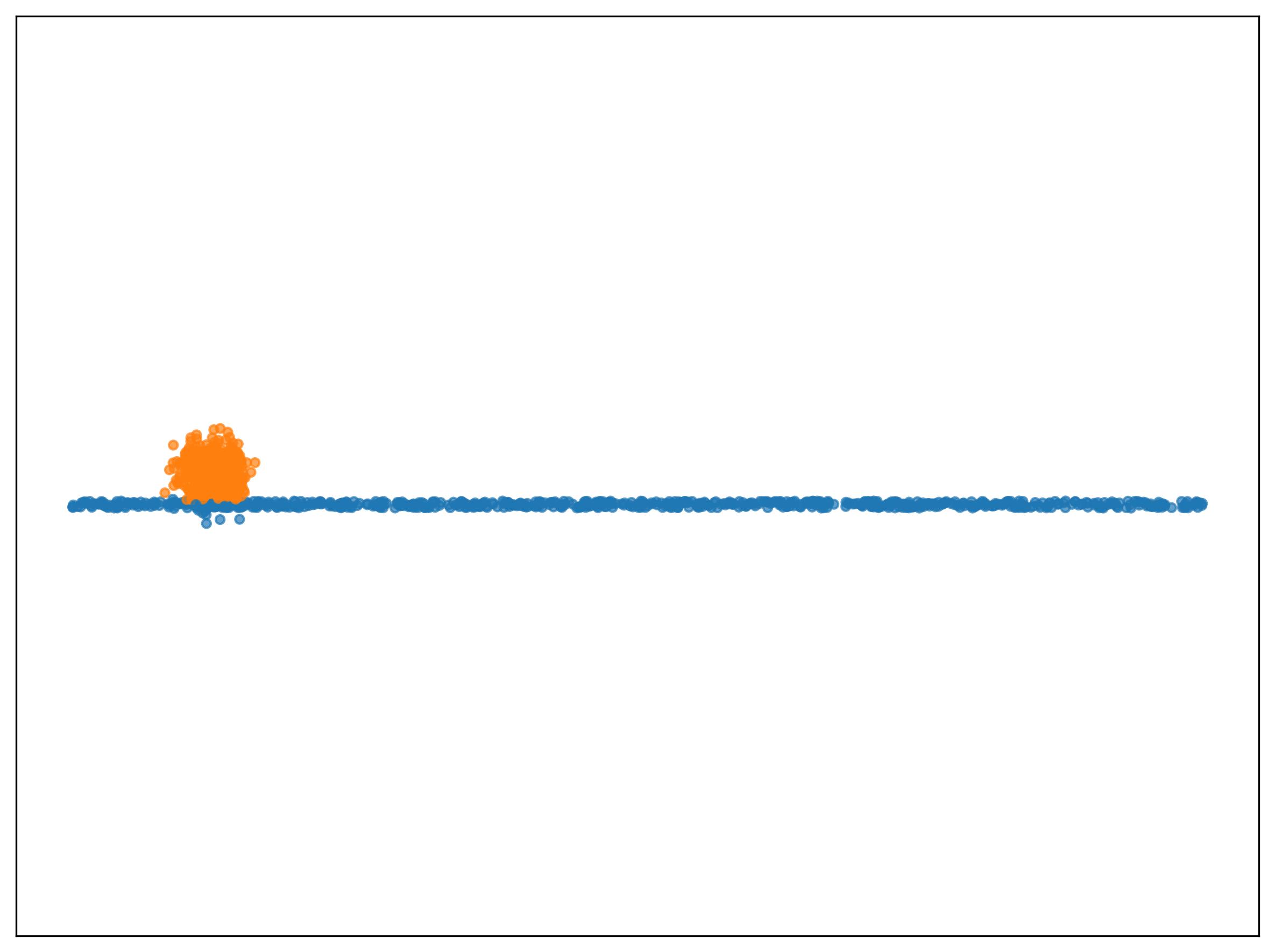} \\
    & & \footnotesize NMI=0.46 & \footnotesize NMI=0.58 & \footnotesize NMI=0.00 & \footnotesize NMI=0.75 & \footnotesize NMI=0.87 \\
    \addlinespace[4pt]
    
    \bottomrule
\end{tabular}}

\end{table*}

\subsection{Visualization of latent representations learned by IDEC}
Here we provide the sisualization of latent representations learned by IDEC. Following the original IDEC settings in Table \ref{tab:idec_hidden_vis}, we fixed the hidden layer dimension to 10 and utilized t-SNE for 2D visualization. The results demonstrate that IDEC fails to effectively learn the latent representations required by Definition \ref{def-requirements}, indicating its limitation in capturing the desired cluster structure.

\begin{table*}[htbp]
\centering
\caption{Visualization of latent representations learned by IDEC. In accordance with the settings in the original IDEC paper, we set the dimension of latent space to 10 and subsequently employ t-SNE \cite{maaten2008tsne} to reduce the dimensionality to 2 for visualization.}
\label{tab:idec_hidden_vis}
\setlength{\tabcolsep}{1.5pt} 
\renewcommand{\arraystretch}{0.2}
\begin{tabular}{c c c c}
    \toprule
    & \textbf{2Crescents} & \textbf{Diff-Sizes} & \textbf{AC} \\
    \midrule

    \rotatebox{90}{Ground Truth} &
    \includegraphics[width=0.25\textwidth]{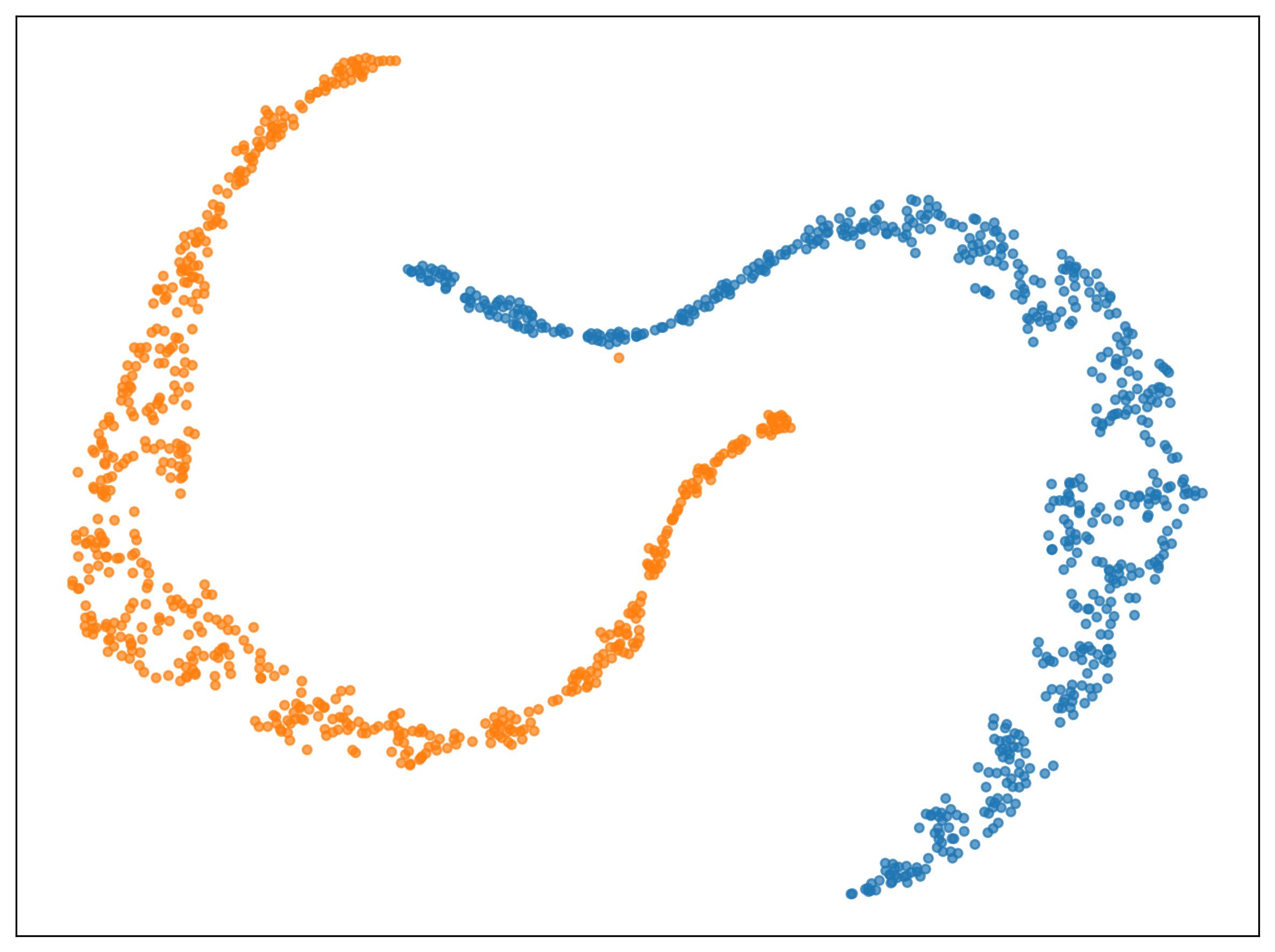} &
    \includegraphics[width=0.25\textwidth]{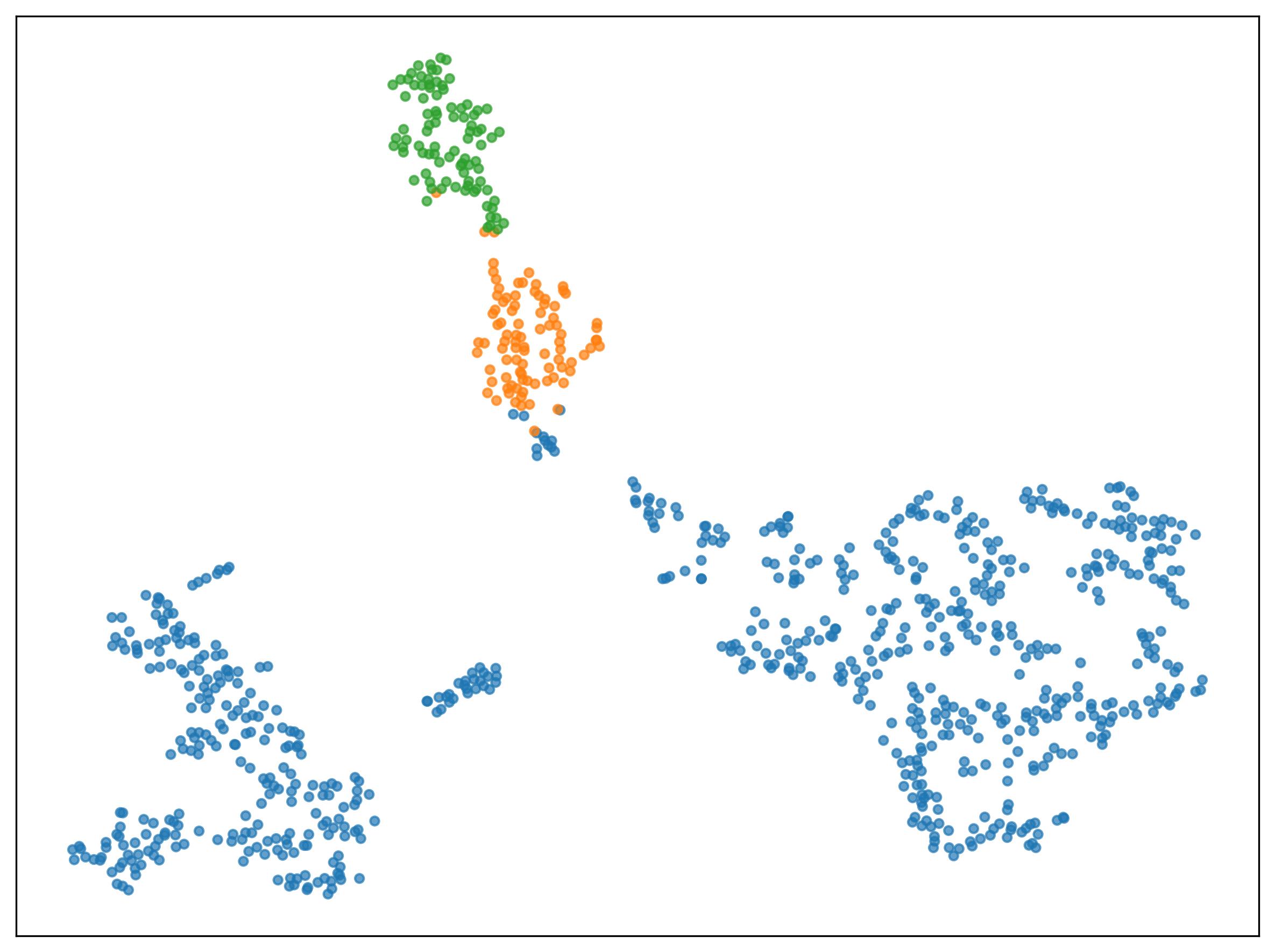} &
    \includegraphics[width=0.25\textwidth]{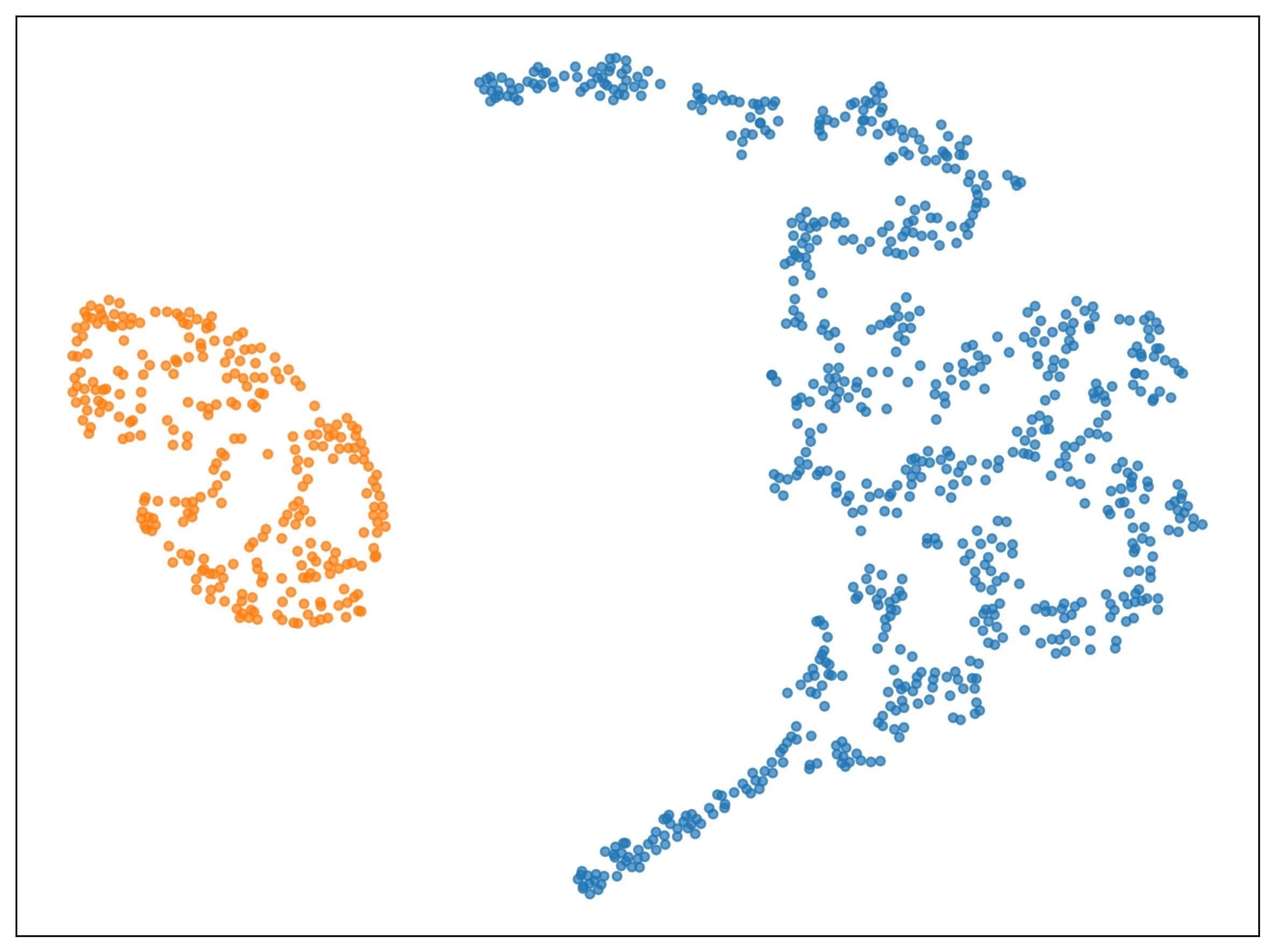} \\
    
    \addlinespace[4pt] 

    \rotatebox{90}{Predicted} &
    \includegraphics[width=0.25\textwidth]{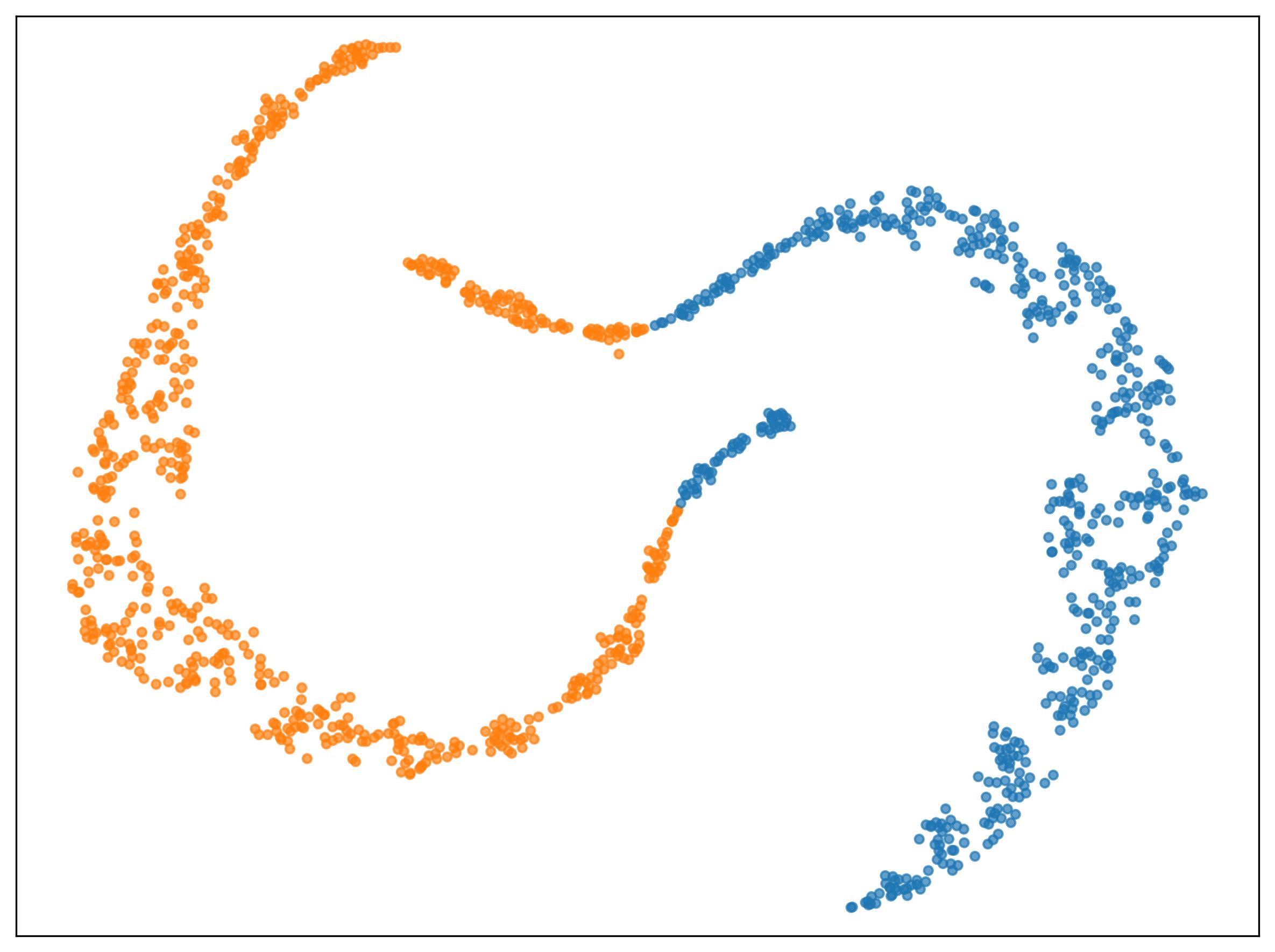} &
    \includegraphics[width=0.25\textwidth]{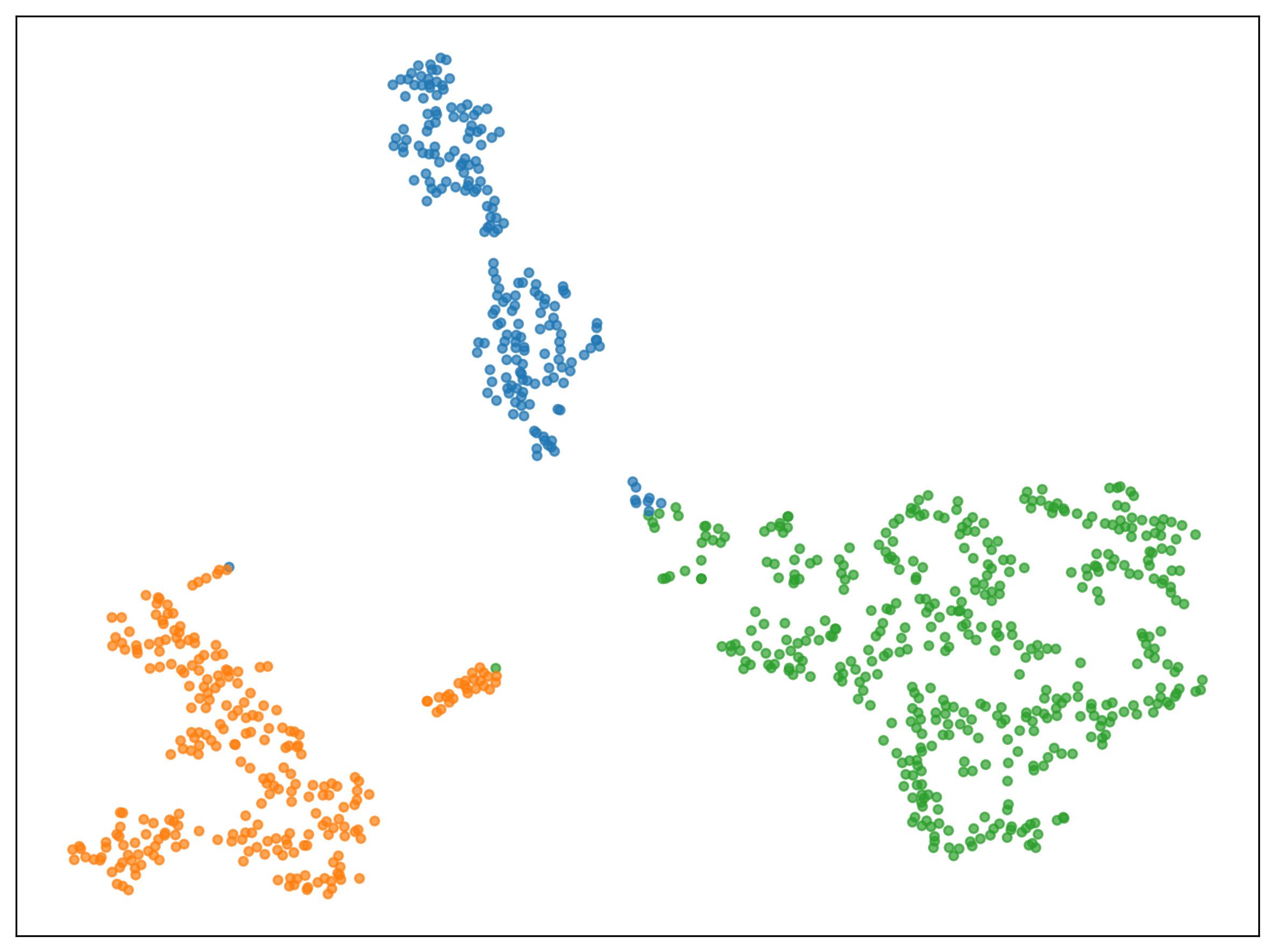} &
    \includegraphics[width=0.25\textwidth]{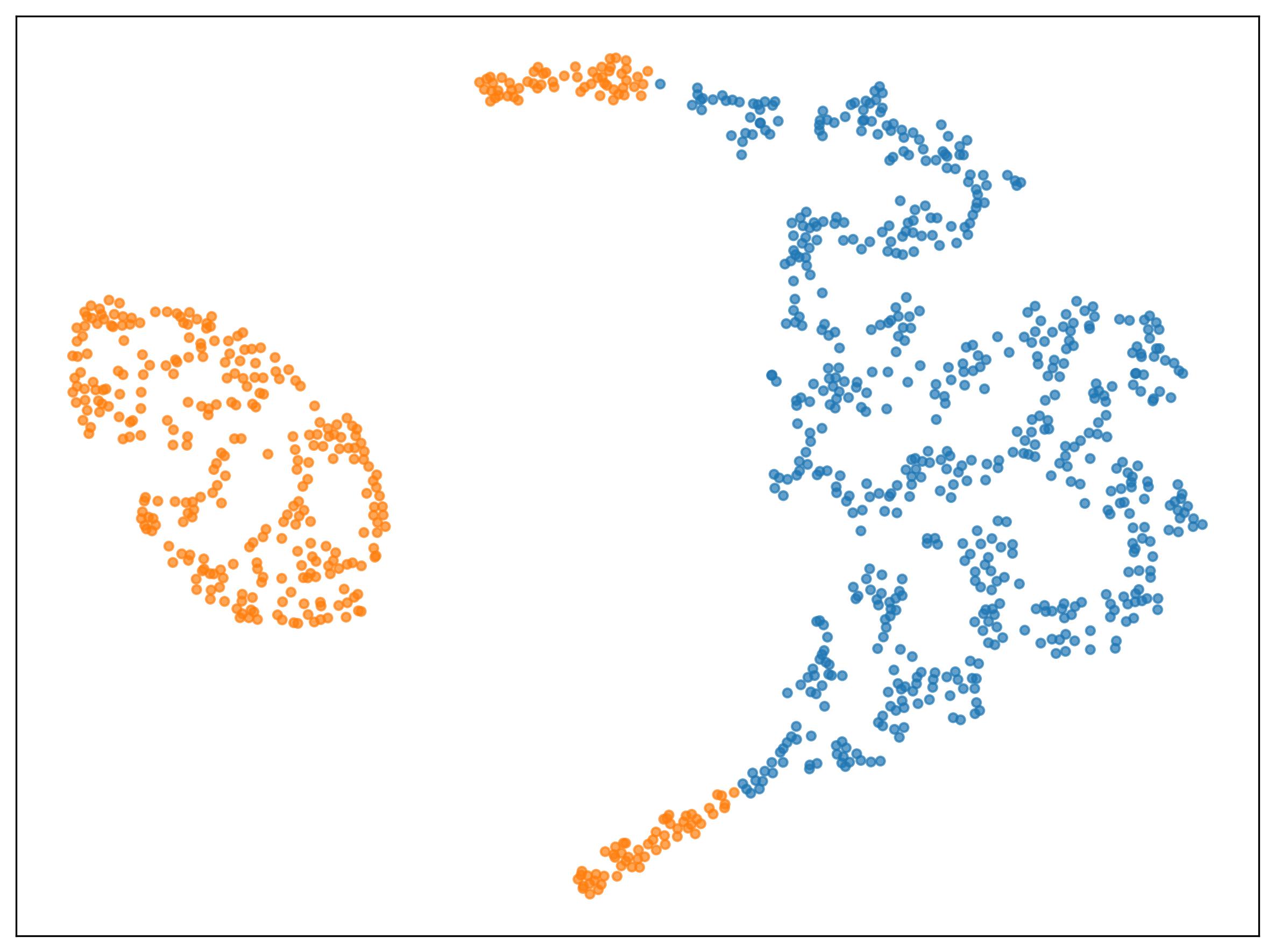} \\

    \bottomrule
\end{tabular}

\end{table*}

\end{document}